\documentclass[twoside,11pt]{article}

\usepackage{jmlr2e}
\usepackage{xcolor}
\usepackage{amsmath}
\usepackage{amssymb}
\usepackage{bm}
\usepackage{hyperref}
\usepackage{algorithm}
\usepackage{algorithmic}
\usepackage{subfigure}
\usepackage{graphicx}
\usepackage{hyperref}
\usepackage{subfigure}
\usepackage{setspace}
\usepackage{amsmath}

 \hypersetup{
     colorlinks=true,
     linkcolor=red,
     filecolor=blue,
     citecolor = green!70!black,
     urlcolor=cyan,
     }
\newcommand*{\QEDA}{\hfill\ensuremath{\blacksquare}}

\firstpageno{1}

\begin{document}

\title{Randomized Spectral Co-Clustering for Large-Scale Directed Networks}

\author{\name Xiao Guo \email xiaoguo.stat@gmail.com \\
        \addr Center for Modern Statistics\\
        School of Mathematics \\
         Northwest University,
         Xi'an, China\\
\name Yixuan Qiu  \email yixuanq@gmail.com \\
        \addr Department of Statistics \\
         Carnegie Mellon University,
         Pittsburgh, USA  \\
\name Hai Zhang \email zhanghai@nwu.edu.cn \\
        \addr School of Mathematics \\
         Northwest University,
         Xi'an, China\\
\name Xiangyu Chang\thanks{Corresponding author. The authors gratefully acknowledge the support of National Natural Science Foundation of China (NSFC, 11771012, 61502342 and U1811461)}\email xiangyuchang@xjtu.eud.cn\\
         \addr Center for Intelligent Decision-Making and Machine Learning\\
         School of Management\\ Xi'an
        Jiaotong University,
           Xi'an, China}
\editor{}

\maketitle

\begin{abstract}
Directed networks are broadly used to represent asymmetric relationships among units. Co-clustering aims to cluster the senders and receivers of directed networks simultaneously. In particular, the well-known spectral clustering algorithm could be modified as the spectral co-clustering to co-cluster directed networks. However, large-scale networks pose great computational challenges to it. In this paper, we leverage sketching techniques and derive two randomized spectral co-clustering algorithms, one \emph{random-projection-based} and the other \emph{random-sampling-based}, to accelerate the co-clustering of large-scale directed networks. We theoretically analyze the resulting algorithms under two generative models -- the stochastic co-block model and the degree-corrected stochastic co-block model, and establish their approximation error rates and misclustering error rates, indicating better bounds than the state-of-the-art results of co-clustering literature. Numerically, we design and conduct simulations to support our theoretical results and test the efficiency of the algorithms on real networks with up to millions of nodes. A publicly available R package \textsf{RandClust} is developed for better usability and reproducibility of the proposed methods.
\end{abstract}


\section{Introduction}

Recent advances in computing and measurement technologies have led to
an explosion of large-scale network data~\citep{newman2018networks}. Networks can describe symmetric (undirected) or asymmetric (directed) relationships among interacting units in various fields, ranging from biology and informatics to social science and finance~\citep{goldenberg2010survey}.
To extract knowledge from complex network structures, many clustering techniques,
also known as community detection algorithms, are widely used to group together
nodes with similar patterns \citep{fortunato2010community}.
In particular, as asymmetric relationships are essential to the organization of networks, clustering \emph{directed networks} is receiving more and more attentions \citep{chung2005laplacians,boley2011commute,rohe2016co,dhillon2001co}. For large-scale directed network data, an appealing clustering algorithm should have not only the statistical guarantee but also the computational advantage.

To accommodate and explore the asymmetry in directed networks, the notion of \emph{co-clustering} was introduced in \citet{rohe2016co,dhillon2001co}, and such an idea can be traced back to \citet{hartigan1972direct}. Let $A\in\{0,1\}^{n\times n}$ be the network adjacency matrix such that $A_{ij}=1$ if there is an edge from node $i$ to node $j$, and $A_{ij}=0$ otherwise. By convention, we also assume $A_{ii}=0$. Then the $i$th row and column of $A$ represent the outgoing and incoming edges for node $i$, respectively. Co-clustering refers to simultaneously clustering both the rows and the columns of $A$ so that the nodes in a row cluster share similar sending patterns, and the nodes in a column cluster share similar receiving patterns. Hence, we will also refer to sending and receiving clusters to row and column clusters, respectively.
Compared to the standard clustering where only one set of clusters is obtained,
co-clustering a directed network yields two possibly different sets of clusters, which provide more insights and improve our understandings on the organization of directed networks.

Spectral clustering \citep{von2007tutorial} is a natural and interpretable algorithm to group undirected networks, which first performs the eigendecomposition on a matrix representing the network, for example, the adjacency matrix $A$, and then runs $k$-means or other similar algorithms to cluster the resulting leading eigenvectors.
Considering the asymmetry in directed networks, the standard spectral clustering algorithm has been modified to the \emph{spectral co-clustering}, in which the eigendecomposition is replaced by the singular value decomposition (SVD), and the $k$-means is implemented on the left and right leading singular vectors, respectively. As the leading left and right singular vectors approximate the row and column spaces of $A$, it is expected that the resulting two sets of clusters contain nodes with similar sending and receiving patterns, respectively. A concrete version of the aforementioned algorithm is introduced in \citet{rohe2016co}.

The spectral co-clustering is easy to be implemented, and has been shown to have many nice properties \citep{von2007tutorial,rohe2016co}. However, large-scale directed networks, namely, networks with a huge number of nodes or dense edges -- say millions of nodes and tens of millions of edges, pose great challenges to the computation of spectral co-clustering. How to improve the efficiency of spectral co-clustering while maintaining a controllable accuracy becomes an urgent need. In this paper, we consider the problem of co-clustering large-scale directed networks based on randomization techniques, a popular approach to reducing the size of data with limited information loss~\citep{mahoney2011randomized,woodruff2014sketching,drineas2016randnla}. Randomization techniques have been widely used in machine learning to speed up fundamental problems such as the least-squares regression and low-rank matrix approximation (see~\citet{drineas2006sampling,meng2013low,nelson2013osnap,pilanci2016iterative,clarkson2017low,halko2011finding,haishan2017fast,martinsson2016randomized}, among many others). The basic idea is to compromise the size of a data matrix (or tensor) by sampling a small subset of the matrix entries or forming linear combinations of the rows or columns. The entries or linear combinations are carefully chosen to preserve the major information contained in the matrix. Hence, randomization techniques may provide a beneficial way to aid the spectral co-clustering of large-scale directed network data.

For a network with community structures, its adjacency matrix $A$ is low-rank in nature, so the randomization for low-rank matrix approximation can be readily used to accelerate the SVD of $A$ \citep{halko2011finding,martinsson2016randomized,witten2015randomized}. We investigate two specific strategies, namely, the random-projection-based and the random-sampling-based SVD. The random projection strategy compresses the original matrix $A$ into a smaller one, whose rows (columns) are random linear combinations of the rows (columns) of $A$. In this way, the dimension of $A$ is largely reduced, and the corresponding SVD is thus sped up. As for the random sampling strategy, the starting point is that there exist fast iterative algorithms to compute the partial SVD of a sparse matrix, such as orthogonal iteration and Lanczos iteration \citep{baglama2005augmented,calvetti1994implicitly}, whose time complexity is generally proportional to the number of non-zero elements of $A$. Therefore, a good way of accelerating the SVD of $A$ is to first sample the elements of $A$ to obtain a sparser matrix, and then use fast iterative algorithms to compute its SVD. As a whole, the spectral co-clustering with the classical SVD therein replaced by the randomized SVD is called the {\it randomized spectral co-clustering}.

Given the fast randomization techniques, it is also critical to study the statistical accuracy of the resulting algorithms under certain generative models. To this end, we assume the directed network is generated from the \emph{stochastic co-block model} (ScBM) or the \emph{degree-corrected stochastic co-block model} (DC-ScBM)~\citep{rohe2016co}. These two models assume the nodes are partitioned into two sets of non-overlapping blocks, one corresponding to the row cluster and the other to the column cluster. Generally, nodes in the same row (column) cluster are stochastically equivalent senders (receivers). That is, two nodes send out (receive) an edge to (from) a third node with the same probability if these two nodes are in the same row (column) cluster.
The difference of these two models lies in that the degree-corrected model~\citep{karrer2011stochastic,rohe2016co} considers the degree heterogeneity arising in real-life networks. The statistical error of the randomized spectral co-clustering is then studied under these two settings.

The merits of the current work lie in the following aspects:
\begin{itemize}
\item We analyze the true singular vector structure of population adjacency matrices generated by ScBMs and DC-ScBMs systematically. The results explain why the spectral co-clustering algorithms work well for directed networks and provide insights on designing the co-clustering algorithms for networks with and without degree heterogeneity. Different from most existing works of literature, we do not assume the row clusters and column clusters have the same number nor assume the ScBMs are of full-rank. We also provide insightful conditions which would lead to the success of spectral co-clustering-based algorithms.

\item We study the approximation and clustering performance of the two randomization based algorithms from a statistical point of view. The results provide statistical insights into the randomized algorithms, which are new in randomization works of literature. For the approximation error, we show that under mild conditions, {\color{red}the minimax optimal error rate is attained}, although we only use sketched data. More interestingly, in terms of misclustering error rates, we even obtain an improved upper bound for randomization methods relative to the non-randomized ones as in \citet{rohe2012co} and \citet{rohe2016co}, due to the use of more advanced technical tools.

\item We evaluate the randomization based spectral co-clustering algorithms on several large-scale networks with up to millions of nodes and tens of millions of edges. They show great efficiency with most running times less than ten seconds on a regular personal computer while yielding satisfactory clustering performance. The core algorithm is publicly available via the R package \textsf{RandClust}\footnote{\url{https://github.com/XiaoGuo-stat/RandClust}}.

\end{itemize}

\subsection{Related works}
Randomization techniques have been widely used to speed up SVD and spectral-clustering-based algorithms; see \citet{halko2011finding,witten2015randomized,martinsson2016randomized,erichson2019randomized,tremblay2016compressive,tremblay2020approximating,liao2020sparse}, among others. The novelty of this paper is that we study the effect of randomization from a statistical perspective. We analyze carefully the approximation error of the randomized adjacency matrix under ScBMs and DC-ScBMs, and find that the approximation error essentially attains the statistical minimax optimal rate under these two models. In particular, under the random sampling scheme, we make full use of the low-rank assumption of population adjacency matrix to deduce an approximation error that can not be obtained by simply combining the results in randomization and SBMs' works of literature. On the other hand, recent years have witnessed a few works studying the randomized algorithms under various statistical models, such as linear regression models, logistic regression models, and constrained regression models; see for example \citet{ma2015statistical,raskutti2016statistical,wang2019information,wang2017sketched,pilanci2016iterative,li2019randomized}. To the best of our knowledge, this is one of the first couples of works to study the randomization under the statistical network models. Note that \citet{zhang2020randomized} studied the randomized spectral clustering algorithms for large-scale undirected networks and analyzed the theoretical properties under the framework of the stochastic block models \citep{holland1983stochastic}. Compared with undirected networks, directed networks contain more information and bring the asymmetry that needs to be accommodated. As will be seen in Lemma \ref{lemma1}, \ref{lemma2}, \ref{lemma3} and \ref{lemma4}, the underlying row and column clusters correspond to distinct singular vector structures. Accordingly, our theoretical results (see Theorem \ref{rpromis} for example) show that the estimated row and column clusters perform differently, and in principle, the nodes would be more easily clustered if their target cluster number is the same as the target rank of ScBMs and DC-SsBMs.

Spectral clustering has also been widely studied under various statistical network models, see \citet{rohe2016co,rohe2012co,lei2015consistency,yun2016optimal,su2017strong,qin2013regularized,abbe2017community,tang2017asymptotically,arroyo2021overlapping}, among many others. In particular, \citet{rohe2016co} and its earlier version \citet{rohe2012co} are seminal works on spectral co-clustering of ScBMs. Compared with previous works, the merits of this work are as follows. First, we provide delicate analysis of the true singular structure of the population adjacency matrix and give sufficient and interpretable conditions on when the population-wise spectral-clustering-based algorithms could succeed (see Lemma \ref{lemma1}, \ref{lemma2}, \ref{lemma3} and \ref{lemma4}). Note that this is rarely mentioned in \citet{rohe2016co,rohe2012co} and previous literature. More importantly, we argue in Theorem \ref{lemmarank} that the extension of this work from the most considered full-rank ScBMs to rank-deficient ScBMs is possible, which is also not common in previous works of literature. Second, as we utilize more advanced techniques, the resulting misclustering bounds are tighter than those in \citet{rohe2016co} and \citet{rohe2012co}, although the latter two studied the non-randomized spectral co-clustering; see the following Table \ref{tabletheory} and find more thorough discussions in Section \ref{relatedwork}. Last but not least, we apply the recently developed techniques for the entry-wise perturbation bound of eigenvectors \citep{abbe2020entrywise} to study the effect of random sampling on the spectral clustering with two underlying clusters, which has been proved to achieve the statistical minimax optimal misclustering error rate without randomization. Our analysis provides insightful results; see Theorem \ref{limits}.
\begin{table}[!htbp]
\centering
\footnotesize
\caption{ A brief comparison of the misclustering error rates and the corresponding conditions in this work and in \cite{rohe2016co} and \cite{rohe2012co}. $K$, $n$, $\alpha_n$, $p$ denote the number of clusters, number of nodes, maximum link probability of edges and the sampling rate in the random sampling scheme, respectively. }
\def\arraystretch{1.5}
\begin{tabular}{lcc}
\hline
&Corollary C.1 in \citet{rohe2016co}&Corollary 4.1 in \citet{rohe2012co}  \\
Bounds&$O(K^2{\rm log} n/n)$&$o(K^3{\rm log} n/\alpha_n^4)$\\
Conditions&$\alpha_n=O(1)$  &$K=O(n^{1/4}/{\rm log} n)$\\
\hline
&Theorem \ref{rpromis}&Theorem \ref{rsammis} \\
Bounds&${O(K^2/(n\alpha_n))}$&$O(K^2/{(pn\alpha_n)})$\\
Conditions&(\ref{C2})&(\ref{C2}), $p>1/2$\\
\hline
\end{tabular}
\label{tabletheory}
\end{table}

The modern computation of SVD can be traced back to the 1960s, when the seminal works \citet{golub1965calculating,golub1970singular} provided the basis for the \emph{EISPACK} and \emph{LAPACK} routines, though the randomized matrix decomposition is a relatively young field. For computing partial SVD of matrices, iterative algorithms flourished; see \citep{calvetti1994implicitly,baglama2005augmented,jia2003implicitly,jia2010refined,wu2015preconditioned,wu2017primme_svds}. Another branch of
algorithms are stochastic and incremental variants of the deterministic iterative algorithms \citep{oja1985stochastic,arora2013stochastic,shamir2015stochastic,shamir2016convergence,xu2018accelerated}. Compared with iterative algorithms, the rationality of random-sampling-based scheme is straightforward: we accelerate the iterative method of \citet{baglama2005augmented} by sampling the original matrix at the price of accuracy. The random-sample-based scheme can be generalized by using more advanced deterministic or stochastic iterative methods as the starting algorithm. On the other hand, for the random-projection-based scheme, its advantage over iterative methods lies in the following aspects. First, the random-projection-based method enables distributed computing since the matrix multiplications therein can be parallelized. Second, it is communication efficient because only a few passes over the input matrix are required. Overall, as is evidenced in Section \ref{realdata}, randomized methods are more efficient than iterative algorithms while maintaining good accuracy on large-scale networks. For reference, we summarize the time complexities of mentioned methods in Table \ref{tablenumerical}.

\begin{table}[!htbp]
\centering
\footnotesize
\caption{A summary of the time complexities of the randomized methods in this work and other methods for computing the SVD. $n,K$ denote the number of nodes and the target rank, respectively. $q$ denotes the power parameter and $r,s$ denote the oversampling parameters. $T_0$ and $T_1$ denote the number of iterations. $\|A\|_0$ and $\|{A}^{\rm rs}\|_0$ represent the number of non-zero elements in the original adjacency matrix and the sparsified matrix.  See Section \ref{algorithms} for more details. }
\def\arraystretch{1.5}
\begin{tabular}{lcc}
\hline
Method&Full SVD&Iterative methods\\
Time&$O(n^3)$&$O(\|{A}\|_0{ T_0})$\\
Method&Projection-based SVD&Sampling-based SVD  \\
Time&$O((2q + 1)n^2 (K+{\rm max}(r,s)))$&$O(\|{A}^{\rm rs}\|_0K{ T_1})$\\
\hline
\end{tabular}
\label{tablenumerical}
\end{table}

The remainder of the paper is organized as follows. Section \ref{algorithms} introduces the randomized spectral co-clustering algorithms for co-clustering large-scale directed networks. Section \ref{theory} includes the theoretical analysis of the proposed algorithms under two network models. Section \ref{relatedwork} discusses several theoretical aspects and possible extensions on the proposed methods. Section \ref{experiment} and \ref{realdata} present the experimental results on simulated and real-world data, respectively. Section \ref{conclusion} concludes the paper. Technical proofs are included in the appendix.

\section{Randomized spectral co-clustering}
\label{algorithms}

\subsection{A brief review of prior art}

For directed networks, co-clustering aims to find two possibly different sets of clusters, namely, row clusters and column clusters, to describe and understand the sending pattern and receiving pattern of nodes, respectively. Suppose there are $K^y$ row clusters and $K^z$ column clusters, and without loss of generality, assume $K^y\leq K^z$. Write the partial SVD of $A$, the adjacency matrix, as $A\approx{U}{\Sigma}{V}^\intercal$, where the left singular vectors ${{U}\in \mathbb {R}^{n\times K^y}}$ and right singular vectors ${{V}\in \mathbb {R}^{n\times K^y}}$ approximate the row and column spaces of $A$, respectively. On the other hand, ${U}$ contains the eigenvectors of the symmetric matrix $AA^\intercal$, whose $(i,j)$ entry corresponds to the number of common children of nodes $i$ and $j$. Similarly, $V$ represents the eigenvectors of $A^\intercal A$, whose $(i,j)$ entry is the number of common parents of $i$ and $j$. Therefore, ${U}$ and ${V}$ contain the sending and receiving information of each node, and clustering ${U}$ and ${V}$ respectively would yield clusters with nodes sharing similar sending and receiving patterns.

Based on the explanations above, the well-known spectral clustering is a good paradigm for co-clustering directed networks \citep{hartigan1972direct,rohe2012co,rohe2016co}. We consider the following two variants of spectral co-clustering algorithms, corresponding to different assumptions on the network. The first one is based on the standard spectral clustering, which first computes the SVD of $A$, and then uses $k$-means to cluster the left and right singular vectors of $A$, respectively (Algorithm \ref{spectral}, SCC). This algorithm is well-suited to networks whose nodes have approximately equal degrees. Whereas for networks whose nodes have heterogeneous degrees, the following algorithm (Algorithm \ref{spectralmedian}, SsCC) is preferred. It first computes the SVD of $A$ and then normalizes the non-zero rows of the left and right singular vectors such that the resulting rows have Euclidean norm 1. The zero rows are remained the same. The $k$-means clustering is then performed on the normalized rows of the left and right singular vectors, respectively. The normalization step aims to balance the importance of each node to facilitate the subsequent clustering procedures, which was also studied in \citet{rohe2012co,rohe2016co,lei2015consistency}, among others. As we will see in Section \ref{theory}, this step is essential for co-clustering networks with degree heterogeneity.

\begin{algorithm}
\small

\renewcommand{\algorithmicrequire}{\textbf{Input:}}

\renewcommand\algorithmicensure {\textbf{Output:} }

\caption{Spectral co-clustering with $k$-means}

\label{spectral}

\begin{algorithmic}[1]

\REQUIRE ~\\

Adjacency matrix $A\in \mathbb{R}^{n\times n}$ of a directed network, number of row clusters $K^y$, and number of column clusters $K^z$ ($K^y\leq K^z$).\\

~\\
\STATE Compute the partial SVD of $A$, with left and right singular vectors ${{U}\in \mathbb {R}^{n\times K^y}}$ and ${{V}\in \mathbb {R}^{n\times K^y}}$.\\
\STATE Run $k$-means on ${U}$ with ${K^y}$ target clusters and on ${V}$ with ${K^z}$ clusters.  \\
\STATE Output the co-clustering results.
 \\
\end{algorithmic}
\end{algorithm}

\begin{algorithm}
\small

\renewcommand{\algorithmicrequire}{\textbf{Input:}}

\renewcommand\algorithmicensure {\textbf{Output:} }

\caption{Spectral co-clustering with spherical $k$-means}

\label{spectralmedian}

\begin{algorithmic}[1]

\REQUIRE ~\\

Adjacency matrix $A\in \mathbb{R}^{n\times n}$ of a directed network, number of row clusters $K^y$, and number of column clusters $K^z$ ($K^y\leq K^z$).\\

~\\
\STATE Compute the partial SVD of $A$, with left and right singular vectors ${{U}\in \mathbb {R}^{n\times K^y}}$ and ${{V}\in \mathbb {R}^{n\times K^y}}$.\\
\STATE Construct $U'$ and $V'$, whose rows are normalized rows of $U$ and $V$, respectively. The zero rows are remained the same.\\
\STATE Run $k$-means on ${U'}$ with ${K^y}$ target clusters and on ${V'}$ with ${K^z}$ clusters. \\
\STATE Output the co-clustering results.
\\
\end{algorithmic}
\end{algorithm}

Now we discuss the time complexity of Algorithm \ref{spectral} and \ref{spectralmedian}. It is well-known that the classical full SVD generally takes $O(n^3)$ time, which is time-consuming when  $n$ is large. But in fact, only the partial SVD of $A$ is needed, which can be done by fast iterative methods \citep{calvetti1994implicitly,baglama2005augmented}. They generally take $O(n^2K^yT_0)$ time, where $T_0$ is the iteration number corresponding to a certain error, and it can be large when $n$ is large. For $k$-means, finding its optimal solution is NP-hard, and hence efficient heuristic algorithms are commonly employed. In this paper, we use the Lloyd's algorithm to solve  $k$-means, whose time complexity is proportional to $n$. Alternatively, one can use a more delicate $(1+\epsilon)$-approximate $k$-means \citep{kumar2004simple} for a good approximate solution within a constant fraction of the optimal value. Based on the discussions above, the time complexities of Algorithm \ref{spectral} and \ref{spectralmedian} are dominated by the SVD, which encourages the use of randomization techniques to speed up the computation of SVD for further improving the spectral co-clustering.

\subsection{Random-projection-based spectral co-clustering (RP-SCC)}

The basic idea of the \emph{{{R}}andom-{P}rojection-based Spectral Co-Clustering} (RP-SCC) is to compress the adjacency matrix $A$ into a smaller matrix, and then apply a standard SVD to the compressed one, thus saving the computational cost. The approximate SVD of the original $A$ can be recovered by postprocessing the SVD of the smaller matrix \citep{halko2011finding,martinsson2016randomized,witten2015randomized}.

For an asymmetric matrix $A$ with a target rank $K^y$, the objective is to find orthonormal bases $Q,T\in \mathbb R^{n\times K^y}$ such that
$$A\approx QQ^{\intercal}ATT^{\intercal}:={A}^{\rm rp}.$$
It is not hard to see that $QQ^\intercal$ projects the column vectors of $A$ to the column space of $Q$, and $TT^\intercal$ projects the row vectors of $A$ to the row space of $T$ (or the column space of $T^\intercal$). Therefore, $Q$ and $T$ approximate the column and row spaces of $A$, respectively. In randomization methods, $Q$ and $T$ can be built via \emph{random projection} \citep{halko2011finding}. Take $Q$ as an example, one first constructs an $n\times K^y$ random matrix whose columns are random linear combinations of the columns of $A$, and then orthonormalizes the $K^y$ columns using the QR decomposition to obtain the orthonormal matrix $Q$. Once $Q$ and $T$ are constructed, the standard SVD is performed on $Q^{\intercal}AT$, and the approximate SVD of $A$ can be achieved by left-multiplying $Q$ and right-multiplying $T$. The whole procedure of the random-projection-based SVD can be summarized as the following steps:
\begin{itemize}
\item\emph{{Step 1:}} Construct two test matrices $\Omega,\Gamma\in \mathbb R^{n\times K^y}$ with independent standard Gaussian entries.
\item\emph{{Step 2:}} Obtain $Q$ and $T$ via the QR decomposition $A\Omega\rightarrow QR_1$ and $A^\intercal\Gamma\rightarrow TR_2$.
\item\emph{{Step 3:}} Compute SVD of $Q^{\intercal}AT\rightarrow U_s\Sigma V_s^\intercal.$
\item\emph{{Step 4:}} Output the approximate SVD of $A$ as $A\approx{U}^{\rm rp}\Sigma ({V}^{\rm rp})^\intercal$, where ${U}^{\rm rp}:=QU_s$ and ${V}^{\rm rp}:=TV_s$.
\end{itemize}
To fix ideas, RP-SCC generally refers to spectral co-clustering with the SVD therein replaced by the random-projection-based SVD. While in some places to follow, RP-SCC and RP-SsCC refer particularly to random-projection-based SVD coupled with Algorithm \ref{spectral} and \ref{spectralmedian}, respectively.

In actual implementation, the \emph{oversampling} and \emph{power iteration} schemes can be used to improve the performance of the randomized SVD \citep{halko2011finding,martinsson2016randomized}. Oversampling uses extra $r$ and $s$ ($K^y+r$ and $K^y+s$ in total) random projections to form the sketch matrices $A\Omega$ and $A^\intercal\Gamma$ in \emph{Step 2}, which reduce the information loss when the rank of $A$ is not exactly $K^y$. The power iteration scheme employs $(AA^\intercal)^qA\Omega$ and $(A^\intercal A)^qA^\intercal\Gamma$ instead of $A\Omega$ and $A^\intercal\Gamma$ in \emph{Step 2}. This treatment improves the quality of the sketch matrix when the singular values of $A$ are not rapidly decreasing.

The time complexity of RP-SCC is dominated by the matrix multiplication operations in \emph{Step 2}, which generally take $O((2q + 1)n^2 (K^y+{\rm max}(r,s)))$ time. Note that the classical SVD in \emph{Step 3} is cheap as the matrix dimension is as low as $K^y+{\rm max}(r,s)$. In addition, the time of \emph{Step 2} can be further improved if one uses structured random test matrices or performs the matrix multiplications in parallel. The random-projection-based SVD is numerically stable, and comes with its good theoretical guarantee \citep{halko2011finding,martinsson2016randomized,witten2015randomized}.

\subsection{Random-sampling-based spectral co-clustering (RS-SCC)}

The \emph{{R}andom-Sampling-based Spectral Co-Clustering} (RS-SCC) is based on the fact that real-world networks are often sparse~\citep{watts1998collective,chang2019popularity}, meaning that the number of non-zero elements in the adjacency matrix $A$ is $O(n^\alpha)$ with $0<\alpha<2$. It is known that the time complexity of fast iterative algorithms of SVD is proportional to the number of non-zero elements of the matrix~\citep{calvetti1994implicitly,baglama2005augmented}. Consequently, RS-SCC makes the SVD of $A$ more efficient by randomly sampling the elements of $A$, followed by a fast iterative algorithm to compute the leading singular vectors of the sparsified matrix. The SVD of $A$ can then be approximated by that of the sparsified matrix.

We use the following simple strategy to construct the sparsified matrix ${A}^{\rm rs}$: each element of $A$ is sampled with equal probability $p$, and the elements that are not sampled are forced to be zero. Formally, for each pair of $(i,j)$,
\[{A}_{ij}^{\rm rs}=\begin{cases}
\label{2.1}
\frac{A_{ij}}{p}, & \mbox{if }\; (i,j) { \mbox{ is selected},} \\
0,& \mbox{if } \;(i,j) {\mbox{ is not selected}},\tag{2.1}
\end{cases}\]
where $A_{ij}$ is divided by $p$ to remove bias as we will see in Section \ref{theory}. If the sampling probability $p$ is not too small, then ${A}^{\rm rs}$ is close to $A$ with little information loss. With $A^{\rm rs}$ at hand, the random-sampling-based SVD follows:
\begin{itemize}
\item\emph{{Step 1:}} Form the sparsified matrix ${A}^{\rm rs}$ via (\ref{2.1}).
\item\emph{{Step 2:}} Compute the partial SVD of ${A}^{\rm rs}$ using the fast iterative algorithm in \citet{calvetti1994implicitly} or \citet{baglama2005augmented} such that ${A}^{\rm rs}\approx U^{\rm rs}_{n\times K^y}\Sigma_{K^y\times K^y}(V^{\rm rs})^\intercal_{K^y\times n}$.
\end{itemize}
Generally, RS-SCC refers to spectral co-clustering with the SVD therein replaced by the random-sampling-based SVD, while in some places to follow, we may use RS-SCC and RS-SsCC to distinguish Algorithm \ref{spectral} and \ref{spectralmedian}.

The time complexities of \emph{Step 1} and \emph{Step 2} are approximately $O(\|A\|_0)$ and $O(\|{A}^{\rm rs}\|_0K^y{ T_1})$, where $\|A\|_0$ denotes the number of non-zero elements in $A$, and ${T_1}$ is the number of iterations. The number of edges in a real-world network is typically far below $O(n^2)$, thus making RS-SCC rather efficient.

\section{Theoretical analysis}
\label{theory}

\subsection{Preliminaries}
We analyze the theoretical properties of the randomized spectral co-clustering algorithms under two generative models, ScBM and DC-ScBM \citep{rohe2016co}. In ScBM, nodes in a common row cluster are stochastically equivalent senders in the sense that they send out an edge to a third node with equal probabilities. Similarly, nodes in a common column cluster are stochastically equivalent receivers, as they receive an edge from a third node with equal probabilities. In DC-ScBM, however, the probabilities depend not only on the row or column clusters but also on propensity parameters for each node.

To provide the formal definitions of these two models, we first introduce the following notation. Recall that for a directed network $A\in\mathbb R^{n\times n}$, we assume there exist $K^y$ row clusters and $K^z$ column clusters with $K^y \leq K^z$. For $i=1,...,n$, let $g_i^y\in\{1,...,K^y\}$ and $g_i^z\in\{1,...,K^z\}$ denote the assignments of the row cluster and column cluster of node $i$, respectively. Alternatively, the cluster assignments can be represented by membership matrices defined as follows. Let $\mathbb{M}_{n,K}$ be the set of all $n\times K$ matrices that have exactly one 1 and $K-1$ 0's in each row, and $Y\in \mathbb{M}_{n,K^y}$ and $Z\in \mathbb{M}_{n,K^z}$ are two matrices such that $Y_{ig_i^y}=1$ and $Z_{ig_i^z}=1$ for each $i$. $Y$ and $Z$ are then called row and column membership matrices, respectively. For $1\leq k\leq K^y$, let $G_k^y=\{1\leq i\leq n:g_i^y=k\}$ be the set of nodes belonging to row cluster $k$, and denote by $n_k^y=|G_k^y|$ its size. Similarly, for $1\leq k\leq K^z$, define $G_k^z=\{1\leq i\leq n:g_i^z=k\}$ and $n_k^z=|G_k^z|$ for column cluster $k$. For any matrix $B$ and proper index sets $I$ and $J$, $B_{I\ast}$ and $B_{\ast J}$ denote the sub-matrices of $B$ that consist of the rows in $I$ and columns in $J$, respectively. $\|B\|_{\tiny {\rm F}}$, $\|B\|_2$, and $\|B\|_{\infty}$ are the Frobenius norm, spectral norm, and the element-wise maximum absolute value of $B$, respectively. Finally, ${\rm diag}(B)$ denotes a diagonal matrix whose diagonal entries are the same as those of $B$.

\subsection{Stochastic co-block model}
We use the following definition for ScBM.
\begin{definition}[ScBM, \citealp{rohe2016co}]
Let $Y\in \mathbb M_{n,K^y}$ and $Z\in \mathbb M_{n,K^z}$ be the row and column membership matrices, respectively. Let $B\in [0,1]^{K^y\times K^z}$ be the connectivity matrix whose $(k,l)$th element is the probability of a directed edge from any node in the row cluster $k$ to any node in the column cluster $l$. Given $(Y,Z,B)$, each element of the network adjacency matrix $A=(a_{ij})_{1 \leq i,j\leq n}$ is generated independently as $a_{ij}\sim{\rm Bernoulli}(B_{g_i^yg_j^z})$ if $i\neq j$, and $a_{ij}=0$ if $i= j$.
\end{definition}

Throughout this subsection, we would consider the ScBM parameterized by $(Y,Z,B)$. Note that when $Y=Z$, and $B$ and $A$ are symmetric, then ScBMs reduce to the stochastic block models (SBMs) \citep{holland1983stochastic}. Define $P=YBZ^\intercal$, and let $\sigma_n$ and $\gamma_n$ denote its maximum and minimum non-zero singular values, respectively. The formulation of $P$ makes sense throughout this subsection. It is easy to see that $P$ is the population version of $A$ in the sense that $\mathbb E(A)=P-{\rm diag}(P)$. We assume throughout this subsection that ${\rm rank}(P)={\rm rank}(B)=K^y$, though relaxing this assumption to ${\rm rank}(P)={\rm rank}(B)\leq K^y$ is also feasible as we will discuss in Section \ref{relatedwork}. The following Lemma \ref{lemma1} reveals the structure of the singular vectors of $P$.
\begin{lemma}
\label{lemma1}
Denote the SVD of the population matrix $P=YBZ^\intercal$ by $\bar{U}_{n\times K^y}\bar{\Sigma}_{K^y\times K^y}\bar{V}^\intercal_{K^y\times n}$. Define $\Delta_y={\rm diag}(\sqrt{n_1^y},...,\sqrt{n_{K^y}^y})$, $\Delta_z={\rm diag}(\sqrt{n_1^z},...,\sqrt{n_{K^z}^z})$, and denote the SVD of $\Delta_yB\Delta_z$ by $L_{K^y\times K^y}D_{K^y\times K^y}R^\intercal_{K^y\times K^z}$. Then the following arguments hold for any $1\leq i\neq j\leq  n$.

(1) If $Y_{i\ast}=Y_{j\ast}$, then $\bar{U}_{i\ast}=\bar{U}_{j\ast}$; otherwise
$$\|\bar{U}_{i\ast}-\bar{U}_{j\ast}\|_2=\sqrt{(n_{g_i^y}^y)^{-1}+(n_{g_j^y}^y)^{-1}}.$$

(2) If $Z_{i\ast}=Z_{j\ast}$, then $\bar{V}_{i\ast}=\bar{V}_{j\ast}$; otherwise
$$\|\bar{V}_{i\ast}-\bar{V}_{j\ast}\|_2=\left\|\frac{R_{g_{i}^z\ast}}{\sqrt{n_{g_i^z}^z}}-\frac{R_{g_{j}^z\ast}}{\sqrt{n_{g_j^z}^z}}\right\|_2.$$
Moreover, if $\Delta_z^{-1}R$'s rows are mutually distinct such that there exists a deterministic sequence $\{\xi_n\}_{n\geq 1}$ satisfying
\begin{align}
\label{C1}
\min_{1\leq k\neq l\leq K^z}\left\|\frac{R_{k\ast}}{\sqrt{n_{k}^z}}-\frac{R_{l\ast}}{\sqrt{n_{l}^z}}\right\|_2\geq \xi_n >0,\tag{C1}
\end{align}
then $\|\bar{V}_{i\ast}-\bar{V}_{j\ast}\|_2\geq \xi_n >0$.
\end{lemma}

The following lemma provides an explicit condition on $B$ which suffices for \eqref{C1}.

\begin{lemma}
\label{lemma2}
Under the same parameter setting as in Lemma \ref{lemma1}, if the columns of $B$ are mutually distinct such that
$$\min_{1\leq k\neq l\leq K^y}\|B_{\ast k}-B_{\ast l}\|_2\geq \mu_n $$
for some $\mu_n>0$, then (\ref{C1}) holds with $\xi_n=\mu_n \cdot{\underset{1\le k\le K^y}{{\rm min}} ({n^y_k})^{1/2}}/\sigma_n$, where $\sigma_n$ is the maximum singular value of $P$.
\end{lemma}

Lemma \ref{lemma1} and \ref{lemma2} provide the following important insights. The left singular vectors $\bar{U}$ of $P$ reveal the true row clusters in the sense that two rows of $\bar{U}$ are identical if and only if the corresponding nodes are in the same row cluster. In addition, for two nodes in distinct row clusters, the distance between their corresponding rows of $\bar{U}$ is determined by their row cluster sizes. However, the story for the column clusters is slightly different. Nodes in common column clusters have equal rows in $\bar{V}$, but the converse is generally not true which is caused by the fact that $K^z\geq K^y={\rm rank}(B)$. Nonetheless, in Lemma \ref{lemma2}, we see that if the columns of $B$ are mutually distinct, then the converse is also true. In particular, a larger minimum distance of the column pairs in $B$ would possibly lead to a larger minimum distance of the rows pairs in $\bar{V}$. Based on these facts, one would expect that the spectral co-clustering algorithms (Algorithm \ref{spectral} and \ref{spectralmedian}) would estimate the true underlying clusters well if the singular vectors of $A$ are close enough to those of $P$, which by the Davis-Kahan-Wedin theorem \citep{o2018random} would hold if $A$ and $P$ are close in some sense. Moreover, $A$ is approximated by ${A}^{\rm rp}$ in RP-SCC and by ${A}^{\rm rs}$ in RS-SCC, respectively, so in subsequent sections, we first study the deviation of ${A}^{\rm rp}$ and ${A}^{\rm rs}$ from $P$, and then examine the clustering performance of the proposed methods.

\subsubsection{Performance of RP-SCC in ScBMs}

First, Theorem \ref{rproappro} quantifies the spectral deviation of ${A}^{\rm rp}$ from $P$.

\begin{theorem}\label{rproappro}
Let ${A}^{\rm rp}=QQ^{\intercal}ATT^{\intercal}$ be the random projection approximation to $A$ with target rank $K^y$. Assume that the oversampling parameters $(r,s)$ satisfy $K^y+r\leq n$ and $K^y+s\leq n$, and the test matrices have i.i.d. standard Gaussian entries. If
\begin{equation}
\label{C2}{\rm max}_{kl}B_{kl}\leq \alpha_n \;{\rm for\; some}\; \alpha_n\geq c_0\,{\rm log}n/n \;{\rm and}\; c_0>0,\tag{C2}
\end{equation}
and
\begin{equation}
\label{C3} \min (r,s)\geq 4,\ \max (r{\rm log}r, s{\rm log}s)\leq n,\ q=c_1\cdot n^{1/\tau}\tag{C3}
\end{equation}
for some constant $c_1>0$ and $\tau>0$,
then for any $\epsilon>0$, there exists a constant $c_2=c_2(c_0,c_1,\tau,\epsilon)$ such that
\begin{equation}
\label{3.1}\|{A}^{\rm rp}-P\|_2\leq c_2\sqrt{n\alpha_n}, \tag{3.1}
\end{equation}
with probability at least $1-6r^{-r}-6 s^{-s}-3n^{-\epsilon}$.
\end{theorem}

Theorem \ref{rproappro} implies that the randomized adjacency matrix ${A}^{\rm rp}$ concentrates around $P$ at the rate of $\sqrt{n\alpha_n}$, where $n\alpha_n$ can be regarded as the upper bound of the expected degree in the network $A$. (\ref{C2}) prevents the network from being too sparse, and it is a common requirement in SBMs literature; see \citet{lei2015consistency}, among others. (\ref{C3}) ensures that the error caused by the random projection, namely, $\|A^{\rm rp}-A\|_2$, is dominated by the error caused by the ScBMs, namely, $\|A-P\|_2$. The bound (\ref{3.1}) achieves the statistical minimax optimal error rate \citep{gao2017achieving}. In this sense, the random projection pays no price under the framework of ScBMs.

The next theorem provides an upper bound for the proportion of misclustered nodes.

\begin{theorem}\label{rpromis}
Let ${{Y}}^{\rm rp}\in \mathbb M_{n,K^y}$ and ${{Z}}^{\rm rp}\in \mathbb M_{n,K^z}$ be the estimated membership matrices of RP-SCC. Suppose {(\ref{C2})} and (\ref{C3}) hold and other parameter settings are the same with those in Theorem \ref{rproappro}. The following two arguments hold for ${{Y}}^{\rm rp}$ and ${{Z}}^{\rm rp}$, respectively.

(1) Define $${\tau={\rm min}_{l\neq k}\;\sqrt{(n_k^y)^{-1}+(n_l^y)^{-1}}.}$$
If there exists a positive constant $c_3>0$ such that,
\begin{equation}
\label{C4}\frac{{K^y\alpha_n n}}{n^y_k\tau^2\gamma_n^2}\leq c_3,\tag{C4}
\end{equation}
for any $k=1,...,K^y$, then with probability larger than $1-6r^{-r}-6 s^{-s}-3n^{-\epsilon}$ for any $\epsilon>0$, there exists a subset $M^y\in \{1,...,n\}$ satisfying
\begin{equation}
\label{3.2} \frac{|M^y|}{n}\leq c_3^{-1}\frac{{K^y\alpha_n}}{\tau^2\gamma_n^2}.\tag{3.2}
\end{equation}
Moreover, for $T^y=\{1,...,n\}\backslash M^y$, there exists a $K^y\times K^y$ permutation matrix $J^y$ such that
\begin{equation}
\label{3.3}{{Y}}^{\rm rp}_{T^y\ast}J^y=Y_{T^y\ast}.\tag{3.3}
\end{equation}

(2) Define $$\delta=\min_{1\leq k\neq l\leq K^z}\|\frac{R_{k\ast}}{\sqrt{n_{k}^z}}-\frac{R_{l\ast}}{\sqrt{n_{l}^z}}\|_2,$$
where recall that $R$ denotes the right singular matrix of $\Delta_yB\Delta_z$. If there exists a positive constant $c_4>0$ such that,
\begin{equation}
\label{C5}\frac{{{K^y}\alpha_n n}}{n^z_k\delta^2\gamma_n^2}\leq c_4,\tag{C5}
\end{equation}
for any $k=1,...,K^z$, then with probability larger than $1-6r^{-r}-6 s^{-s}-3n^{-\epsilon}$ for any $\epsilon>0$, there exists a subset $M^z\in \{1,...,n\}$ satisfying
\begin{equation}
\label{3.4} \frac{|M^z|}{n}\leq c_4^{-1}\frac{{{K^y}\alpha_n}}{\delta^2\gamma_n^2}.\tag{3.4}
\end{equation}
Moreover, for $T^z=\{1,...,n\}\backslash M^z$, there exists a $K^z\times K^z$ permutation matrix $J^z$ such that
\begin{equation}
\label{3.5}{{Z}}^{\rm rp}_{T^z\ast}J^z=Z_{T^z\ast}.\tag{3.5}
\end{equation}
\end{theorem}

Theorem \ref{rpromis} provides upper bounds for the misclustering rates  with respect to row clusters and column clusters, as indicated in (\ref{3.2}) and (\ref{3.4}). Recalling Lemma \ref{lemma1}, we can see that the clustering performance depends on the minimum row distances $\tau$ and $\delta$ of the population singular vectors $\bar{U}$ and $\bar{V}$. As expected, larger distances imply more accurate clusters. (\ref{3.3}) and (\ref{3.5}) imply that nodes in $T^y$ and $T^z$ are correctly clustered into the underlying row clusters and column clusters up to permutations, respectively. (\ref{C4}) and (\ref{C5}) are technical conditions that ensure the validity of the results. They actually ensure that each true cluster has nodes that are correctly clustered. These conditions can be easily met. Moreover, when $K^y=K^z$, Lemma \ref{lemma1} implies that the column clusters and the row clusters behave similarly with similar misclustering error bounds. {In Section \ref{relatedwork}, we will discuss the misclustering error bounds and compare them with the state of art in more detail.}

\subsubsection{Performance of RS-SCC in ScBMs}
We first provide the deviation of ${A}^{\rm rs}$ from $P$ in the sense of the spectral norm.
\begin{theorem}\label{rsamappro}
Let ${A}^{\rm rs}$ be the random sampling approximation to $A$ with sampling probability $p$. Suppose (\ref{C2}) holds, then for any $\nu>0$ and $0<p\leq1$, there exist constants $c_5>0$ and $c_6>0$ such that
\begin{align}
\label{3.6}\|{A}^{\rm rs}-P\|_2\leq  c_{5}\, {\rm max}\Big\{\sqrt{\frac{n\alpha_n}{p}},\;\frac{\sqrt{{\rm log} n}}{p},\, {\Delta(n,\alpha_n,p)}\Big\}, \tag{3.6}
\end{align}
where $$\Delta(n,\alpha_n,p):=\sqrt{\frac{n\alpha_n^2}{p}}\Big(1+p^{1/4}\cdot {\rm max} \big(1,\,\sqrt{\frac{1}{p}-1}\big)\Big),$$
with probability larger than $1-2n^{-\nu}-{\rm exp}\Big(-c_6np\big(1+p^{1/4}\cdot {\rm max} (1,\,\sqrt{\frac{1}{p}-1})^2\big)\Big)$.
\end{theorem}

Theorem \ref{rsamappro} says that {${A}^{\rm rs}$ concentrates around $P$ at the rate shown in (\ref{3.6}). As expected, the rate decreases as $p$ increases. Note that (\ref{3.6}) simplifies to $ O((\sqrt{{n\alpha_n}/{p}}),$ provided that $p>1/2$.
}

In what follows, for notational simplicity, we denote $${\Phi(n,p,\alpha_n)}:={\rm max}\Big\{\sqrt{\frac{n\alpha_n}{p}},\;\frac{\sqrt{{\rm log} n}}{p},\, {\Delta(n,\alpha_n,p)}\Big\}.$$ The next theorem provides an upper bound for the misclustering error rates of RS-SCC under ScBMs.

\begin{theorem}\label{rsammis}
Let ${{Y}}^{\rm rs}\in \mathbb M_{n,K^y}$ and ${{Z}}^{\rm rs}\in \mathbb M_{n,K^z}$ be the estimated membership matrices of RS-SCC. Suppose {(\ref{C2})} holds and other parameter settings are identical with those in Theorem \ref{rsamappro}. The following two arguments hold for ${{Y}}^{\rm rs}$ and ${{Z}}^{\rm rs}$, respectively.

(1) If exists an absolute constant $c_7>0$ such that,
\begin{equation}
\label{C6}\frac{{K^y}\Phi^2(n,p,\alpha_n)}{{n^y_k}\tau^2\gamma_n^2}\leq c_7,\tag{C6}
\end{equation}
for any $k=1,...,K^y$, then with probability larger than $1-2n^{-\nu}-{\rm exp}\Big(-c_6np\big(1+p^{1/4}\cdot {\rm max} (1,\,\sqrt{\frac{1}{p}-1})^2\big)\Big)$ for any $\nu>0$, there exist subsets $M^y\in \{1,...,n\}$ satisfying
\begin{equation}
\label{3.7}\frac{|M^y|}{n}\leq c_7^{-1}\frac{{K^y}\Phi^2(n,p,\alpha_n)}{{n}\tau^2\gamma_n^2}.\tag{3.7}
\end{equation}
Moreover, for $T^y=\{1,...,n\}\backslash M^y$, there exists a $K^y\times K^y$ permutation matrix $J^y$ such that
\begin{equation}
\label{3.8}{{Y}}^{\rm rs}_{T^y\ast}J^y=Y_{T^y\ast}.\tag{3.8}
\end{equation}

(2) If there exists an absolute constant $c_8>0$ such that,
\begin{equation}
\label{C7}\frac{{K^y}\Phi^2(n,p,\alpha_n)}{{n^z_k}\delta^2\gamma_n^2}\leq c_8,\tag{C7}
\end{equation}
for any $k=1,...,K^z$, then with probability larger than $1-2n^{-\nu}-{\rm exp}\Big(-c_6np\big(1+p^{1/4}\cdot {\rm max} (1,\,\sqrt{\frac{1}{p}-1})^2\big)\Big)$ for any $\nu>0$, there exist subsets $M^z\in \{1,...,n\}$ such that
\begin{equation}
\label{3.9} \frac{|M^z|}{n}\leq c_8^{-1}\frac{{K^y}\Phi^2(n,p,\alpha_n)}{{n}\delta^2\gamma_n^2}.\tag{3.9}
\end{equation}
Moreover, for $T^z=\{1,...,n\}\backslash M^z$, there exists a $K^z\times K^z$ permutation matrix $J^z$ such that
\begin{equation}
\label{3.10}{{Z}}^{\rm rs}_{T^z\ast}J^z=Z_{T^z\ast}.\tag{3.10}
\end{equation}
\end{theorem}

The proof of Theorem \ref{rsammis} is similar to that of Theorem \ref{rpromis}, hence we omit it. (\ref{3.7}) and (\ref{3.9}) provide upper bounds for the proportion of the misclustered nodes in the estimated row clusters and column clusters, respectively. As in the random projection scheme, the minimum non-zero row distance in the true singular vectors $\bar{U}$ and $\bar{V}$, i.e., $\tau$ and $\delta$, play an important role in the clustering performance. The nodes outside $M^y$ and $M^z$ are correctly clustered up to permutations (see (\ref{3.8}) and (\ref{3.10})). (\ref{C6}) and (\ref{C7}) are technical conditions which ensure the validity of the results. They have the same effect with those of (\ref{C4}) and (\ref{C5}), and they could be achieved with ease as we will discuss in Section \ref{relatedwork}.

\subsection{Degree-corrected stochastic co-block model}
In ScBMs, the nodes within each row cluster and column cluster are stochastic equivalent. While in real networks, there exists hubs whose edges are far more than those of the non-hub nodes~\citep{karrer2011stochastic}. To model such degree heterogeneity, DC-ScBMs introduce extra parameters {$\theta^y=(\theta^y_1,\theta^y_2,\dots,\theta^y_n)^\top\in \mathbb R^n_+$ and $\theta^z=(\theta^z_1,\theta^z_2,\dots,\theta^z_n)^\top\in \mathbb R^n_+$, }which represent the propensity of each node to send and receive edges. We used the following definition for DC-ScBMs.
\begin{definition}[DC-ScBM, \citealp{rohe2016co}]
\label{defdc}
Let $Y\in \mathbb M_{n,K^y}$ and $Z\in \mathbb M_{n,K^z}$ be the row and column membership matrices, respectively. Let $B\in [0,1]^{K^y\times K^z}$ be the connectivity matrix whose $(k,l)$th element is the probability of a directed edge from any node in the row cluster $k$ to any node in the column cluster $l$. Let $\theta^y\in \mathbb R^n$ and $\theta^z\in \mathbb R^n$ be the node propensity parameters. Given $(Y,Z,B, \theta^y, \theta^z)$, each element of the network adjacency matrix $A=(a_{ij})_{1 \leq i,j\leq n}$ is generated independently as $a_{ij}\sim{\rm Bernoulli}(\theta^y_i\theta^z_jB_{g_i^yg_j^z})$ if $i\neq j$, and $a_{ij}=0$ if $i=j$.
\end{definition}

From the above definition, we see that the probability of an edge from node $i$ to $j$ depends on not only the row cluster and column cluster they respectively lie in, but also the propensity of them to send and receive edges, respectively. Note that $\theta^y$ and $\theta^z$ would make the model non-identifiable except that additional assumptions are enforced. In this paper, we assume $\max_{i\in G_k^y}\theta^y_i=1$ and $\max_{i\in G_k^z}\theta^z_i=1$ for each $k=1,...,K^y$ and $k=1,...,K^z$, respectively.

Throughout this subsection, we would consider the DC-ScBM parameterized by $(\theta^y,\theta^z,\\Y,Z,B)$. Note that when $\theta^y$ and $\theta^z$ take 1 as all their entries, then the DC-ScBM reduce to ScBM. With a slight abuse of notation, define $P={\rm diag}(\theta^y)YBZ^\intercal{\rm diag}(\theta^z)$, which is actually the population version of $A$ in the sense that $\mathbb E(A)=P-{\rm diag}(P)$. The formulation of $P$ would be used throughout  this subsection. Assume ${\rm rank}(P)={\rm rank}(B)=K^y$, and denote the maximum and minimum singular values of $P$ by $\sigma_n$ and $\gamma_n$, respectively.

Before analyzing the singular structure of $P$, we now introduce some notations. Let {$\phi_k^y$ and $\phi_k^z$} be $n\times1$ vectors that consistent with $\theta^y$ and $\theta^z$ respectively on $G_k^y$ and $G_k^z$ and zero otherwise. {Thus, $\sum_{k=1}^{K^y}\phi_k^y=\theta^y$ and $\sum_{k=1}^{K^z}\phi_k^z=\theta^z$.} Let $\Psi^y={\rm diag} (\|\phi^y_1\|_2,...,\|\phi^y_{K^y}\|_2)$, $\Psi^z={\rm diag} (\|\phi^z_1\|_2,...,\|\phi^z_{K^z}\|_2)$ and ${\Psi^y {B} \Psi^z=\tilde{B}}$. Define {$\tilde{\theta}^y$ and $\tilde{\theta}^z$} be $n\times 1$ vectors such that their $i$th elements are $\theta^y_i/\|\phi^y_{g_i}\|_2$ and $\theta^z_i/\|\phi^z_{g_i}\|_2$, respectively. The next lemma reveals the singular structure in $P$.

\begin{lemma}
\label{lemma3}
Denote the SVD of the population matrix $P={\rm diag}(\theta^y)YBZ^\intercal{\rm diag}(\theta^z)$ by $\bar{U}_{n\times K^y}\\\bar{\Sigma}_{K^y\times K^y}\bar{V}^\intercal_{K^y\times n}$. Denote the SVD of $\tilde{B}$ by $H_{K^y\times K^y}D_{K^y\times K^y}J_{K^y\times K^z}^\intercal$. For any two vectors $a$ and $b$, ${\rm cos}(a,b)$ is defined to be $a^\intercal b/\|a\|_2\|b\|_2$. Then the following arguments hold for any $1\leq i\neq j\leq  n$.

(1) $\bar{U}_{i\ast}=\tilde{\theta}^y_iH_{k\ast}$ for $i\in G_k^y$, where $H$ is an $K^y\times K^y$ orthonormal matrix. As a result, if $Y_{i\ast}=Y_{j\ast}$, then ${\rm cos}(\bar{U}_{i\ast},\bar{U}_{j\ast})=1$; otherwise ${\rm cos}(\bar{U}_{i\ast},\bar{U}_{j\ast})=0$.

(2) $\bar{V}_{i\ast}=\tilde{\theta}^z_iJ_{k\ast}$ for $i\in G_k^z$, where $J$ is an $K^z\times K^y$ matrix with orthonormal columns. As a result, if $Z_{i\ast}=Z_{j\ast}$, then ${\rm cos}(\bar{V}_{i\ast},\bar{V}_{j\ast})=1$; otherwise $${\rm cos}(\bar{V}_{i\ast},\bar{V}_{j\ast})={\rm cos}((\tilde{B}_{\ast g_i^z})^\intercal {H}\bar{\Sigma}^{-1},(\tilde{B}_{\ast g_j^z})^\intercal {H}^{-1}\bar{\Sigma}^{-1}).$$ Moreover, if there exists a deterministic sequence $\{\xi'_n\}_{n\geq 1}< 1$ such that
\begin{equation}
\label{C8}{\rm cos}((\tilde{B}_{\ast g_i^z})^\intercal {H}\bar{\Sigma}^{-1},(\tilde{B}_{\ast g_j^z})^\intercal {H}\bar{\Sigma}^{-1})\leq \xi'_n,\tag{C8}
\end{equation}
then ${\rm cos}(\bar{V}_{i\ast},\bar{V}_{j\ast})\leq \xi'_n$.
\end{lemma}
The following lemma provides sufficient conditions for (\ref{C8}) to hold.
\begin{lemma}
\label{lemma4}
Under the same parameter setting as in Lemma \ref{lemma3}, if the columns of $\tilde{B}$ are {\rm not} mutually proportional such that there exists a deterministic sequence $\{\zeta_n\}_{n\geq 1}<1$ satisfying
$$\max_{1\leq k\neq l\leq K^z} {\rm cos}(\tilde{B}_{\ast k},\tilde{B}_{\ast l})\leq \zeta_n,$$
and $\underline{\iota}_n \leq \min_{1\leq k \leq K^z}\|\tilde{B}_{\ast k}\|_2 \leq \max_{1\leq k \leq K^z}\|\tilde{B}_{\ast k}\|_2\leq \overline{\iota}_n$, then (\ref{C8}) holds with
$$\xi'_n=\sqrt{1-\left(\frac{\underline{\iota}_n}{\overline{\iota}_n}\cdot\frac{\sigma_{\min}(H)}{\sigma_{n}(H)}\cdot\frac{\sigma_{\min}(P)}{\sigma_{n}(P)}\right)^2(1-\zeta_n)^2},$$
\end{lemma}
where $\sigma_{\rm min}(\cdot)$ and $\sigma_n(\cdot)$ denote the minimum and maximum non-zero singular value of matrices.

Lemma \ref{lemma3} says that {the directions of two rows in $\bar{U}$ ($\bar{V}$) are the same if and only if the corresponding nodes lie in the same row (column) cluster. For example, if node $i$ and node $j$ are in the same row cluster $k$, then $\bar{U}_{i\ast}$ and $\bar{U}_{i\ast}$ both have direction $H_{k\ast}$.} On the other hand, if two nodes are in different row (column) clusters, there exists an angle between the corresponding rows of $\bar{U}$ ($\bar{V}$). In particular, two rows of $\bar{U}$ are perpendicular if the corresponding nodes lie in different row clusters. While on the column's side, if two nodes are in different column clusters, then the angle between the corresponding rows of $\bar{V}$ depends generally on the angles between the indicated rows in a ``normalized'' connectivity matrix $\tilde{B}^\intercal H\bar{\Sigma}^{-1}$. In Lemma \ref{lemma4}, we provide understandable and reasonable conditions which lead to an explicit bound of the angle. Except for these facts, Lemma \ref{lemma3} essentially explains why a normalization step is needed in Algorithm \ref{spectralmedian} before the $k$-means. It is well-known that $k$-means clusters nodes together if they are close in the sense of Euclidean distance. The normalization step forces any two rows of $\bar{U}$ or $\bar{V}$ to lie in the same position if the corresponding nodes are in the same row cluster or column cluster. In such a way, the $k$-means could succeed when applied to the sample version singular vectors.

In Theorem \ref{rproappro} and \ref{rsamappro}, we have proved that the randomized adjacency matrices ${A}^{\rm rp}$ and ${A}^{\rm rs}$ concentrate around the population $P$ under the ScBMs, where we actually did not make use of the explicit structure of $P$ but only the facts that $P$ is the population of $A$, and $P$ is of rank $K^y$. Hence the same results hold here for the DC-ScBMs. Next, we use these results combining with Lemma \ref{lemma3} to analyze the clustering performance of RP-SsCC and RS-SsCC.

\subsubsection{Performance of RP-SsCC in DC-ScBMs}
The next theorem quantifies the clustering performance of RP-SsCC.

\begin{theorem}\label{rpromisdegree}
Let ${{Y}}^{\rm rp}\in \mathbb M_{n,K^y}$ and ${{Z}}^{\rm rp}\in \mathbb M_{n,K^z}$ be the estimated membership matrices of RP-SsCC. The other parameters are the same with those in Theorem \ref{rproappro}. Suppose {(\ref{C2})} and (\ref{C3}) hold and recall that the minimum and maximum non-zero singular value of $P$ are $\gamma_n$ and $\sigma_n$, respectively. The following two arguments hold for ${{Y}}^{\rm rp}$ and ${{Z}}^{\rm rp}$, respectively.

(1) Define $${\kappa^y:=\max_i\;(\tilde{\theta}_i^y)^{-2}}.$$
If there exists an absolute constant $c_9>0$ such that,
\begin{equation}
\label{C9}\frac{n\kappa^y K^y \alpha_n}{\gamma_n^2{n^y_k}}\leq c_9,\tag{C9}
\end{equation}
for any $k=1,...,K^y$, then with probability larger than $1-6r^{-r}-6 s^{-s}-3n^{-\epsilon}$ for any $\epsilon>0$ there exist subsets $M^y\in\{1,...,n\}$ satisfying
\begin{equation}
\label{3.11}\frac{|M^y|}{n}\leq c_9^{-1}\frac{\kappa^y K^y\alpha_n}{\gamma_n^2}.\tag{3.11}
\end{equation}
Moreover, for $T^y=\{1,...,n\}\backslash M^y$, there exists a $K^y\times K^y$ permutation matrix $J^y$ such that
\begin{equation}
\label{3.12}{{Y}}^{\rm rp}_{T^y\ast}J^y=Y_{T^y\ast}.\tag{3.12}
\end{equation}

(2) Define $$\kappa^z:=\max_i \;(\tilde{\theta}_i^z)^{-2}\|(\tilde{B}_{\ast g_i})^\intercal H\bar{\Sigma}^{-1}\|_2^{-2},$$
and
$$\eta(P)=\underset{k\neq l}{\rm max}\;{\rm cos}(\tilde{B}_{\ast k})^\intercal H{\bar{\Sigma}}^{-1},(\tilde{B}_{\ast l})^\intercal H\bar{\Sigma}^{-1}),$$
where recall that $H$ is the left singular matrix of $\tilde{B}$.
If there exists an absolute constant $c_{10}>0$ such that,
\begin{equation}
\label{C10}\frac{n\kappa^z K^y \alpha_n}{{(1-\eta(P))}\gamma_n^2 n^z_k}\leq c_{10},\tag{C10}
\end{equation}
for any $k=1,...,K^z$, then with probability larger than $1-6r^{-r}-6 s^{-s}-3n^{-\epsilon}$ for any $\epsilon>0$ there exist subsets $M^z\in \{1,...,n\}$ satisfying
\begin{equation}
\label{3.13}\frac{|M^z|}{n}\leq c_{10}^{-1}\frac{\kappa^z K^y \alpha_n}{{(1-\eta(P))}\gamma_n^2}.\tag{3.13}
\end{equation}
Moreover, for $T^z=\{1,...,n\}\backslash M^z$, there exists a $K^z\times K^z$ permutation matrix $J^z$ such that
\begin{equation}
\label{3.14}{{Z}}^{\rm rp}_{T^z\ast}J^z=Z_{T^z\ast}.\tag{3.14}
\end{equation}
\end{theorem}

The quantity $\kappa_y$ ($\kappa_z$) can be thought of as the maximum node heterogeneity in sending (receiving) edges across all row (column) clusters, respectively. The quantity $\eta(P)$ indicates the minimum non-zero angles among the rows of the population singular vectors $\bar{V}$ (see (2) of Lemma \ref{lemma3}), valued by cosine. (\ref{3.12}) and (\ref{3.13}) provide upper bounds for the proportion of the misclustered nodes with respect to row and column clusters, respectively. It can be seen that larger node degree heterogeneity may lead to poorer clustering performance. And different from the row clusters, the performance of the estimated column clusters additionally depend on $\eta(P)$. As expected, smaller $\eta(P)$ indicates better clustering performance. (\ref{3.12}) and (\ref{3.14}) indicate that nodes lying in $T^y$ and $T^z$ are correctly clustered into the underlying row clusters and column clusters up to some permutations. (\ref{C9}) and (\ref{C10}) are conditions ensuring that each true cluster has nodes that are correctly clustered.

\subsubsection{Performance of RS-SsCC in DC-ScBMs}
The next theorem reflects the clustering performance of RS-SsCC.
\begin{theorem}\label{rsammisdegree}
Let ${{Y}}^{\rm rs}\in \mathbb M_{n,K^y}$ and $\hat{{Z}}^{\rm rs}\in \mathbb M_{n,K^z}$ be the estimated membership matrices of RS-SsCC. The other parameters are the same with those in Theorem \ref{rsamappro}. Suppose {(\ref{C2})} holds and recall that the minimum non-zero singular value of $P$ is $\gamma_n$. Recall that $${\Phi(n,p,\alpha_n)}:={\rm max}\Big\{\sqrt{\frac{n\alpha_n}{p}},\;\frac{\sqrt{{\rm log} n}}{p},\, {\Delta(n,\alpha_n,p)}\Big\},$$
where $$\Delta(n,\alpha_n,p):=\sqrt{\frac{n\alpha_n^2}{p}}\Big(1+p^{1/4}\cdot {\rm max} \big(1,\,\sqrt{\frac{1}{p}-1}\big)\Big).$$
Then the following two results hold for ${{Y}}^{\rm rs}$ and ${{Z}}^{\rm rs}$, respectively.

(1) If there exists an absolute constant $c_{11}>0$ such that,
\begin{equation}
\label{C11}\frac{\kappa^y K^y \Phi^2(n,\alpha_n,p)}{\gamma_n^2 n_k^y}\leq c_{11},\tag{C11}
\end{equation}
for any $k=1,...,K^y$, then with probability larger than $1-2n^{-\nu}-{\rm exp}\Big(-c_6np\big(1+p^{1/4}\cdot {\rm max} (1,\,\sqrt{\frac{1}{p}-1})^2\big)\Big)$ for any $\nu>0$, there exist subsets $M^y\in \{1,...,n\}$ for $k=1,...,K^y$ satisfying
\begin{equation}
\label{3.15}\frac{|M^y|}{n}\leq c_{11}^{-1}\frac{\kappa^yK^y\Phi^2(n,\alpha_n,p)}{\gamma_n^2{n}}.\tag{3.15}
\end{equation}
Moreover, for $T^y=\{1,...,n\}\backslash M^y$, there exists a $K^y\times K^y$ permutation matrix $J^y$ such that
\begin{equation}
\label{3.16}{{Y}}^{\rm rs}_{T^y\ast}J^y=Y_{T^y\ast}.\tag{3.16}
\end{equation}

(2) There exists an absolute constant $c_{12}>0$ such that, if
\begin{equation}
\label{C12}\frac{\kappa^zK^y\Phi^2(n,\alpha_n,p)}{({1-\eta(P)})\gamma_n^2{n^z_k}}\leq c_{12},\tag{C12}
\end{equation}
for any $k=1,...,K^z$, then with probability larger than $1-2n^{-\nu}-{\rm exp}\Big(-c_6np\big(1+p^{1/4}\cdot {\rm max} (1,\,\sqrt{\frac{1}{p}-1})^2\big)\Big)$ for any $\nu>0$, there exist subsets $M^z\in \{1,...,n\}$ for $k=1,...,K^z$ such that
\begin{equation}
\label{3.17}\frac{|M^z|}{n}\leq c_{12}^{-1}\frac{\kappa^zK^y\Phi^2(n,\alpha_n,p)}{{(1-\eta(P))}\gamma_n^2{n}}.\tag{3.17}
\end{equation}
And for $T^z=\{1,...,n\}\backslash M^z$, there exists a $K^z\times K^z$ permutation matrix $J^z$ such that
\begin{equation}
\label{3.18}{{Z}}^{\rm rs}_{T^z\ast}J^z=Z_{T^z\ast}.\tag{3.18}
\end{equation}
\end{theorem}

We omit the proof of Theorem \ref{rsammisdegree} since it is similar to that of Theorem \ref{rpromisdegree}. (\ref{3.15}) and (\ref{3.17}) provide upper bounds for the proportion of misclustered nodes with respect to the row clusters and column clusters, respectively.  Similar to the results of the random projection paradigm, the clustering performance depends on the degree heterogeneity. And for the column cluster, it additionally depends on the minimum non-zero angles among the rows in the population singular vector $\bar{V}$. (\ref{C11}), (\ref{C12}), (\ref{3.16}) and (\ref{3.18}) have the same effect and meaning with those in Theorem \ref{rpromisdegree}.

\section{Discussions}
\label{relatedwork}
We discuss the theoretical aspects of the proposed randomized methods.

\paragraph{On the misclustering error rates and comparison with the state of the art.} Note that the misclustering error rates depend on one unknown parameter, namely, the minimum non-zero singular value of the population matrix. Here we specify the bounds and compare them with the state of art by consider the following four-parameter ScBM. The underlying number of row clusters and column clusters are the same, i.e., $K^y=K^z=K$. Each cluster has a balanced size $n/K$. For any pair of nodes $(i,j)$, a directed edge from $i$ to $j$ is generated with probability $\alpha_n$ if the row cluster of $i$ is identical to the column cluster of $j$, and with probability $\alpha_n(1-\lambda)$ otherwise. Formally,
\begin{equation}
P=YBZ^\intercal=Y(\alpha_n\lambda I_K+\alpha_n(1-\lambda)1_K1_K^\intercal)Z^\intercal.\nonumber
\end{equation}
In this case, the minimum non-zero singular value of $P$ is $n\alpha_n \lambda/K$ \citep{rohe2011spectral}, the row and column clusters has the same misclustering error rates, and there is no degree heterogeneity. In what follows, we examine the misclustering rate in Theorem \ref{rpromis} (\ref{rpromisdegree}) and \ref{rsammis} (\ref{rsammisdegree}), respectively. The misclustering error rates mentioned below have been summarized in Table \ref{tabletheory}.
\begin{itemize}
    \item The bound in (\ref{3.2}) reduces to ${O(K^2/(n\alpha_n))}$ under the four-parameter ScBM. If $\alpha_n=O({\rm log }n/n)$, then $O(K^2/(n\alpha_n))$ vanishes as $n$ increases provided that $K=o(\sqrt{{\rm log} n})$ and (\ref{C4}) is automatically satisfied. While in \citet{rohe2012co} (see Corollary 4.1 therein), the misclustering rate is $o(K^3{\rm log} n/\alpha_n^4)$, and $K=O(n^{1/4}/{\rm log} n)$ is required to make the results hold. In addition, in \citet{rohe2016co} (see Corollary C.1 therein), the misclustering rate is $O(K^2{\rm log} n/n)$ provided that $\alpha_n$ is fixed, which is not better than the $O(K^2/n)$ in our case.
   \item The bound in (\ref{3.7}) simplifies to $O(K^2/(pn\alpha_n))$ provided that $p>1/2$, which is tighter than those in \citet{rohe2012co} and \citet{rohe2016co}.
\end{itemize}
The major reason why the randomized algorithms lead to even better misclustering error rates than the non-randomized algorithms do in \citet{rohe2016co} and \citet{rohe2012co} is that we derive the approximation bounds of $\|A^{\tiny{\rm rp}}-P\|_2$ and $\|A^{\tiny{\rm rs}}-P\|_2$ on the basis of the tightest concentration bound of $\|A-P\|_2$ \citep{lei2015consistency,chin2015stochastic}, which was originally developed using  combinatorial arguments \citep{feige2005spectral}.

\paragraph{On the information-theoretic threshold and the optimal misclustering rate.} As we have mentioned, the approximation error rates of the randomized adjacency matrices achieve the minimax optimal rates established in \citet{zhang2016minimax}. We discuss the misclustering error rate in what follows.
It should be noted that the original spectral (co-)clustering generally could not attain the minimax optimal misclustering error rate under SBMs \citep{zhang2016minimax} except that the number of clusters $K=2$ \citep{abbe2020entrywise}, though \citet{gao2017achieving} developed a refined spectral clustering algorithm which could attain the optimal rates but with time complexity $O(n^3)$. Hence our randomized version inherits such limitations when $K>2$. Next we examine the case $K=2$ in more detail.

Consider the following two blocked symmetric ScBM (i.e., SBM) ${\rm SBM}(n,a\frac{{\rm log} n}{n},b\frac{{\rm log} n}{n}, J)$. Let $n$ be even, $J\subseteq [n]$ with $|J|=n/2$. Each entry $A_{ij}(i<j)$ of the symmetric adjacency matrix $A=(A_{ij})$ is independently generated as $\mathbb P(A_{ij}=1)=a\frac{{\rm log} n}{n}$ if $i\sim j$, and $\mathbb P(A_{ij}=1)=b\frac{{\rm log} n}{n}$ otherwise, where $i\sim j$ means that $i\in J,j\in J$ or $i\in J^c,j\in J^c$.
Under this regime, the population adjacency matrix $P$ is a rank-2 matrix with two nonzero eigenvalues $\lambda_1^\ast =(a+b){\rm log}n/2$ and $\lambda_2^\ast =(a-b){\rm log}n/2$ whose corresponding eigenvectors are
$u^\ast_1=\frac{1}{\sqrt{n}}\mathbb I_n$ and $u^\ast_2=\frac{1}{\sqrt{n}}\mathbb I_J-\frac{1}{\sqrt{n}}\mathbb I_J^c$.

Let $z\in\mathbb R^n$ be the true labels, specifically, $z_i=1$ if $i\in J$ and $z_i=-1$ if $i\in J^c$.
Then, the clustering aims to estimate the unknown $z$ by $\hat{z}\in \mathbb R^n$. Under the ${\rm SBM}(n,a\frac{{\rm log} n}{n},b\frac{{\rm log} n}{n}, J)$, \citet{abbe2015exact} and \citet{mossel2015consistency} proved that exact recovery ($\hat{z}$ equal to $z$ or $-z$ with probability tending to 1) is information-theoretically possible if and only if $\sqrt{a}-\sqrt{b}>\sqrt{2}$. On the other hand, when the exact recovery is impossible, that is $\sqrt{a}-\sqrt{b}\in(0,\sqrt{2}]$, \citet{zhang2016minimax} provided the following minimax misclustering error rate \citep{abbe2020entrywise}
$$\inf_{\hat{z}}\sup \mathbb E r(\hat{z},z)=\exp \left(-(1+o(1))\cdot(\sqrt{a}-\sqrt{b})^2\frac{{\rm log} n}{2}\right), $$
where the misclustering rate $r(\hat{z},z)$ is defined as
\begin{equation}
\label{4.1}
r(\hat{z},z)=\min_{s\in\{\pm 1\}}n^{-1}\sum_{i=1}^n\mathbb I _{\hat{z}_i\neq sz_i}.\nonumber
\tag{4.1}
\end{equation}
By developing technical tools for entry-wise perturbation bound of eigenvectors, \citet{abbe2020entrywise} proved that the vanilla spectral method based on the adjacency matrix (first compute $u_2$, the eigenvector of $A$ corresponding to its second largest eigenvalue $\lambda_2$; then set $\hat{z}={\rm sgn} (u_2)$) can achieve exact recovery when it is information-theoretic possible, and can attain the optimal minimax misclutering rate otherwise.

To see whether such optimal results could be obtained or how much distortion would be produced by the randomized spectral clustering algorithms. We consider the vanilla spectral method based on the sparisified adjacency matrix $A^{\rm rs}$ (see (\ref{2.1})). With a slight abuse of notation, let $u_2$ be the eigenvector of $A^{\rm rs}$ associated with its second largest $\lambda_2$, and the estimated labels are obtained by $\hat{z}={\rm sgn}(u_2)$. We show in the following theorem that the vanilla spectral method based on $A^{\rm rs}$ could yield near-optimal results while bringing some distortion, which is caused by the random sampling of $A$. In particular, the exact recovery succeeds when $\sqrt{a}-\sqrt{b}>\sqrt{2/p}$, which is a little more stringent than the information-theoretic threshold $\sqrt{2}$. Moreover, when $\sqrt{a}-\sqrt{b}\in(0,\sqrt{2/p}]$, the misclustering error rate turns out to be $n^{-(1+o(1))(p(\sqrt{a}-\sqrt{b})^2/2)}$, which is also slightly inferior to the minimax optimal rate $n^{-(1+o(1))((\sqrt{a}-\sqrt{b})^2/2)}$.

\begin{theorem}
\label{limits}
(1) If $\sqrt{a}-\sqrt{b}>\sqrt{2/p}$, then there exists $\eta=\eta(a,b,p)>0$ and $s\in\{\pm 1\}$ such that with probability $1-o(1)$,
$$\sqrt{n}\min_{i\in [n]} s z_i(u_2)_i\geq \eta.$$
As a consequence, the vanilla spectral method based on the sparsified matrix $A^{\rm rs}$ achieves exact recovery.

(2) Let the misclustering rate $r(\hat{z},z)$ be defined in (\ref{4.1}). If $\sqrt{a}-\sqrt{b}\in (0,\sqrt{2/p}]$, then
$$\mathbb Er(\hat{z},z)\leq n^{-(1+o(1))(p(\sqrt{a}-\sqrt{b})^2/2)}.$$
\end{theorem}

\paragraph{On the extensions to rank-deficient ScBMs.} In this work, we mainly consider the ScBMs (DC-SsBMs) with $B_{K^y\times K^z}$ being of full row rank (recall that $K^y\leq K^z$), where the coupling of the target rank and the smaller target cluster size would make the result nice and interpretable. In real applications, however, the number of clusters may be larger than the dimension \citep{tang2017asymptotically}. We argue that all the results could be generated to rank-deficient case. We provide in the following theorem some inspirations.
\begin{theorem}
\label{lemmarank}
Consider an ScBM parameterized by $(Y,Z,B)$, where $B$ is rank deficient with ${\rm rank}(B)=K'< K^y$. $P=YBZ^\intercal$ is the population adjacency matrix with its SVD being $\bar{U}_{n\times K'}\bar{\Sigma}_{K'\times K'}\bar{V}^\intercal_{K'\times n}$. Define $\Delta_y={\rm diag}(\sqrt{n_1^y},...,\sqrt{n_{K^y}^y})$, $\Delta_z={\rm diag}(\sqrt{n_1^z},...,\sqrt{n_{K^z}^z})$, $\bar{B}=B\Delta_z$, and denote the SVD of $\Delta_yB\Delta_z$ by $L_{K^y\times K'}D_{K'\times K'}R^\intercal_{K'\times K^z}$. If $Y_{i\ast}=Y_{j\ast}$, then $\bar{U}_{i\ast}=\bar{U}_{j\ast}$; otherwise
if the rows of $\bar{B}$ are mutually distinct such that
$$\min_{1\leq k\neq l\leq K^y}\|\bar{B}_{k\ast}-\bar{B}_{l\ast}\|_2\geq \nu_n,$$
and $0<\bar{\Sigma}_{ii}\leq  \mu_n$, then
$$\|\bar{U}_{i\ast}-\bar{U}_{j\ast}\|_2=\left\Vert\frac{L_{g_{i}^y\ast}}{\sqrt{n_{g_i^y}^y}}-\frac{L_{g_{j}^y\ast}}{\sqrt{n_{g_j^y}^y}}\right
\Vert_2\geq \frac{\nu_n}{\mu_n}.$$
\end{theorem}
Theorem \ref{lemmarank} shows that the population left singular vectors are well-separated provided that the rows of $\bar{B}$ are mutually distinct. Therefore, the analysis on the clustering performance naturally follows with extra conditions and notations.

\section{Numerical studies}
\label{experiment}
We evaluate the finite sample performance of RP-SCC (RP-SsCC) and RS-SCC (RS-SsCC), and compare them with SCC (SsCC).

\begin{figure}[h]{}
\centering
\subfigure[Sending clusters]{\includegraphics[height=1.8cm,width=5cm,angle=0]{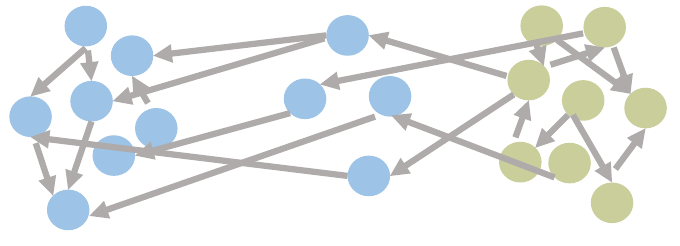}}\hspace{2cm}
\subfigure[Receiving clusters]{\includegraphics[height=1.8cm,width=5cm,angle=0]{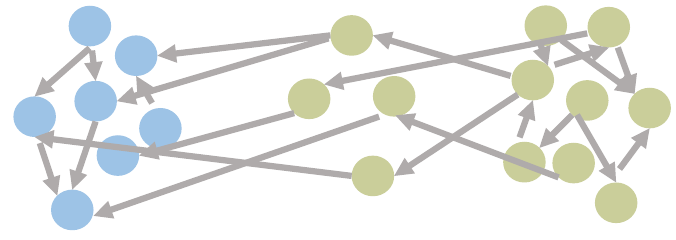}}\\
\caption{Illustration for networks under model set-up 1. Colors indicate clusters.}\label{m1network}
\end{figure}

\begin{figure}[h]{}
\centering
\subfigure[Sending clusters]{\includegraphics[height=4cm,width=4cm,angle=0]{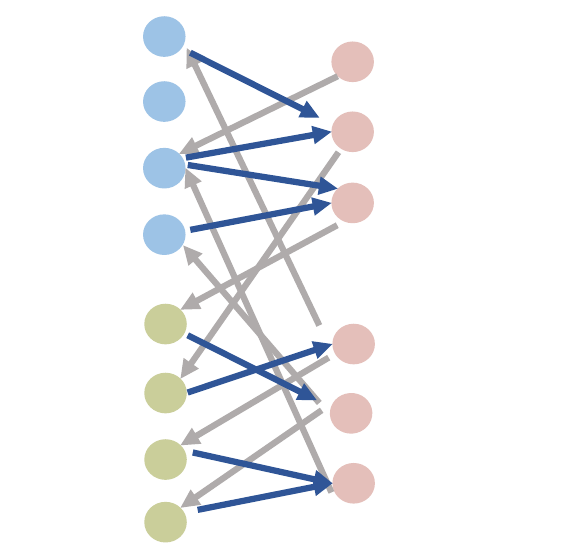}}\hspace{1cm}
\subfigure[Receiving clusters]{\includegraphics[height=3.8cm,width=3.8cm,angle=0]{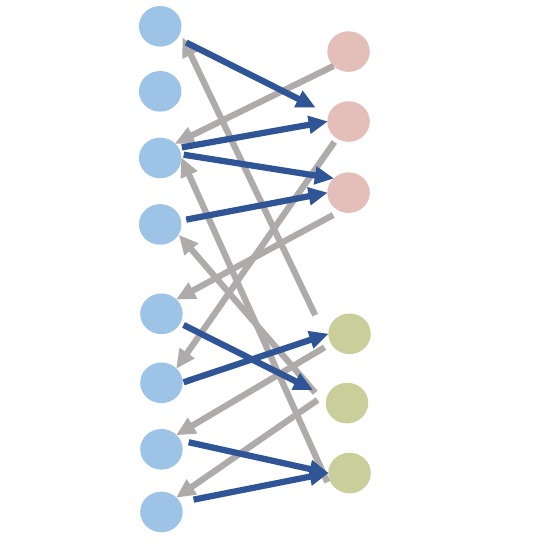}}\\
\caption{Illustration for networks under model set-up 2. Colors indicate clusters.}\label{m2network}
\end{figure}

\begin{figure}[h]{}
\centering
\subfigure[Sending clusters]{\includegraphics[height=3.3cm,width=3.5cm,angle=0]{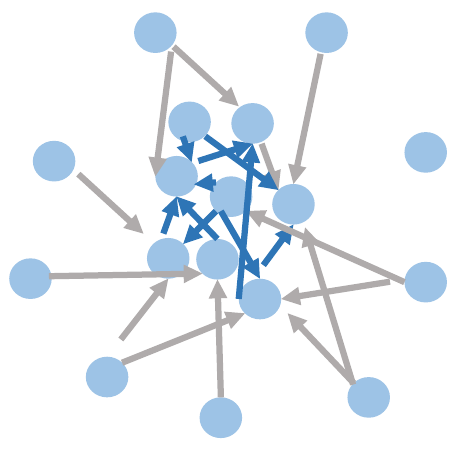}}\hspace{2cm}
\subfigure[Receiving clusters]{\includegraphics[height=3.3cm,width=3.5cm,angle=0]{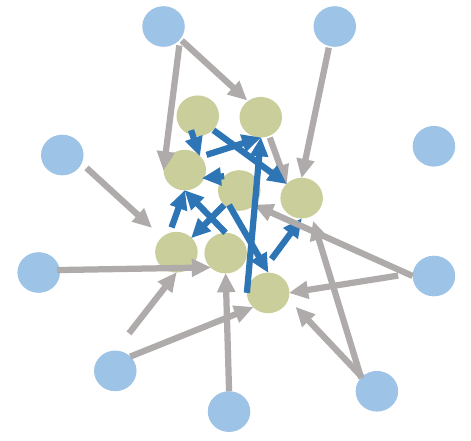}}\\
\caption{Illustration for networks under model set-up 3. Colors indicate clusters.}\label{m3network}
\end{figure}

\begin{figure}[h]{}
\centering
\subfigure[Sending clusters]{\includegraphics[height=3.5cm,width=3.5cm,angle=0]{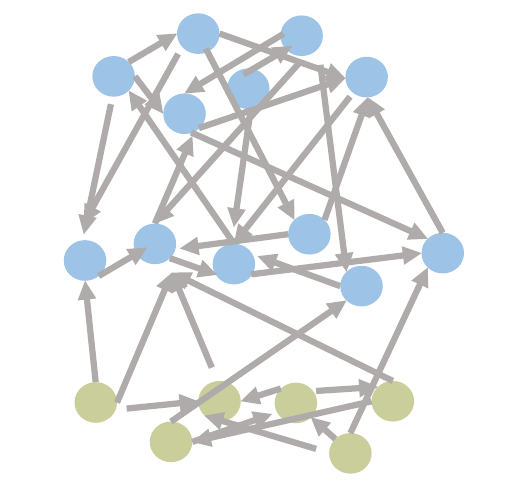}}\hspace{1.4cm}
\subfigure[Receiving clusters]{\includegraphics[height=3.5cm,width=3.5cm,angle=0]{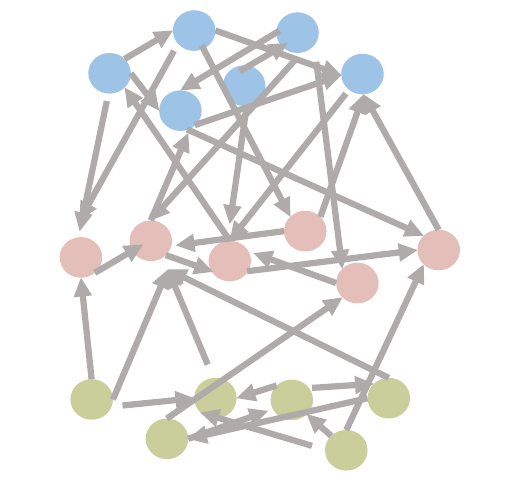}}\\
\caption{Illustration for networks under model set-up 4. Colors indicate clusters.}\label{m4network}
\end{figure}

\begin{figure}[h]{}
\centering
{\includegraphics[height=3.7cm,width=4cm,angle=0]{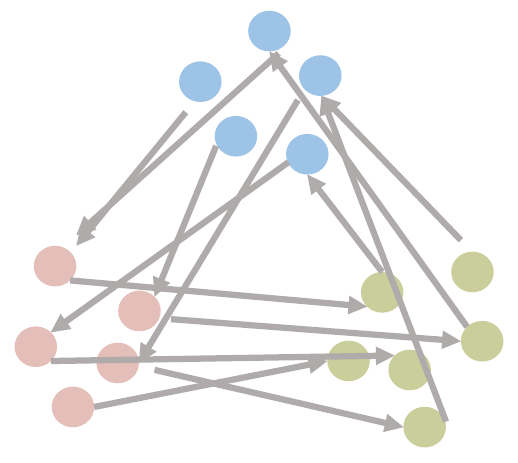}}\\
\caption{Illustration for networks under model set-up 5. Sending and receiving clusters are the same. Colors indicate clusters. }\label{m5network}
\end{figure}

\begin{figure}[h]{}
\centering
{\includegraphics[height=4cm,width=4cm,angle=0]{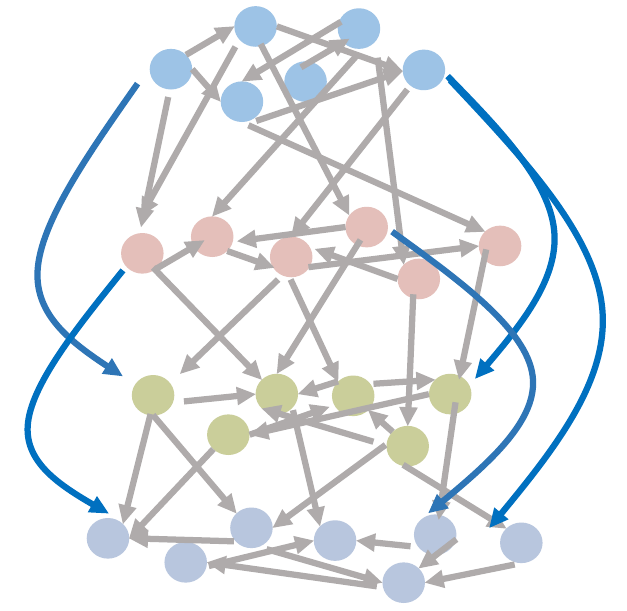}}\\
\caption{Illustration for networks under model set-up 6. Sending and receiving clusters are the same. Colors indicate clusters. }\label{m6network}
\end{figure}

\begin{figure}[!htbp]{}
\centering
\subfigure[Model set-up 1]{\includegraphics[height=4.1cm,width=4.1cm,angle=0]{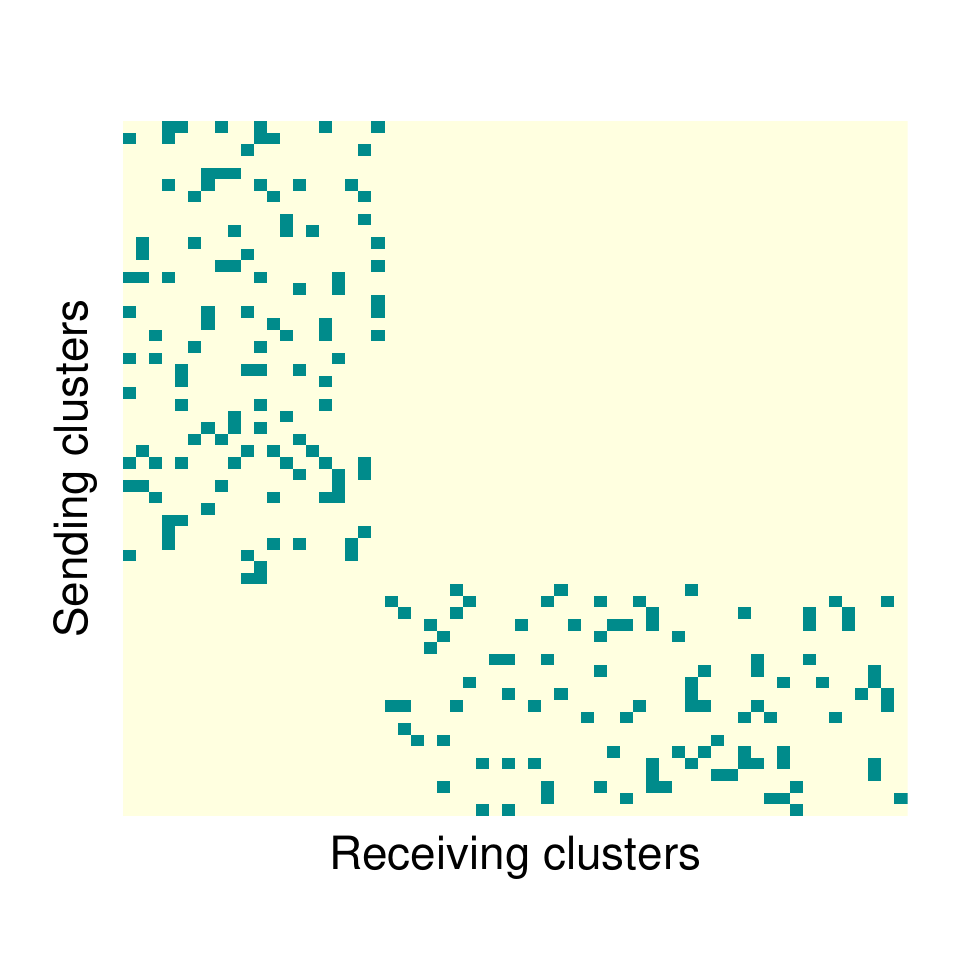}}
\subfigure[Model set-up 2]{\includegraphics[height=4.1cm,width=4.1cm,angle=0]{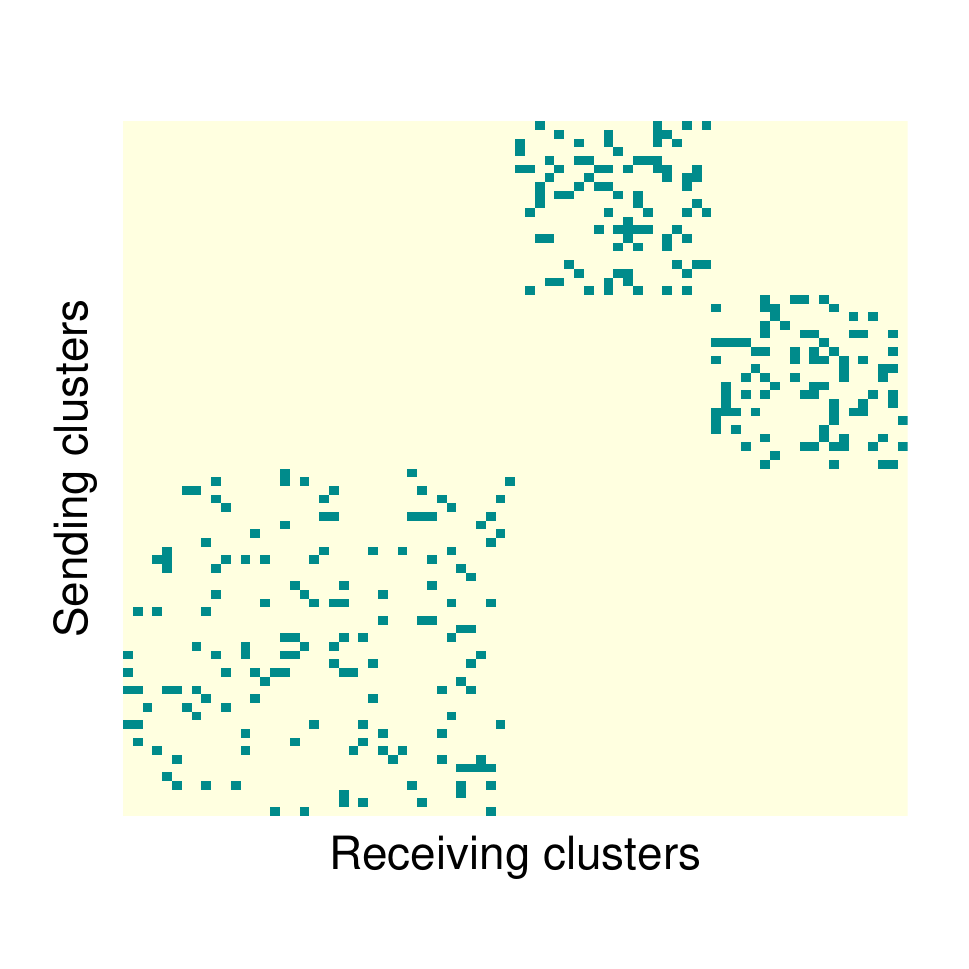}}
\subfigure[Model set-up 3]{\includegraphics[height=4.1cm,width=4.1cm,angle=0]{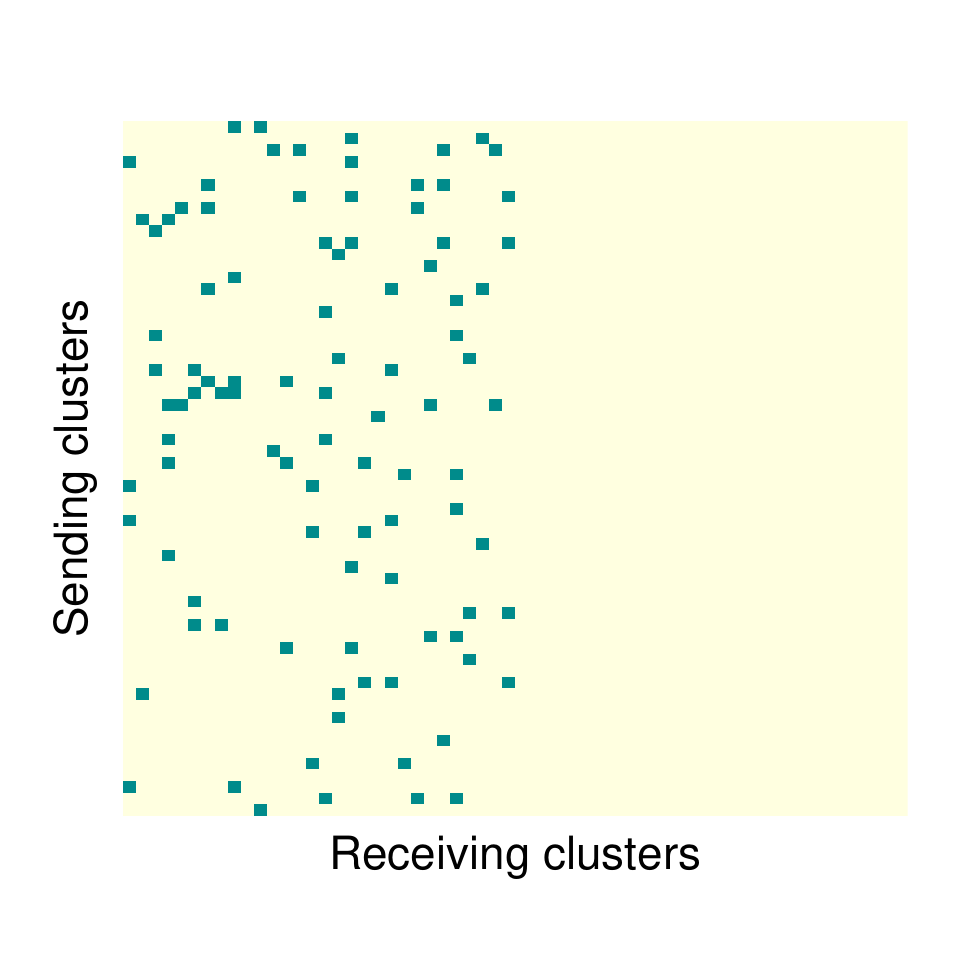}}\\
\subfigure[Model set-up 4]{\includegraphics[height=4.1cm,width=4.1cm,angle=0]{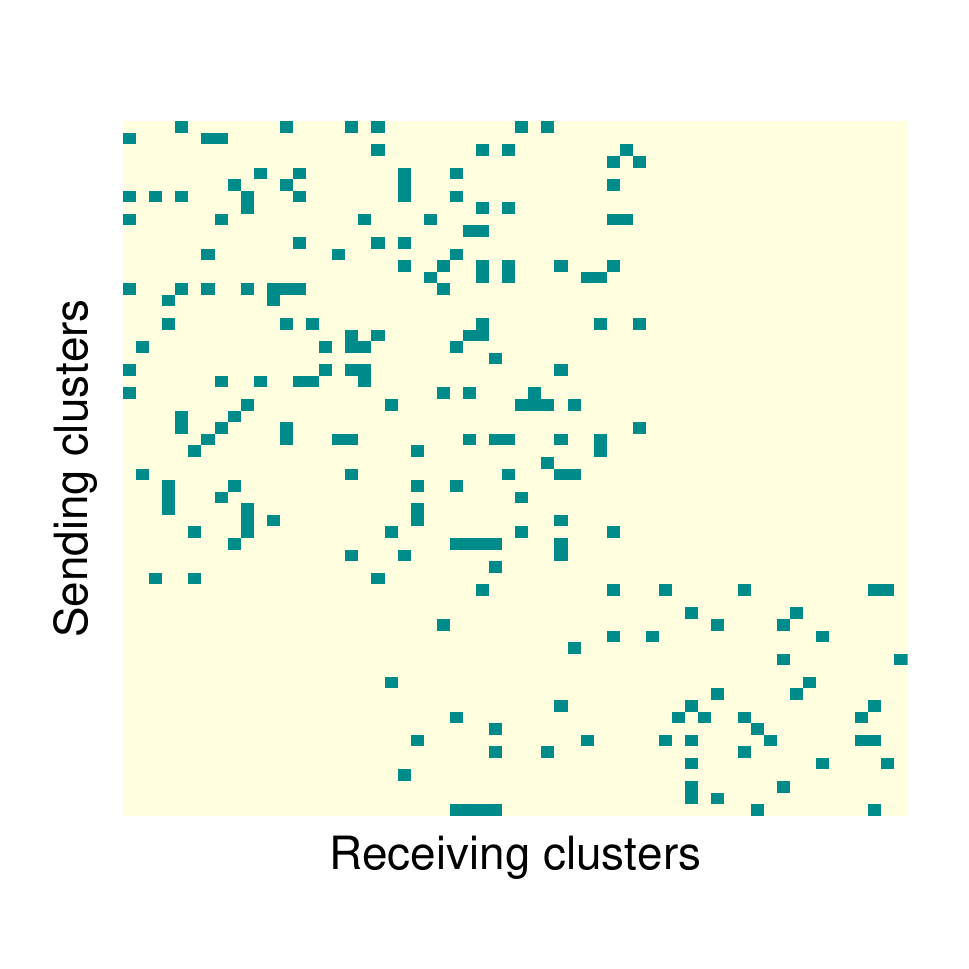}}
\subfigure[Model set-up 5]{\includegraphics[height=4.1cm,width=4.1cm,angle=0]{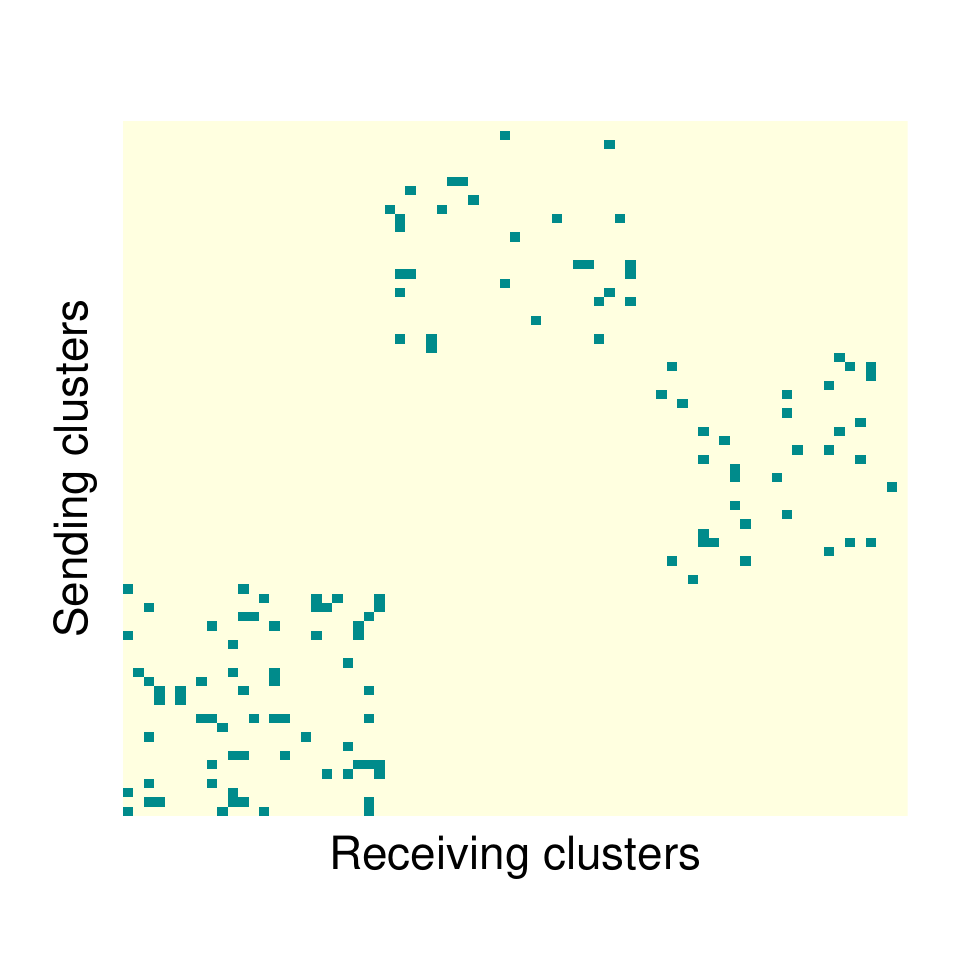}}
\subfigure[Model set-up 6]{\includegraphics[height=4.1cm,width=4.1cm,angle=0]{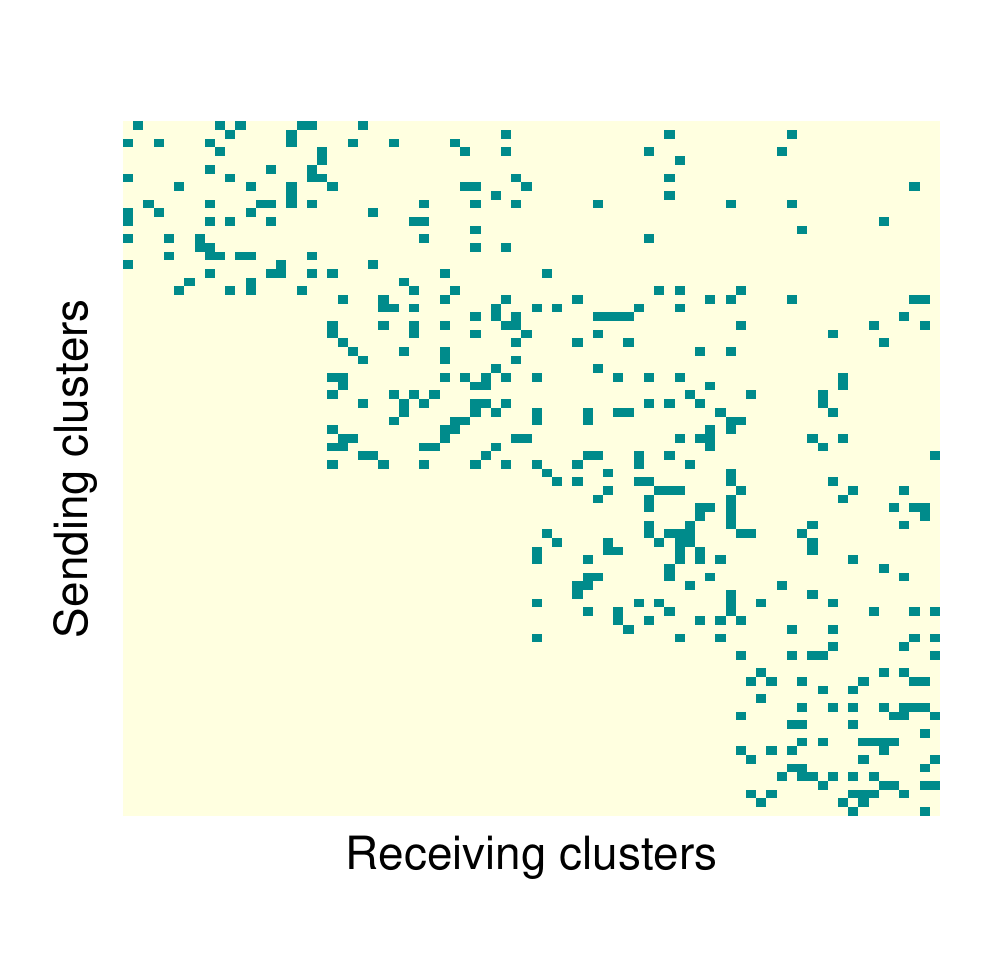}}
\subfigure[Model set-up 7]{\includegraphics[height=4.1cm,width=4.1cm,angle=0]{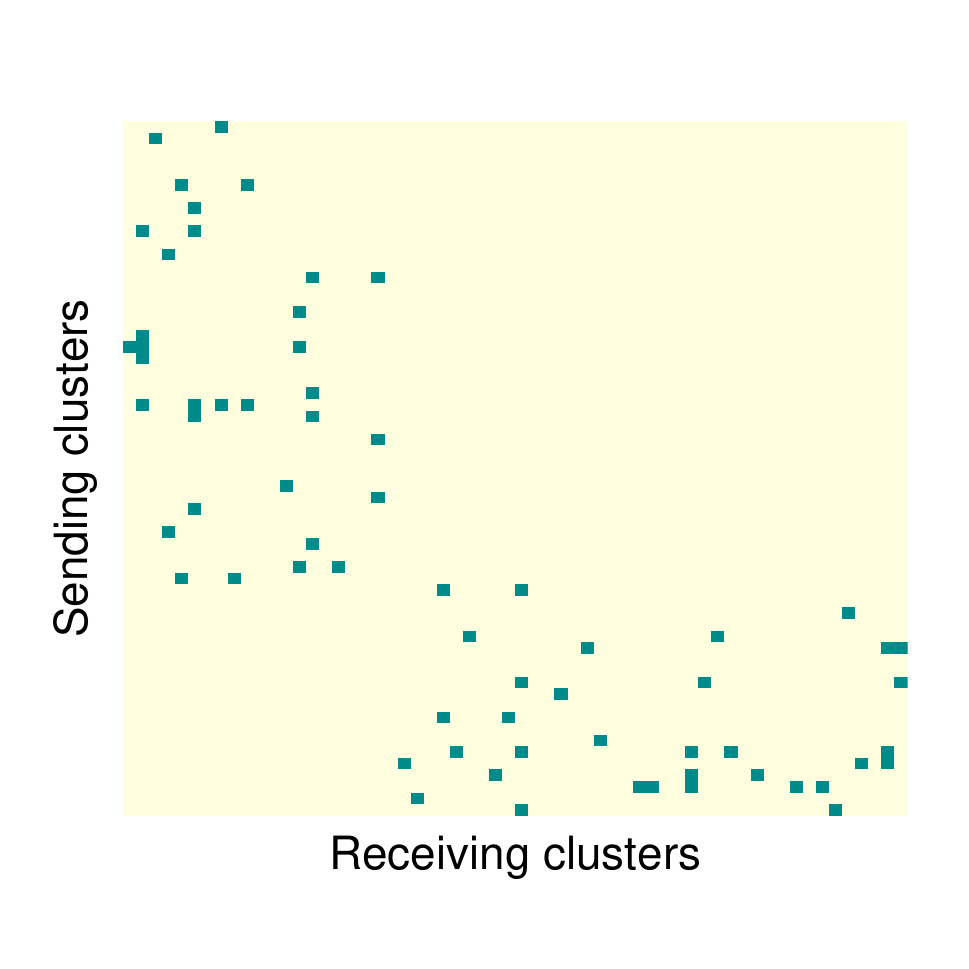}}
\subfigure[Model set-up 8]{\includegraphics[height=4.1cm,width=4.1cm,angle=0]{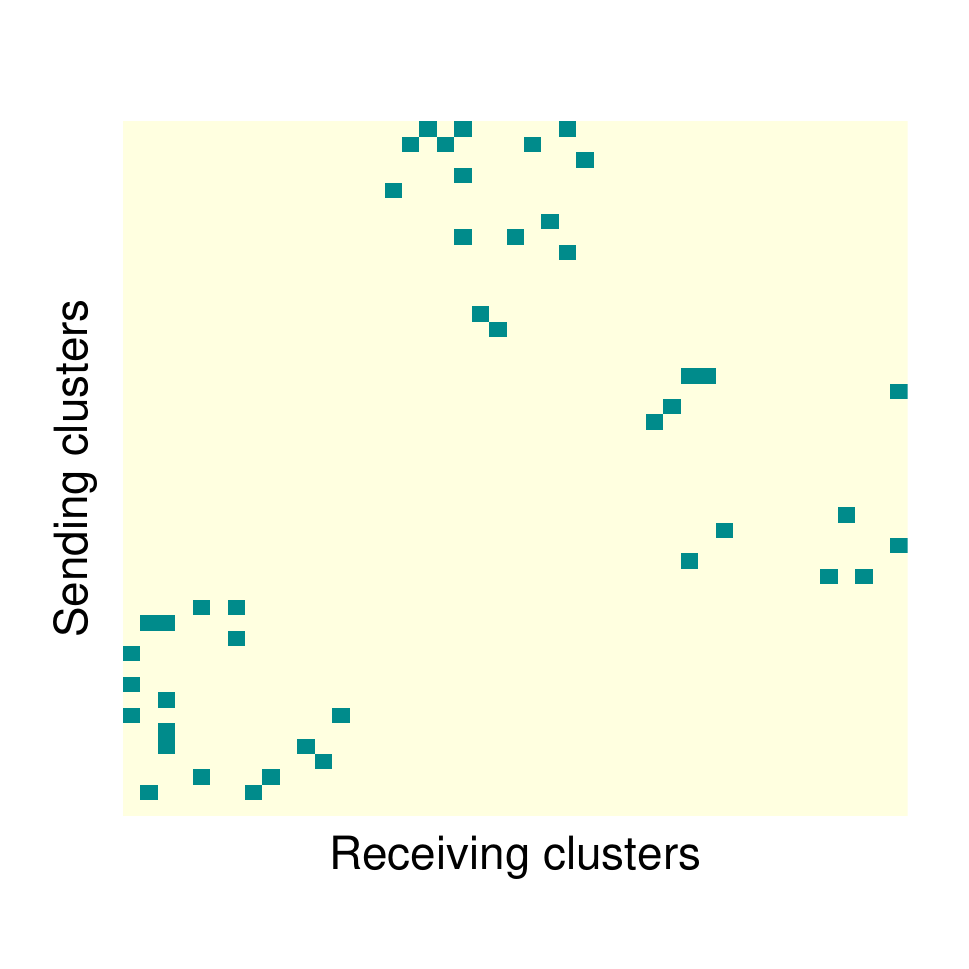}}
\caption{Matrix representation of eight model set-ups considered in simulations. Darker entries are 1's and lighter entries are 0's. The column-wise and row-wise block structures reveal the sending and receiving clusters, respectively.  }\label{matrix}
\end{figure}

In accordance with the theoretical results, we use the following two measures to examine the empirical performance of the three methods. The first is the approximation error, defined by $\|\tilde{A}-P\|_2$, where $\tilde{A}$ can be $A$, ${A}^{\rm \tiny rp}$, or ${A}^{\rm \tiny rs}$. The second is the misclustering error rate with respect to the row clusters and column clusters, defined by ${\rm min}_{J\in E_{K^y}} \frac{1}{2n}\|\tilde{Y}J-Y\|_0$ and ${\rm min}_{J\in E_{K^z}} \frac{1}{2n}\|\tilde{Z}J-Z\|_0$, respectively, where $J$ stands for the permutation matrix, $\tilde{Y}$ can be ${Y}^{\rm \tiny rp}$, ${Y}^{\rm \tiny rs}$, or the estimated row membership matrix of SCC, and $\tilde{Z}$ is similarly defined with respect to the column clusters. We consider the following eight model set-ups, each of which is designed to imitate typical directed network structures. See Figure \ref{m1network}-\ref{m6network} and \ref{matrix} for illustrations of the topological structures and matrix representation for each kind of network, respectively. For each model set-up, we
consider two parameter settings for the link probability matrix $B$. One corresponds to the fixed $B$ setting, that is to say, each element of $B$ is kept a constant as the network size $n$ increases. The other corresponds to the  more challenging high-dimensional setting, namely, each element of $B$ is vanishing as $n$ increases, which is a more realistic setting.

\paragraph{Model set-up 1 (Networks with `transmission' nodes)} In such networks, there exists a set of nodes which only receive edges from one set of nodes (set 1) and send edges to another set of nodes (set 2), and hence termed as `transmission nodes'. While for nodes from set 1 or set 2, except those edges linked to `transmission nodes', they send to and receive from nodes within their own set. As a result, the sending clusters and receiving clusters are different. In this model-set up, $K^y=K^z=2$, and we consider the following two cases for the link probability matrix $B$: \begin{equation*}
      B_1:= \left[\begin{matrix}
     0.05&  0\\
     0& 0.05
      \end{matrix}\right], \quad
       B_2:= \left[\begin{matrix}
     \frac{{\rm log} n}{n}&  0\\
     0& \frac{{\rm log} n}{n}
      \end{matrix}\right].
      \end{equation*}

\paragraph{Model set-up 2 (Bipartite networks)} In such networks, edges only exists between nodes from different clusters. The sending and receiving clusters could be different by incorporating different sending and receiving patterns. In this model-set up, $K^y=K^z=3$, and we consider the following two cases for the link probability matrix $B$:
\begin{equation*}
      B_1:= \left[\begin{matrix}
     0&  0.2&0\\
     0& 0&0.2\\
     0.1&0&0
      \end{matrix}\right], \quad
       B_2:= \left[\begin{matrix}
       0&  \frac{1.5{\rm log} n}{n}&0\\
     0& 0&\frac{1.5{\rm log} n}{n}\\
     \frac{{\rm log} n}{n}&0&0
      \end{matrix}\right].
      \end{equation*}
\paragraph{Model set-up 3 (Networks with core-periphery structure)} In such networks, there exists a set of nodes (termed as core nodes) which send and receive edges with each other, and they also receive edges from another set of nodes (termed as periphery nodes) which has no incoming edges. Thus, the sending and receiving clusters are different. In this model-set up, $K^y=1$ and $K^z=2$, and we consider the following two cases for the link probability matrix $B$:

 \begin{equation*}
      B_1:= \left[\begin{matrix}
     0.05&0
      \end{matrix}\right], \quad
       B_2:= \left[\begin{matrix}
     \frac{1.8{\rm log} n}{n}&0
      \end{matrix}\right].
      \end{equation*}
\paragraph{Model set-up 4 (Multi-layer networks)} In such networks, nodes with in each ``layer'' share similar sending \emph{and} receiving patterns, while nodes across different layers might also share similar sending \emph{or} receiving patterns. Thus considering the directions of edges, the sending and receiving clusters may be different. In this model-set up, $K^y=2$ and $K^z=3$, and we consider the following two cases for the link probability matrix $B$:
\begin{equation*}
      B_1:= \left[\begin{matrix}
     0.1&0.1&0\\
     0&0.04&0.1
      \end{matrix}\right], \quad
       B_2:= \left[\begin{matrix}
     \frac{1}{\sqrt{n}}& \frac{1}{\sqrt{n}}&0\\
     0& \frac{1}{\sqrt{n}}& \frac{1}{\sqrt{n}}
      \end{matrix}\right].
      \end{equation*}
\paragraph{Model set-up 5 (Networks with `message loop')} In such networks, edges only exist between nodes belonging to different clusters. The information flow from one cluster to another and finally forms a `message loop'. The sending and receiving clusters are the same but as we can imagine, clustering based on symmetrized adjacency matrix would not perform well. In this model-set up, $K^y=K^z=3$, and we consider the following two cases for the link probability matrix $B$:
\begin{equation*}
      B_1:= \left[\begin{matrix}
     0&0.05&0\\
     0&0&0.05\\
     0.1&0&0
      \end{matrix}\right], \quad
       B_2:= \left[\begin{matrix}
     0& \frac{1}{2\sqrt{n}}&0\\
     0& 0& \frac{1}{2\sqrt{n}}\\
    \frac{1}{2\sqrt{n}}& 0& 0
      \end{matrix}\right].
      \end{equation*}

\paragraph{Model set-up 6 (Networks of `message passing')} The network structure is similar to that in model set-up 4 except that the edges across different layers start from upper layers down to lower layers, just like passing messages. The sending and receiving clusters are the same but as we can imagine, treating these networks as undirected networks does not work well. In this model-set up, $K^y=K^z=4$, and we consider the following two cases for the link probability matrix $B$:
\begin{equation*}
      B_1:= \left[\begin{matrix}
     0.2&0.1&0.05&0.01\\
     0&0.2&0.1&0.05\\
     0&0&0.2&0.1\\
     0&0&0&0.2
      \end{matrix}\right], \quad
       B_2:= \left[\begin{matrix}
    \frac{2{\rm log} n}{{n}} & \frac{{\rm log} n}{{2n}} &\frac{{\rm log} n}{{2n}}&\frac{{\rm log} n}{{2n}}\\
     0& \frac{2{\rm log} n}{{n}}&\frac{{\rm log} n}{{2n}}&\frac{{\rm log} n}{{2n}}\\
   0&0& \frac{2{\rm log} n}{{n}} & \frac{{\rm log} n}{{2n}}\\
  0& 0&0& \frac{2{\rm log} n}{{n}}
      \end{matrix}\right].
      \end{equation*}
\paragraph{Model set-up 7 (Networks in model set-up 1 with degree heterogeneity)} The basic set-up is similar to model 1 except that we also incorporate the degree heterogeneity. See also Figure \ref{m1network} for the topological structure example. In this model-set up, $K^y=K^z=2$, each $\theta^y_i$ is generated i.i.d. to be 0.3 w.h.p. 0.8 and 1 w.h.p. 0.2, and $\theta^z_i$ is generated in a similar way. We consider the following two cases for the link probability matrix $B$:
 \begin{equation*}
 B_1:= \left[\begin{matrix}
     0.3&  0\\
     0& 0.3
      \end{matrix}\right], \quad
       B_2:= \left[\begin{matrix}
     \frac{3}{\sqrt{n}}&  0\\
     0&  \frac{3}{\sqrt{n}}
      \end{matrix}\right].
      \end{equation*}

\paragraph{Model set-up 8 (Networks in model set-up 5 with degree heterogeneity)} The basic set-up is similar to model 5 except that the degree heterogeneity is also encoded. See also Figure \ref{m5network} for the topological structure example. In this model-set up, $K^y=K^z=3$, the entries of $\theta^y$ are 0.5 except the first three ones, and $\theta^z=\theta^y$. We consider the following two cases for the link probability matrix $B$:
\begin{equation*}
      B_1:= \left[\begin{matrix}
     0&0.2&0\\
     0&0&0.2\\
     0.2&0&0
      \end{matrix}\right], \quad
       B_2:= \left[\begin{matrix}
     0& \frac{2}{\sqrt{n}}&0\\
     0& 0& \frac{2}{\sqrt{n}}\\
    \frac{2}{\sqrt{n}}& 0& 0
      \end{matrix}\right].
      \end{equation*}

In the random projection scheme, the oversampling parameter is 10, the power parameter is 2, and the test matrices are generated with i.i.d. standard Gaussian entries. In the random sampling scheme, the sampling rate is 0.7. We use the R package \textsf{{irlba}} to compute the singular vector iteratively after the sampling step. We evaluate how the approximation error for $P$, the estimation error for the row clusters $Y$, and the estimation error for the column clusters $Z$ alter as the network size $n$ increases, respectively. For the sake of readability, we only display the averaged results together with the standard deviations over 20 replications of model set-up 1 in Figure \ref{m1case1}-\ref{m1case2}.
The results corresponding the remaining model set-ups can be found in the appendix.
We can make the following observations from the results. First, for the approximation error, the three methods show similar tendencies as the sample size increases, all grow at rate $o(n)$, indicating that $\tilde{A}$'s concentrate around the population $P$. The slight differences come from pre-constants because we have shown in Section \ref{theory} that $\|\tilde{A}-P\|_2$ attains the order-wise minimax optimal rate with large probability. Second, for the misclustering error, all three methods yield decreasing misclustering rates as $n$ increases. The standard deviations generally decrease as $n$ increases, though bad clustering performances would lead to small deviations of all methods when $n$ is relatively small. The RP-SCC (RP-SsCC) and RS-SCC (RS-SsCC) perform just slightly worse than SCC (SsCC), especially when $n$ is large, which is the focus of this work. In particular, RP-SCC (RP-SsCC) generally leads to better clustering performances than RS-SCC (RS-SsCC) does, and more interestingly, the former are even more stable than SCC (SsCC); see Figure \ref{m1case2} and \ref{m6case1} for example. In addition, one may find that the performance of the estimated row clusters are better than that of the column clusters in mode set-up 3 and 4, which is because $K^y< K^z$ therein and is consistent with our theoretical results. Finally, we note that the misclustering error might decrease slowly. This because when the link probability is $O({\rm log} n/n)$, the misclustering rate is as slow as $O({\rm log} n)$, indicated by our theory. In addition, the theoretical bounds hold in the sense of probability.

\begin{figure}[!htbp]{}
\centering
\subfigure[]{\includegraphics[height=4.1cm,width=4.3cm,angle=0]{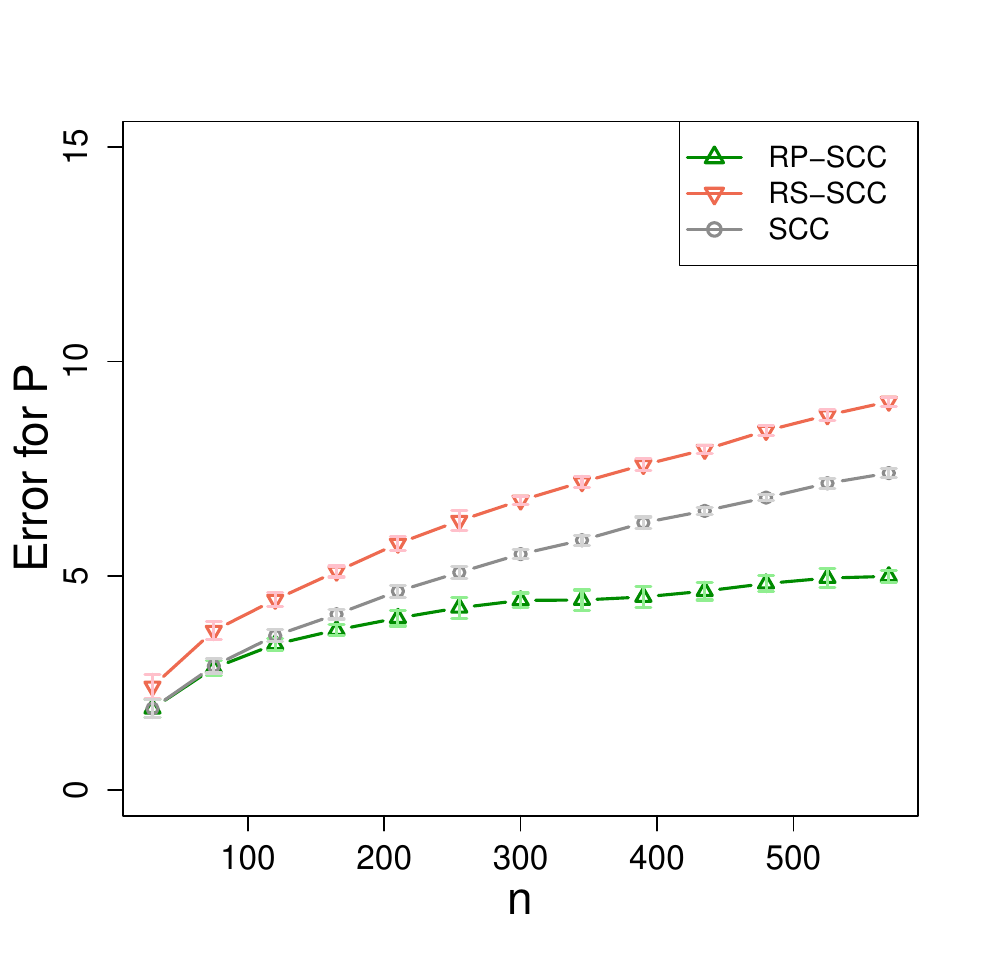}}
\subfigure[]{\includegraphics[height=4.1cm,width=4.3cm,angle=0]{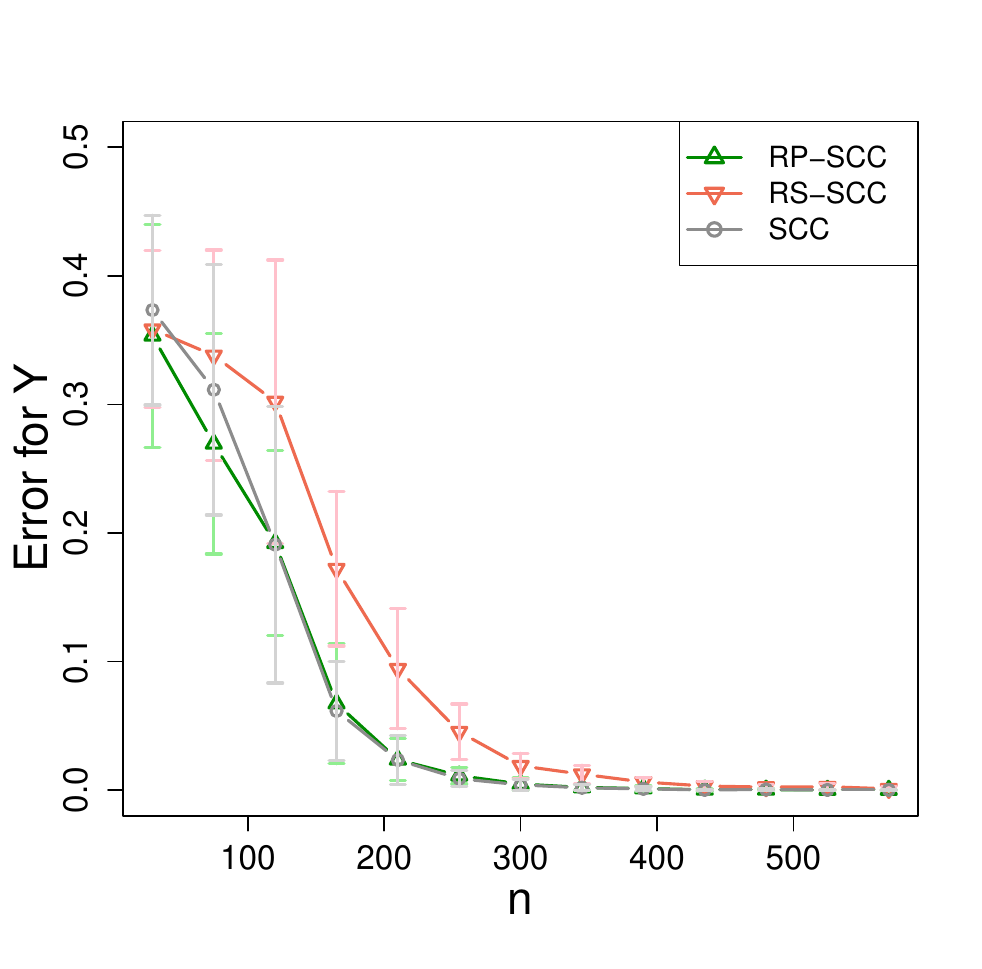}}
\subfigure[]{\includegraphics[height=4.1cm,width=4.3cm,angle=0]{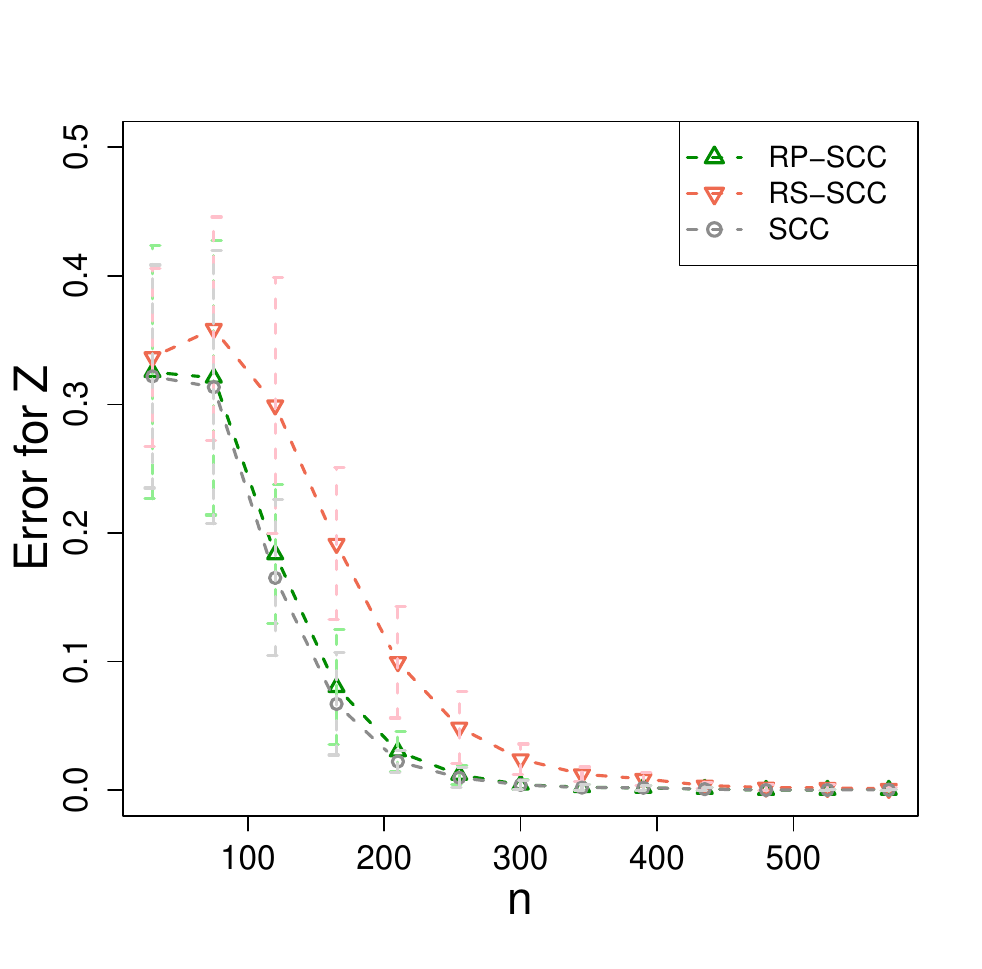}}
\caption{Simulation results of case 1 under {model set-up 1}.}\label{m1case1}
\end{figure}

\begin{figure}[!htbp]{}
\centering
\subfigure[]{\includegraphics[height=4.1cm,width=4.3cm,angle=0]{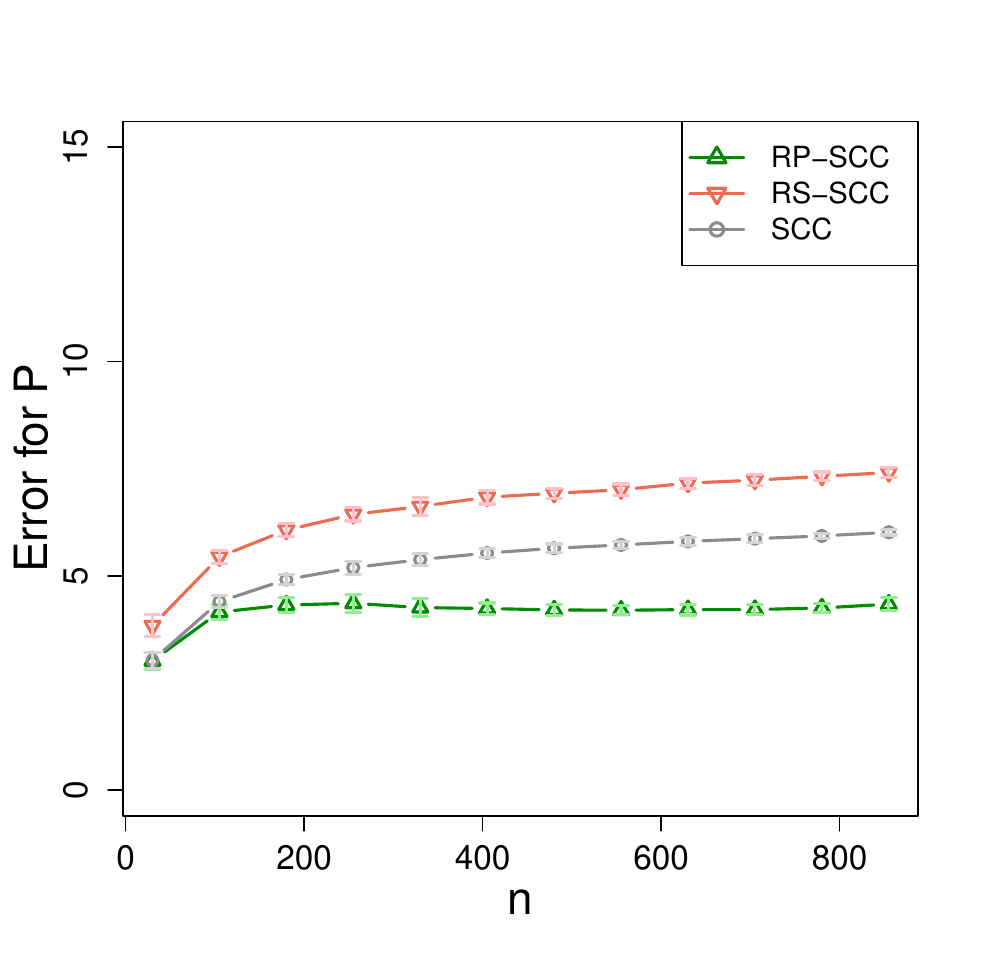}}
\subfigure[]{\includegraphics[height=4.1cm,width=4.3cm,angle=0]{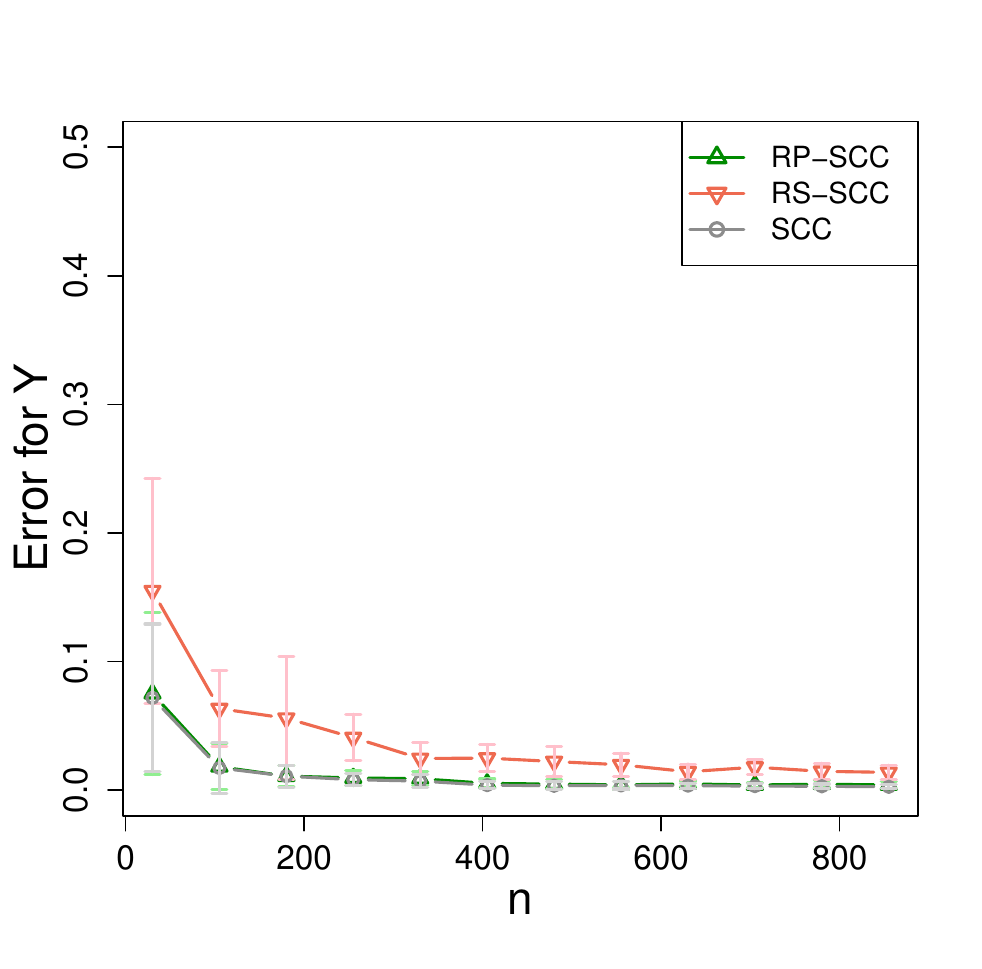}}
\subfigure[]{\includegraphics[height=4.1cm,width=4.3cm,angle=0]{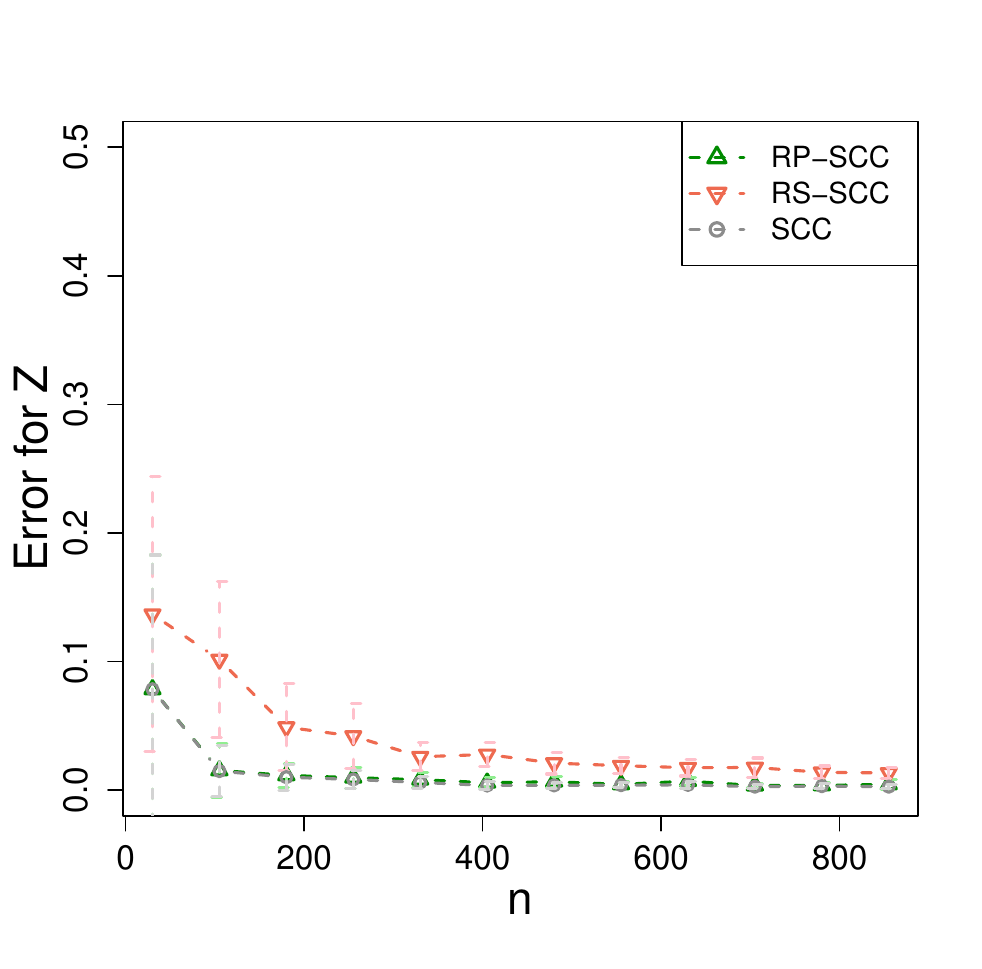}}
\caption{Simulation results of case 2 under {model set-up 1}.}\label{m1case2}
\end{figure}

\section{Real data analysis}
\label{realdata}
We empirically evaluate the randomized spectral co-clustering algorithms on real network datasets, considering both the clustering accuracy and the computational efficiency.
\subsection{Accuracy comparison on small-scale networks}
We compare the clustering performance of the proposed methods RP-SCC (RP-SsCC) and RS-SCC (RS-SsCC) with SCC and two iterative-algorithm-based spectral co-clustering algorithms, denoted as \textsf{svds} and \textsf{irlba}. Specifically, \textsf{svds} and \textsf{irlba} use respectively the implicitly restarted Lanczos algorithm \citep{calvetti1994implicitly} (\textsf{svds} in R package \textsf{RSpectra} \citep{rspectra}) and the augmented implicitly restarted Lanczos bidiagonalization algorithm \citep{baglama2005augmented} (\textsf{irlba} in R package \textsf{irlba} \citep{irlba}) to compute the SVD in SCC. To fix ideas, we call the four methods including RP-SCC, RS-SCC, \textsf{svds} and \textsf{irlba} the approximated methods for SCC (SsCC) in what follows.
For RP-SCC (RP-SsCC), the oversampling parameter is 10, the power parameter is 2, and the test matrices are generated with i.i.d. standard Gaussian entries. For RS-SCC (RS-SsCC), the sampling rate is 0.2. We use the R package \textsf{{irlba}} to compute the singular vector iteratively after the sampling step. For \textsf{{svds}} and \textsf{irlba}, the tolerance parameter is set to be $10^{-5}$.

We consider two directed networks, one is the statisticians citation network \citep{ji2016coauthorship} and the other is the European email network \citep{yin2017local}.

\paragraph{Statisticians citation network}
This network describes the citation relationships between statisticians who published at least one paper in four top journals of statistics from 2003 to the first half of 2012. If author $i$ cited at least one paper written by author $j$, then there is a directed edge from node $i$ to node $j$. The largest component of this network results in 2,654 nodes and 21,568 edges.

\begin{figure}[!htbp]{}
	\centering
	\subfigure{\includegraphics[height=7cm,width=8cm,angle=0]{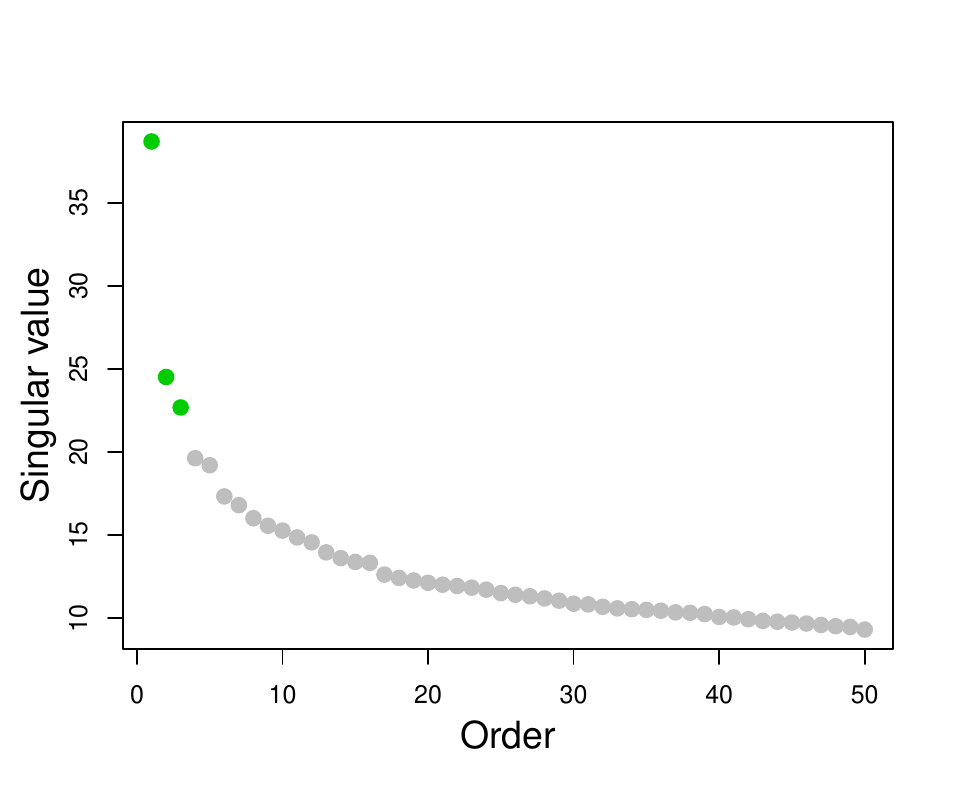}}
	\caption{The top 50 singular values of the adjacency matrix of the statisticians citation network.}\label{citationsv}
\end{figure}

\begin{figure}[!htbp]{}
	\centering
	\subfigure[SCC]{\includegraphics[height=4.8cm,width=5cm,angle=0]{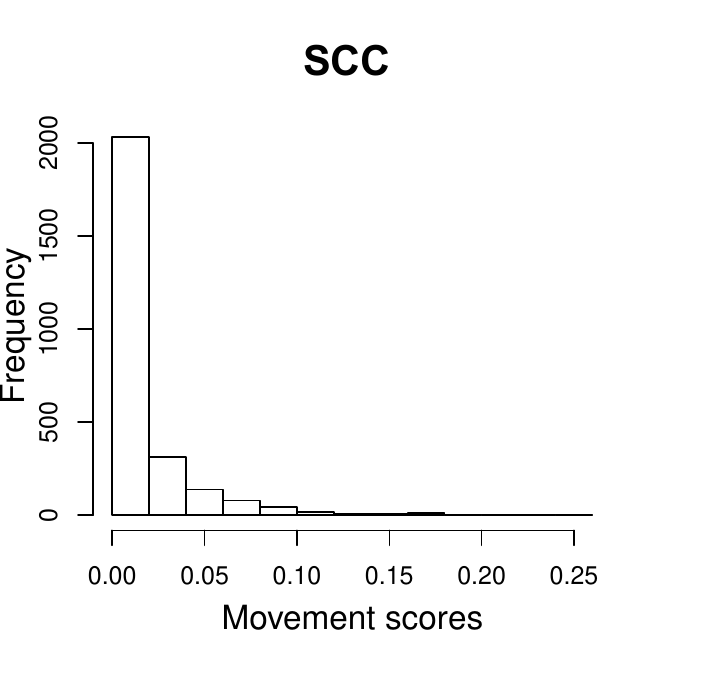}}
    \subfigure[RP-SCC]{\includegraphics[height=4.8cm,width=5cm,angle=0]{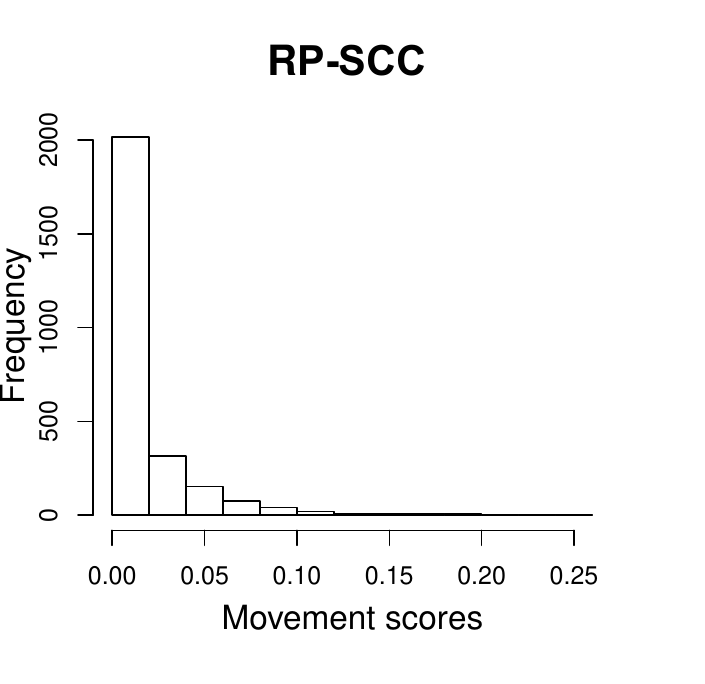}}
    \subfigure[RS-SCC]{\includegraphics[height=4.8cm,width=5cm,angle=0]{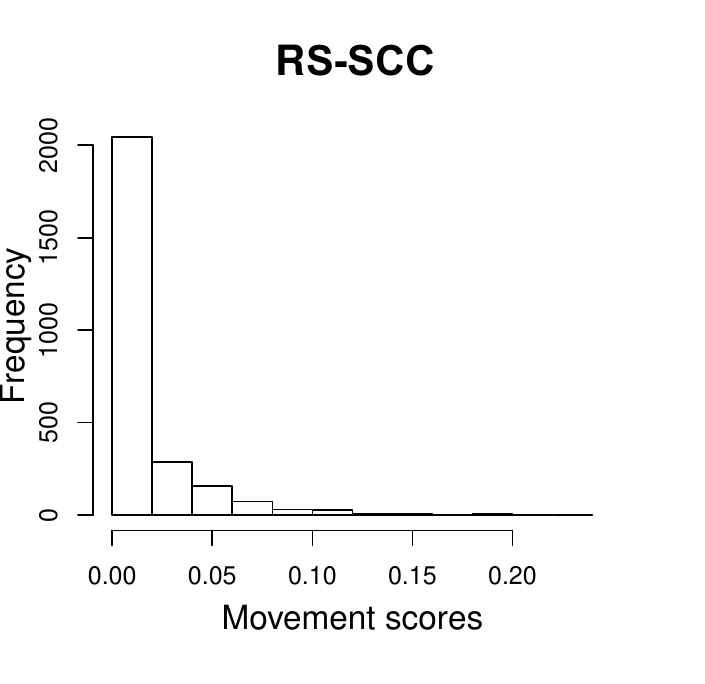}}
     \subfigure[\textsf{svds}]{\includegraphics[height=4.8cm,width=5cm,angle=0]{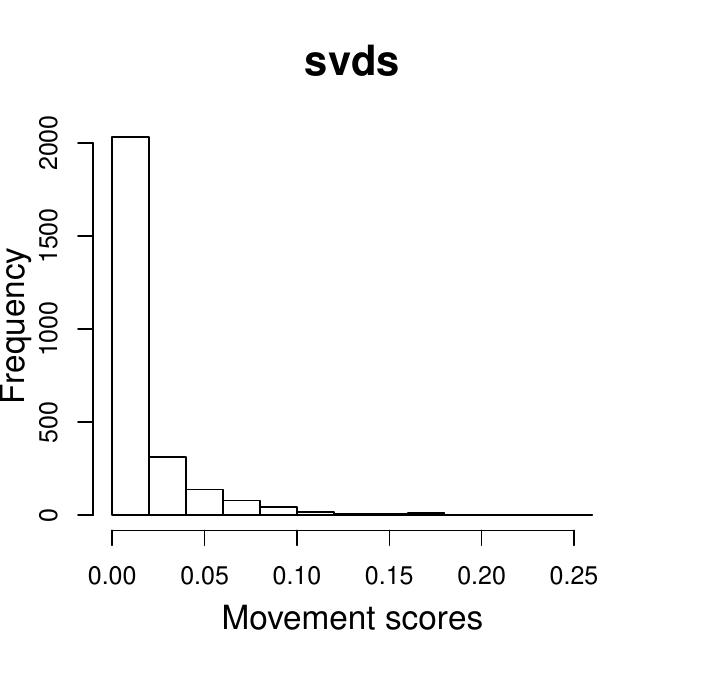}}
    \subfigure[\textsf{irlba}]{\includegraphics[height=4.8cm,width=5cm,angle=0]{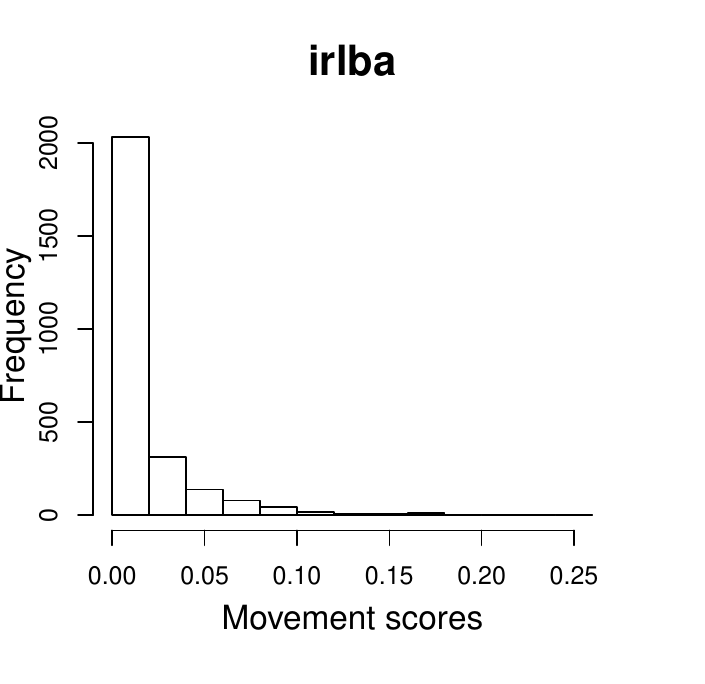}}
	\caption{Histogram of movement scores of different methods for the statisticians citation network.}\label{citationmove}
\end{figure}

To decide the target rank, we evaluate the top 50 singular values of the associated adjacency matrix $A$. As indicated in Figure \ref{citationsv}, there is an eigen-gap between the third and fourth singular values, suggesting that the target rank is 3 \citep{rohe2016co}. Before doing clustering, we first evaluate the similarities and patterns of singular vectors found by different methods. Define the \emph{movement score} for node $i$ to be the Euclidean norm of difference between the $i$th row of the right and left (approximated) singular vectors. Figure \ref{citationmove} shows the histograms of the movement scores by five methods. Five histograms turn out to be very similar, indicating that five methods lead to similar singular vectors to some extent. In addition, the asymmetric nature of this citation network is again evidenced since there exists nodes with movement score away from zero.

Now, we evaluate the clustering performance of these methods. We actually test two set of algorithms, one based on SCC (Algorithm \ref{spectral}) and the other based on SsCC (Algorithm \ref{spectralmedian}). We use the \emph{silhouette method} to select respectively the target number of row clusters $(K^y)$ and column clusters ($K^z$). In our setting, the silhouette method typically evaluates the clustering performance of the $k$-means output with the original left and right singular vectors as the input with respect to different $k$ via the average silhouette width, for which larger value indicates better performance. With the selected number of row clusters and column clusters, we compare the five methods using the ARI \citep{hubert1985comparing,manning2010introduction} between the SCC (SsCC) and the four SCC (SsCC) based approximate methods. Larger ARI reflects more consistency of the clustering results of the compared method pairs.

\begin{figure}[!htbp]{}
	\centering
	\subfigure[Sending clusters]{\includegraphics[height=6.5cm,width=6.5cm,angle=0]{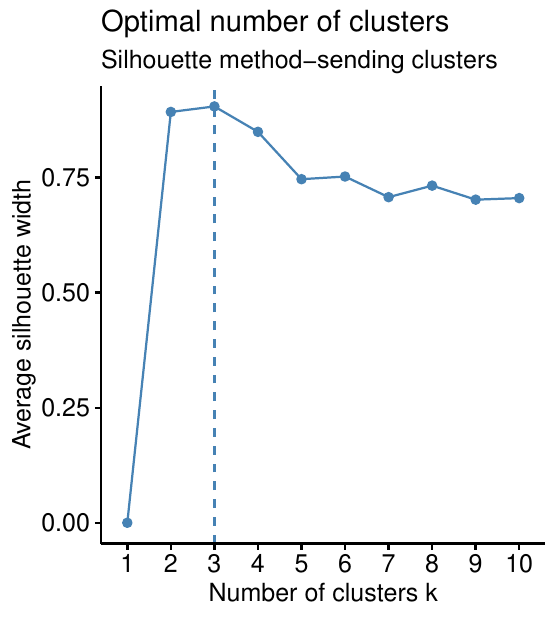}}\hspace{1.3cm}
    \subfigure[Receiving clusters]{\includegraphics[height=6.5cm,width=6.5cm,angle=0]{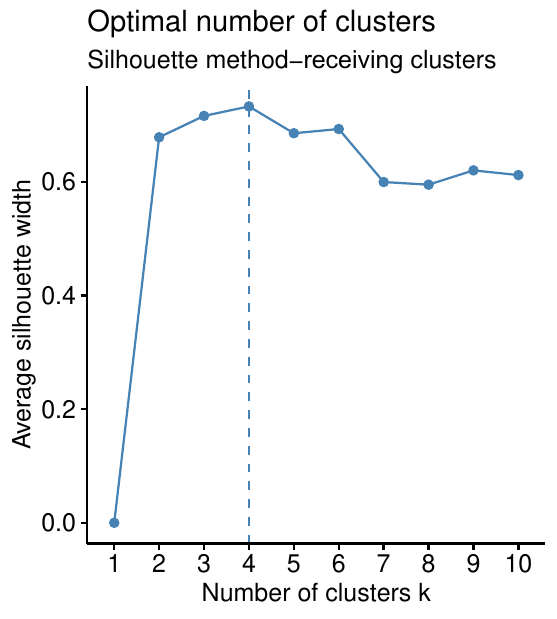}}
	\caption{Optimal number of clusters of the statisticians citation network selected by the average silhouette method based on the SCC.}\label{citationclusternumber}
\end{figure}

\begin{figure}[!htbp]{}
	\centering
	\subfigure[Sending clusters]{\includegraphics[height=5.5cm,width=5.3cm,angle=0]{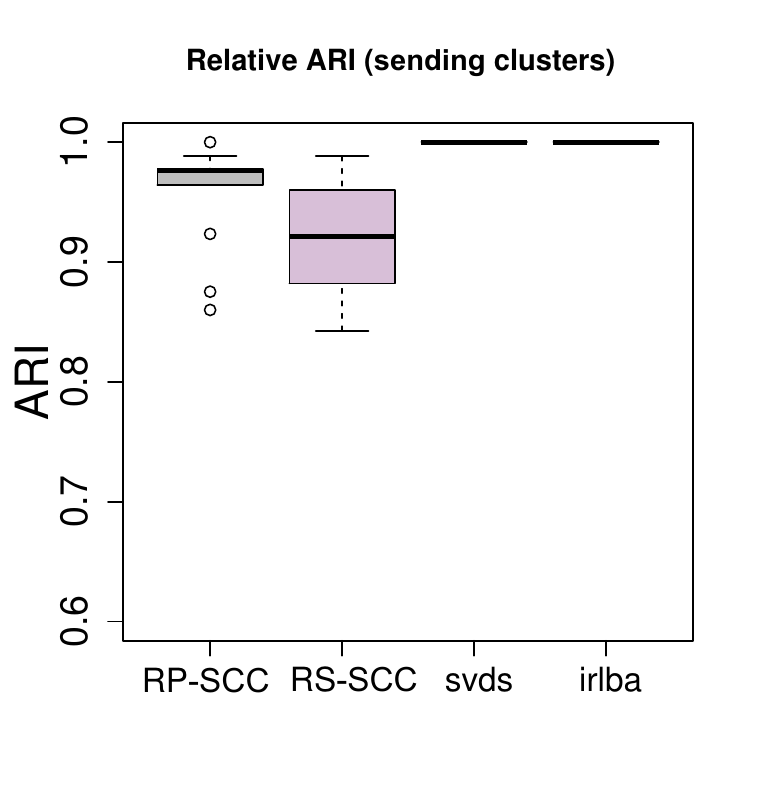}}\hspace{1.3cm}
    \subfigure[Receiving clusters]{\includegraphics[height=5.5cm,width=5.3cm,angle=0]{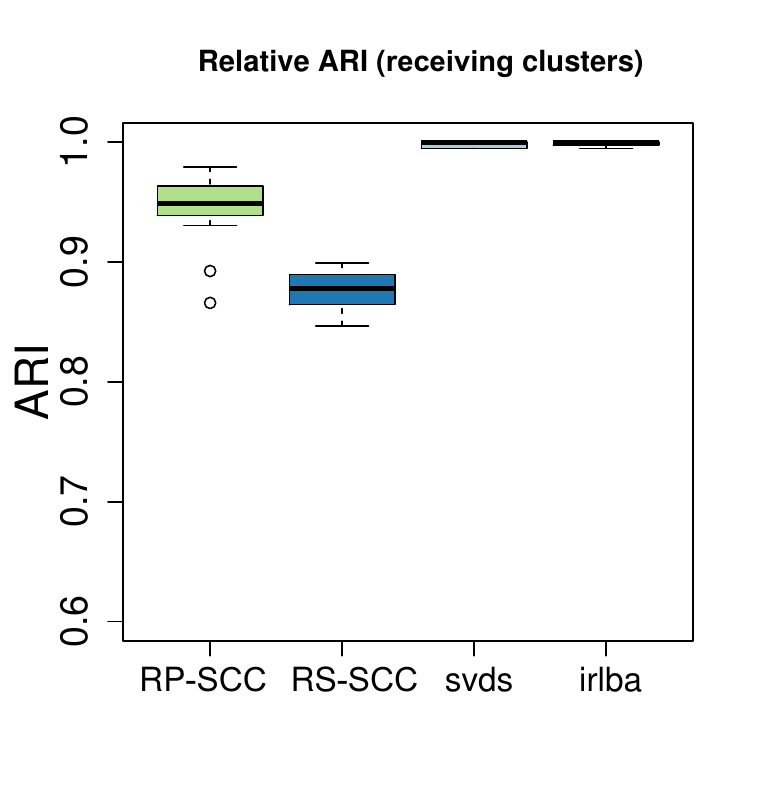}}
	\caption{Relative ARI between the SCC and SCC-based four approximate methods on the statisticians citation network.}\label{citationbox}
\end{figure}

\begin{figure}[!htbp]{}
	\centering
	\subfigure[SCC]{\includegraphics[height=5.5cm,width=4.9cm,angle=0]{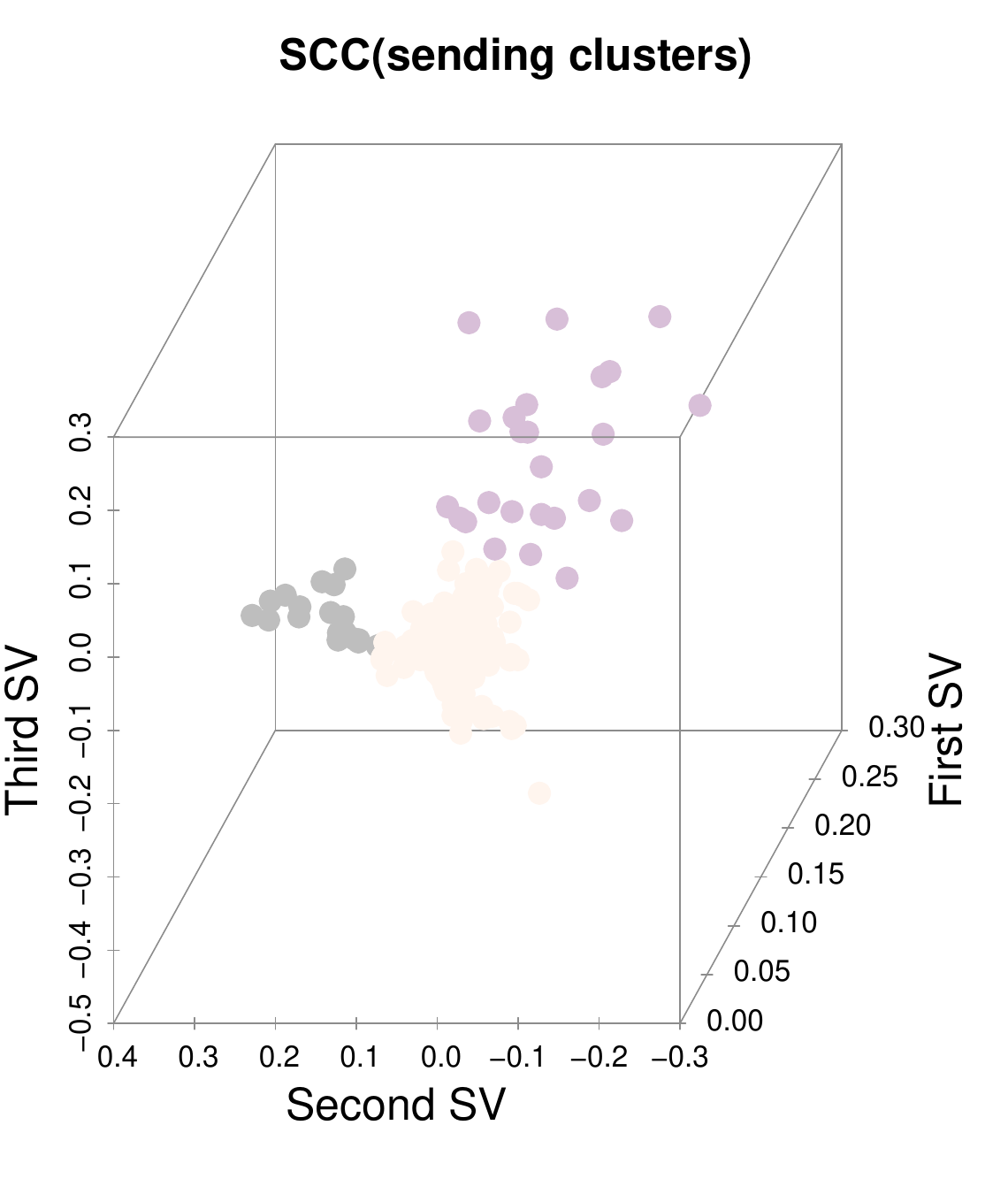}}
    \subfigure[RP-SCC]{\includegraphics[height=5.5cm,width=4.9cm,angle=0]{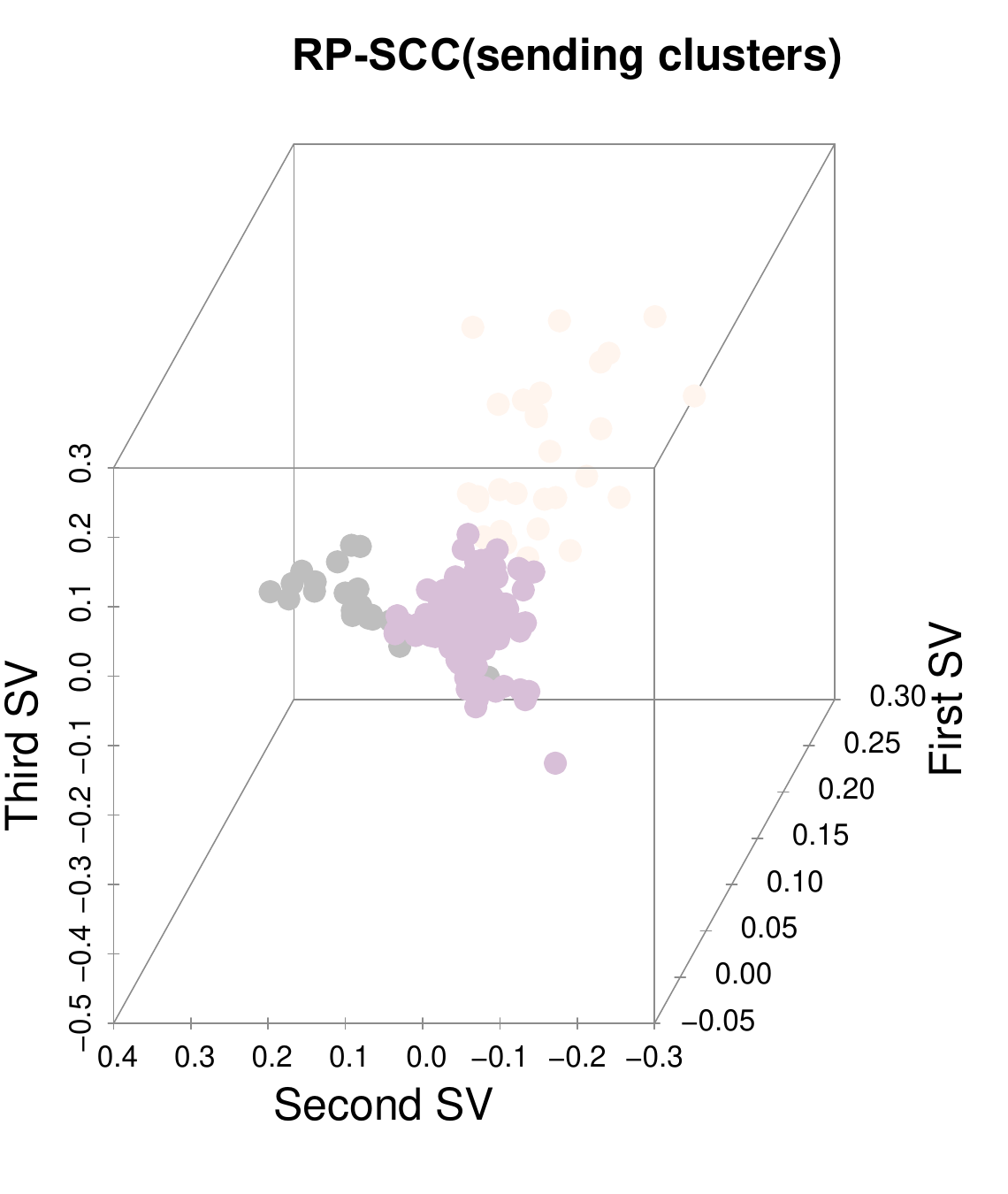}}
    \subfigure[RS-SCC]{\includegraphics[height=5.5cm,width=4.9cm,angle=0]{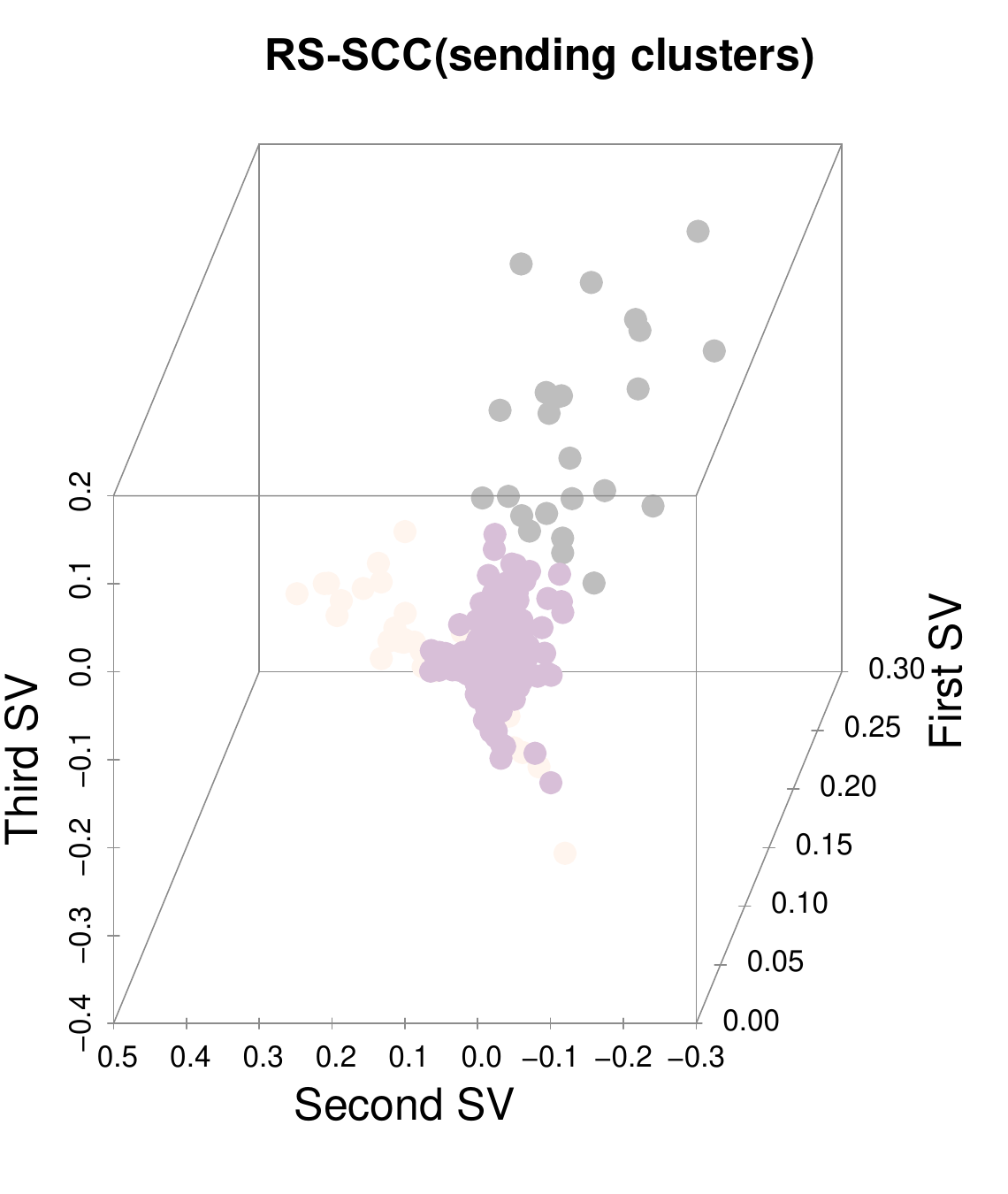}}
    \subfigure[\textsf{svds}]{\includegraphics[height=5.5cm,width=4.9cm,angle=0]{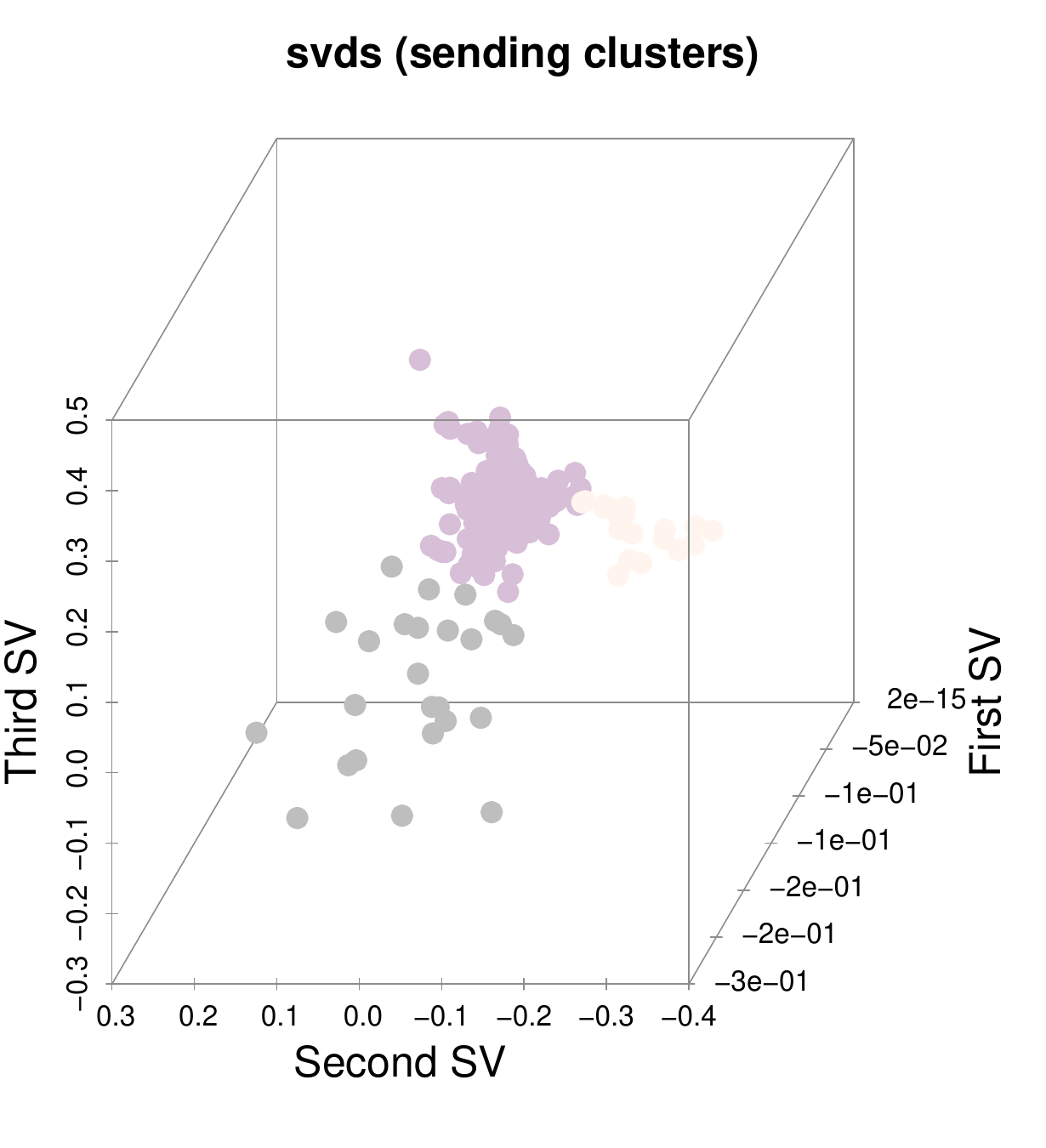}}
    \subfigure[\textsf{irlba}]{\includegraphics[height=5.5cm,width=4.9cm,angle=0]{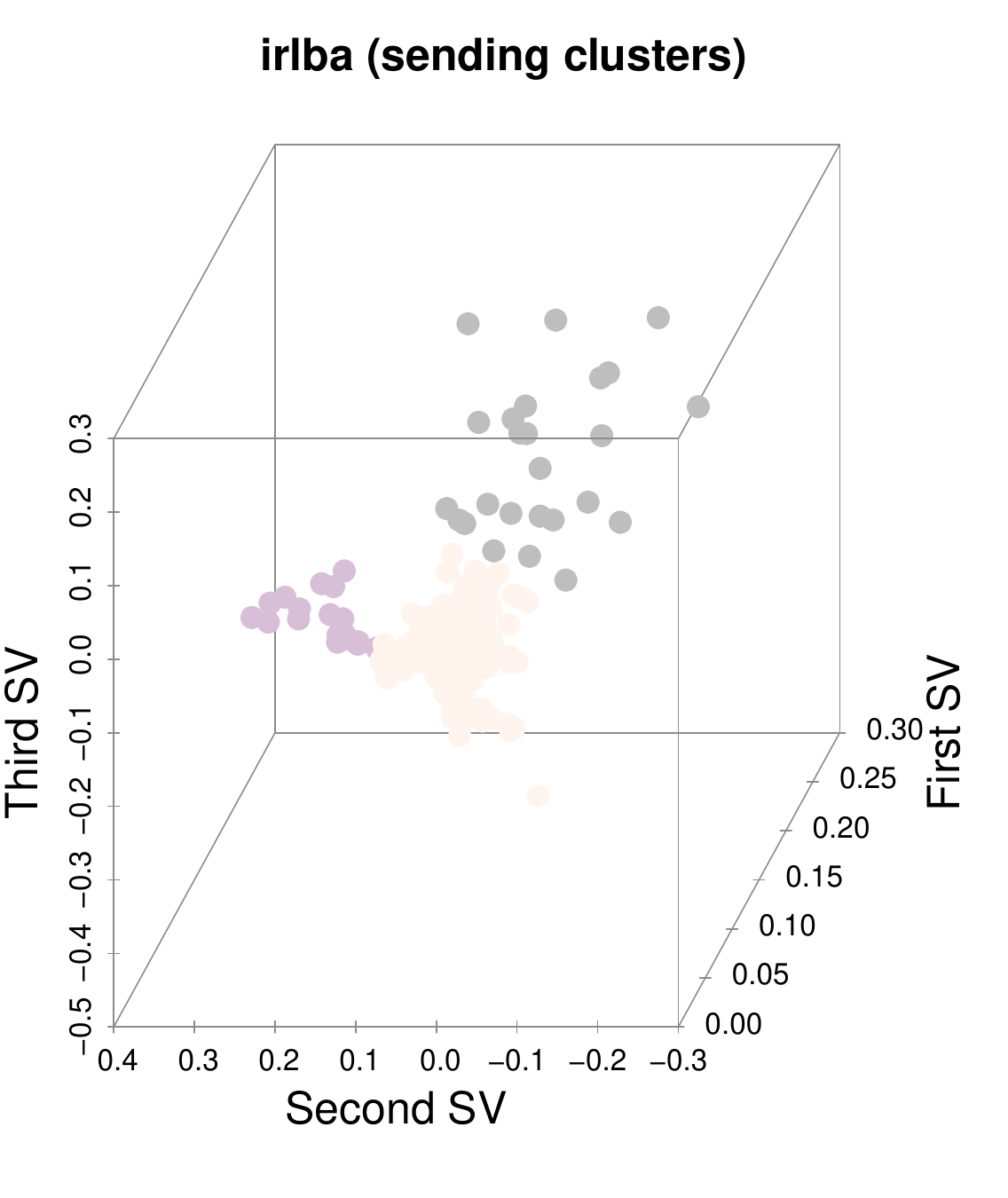}}
	\caption{Sending clusters of the statisticians citation network detected by SCC and four SCC-based approximate algorithms.}\label{citationpointsrow}
\end{figure}

\begin{figure}[!htbp]{}
	\centering
	\subfigure[SCC]{\includegraphics[height=5.5cm,width=4.9cm,angle=0]{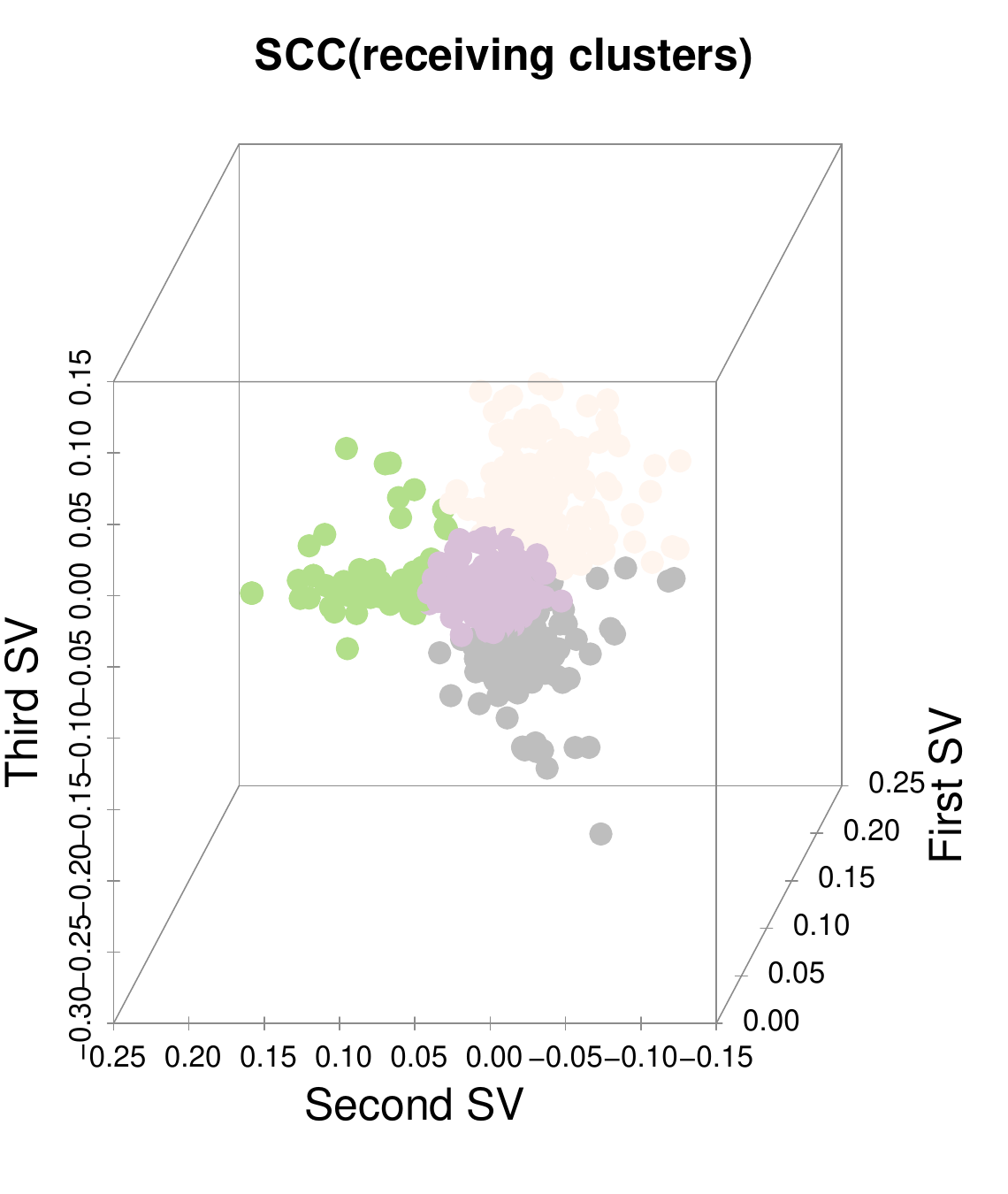}}
    \subfigure[RP-SCC]{\includegraphics[height=5.5cm,width=4.9cm,angle=0]{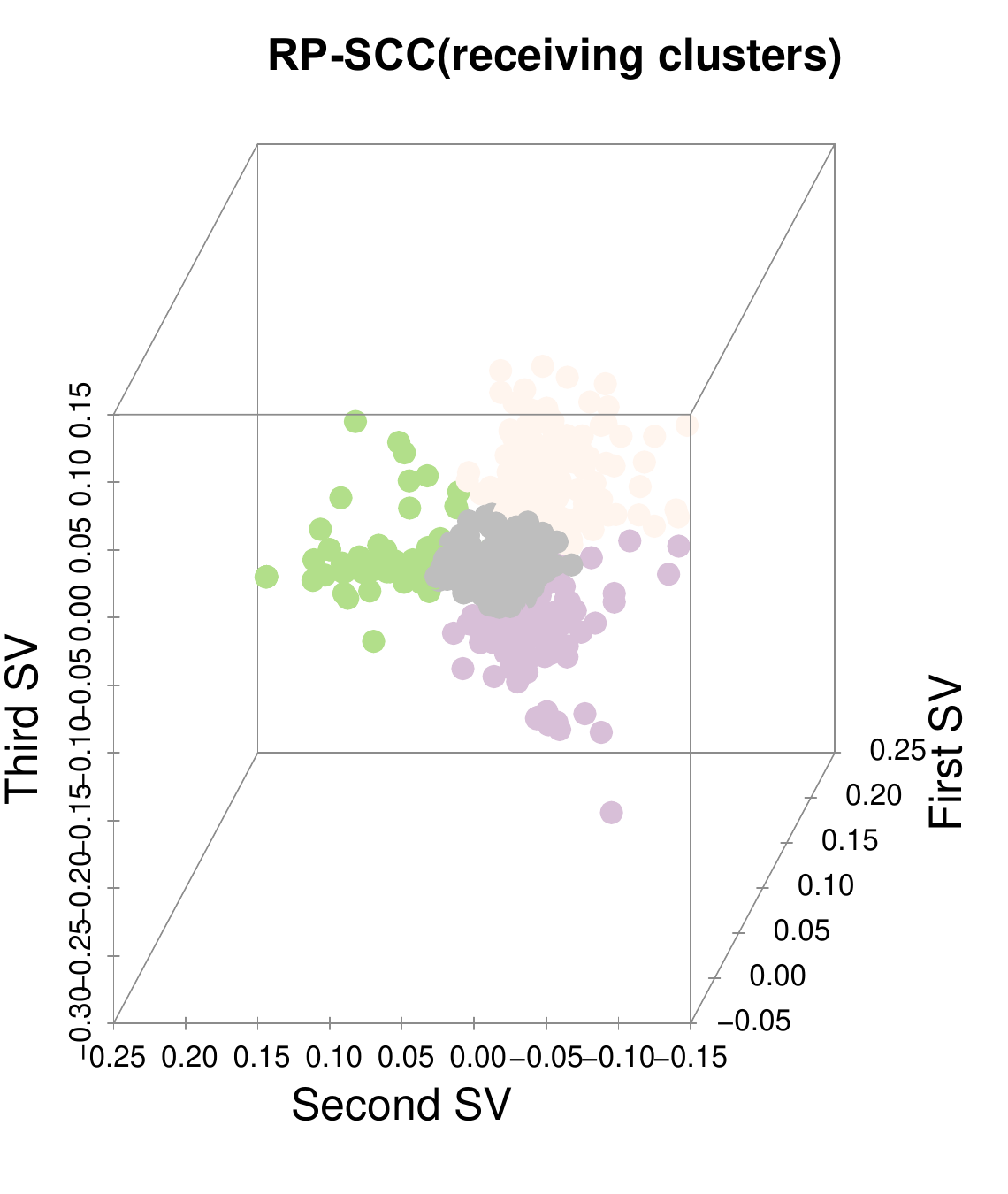}}
    \subfigure[RS-SCC]{\includegraphics[height=5.5cm,width=4.9cm,angle=0]{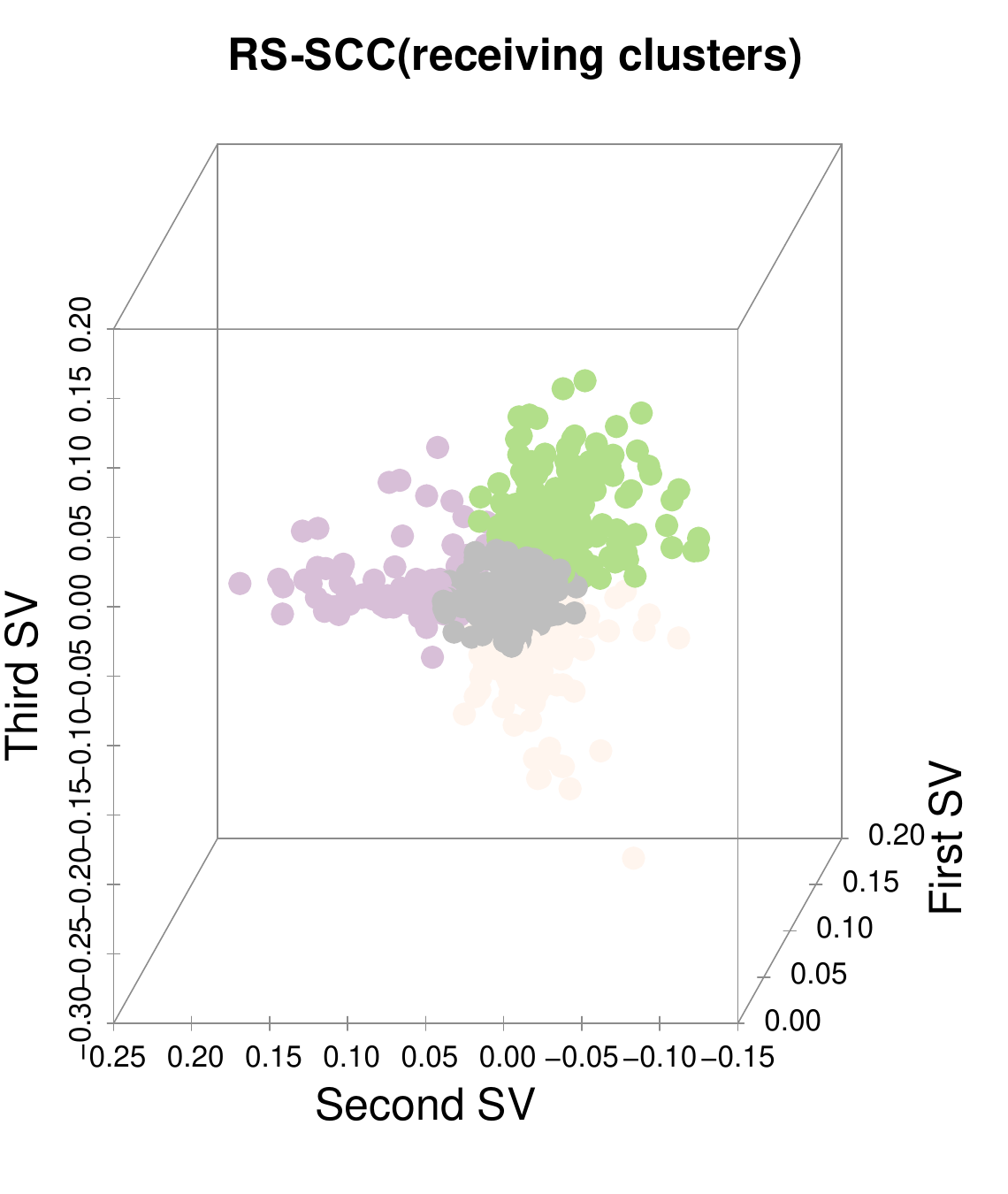}}
    \subfigure[\textsf{svds}]{\includegraphics[height=5.5cm,width=4.9cm,angle=0]{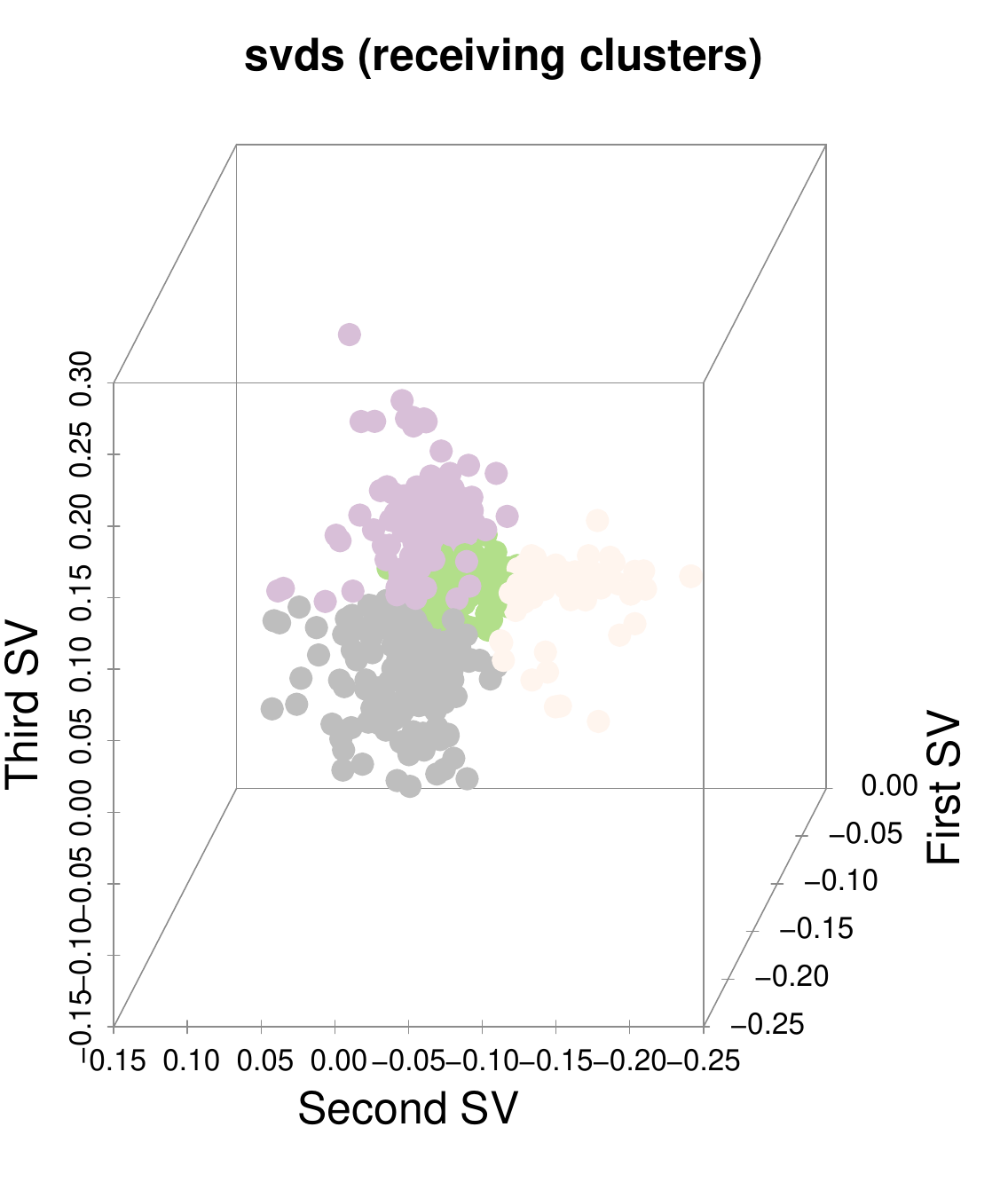}}
    \subfigure[\textsf{irlba}]{\includegraphics[height=5.5cm,width=4.9cm,angle=0]{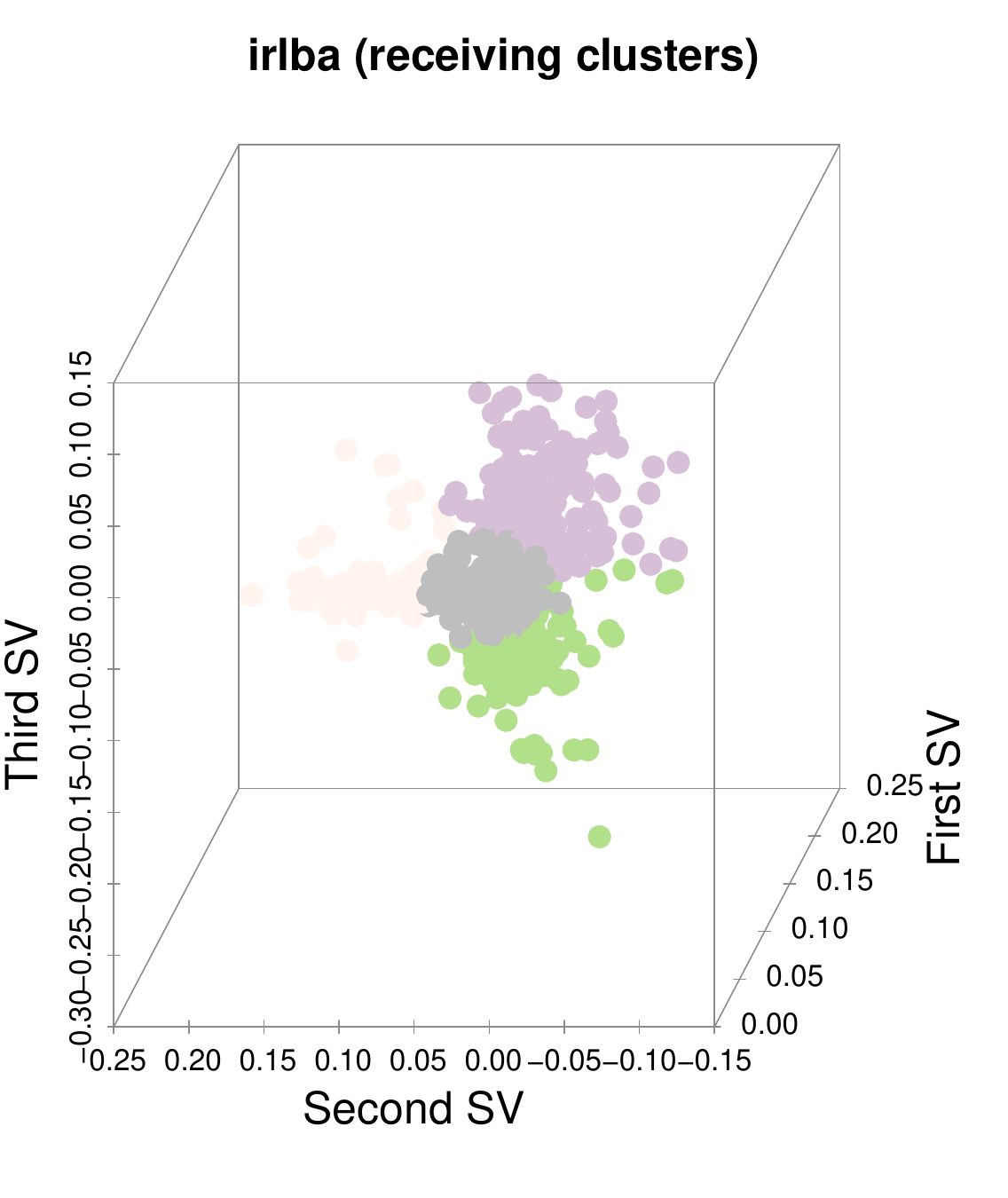}}
	\caption{Receiving clusters of the statisticians citation network detected by SCC and four SCC-based approximate algorithms.}\label{citationpointscolumn}
\end{figure}

For the SCC-based algorithms, the number of sending clusters and receiving clusters turn out to be 3 and 4, respectively; see Figure \ref{citationclusternumber} for details. The box plots of relative ARI over 20 replications are shown in Figure \ref{citationbox}. For both side of clusters, the average ARI's of all methods are larger than 0.85, showing that these methods could well approximate the SCC, though the randomization-based methods are slightly inferior to the iterative algorithm-based methods. We also display the embedding of nodes provided by their corresponding components of the first three left and right singular vectors in Figure \ref{citationpointsrow} and \ref{citationpointscolumn}, where colors indicate clusters. We see that all methods yield similar clusters up to certain rotations of singular vectors and clusters. In addition, the sending clusters (authors cited by others) are more concentrated than the receiving clusters (authors citing others), which agrees with the common logic.

\begin{figure}[!htbp]{}
	\centering
	\subfigure[Sending clusters]{\includegraphics[height=6.5cm,width=6.5cm,angle=0]{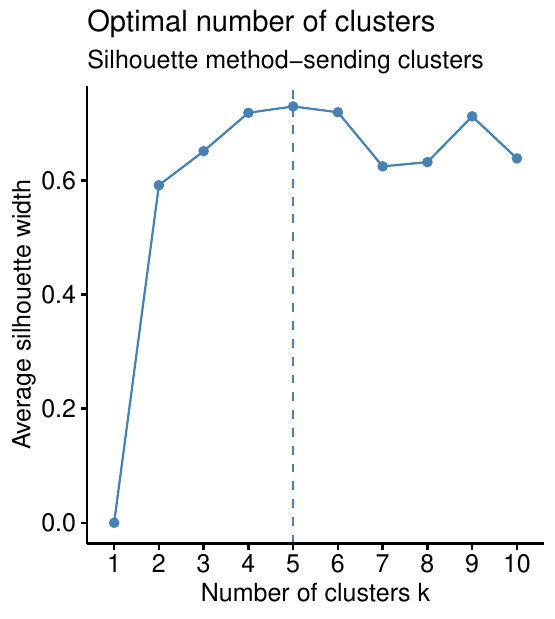}}\hspace{1.3cm}
    \subfigure[Receiving clusters]{\includegraphics[height=6.5cm,width=6.5cm,angle=0]{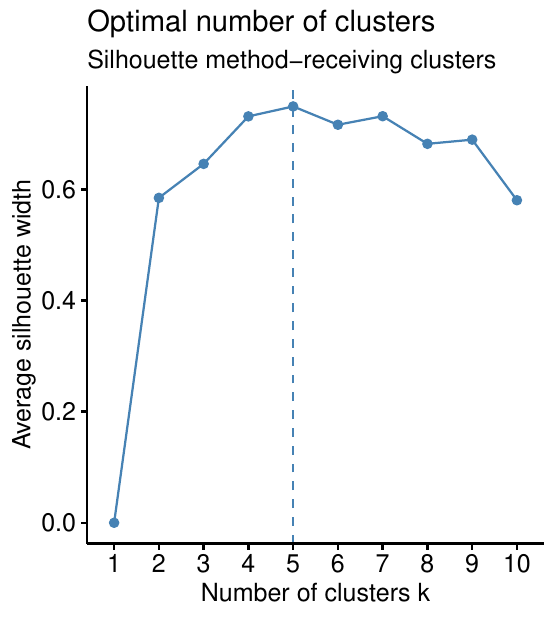}}
	\caption{Optimal number of clusters of the statisticians citation network selected by the average silhouette method based on the SsCC.}\label{citationclusternumberdc}
\end{figure}

\begin{figure}[!htbp]{}
	\centering
	\subfigure[Sending clusters]{\includegraphics[height=5.5cm,width=5.3cm,angle=0]{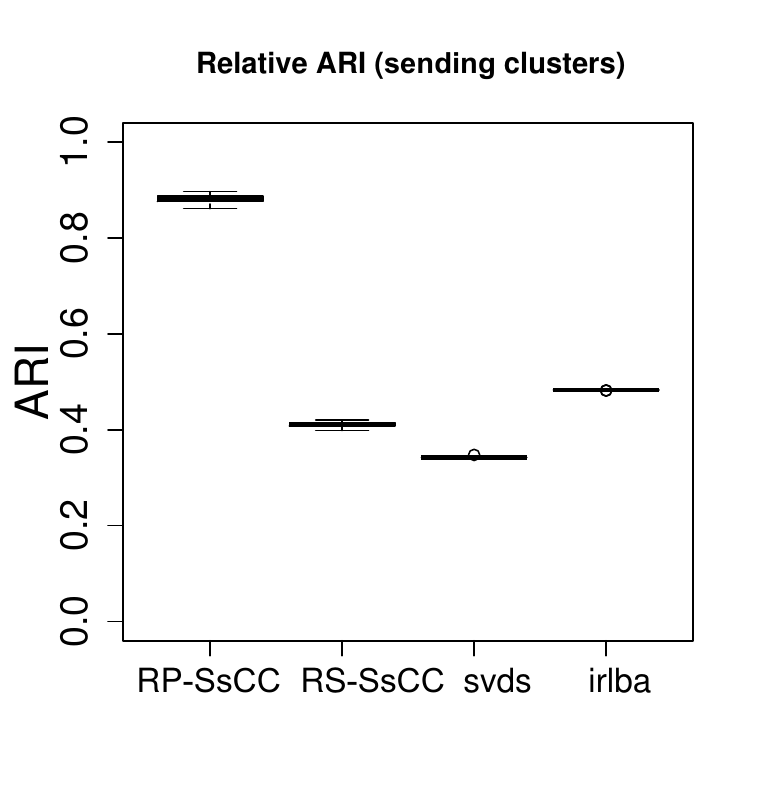}}\hspace{1.3cm}
    \subfigure[Receiving clusters]{\includegraphics[height=5.5cm,width=5.3cm,angle=0]{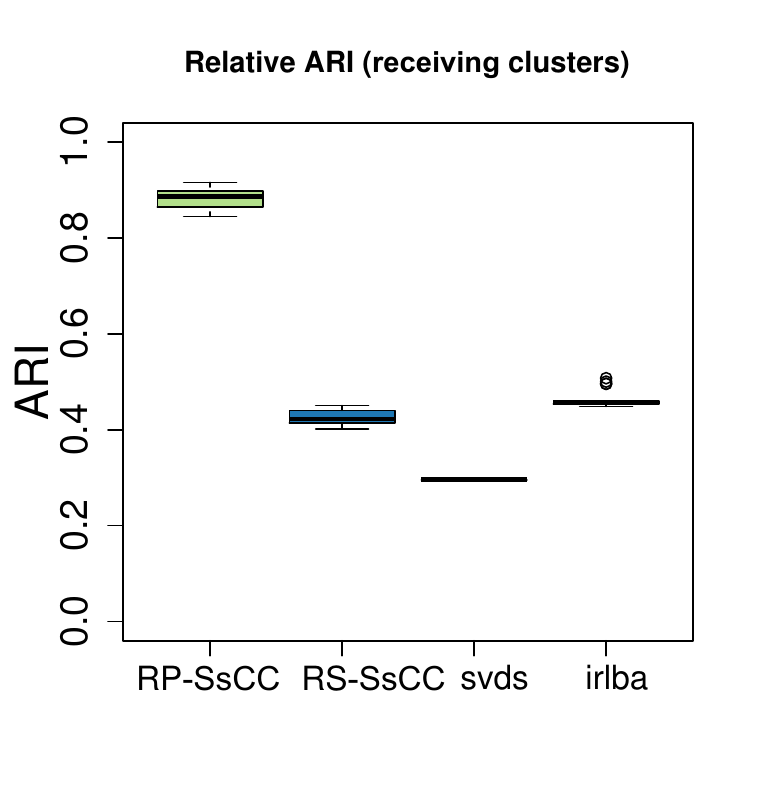}}
	\caption{Relative ARI between the SsCC and SsCC-based four approximate methods on the statisticians citation network.}\label{citationboxdc}
\end{figure}

For the SsCC-based algorithms, Figure \ref{citationclusternumberdc} shows that the optimal number of sending clusters and receiving clusters are both 5. Recall that the target rank is 3, and thus this set-up corresponds to the rank-deficient model, which has been discussed in Section \ref{relatedwork}. Figure \ref{citationboxdc} displays the box plots of relative ARI. For both side of clusters, it turns out that except the random-projection-based method, other three methods seem to perform poorly. This does not contradict with the results of SCC-based algorithms. Because the normalization operator is not stable to noise, two close singular vectors could be far from each other after normalization. Our theory also indicate the hardness of clustering under ScBMs. Nevertheless, the random-projection-based method shows great clustering performance and certain degree of robustness.


\paragraph{European email network}
This network was generated using the email data from a large European research institution. If person $i$ sent at least one email to person $j$, then there is a directed edge from person $i$ to $j$. The largest component of this network results in 986 nodes and 24,929 edges. All treatments are similar to that for the statisticians citation network, hence we mainly discuss the results. All the mentioned figures are in the appendix.

We choose the target rank to be 2 by evaluating the eigen-gap of the original singular values (Figure \ref{emailsv}). The histogram of movement score (Figure \ref{emailmove}) shows that five methods lead to similar singular vectors.

For the SCC-based methods, the optimal number of both sending clusters and receiving clusters turn out to be 2 (Figure \ref{emailclusternumber}).  The average relative ARI of SCC-based methods are all above 0.9 (Figure \ref{emailbox}), suggesting that approximated methods are comparable to SCC. The two-dimensional embedding of estimated singular vectors (Figure \ref{emailpointsrow} and \ref{emailpointscolumn}) show similar pattern across different methods.

For the SsCC-based methods, the optimal number of both sending clusters and receiving clusters turn out to be 4 (Figure \ref{emailclusternumberdc}), thus corresponding to rank-deficient models. Regarding the clustering performance (\ref{emailboxdc}), the random-projection-based method and \textsf{irlba} turn out to be superior than the other two. The reason that lead to the deterioration of approximated methods might due to the noise accumulation of the normalization step therein.

%

\subsection{Time comparison and accuracy evaluation on large-scale networks}
The main barrier that hinders SCC to handle large-scale directed networks is the SVD computation. Therefore, we compare the computational time of the (approximated) SVD using randomized methods, denoted as RP and RS for short, and iterative methods, \textsf{svds} and \textsf{irlba}. We examine five real networks with their number of nodes ranging from more than seventy thousands to more than two millions. Table \ref{tablenetwork} summarizes the basic information of each network, where the target rank means the number of singular vectors to be computed after we examine the (approximate) eigen-gap of the corresponding adjacency matrices.

For \textsf{{svds}} and \textsf{irlba}, the tolerance parameter is set to be $10^{-5}$. For RP, the power parameter is 1 and the oversampling parameter is 5, which are adequate to improve the approximation quality \citep{halko2011finding}. For RS, the sampling parameter is 0.7, and \textsf{irlba} is used after the sampling procedure. A machine with Intel Core i9-9900K CPU 3.60GHz, 32GB memory, and 64-bit WS operating-system, and R version 4.0.4 is used for all computations. Table \ref{tabletime} shows the median time (milliseconds) of each method for computing the SVD of the corresponding adjacency matrices of five real networks over 20 replications. For RS, we report the time including and excluding the sampling procedure, respectively.

It should be noted that the full SVD failed in all five cases. Among the compared four approximate methods for partial SVD, RP is faster than both \textsf{{svds}} and \textsf{irlba} in all the data sets considered. RP is comparable to the baseline iterative methods on smaller networks, and shows more advantage on large networks, even when the sampling time is included. This provides evidence for the computational superiority of randomized methods.

\begin{figure*}[!htbp]{}
\centering
\subfigure[Epinions]{\includegraphics[height=3.2cm,width=4cm,angle=0]{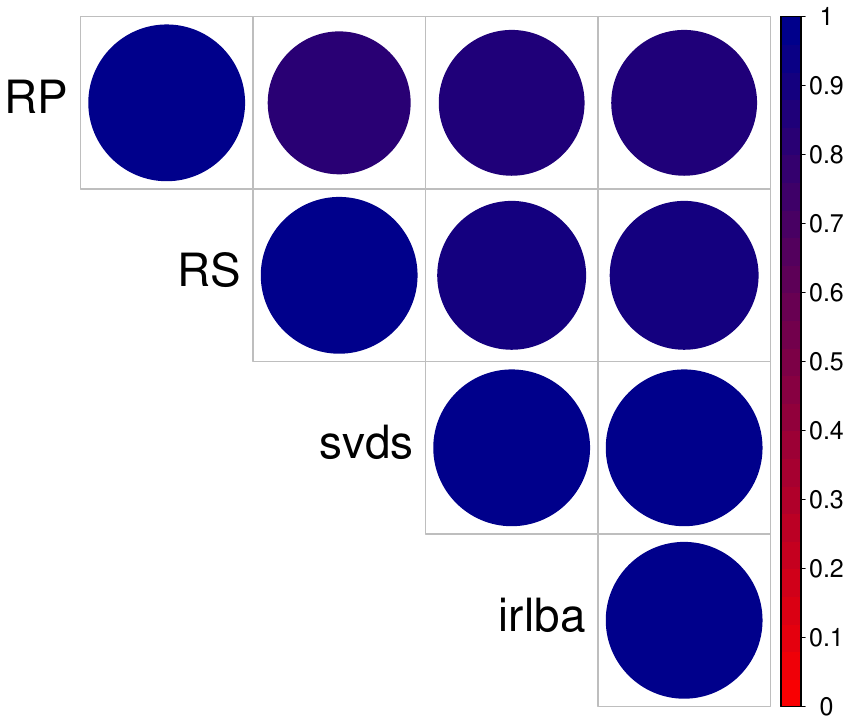}}\hspace{1cm}
\subfigure[Slashdot]{\includegraphics[height=3.2cm,width=4cm,angle=0]{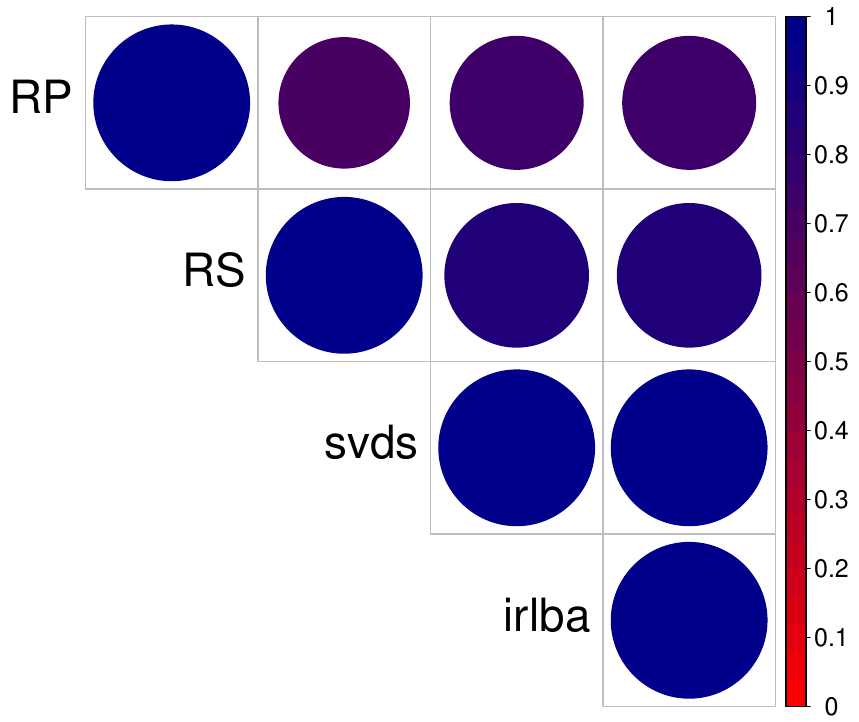}}\hspace{1cm}
\subfigure[Web]{\includegraphics[height=3.2cm,width=4cm,angle=0]{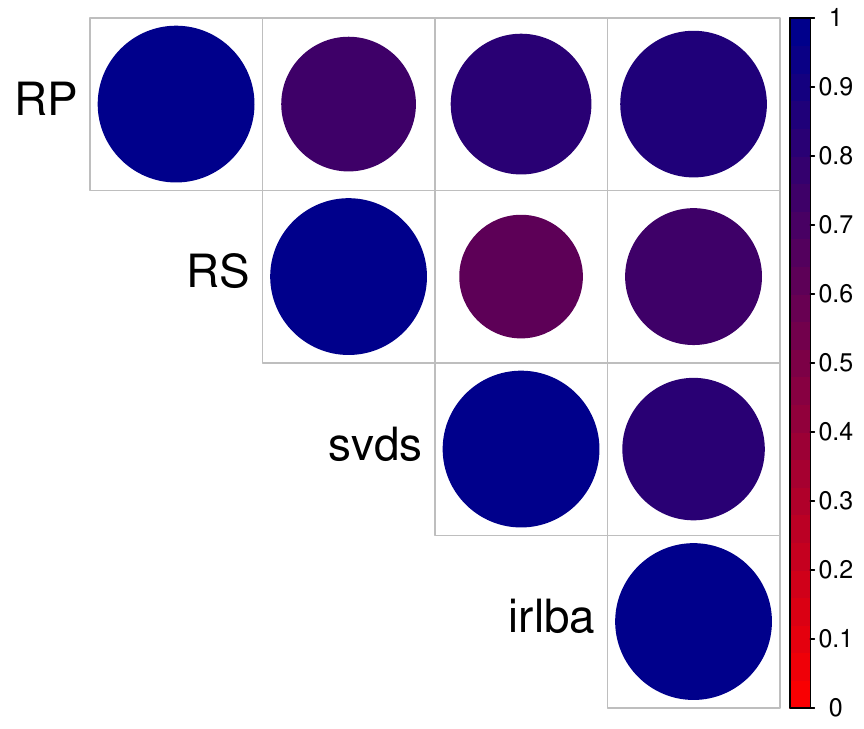}}
\subfigure[Wiki-cat]{\includegraphics[height=3.2cm,width=4cm,angle=0]{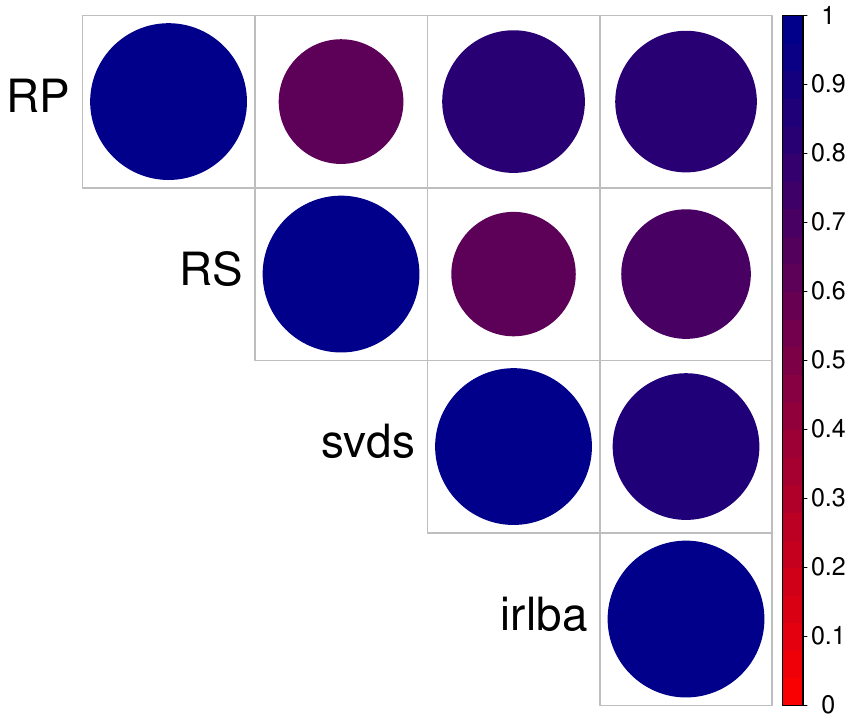}}\hspace{1cm}
\subfigure[Wiki-talk]{\includegraphics[height=3.2cm,width=4cm,angle=0]{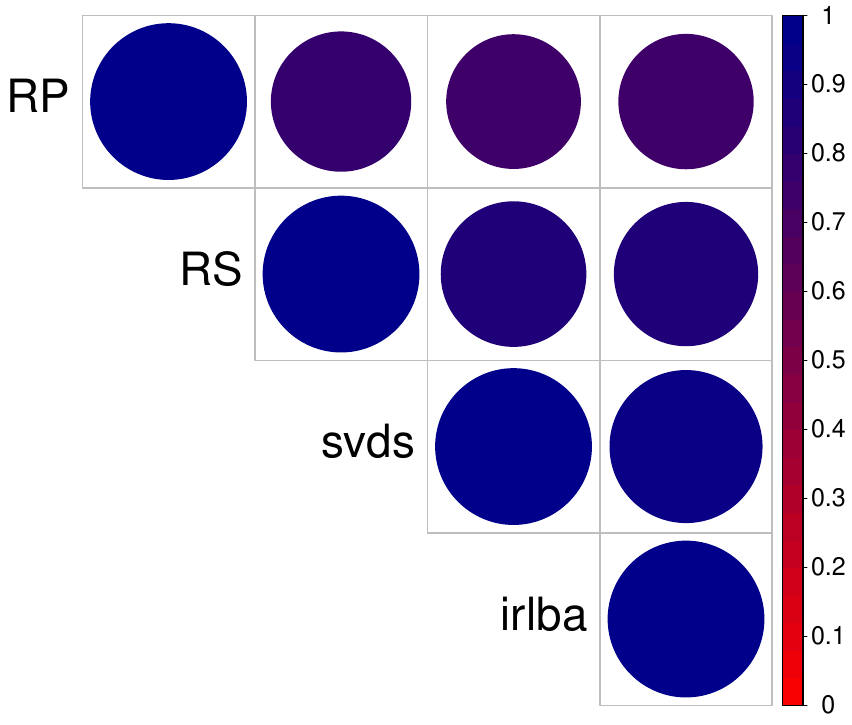}}

\caption{The pairwise comparison of the sending clusters of four methods on five large-scale networks. The relative clustering performance are measured by ARI. Larger ARI, i.e., {larger circles} in the figure, indicates that the clustering results of the two associated methods are more close.}\label{compararirow}
\end{figure*}

\begin{figure*}[!htbp]{}
\centering
\subfigure[Epinions]{\includegraphics[height=2.5cm,width=2.9cm,angle=0]{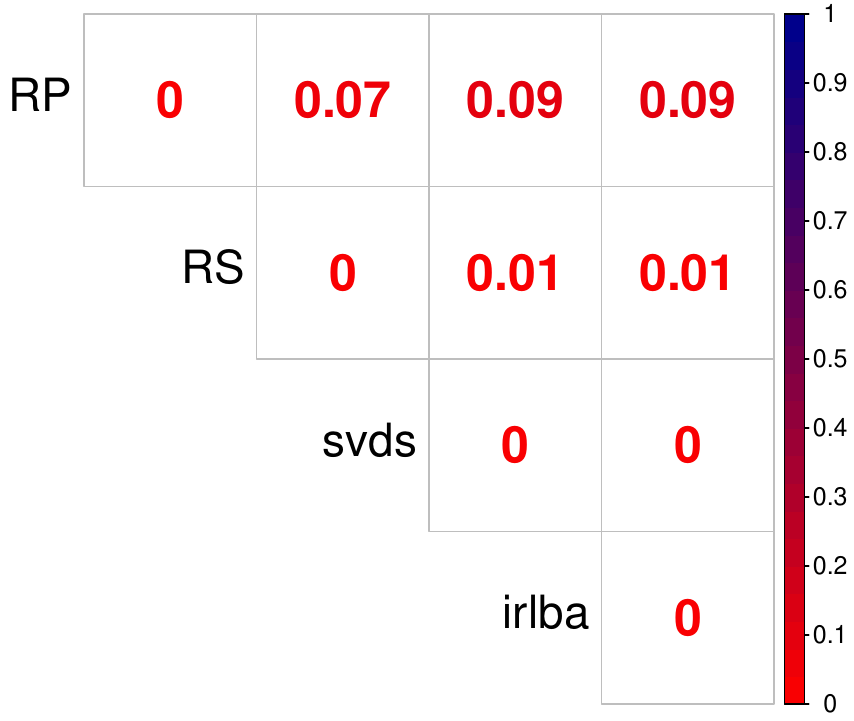}}
\subfigure[Slashdot]{\includegraphics[height=2.5cm,width=2.9cm,angle=0]{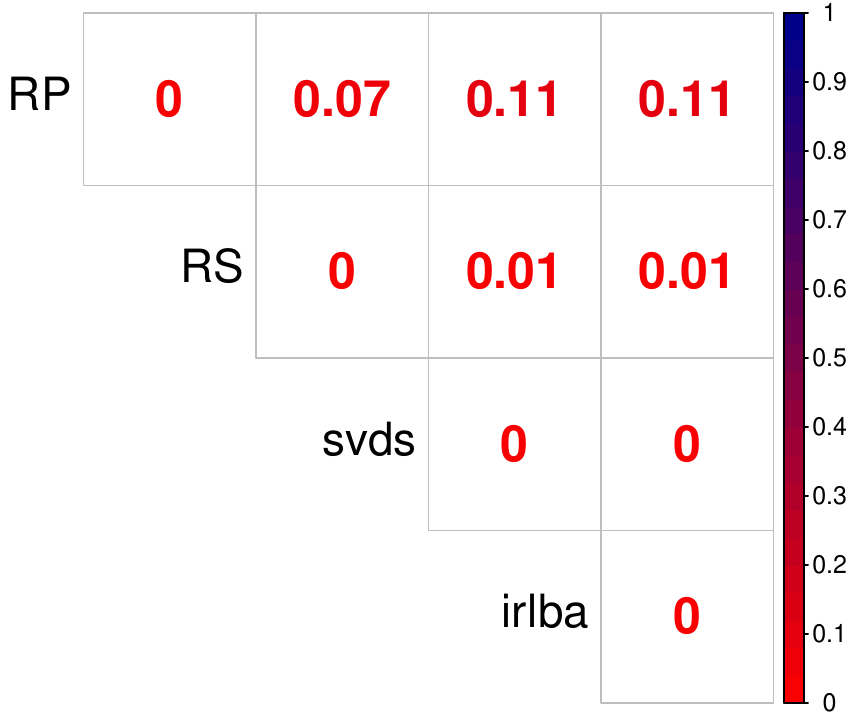}}
\subfigure[Web]{\includegraphics[height=2.5cm,width=2.9cm,angle=0]{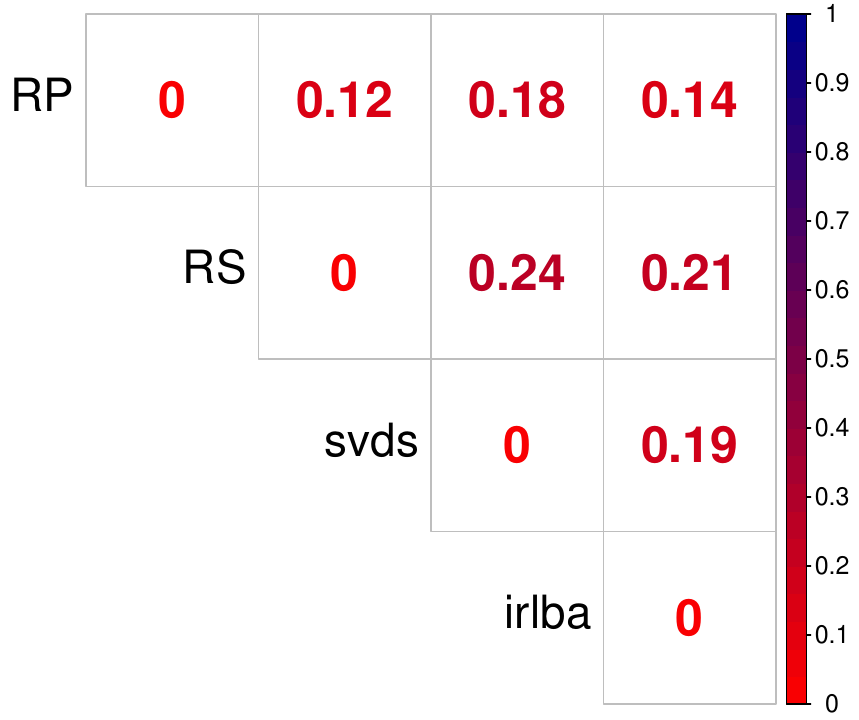}}
\subfigure[Wiki-cat]{\includegraphics[height=2.5cm,width=2.9cm,angle=0]{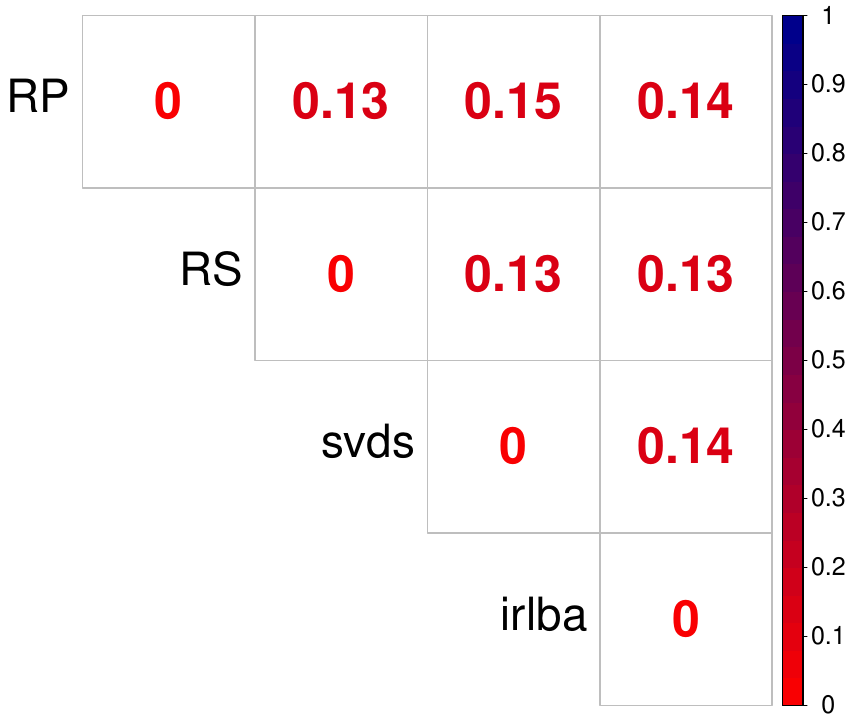}}
\subfigure[Wiki-talk]{\includegraphics[height=2.5cm,width=2.9cm,angle=0]{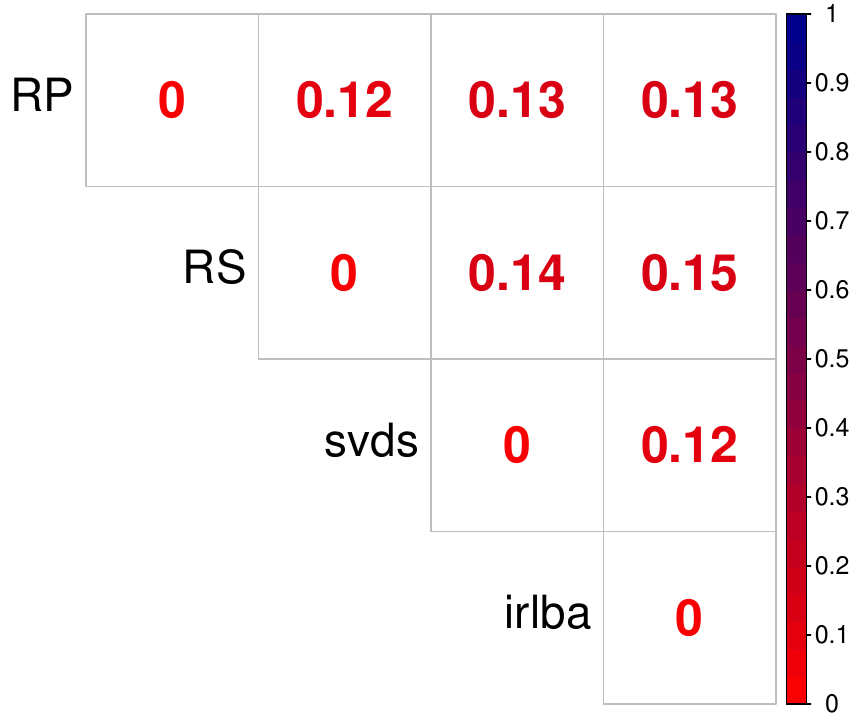}}

\caption{The standard deviations corresponding to the pairwise ARI of row clusters.}\label{comparsdrow}
\end{figure*}

\begin{figure*}[!htbp]{}
\centering
\subfigure[Epinions]{\includegraphics[height=3.2cm,width=4cm,angle=0]{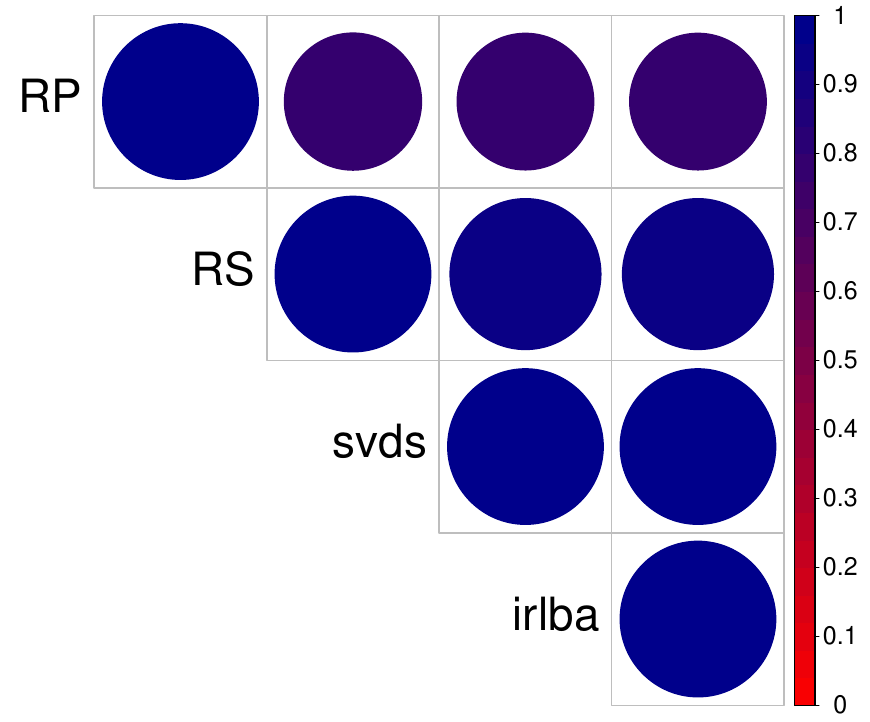}}\hspace{1cm}
\subfigure[Slashdot]{\includegraphics[height=3.2cm,width=4cm,angle=0]{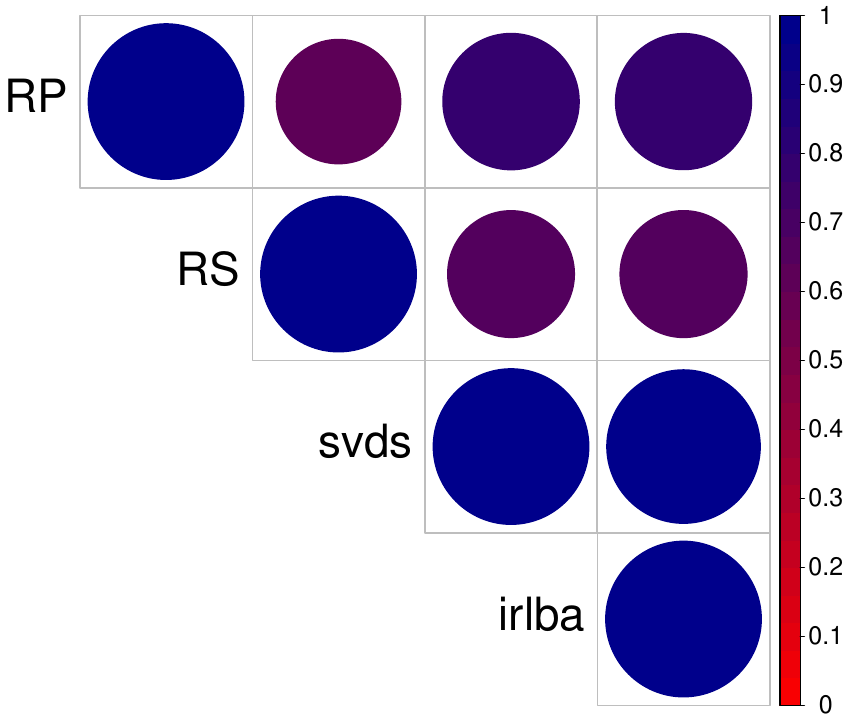}}\hspace{1cm}
\subfigure[Web]{\includegraphics[height=3.2cm,width=4cm,angle=0]{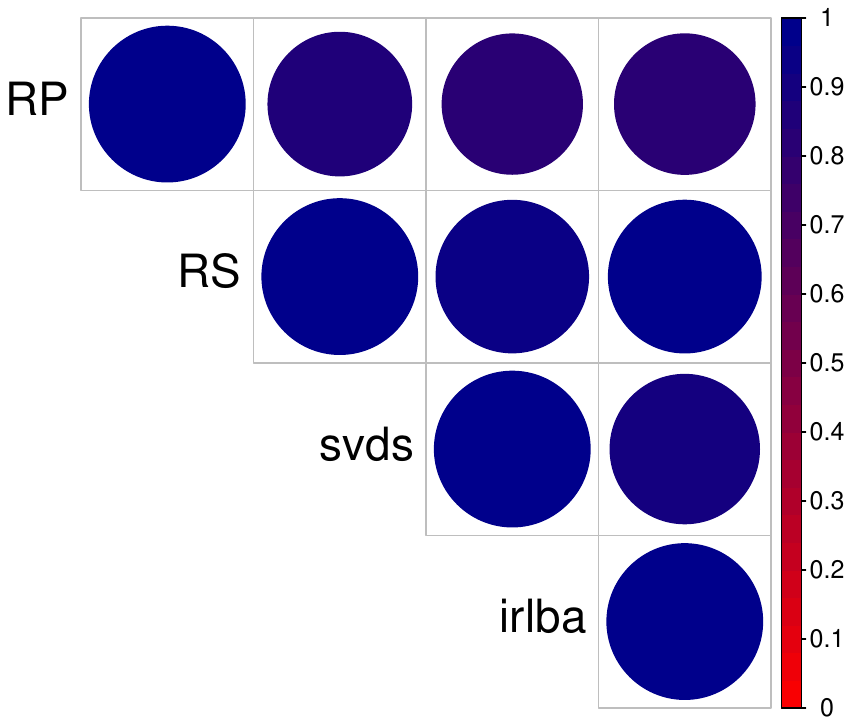}}
\subfigure[Wiki-cat]{\includegraphics[height=3.2cm,width=4cm,angle=0]{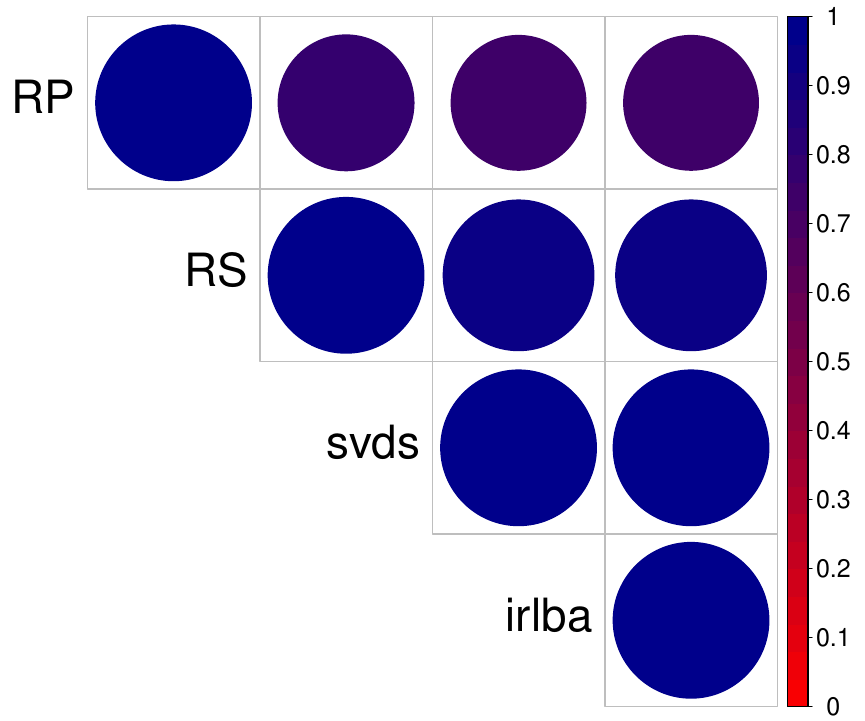}}\hspace{1cm}
\subfigure[Wiki-talk]{\includegraphics[height=3.2cm,width=4cm,angle=0]{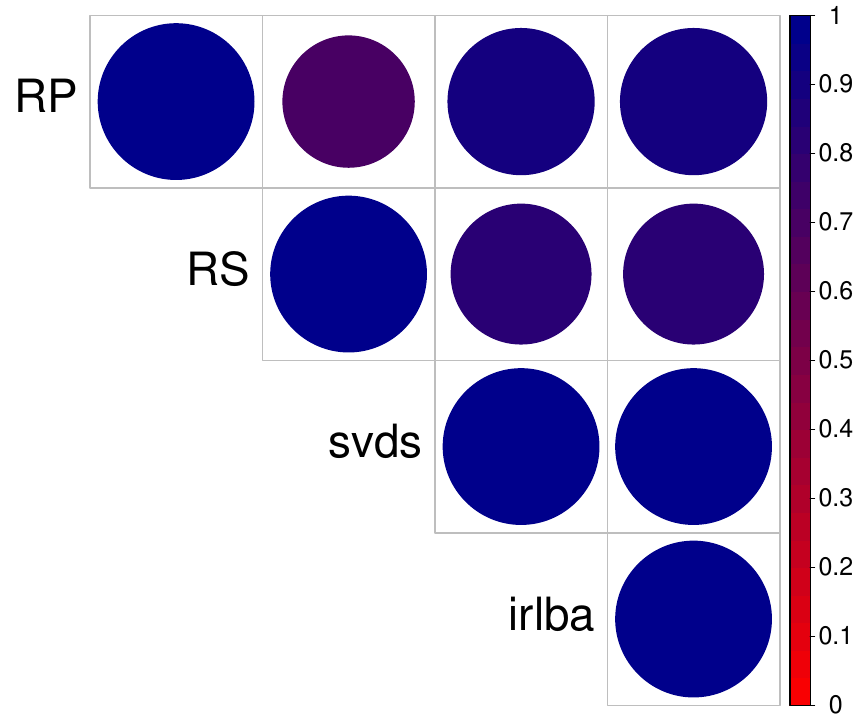}}

\caption{The pairwise comparison of the column clusters of four methods on five large-scale networks. The relative clustering performance are measured by ARI. Larger ARI, i.e., {larger circles} in the figure, indicates that the clustering results of the two associated methods are more close.}\label{compararicol}
\end{figure*}

\begin{figure*}[!htbp]{}
\centering
\subfigure[Epinions]{\includegraphics[height=2.5cm,width=2.9cm,angle=0]{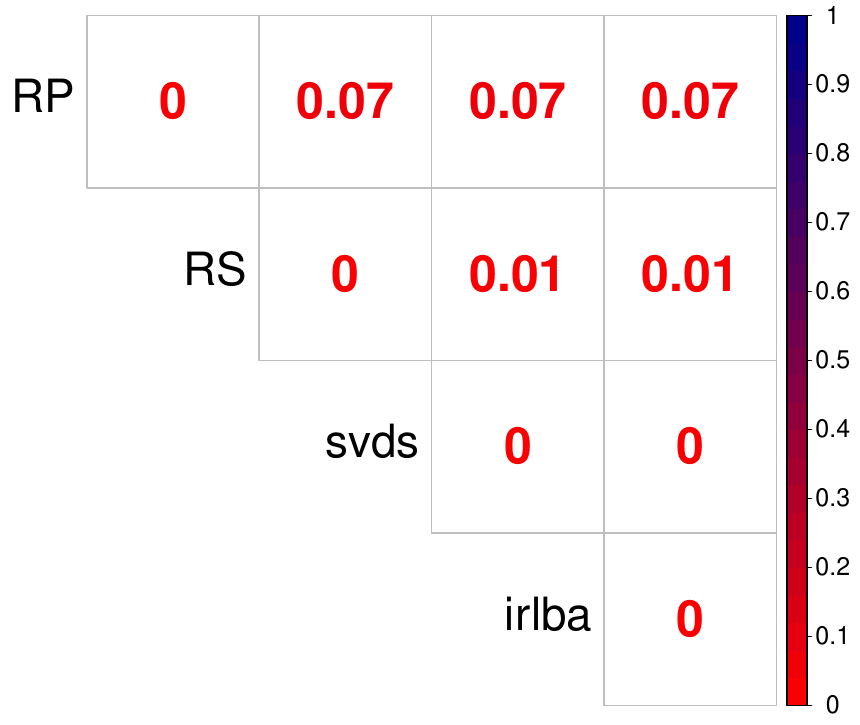}}
\subfigure[Slashdot]{\includegraphics[height=2.5cm,width=2.9cm,angle=0]{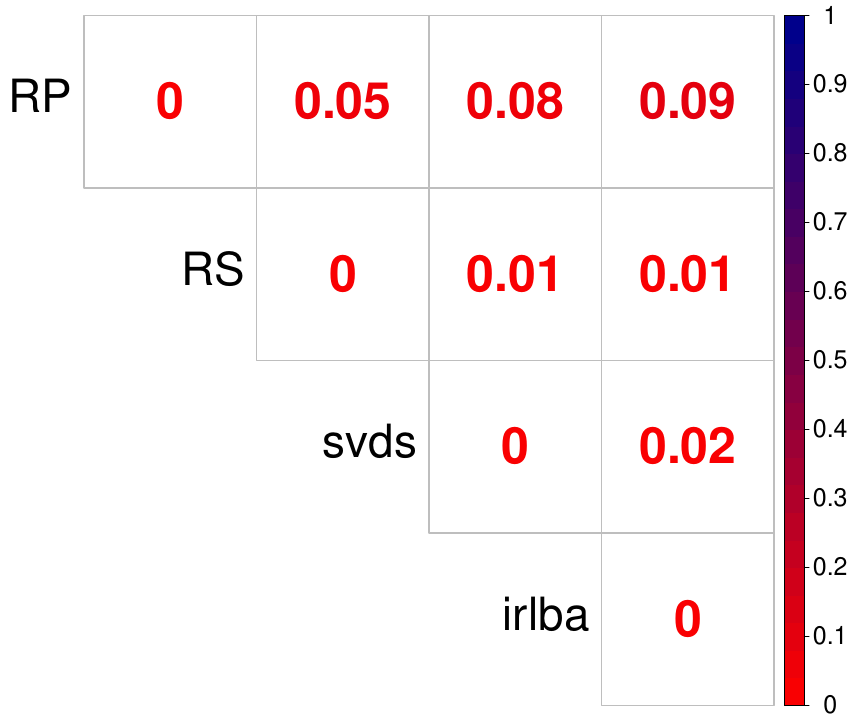}}
\subfigure[Web]{\includegraphics[height=2.5cm,width=2.9cm,angle=0]{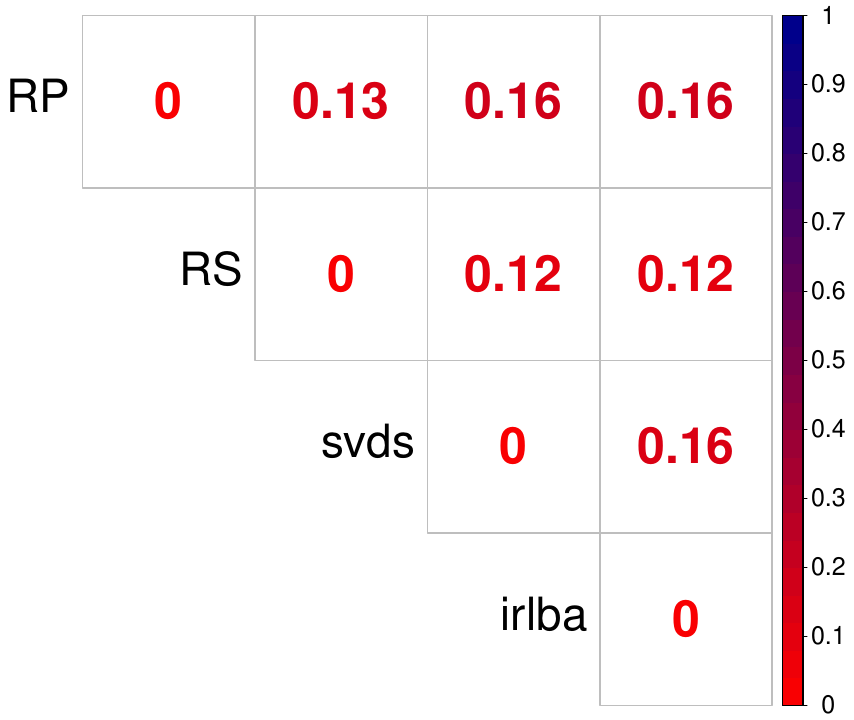}}
\subfigure[Wiki-cat]{\includegraphics[height=2.5cm,width=2.9cm,angle=0]{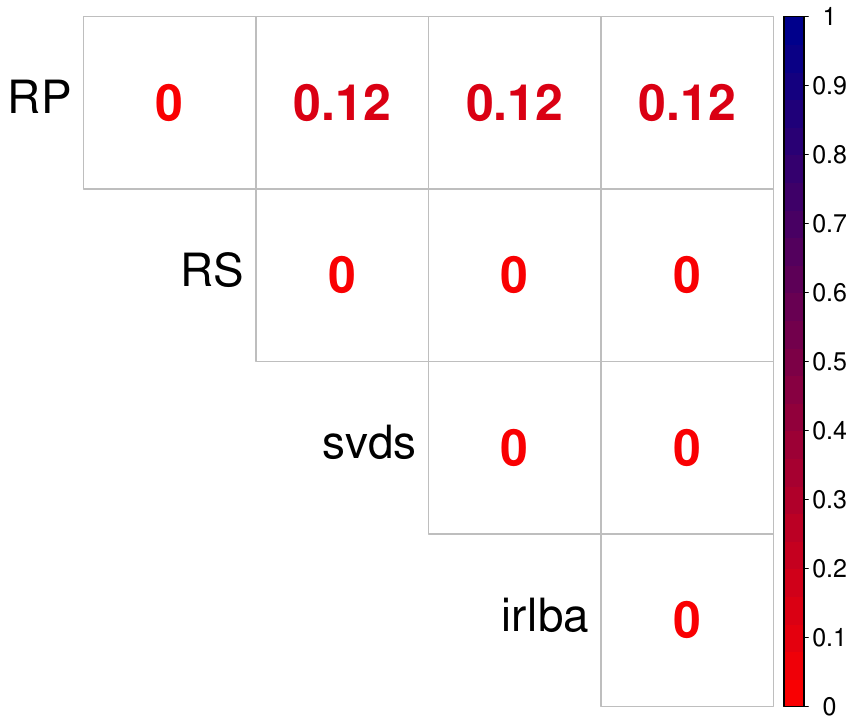}}
\subfigure[Wiki-talk]{\includegraphics[height=2.5cm,width=2.9cm,angle=0]{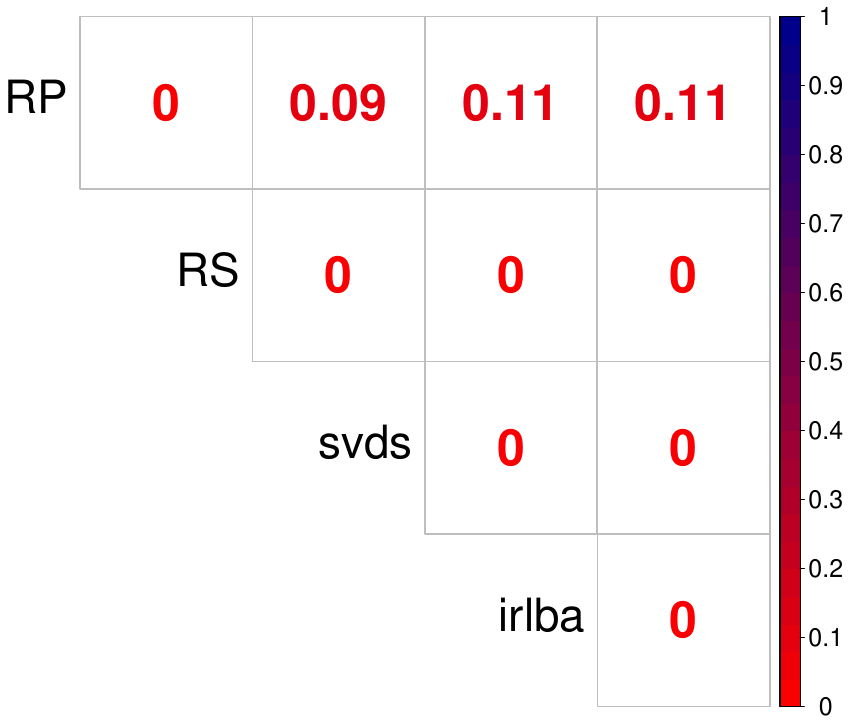}}

\caption{The standard deviations corresponding to the pairwise ARI of column clusters.}\label{comparsdcol}
\end{figure*}

Apart from computational time, we also compare the ARI between each pair of methods, evaluating their similarity of clustering results. Note that the silhouette method and other related methods for selecting the target number of clusters often fail on large-scale networks. Thus we choose the target number of sending and receiving clusters as the target rank roughly. Also note that we have provided evidence that RP-SsCC perform more similarly to SsCC than the other three SsCC-based methods (RS-SsCC, \textsf{svds} and \textsf{irlba}) on small-scale networks. Hence we here focus on comparing the SCC-based methods, denoted for short by RP, RS, \textsf{svds} and \textsf{irlba}. Figure \ref{compararirow} and \ref{compararicol} show the averaged ARI associated with sending and receiving clusters over 20 replications, respectively, and Figure \ref{comparsdrow} and \ref{comparsdcol} show their corresponding standard deviations. We observe that the randomized methods, especially RP, perform similarly to baseline iterative methods with small standard deviations.

Overall, the randomized methods show computational superiority while maintaining satisfactory clustering performance on tested real networks. In real applications, one could balance the accuracy and efficiency via changing the hyper-parameters according to the problems faced.

\begin{table*}[!htbp]
	\centering
	\footnotesize
	\caption{ A summary of the five real large-scale networks.
	}
	\def\arraystretch{1.5}
	\begin{tabular}{lccc}
		\hline
		\textbf{Data}&\textbf{No. of nodes}&\textbf{No. of edges}&\textbf{Target rank}\\
		\hline
		Epinions  social network \citep{richardson2003trust}& 75,877&508,836&3\\
		Slashdot social network \citep{leskovec2009community}&77,360&905,468&5\\
		Berkeley-Stanford web network \citep{leskovec2009community}&654,782&7,499,425&4\\
		Wikipedia top categories network\citep{yin2017local}&1,791,489&28,511,807&5\\
		Wikipedia talk network \citep{leskovec2010predicting}& 2,388,953&5,018,445&3\\
		
		\hline
	\end{tabular}
	\label{tablenetwork}
\end{table*}

\begin{table*}[]
\centering
\footnotesize
\caption{ Median time (milliseconds) of each method for computing the SVD of the corresponding adjacency matrix of five real networks over 20 replications, where for RS, the time with the sampling time included and excluded (shown in the parentheses) are reported, respectively.
}
\def\arraystretch{1.5}
\begin{tabular}{lcccc}
\hline
{Data}&{RP}&{RS}&\textsf{svds}&\textsf{irlba}\\
\hline
Epinions  social network & 29.66 &  86.54(81.28) &  69.58 & 90.54\\
Slashdot social network &  53.75 & 125.62(117.03) &116.68  & 126.95\\
Berkeley-Stanford web network & 454.18 & 999.40(926.08) & 1014.76  &1051.21 \\
Wikipedia top categories network &  2836.68&4727.99(4432.79)&4250.49&5125.12\\
Wikipedia talk network & 1163.52 &1717.09(1663.84) & 1881.16 & 1701.32\\
\hline
\end{tabular}
\label{tabletime}
\end{table*}

\section{Conclusion}
\label{conclusion}
In this paper, we studied how randomization can be used to speed up the spectral co-clustering algorithms for co-clustering large-scale directed networks and how well the resulting algorithms perform under specific network models. In particular, the random-projection-based and random-sampling-based spectral co-clustering algorithms were derived. The clustering performance of these two algorithms was studied under the ScBMs and the DC-ScBMs, respectively. The theoretical bounds are high-dimensional in nature and easy to interpret. The theoretical optimality and possible extensions of the models were also discussed. We numerically compared the randomized algorithms with fast iterative methods for computing the SVD. It turns out that the randomized algorithms are faster than or comparable to the iterative methods at the same time maintaining satisfactory clustering performance. We developed a publicly available R package \textsf{RandClust} for better usability of the proposed methods.

In this work, we focused on the pure spectral clustering without regularization or other refinements. The current theoretical results might be further improved if one uses refined spectral clustering as the starting algorithm and control its time complexity simultaneously. See \citet{qin2013regularized,gao2017achieving} for example. Note that the numbers of clusters were assumed to be known in the theoretical analysis. It would be important to study the selection of target cluster numbers, especially in an efficient way. In addition, it would be interesting to generalize the current framework to bipartite networks \citep{zhou2019analysis}, multi-layer networks \citep{lei2020consistent}, etc.

\appendix
\section{Proofs for ScBMs}
This sections includes the proofs with respect to ScBMs.
\subsection{Proof of Lemma \ref{lemma1}}
Recall $\Delta_y={\rm diag}(\sqrt{n_1^y},...,\sqrt{n_{K^y}^y})$ and $\Delta_z={\rm diag}(\sqrt{n_1^z},...,\sqrt{n_{K^z}^z})$. Then we can write $P$ as
\begin{equation}
\phantomsection
\label{A.1}P=YBZ^\intercal=Y\Delta_y^{-1}\Delta_yB\Delta_z\Delta_z^{-1}Z^\intercal,\tag{A.1}
\end{equation}
where $Y\Delta_y^{-1}$ and $Z\Delta_z^{-1}$ are both {column orthogonal matrices}. Recall that the SVD of $\Delta_yB\Delta_z$ is denoted by $L_{K^y\times K^y}D_{K^y\times K^y}R^\intercal_{K^y\times K^z}$, then (\ref{A.1}) implies
\begin{equation}
\phantomsection
\label{A.2}P=YBZ^\intercal=Y\Delta_y^{-1}LDR^\intercal\Delta_z^{-1}Z^\intercal.\tag{A.2}
\end{equation}
Note that $L$, $R$, $Y\Delta_y^{-1}$ and $Z\Delta_z^{-1}$ are all orthonormal matrices and recall that the SVD of $P$ is $\bar{U}\bar{\Sigma}\bar{ V}^\intercal$, and then we have $\bar{\Sigma}=D$,
\begin{equation}
\phantomsection
\label{A.3}\bar{U}=Y\Delta_y^{-1}L,\tag{A.3}
\end{equation}
and
\begin{equation}
\phantomsection
\label{A.4}\bar{V}=Z\Delta_z^{-1}R.\tag{A.4}
\end{equation}

For $\bar{U}$, since $\Delta_y^{-1}L$ is invertible, $Y_{i\ast}=Y_{j\ast}$ if and only if $\bar{U}_{i\ast}=\bar{U}_{j\ast}$.
In addition, we can easily verify that the rows of $\Delta_y^{-1}L$ are perpendicular to each other and the $k$th row has length $\sqrt{1/n_k^y}$, therefore we have
$$\|\bar{U}_{i\ast}-\bar{U}_{j\ast}\|_2=\sqrt{(n_{g_i^y}^y)^{-1}+(n_{g_j^y}^y)^{-1}},$$
if $g_i^y\neq g_j^y$. The argument (1) follows.

For $\bar{V}$, it is obvious that $Z_{i\ast}=Z_{j\ast}$ can imply $\bar{V}_{i\ast}=\bar{V}_{j\ast}$. While if $Z_{i\ast}\neq Z_{j\ast}$, then by (\ref{A.4}), we have
$$\|\bar{V}_{i\ast}-\bar{V}_{j\ast}\|_2=\|\frac{R_{g_{i}^z\ast}}{\sqrt{n_{g_i^z}^z}}-\frac{R_{g_{j}^z\ast}}{\sqrt{n_{g_j^z}^z}}\|_2\geq \xi_n >0,$$
where the last inequality follows from assumption (\ref{C1}). The argument (2) follows.
\QEDA

\subsection{Proof of Lemma \ref{lemma2}}
In (\ref{A.4}), we already observe $\bar{V}=Z\Delta_z^{-1}R$. Note that $R$ is not invertible, to facilitate further analysis,
we reformulate $\bar{V}$ as $$\bar{V}=Z\Delta_z^{-1}R=ZB^{\intercal} \Delta_y(L^{-1})^\intercal D^{-1},$$
using $\Delta_yB\Delta_z=L_{K^y\times K^y}D_{K^y\times K^y}R^\intercal_{K^y\times K^z}$. Without loss of generality, we assume $g_i^z=k,g_j^z=l(l\neq k)$.
Then, we have
\begin{align}
\phantomsection
\label{A.5}
\|\bar{V}_{i\ast}-\bar{V}_{j\ast}\|_2&=\|(Z_{i\ast}-Z_{j\ast})B^{\intercal} \Delta_y(L^{-1})^\intercal D^{-1}\|_2\nonumber\\
&=\|(B_{\ast k}-B_{\ast l})^{\intercal} \Delta_y(L^{-1})^\intercal D^{-1}\|_2,\nonumber\\
&\geq \|B_{\ast k}-B_{\ast l}\|_2\|\Delta_y(L^{-1})^\intercal\|_{m}\| D^{-1}\|_{m},\nonumber\\
&\geq \mu_n \cdot{\underset{{k=1,...,K^y}}{{\rm min}} ({n^y_k})^{1/2}}/\sigma_n,\nonumber
\tag{A.5}
\end{align}
where $\|M\|_m:={\rm min}_{x:\|x\|_2=1}\|Mx\|_2$, and the last inequality is implied by our condition and the following facts, $\| D^{-1}\|_{m}=1/\sigma_n$ and $\|\Delta_y(L^{-1})^\intercal\|_m\geq \|\Delta_y\|_m\|L^{-1}\|_m={\underset{{k=1,...,K^y}}{{\rm min}} (n^y_k)^{1/2}}$ by the definition of $\Delta_y$ and the orthogonality of $L$. The proof is completed. \QEDA

\subsection{Proof of Theorem \ref{rproappro}}
To begin with, we notice that
\begin{align}
\phantomsection
\label{A.6}
\|{A}^{\rm rp}-P\|_2&=\|QQ^{\intercal}ATT^{\intercal}-P\|\nonumber\\
&\leq \|A-P\|_2+\|QQ^{\intercal}ATT^{\intercal}-A\|_2\nonumber \\
&=:\mathcal I_1+\mathcal I_2.
\tag{A.6}
\end{align}
In the sequel, we discuss $\mathcal I_1$ and $\mathcal I_2$, respectively.

To bound $\mathcal I_1$, namely, the deviation of adjacency matrix from its population, we use the results in \citet{lei2015consistency}. Specifically, under condition (\ref{C2}), there exists a constant $c=c(\epsilon,c_0)$ such that
\begin{align}
\phantomsection
\label{A.7}
\|A-P\|_2\leq c\sqrt{n\alpha_n}.
\tag{A.7}
\end{align}
with probability at least $1-n^{-\epsilon}$ for any $\epsilon>0$.

To bound $\mathcal I_2$, we first notice that
\begin{align}
\phantomsection
\label{A.8}
\mathcal I_2&=\|A-QQ^{\intercal}A+QQ^{\intercal}A-QQ^{\intercal}ATT^{\intercal}\|_2\nonumber \\
&\leq \|A-QQ^{\intercal}A\|_2+\|A-ATT^{\intercal}\|_2,
\tag{A.8}
\end{align}
where in the last inequality we used the facts that $\|AB\|_2\leq \|A\|_2\|B\|_2$ for any matrices $A$ and $B$, and $\|QQ^{\intercal}\|_2\leq 1$.
When $r\geq 4$, $r{\rm log r}\leq n$ and $q\geq1$, the Corollary 10.9 and Theorem 9.2 of \cite{halko2011finding} indicate that the following inequality holds with probability at least $1-6r^{-r}$,
\begin{equation}
\phantomsection
\label{A.9}\|A-QQ^{\intercal}A\|_2\leq \sigma_{K^y+1} (A)(1+11\sqrt{K^y+r}\cdot\sqrt{n})^{\frac{1}{2q+1}},\tag{A.9}
\end{equation}
where $\sigma_{K^y+1} (\cdot)$ denotes the $K+1$th largest eigenvalue of a symmetric matrix. In particular, by Weyl's inequality,
\begin{equation}
\phantomsection
\label{A.10}
\sigma_{K^y+1} (A)=\sigma_{K^y+1} (A)-\sigma_{K^y+1} (P)\leq \|A-P\|_2.
\tag{A.10}
\end{equation}
Hence, with probability at least $1-6r^{-r}-n^{-\epsilon}$,
\begin{equation}
\phantomsection
\label{A.11}\|A-QQ^{\intercal}A\|_2\leq c'\sqrt{n\alpha_n}(1+11\sqrt{K^y+r}\cdot\sqrt{n})^{\frac{1}{2q+1}}\leq c''\sqrt{n\alpha_n}(\sqrt{K^y+r}\cdot\sqrt{n})^{\frac{1}{2q+1}}\leq c\sqrt{n\alpha_n},\tag{A.11}
\end{equation}
where the last inequality follows from the fact that $(\sqrt{K^y+r}\cdot\sqrt{n})^{\frac{1}{2q+1}}=O(1)$ provided that $q=cn^{1/\tau}$ for any $\tau>0$ and $n$ goes to infinity, and note that we usually use $c,c',c''$ to denote constants and they may be different from place to place. Similarly, under condition (\ref{C3}), we have with probability at least $1-6 s^{-s}-n^{-\epsilon}$ that
\begin{equation}
\phantomsection
\label{A.12}\|A-ATT^{\intercal}\|_2\leq c'\sqrt{n\alpha_n}.\tag{A.12}
\end{equation}
As a result, with probability larger than $1-6 r^{-r}-6 s^{-s}-2n^{-\epsilon}$,
\begin{equation}
\phantomsection
\label{A.13}\mathcal I_2\leq c\sqrt{n\alpha_n}.\tag{A.13}
\end{equation}

Finally, combining the bounds for $\mathcal I_1$ and $\mathcal I_2$, we have with probability larger than $1-6r^{-r}-6 s^{-s}-3n^{-\epsilon}$ that
\begin{equation}
\phantomsection
\label{A.14}\|{A}^{\rm rp}-P\|_2\leq c\sqrt{n\alpha_n}.\tag{A.14}
\end{equation}
The proof is completed.
\QEDA

\subsection{Proof of Theorem \ref{rpromis}}
Generally, we will first bound the perturbation of estimated eigenvectors, and then bound the size for nodes corresponding to a large eigenvector perturbation. At last, we use Lemma \ref{lemma1} to show the remaining nodes are clustered properly. To fix ideas, we now recall and {introduce} some notation.
$\bar{U}$ and $\bar{V}$ denote the left and right $K^y$ leading eigenvectors of $P$, respectively. Accordingly, {${U}^{\rm rp}$ and ${V}^{\rm rp}$ denote the left and right $K^y$ leading eigenvectors of ${A}^{\rm rp}$. Likewise, ${\tilde{U}}^{\rm rp }:={{Y}}^{\rm rp}{{X}}_y^{\rm rp}$ and ${\tilde{V}}^{\rm rp }:={{Z}}^{\rm rp}{{X}}_z^{\rm rp}$ denote the output of the random-projection-based spectral clustering, where ${{X}}_y^{\rm rp}$ and ${{X}}_z^{\rm rp}$ denote the centriods.  Next, we discuss the performance of two types of clusters, respectively.
}

(1) The left side. First, by the modified Davis-Kahan-Wedin sine theorem \citep{o2018random} (See Lemma \ref{davis-kahan}), there exists a $K^y\times K^y$ orthogonal matrix $O$ such that,
\begin{equation}
\phantomsection
\label{A.15}
\|{U}^{\rm rp}- \bar{U}O\|_{\tiny \rm F}\leq \frac{2\sqrt{2K^y}}{\gamma_n}\|{A}^{\rm rp}-P\|_2.\tag{A.15}
\end{equation}
And note that
\begin{align}
\phantomsection
\label{A.16}
\|{\tilde{U}}^{\rm rp}- \bar{U}O\|_{\tiny \rm F}^2&=\|{\tilde{U}}^{\rm rp}-{U}^{\rm rp}+{U}^{\rm rp}-\bar{U}O\|_{\tiny \rm F}^2\nonumber \\
& \leq \|\bar{U}O-{U}^{\rm rp}\|_{\tiny \rm F}^2+\|{U}^{\rm rp}-\bar{U}O\|_{\tiny \rm F}^2 \nonumber \\
&=2\|{U}^{\rm rp}-\bar{U}O\|_{\tiny \rm F}^2,\tag{A.16}
\end{align}
where the first inequality follows because we assume that ${\tilde{U}}^{\rm rp}$ is the global solution minimum of the following $k$-means objective and $\bar{U}O$ is a feasible solution,
\begin{equation}
({{Y}}^{\rm rp}, X^{\rm rp})=\underset{{Y\in  \mathbb M_{n,K^y},X\in \mathbb R^{K^y\times K^y}}}{{\rm arg\;min}}\;\|Y X-{U}^{\rm rp}\|_{\rm \tiny F}^2.\nonumber
\end{equation}
So combining (\ref{A.16}) with (\ref{A.15}) and the bound of $\|{A}^{\rm rp}-P\|_2$ in Theorem \ref{rproappro}, we have with probability larger than $1-6 r^{-r}-6 s^{-s}-3n^{-\epsilon}$ that
\begin{align}
\phantomsection
\label{A.17}
\|{\tilde{U}}^{\rm rp}- \bar{U}O\|_{\tiny \rm F}\leq \frac{c_2 4\sqrt{K^y{n\alpha_n}}}{\gamma_n}.
\tag{A.17}
\end{align}
For notational convenience, we denote the RHS of (\ref{A.17}) as ${\rm err}(K^y,n,c_2,\alpha_n,\gamma_n)$ in what follows.

Now, we begin to bound the fraction of misclustered nodes. Recall
\begin{equation}
\phantomsection
\label{A.18}
\tau={\rm min}_{l\neq k}\;\sqrt{(n_k^y)^{-1}+(n_l^y)^{-1}},
\tag{A.18}
\end{equation}
and define
\begin{equation}
\phantomsection
\label{A.19}
M^y=\{i\in \{1,...,n\}:\; \|{\tilde{U}}_{i\ast}^{\rm rp}- (\bar{U}O)_{i\ast}\|_{\tiny \rm F}>\frac{\tau}{2}\},
\tag{A.19}
\end{equation}
where $M^y$ is actually the number of misclustered nodes up to permutations as we will see soon. By the definition of $M^y$, we can see obviously that
\begin{equation}
\phantomsection
\label{A.20}
|M^y|\leq \frac{4\|{\tilde{U}}^{\rm rp}- \bar{U}O\|_{\tiny \rm F}^2}{\tau^2}{\leq} \frac{4\cdot{\rm err}^2(K^y,n,c_1,\alpha_n,\gamma_n)}{\tau^2}.
\tag{A.20}
\end{equation}
Further,
\begin{equation}
\phantomsection
\label{A.21}
\frac{|M^y|}{n}\leq\frac{4\|{\tilde{U}}^{\rm rp}- \bar{U}O\|_{\tiny \rm F}^2}{\tau^2 n}{\leq}\frac{4\cdot{\rm err}^2(K^y,n,c_1,\alpha_n,\gamma_n)}{\tau^2 n}.
\tag{A.21}
\end{equation}

At last, we show that the nodes outside $M^y$ are correctly clustered. First, we have $|M^y|<n_k$ for any $k$ by condition (\ref{C4}). Define $T^y_k\equiv G_k^y\backslash M^y$, where $G_k^y$ denotes the set of nodes within the true cluster $k$. Then $T_k^y$ is not an empty set. Let $T^y=\cup _{k=1}^{K^y}T_k^y$. Essentially, the rows in $(\bar{U}O)_{T^y\ast}$ has a one to one correspondence with those in ${\tilde{U}}_{T^y\ast}^{\rm rp}$. On the one hand, for $i\in T_k^y$ and $j\in T_l^y$ with $l\neq k$,
${\tilde{U}}_{i\ast}^{\rm rp}\neq {\tilde{U}}_{j\ast}^{\rm rp}$, otherwise the following contradiction follows
\begin{align}
\phantomsection
\label{A.24}
\tau&\leq \|(\bar{U}O)_{i\ast}-(\bar{U}O)_{j\ast}\|_2\nonumber\\
&\leq \|(\bar{U}O)_{i\ast}-{\tilde{U}}_{i\ast}^{\rm rp}\|_2+\|(\bar{U}O)_{j\ast}-{\tilde{U}}_{j\ast}^{\rm rp}\|_2 \nonumber\\
&<\frac{\tau}{2}+\frac{\tau}{2},
\tag{A.24}
\end{align}
where the first and last inequality follows from the Lemma \ref{lemma1}(1) and the definition of $M_k^y$ in (\ref{A.19}), respectively. On the other hand, for $i,j\in T_k^y$,
${\tilde{U}}_{i\ast}^{\rm rp}= {\tilde{U}}_{j\ast}^{\rm rp}$, because otherwise ${\tilde{U}}_{T\ast}$ has more than $K^y$ distinct rows which is contradict with the fact that the output size for the left side cluster is $K^y$.

As a result, we have arrived at the conclusion (1) of Theorem \ref{rpromis}.

(2) The right side. First, follow the same lines as in (1), we have with probability larger than $1-6 r^{-r}-6 s^{-s}-3n^{-\epsilon}$ that
\begin{align}
\phantomsection
\label{A.25}
\|{\tilde{V}}^{\rm rp}- \bar{V}O'\|_{\tiny \rm F}\leq \frac{c_2 4\sqrt{K^y{n\alpha_n}}}{\gamma_n}={\rm err}(K^y,n,c_2,\alpha_n,\gamma_n),
\tag{A.25}
\end{align}
where $O'$ is an orthogonal matrix. Here we want to emphasize that $\bar{V}$, ${V}^{\rm rp}$, and ${\tilde{V}}^{\rm rp }$ are all $n\times K^y$, but the population cluster size and target cluster size are both $K^z$. This brings different performance of the right side clusters compared to that of the left side counterpart.

Now we begin to see how the fraction of misclustered nodes corresponding to the right side differs from that corresponding to left side. Denote
\begin{equation}
\phantomsection
\label{A.26}
\delta=\min_{1\leq k\neq l\leq K^z}\|\frac{R_{k\ast}}{\sqrt{n_{k}^z}}-\frac{R_{l\ast}}{\sqrt{n_{l}^z}}\|_2,
\tag{A.26}
\end{equation}
and define
\begin{equation}
\phantomsection
\label{A.27}
M^z=\{i\in \{1,...,n\}:\; \|({\tilde{V}})_{i\ast}^{\rm rp}- (\bar{V}O')_{i\ast}\|_{\tiny \rm F}>\frac{\delta}{2}\},
\tag{A.27}
\end{equation}
where $M^z$ is actually the number of misclustered nodes up to permutations as we will see soon. By the definition of $M^z$, it is easy to see that
\begin{equation}
\phantomsection
\label{A.28}
|M^z|\leq 4\frac{\|{\tilde{V}}_{i\ast}^{\rm rp}- (\bar{V}O')_{i\ast}\|_{\tiny \rm F}^2}{\delta^2}.
\tag{A.28}
\end{equation}
Moreover, we have
\begin{equation}
\phantomsection
\label{A.29}
\frac{|M^z|}{n}\leq 4\frac{\|{\tilde{V}}_{i\ast}^{\rm rp}- (\bar{V}O')_{i\ast}\|_{\tiny \rm F}^2}{\delta^2 n}.
\tag{A.29}
\end{equation}

Finally, we show that the nodes outside $M_k^z$ are correctly clustered up to some permutations. As the left side case, we have $|M_k^z|<n_k^z$ by condition (\ref{C5}). Define $T_k^z\equiv G_k^z\backslash M_k^z$. Then $T_k^z$ is not an empty set. Let $T^z=\cup _{k=1}^{K^z}T_k^z$. Then follow the same lines as those in (1) and note the results in Lemma \ref{lemma1}(2), we can easily show the rows in $(\bar{V}O')_{T^z\ast}$ has a one to one correspondence with those in ${\tilde{V}}_{T^z\ast}^{\rm rp}$. Hence the corresponding nodes are correctly clustered.

Till now, we have proved the results in Theorem \ref{rpromis}.\QEDA

\subsection{Proof of Theorem \ref{rsamappro}}
Let ${G}$ be the adjacency matrix of an Erod\"{o}s-Renyi graph with each edge probability being $0<p<1$, then it is easy to see that   ${{A}^{\rm rs}=\frac{1}{p}G\circ A}$, where $\circ$ denotes the entry-wise multiplication.
Note that
\begin{align}
\phantomsection
\label{A.30}
\|{A}^{\rm rs}-P\|_2&=\|\frac{1}{p}G\circ A-P\|_2\nonumber\\
&=\|\frac{1}{p}G\circ (A-P)+\frac{1}{p}G\circ P-P\|_2 \nonumber\\
&\leq \|\frac{1}{p}G\circ (A-P)\|_2+  \|\frac{1}{p}G\circ P-P\|_2,\nonumber\\
&=\mathcal I_1+\mathcal I_2.\nonumber\tag{A.30}
\end{align}
In the sequel, we discuss $\mathcal I_1$ and $\mathcal I_2$, respectively.

First, We bound $\mathcal I_1$ using Lemma  \ref{propsition1}, which provides the a spectral-norm bound of a random matrix with independent and bounded entries. In particular, we proceed by conditioning on $A-P\equiv W$. Write $(G\circ W)_{ij}=g_{ij}W_{ij}$, where $g_{ij}\sim {\rm Bernoulli}(p)$. By simple calculations, we have,
\begin{align}
\phantomsection
\label{A.31}
\sigma_1&:={\rm max}_i \sqrt{\mathbb E(\sum_jg_{ij}^2W_{ij}^2| W)}={\rm max}_i \sqrt{\sum_jW_{ij}^2 \mathbb E( g_{ij}^2| W)}\nonumber\\
&\leq{\rm max}_i\sqrt{p}\sqrt{\|W_{i\ast}\|_2^2}\leq \sqrt{p}\|W\|_2.\nonumber
\tag{A.31}
\end{align}
Analogously, (\ref{A.31}) also holds for $$\sigma_2:={\rm max}_j \sqrt{\mathbb E(\sum_ig_{ij}^2W_{ij}^2| W)}.$$ With these bounds, we have by Lemma \ref{propsition1} that with probability $1-n^{\nu}$, there exists constant $c(\nu)$ such that,
\begin{align}
\phantomsection
\label{A.32}
\mathcal I_1\leq \frac{1}{p} c\,{\rm max}(\sqrt{p}\|W\|_2,\,\sqrt{{\rm log}n}).
\tag{A.32}
\end{align}
Further, by the concentration bound $\|A-P\|_2$ in \citet{lei2015consistency}, we have by condition (\ref{C2}) that,
\begin{align}
\phantomsection
\label{A.33}
\|W\|_2=\|A-P\|_2\leq c' \sqrt{n\alpha_n},
\tag{A.33}
\end{align}
with probability larger than $1-n^{-\nu}$. Note that we use $c,c',c''$ to represent the generic constants and they may be different from line to line. Combining (\ref{A.33}) with (\ref{A.32}), we have with probability larger than $1-2n^{-\nu}$ that,
\begin{align}
\phantomsection
\label{A.34}
\mathcal I_1\leq c''\,{\rm max} (\sqrt{\frac{n\alpha_n}{p}},\;\frac{\sqrt{{\rm log} n}}{p}).
\tag{A.34}
\end{align}

Second, we bound $\mathcal I_2$. We will use Lemma \ref{propsition2} which provides bounds on the spectral deviation of a random matrix from its expectation. Specifically, $B$ and $X$ in Lemma correspond to $P$ and $\frac{1}{p}G\circ P$ in our case. It is easy to see that $\mathbb E(X)=B$ and ${\rm max}_{jk}|X_{jk}|\leq \alpha_n/p$. Moreover, we have $${\rm Var}X_{jk}\leq P_{jk}^2/p,$$ and
\begin{align}
\phantomsection
\label{A.35}
\mathbb E(X_{jk}-P_{jk})^4&\leq {\rm Var}X_{jk}\cdot\|X_{jk}-P_{jk}\|_{\infty}^2\nonumber\\
&\leq \frac{P_{jk}^2}{p}\cdot {\rm max} \big(P_{jk},\,\frac{P_{jk}}{p}-P_{jk}\big)^2\nonumber\\
&=\frac{P_{jk}^4}{p}\cdot {\rm max} \big(1,\,(\frac{1}{p}-1)\big)^2.
\tag{A.35}
\end{align}
Therefore, by Lemma \ref{propsition2} and the fact that $P_{ij}\leq \alpha_n$, we have
\begin{align}
\phantomsection
\label{A.36}
\mathcal I_2&\leq c\Big(2\alpha_n\sqrt{\frac{n}{p}}+\alpha_n \frac{\sqrt{n}}{p^{1/4}}{\rm max} \big(1,\,\sqrt{\frac{1}{p}-1}\big)\Big)\nonumber\\
&\leq c'\sqrt{\frac{n\alpha_n^2}{p}}\Big(1+p^{1/4}\cdot {\rm max} \big(1,\,\sqrt{\frac{1}{p}-1}\big)\Big ), \nonumber
\tag{A.36}
\end{align}
with probability larger than $1-{\rm exp}\Big(-c''np(1+p^{1/4}\cdot {\rm max} \big(1,\,\sqrt{\frac{1}{p}-1}\big)^2\Big)$.

Finally, combining (\ref{A.36}) with (\ref{A.34}), we will obtain the conclusion in Theorem \ref{rsamappro}. \QEDA

\section{Proofs for DC-ScBMs}
This section includes the proofs with respect to DC-ScBMs.
\subsection{Proof of Lemma \ref{lemma3} }
Define $\tilde{Y}$ and $\tilde{Z}$ be normalized membership matrices such that $\tilde{Y}(i,k)=\tilde{\theta}^y_i$ if $i\in G_k^y$ and $\tilde{Y}(i,k)=0$ otherwise, and accordingly $\tilde{Z}(i,k)=\tilde{\theta}^z_i$ if $i\in G_k^z$ and $\tilde{Z}(i,k)=0$ otherwise.
Then it is easy to see $\tilde{Y}^\intercal \tilde{Y}=I $ and $\tilde{Z}^\intercal \tilde{Z}=I $. Let $\Psi^y={\rm diag} (\|\phi^y_1\|_2,...,\|\phi^y_{K^y}\|_2)$ and $\Psi^z={\rm diag} (\|\phi^z_1\|_2,...,\|\phi^z_{K^z}\|_2)$. Then after some rearrangements, we can see that
\begin{equation}
\phantomsection
\label{B.1}
{\rm diag}(\theta^y)Y=\tilde{Y}\Psi^y,\;{\quad}\;{\rm diag}(\theta^z)Z=\tilde{Z}\Psi^z.
\tag{B.1}
\end{equation}
Thus,
\begin{equation}
\phantomsection
\label{B.2}
P={\rm diag}(\theta^y)YBZ^\intercal{\rm diag}(\theta^z)=\tilde{Y}\Psi^yB\Psi^z\tilde{Z}^\intercal.
\tag{B.2}
\end{equation}
Denote the SVD of $\Psi^yB\Psi^z$ as
\begin{equation}
\phantomsection
\label{B.3}
\Psi^yB\Psi^z=H_{K^y\times K^y}D_{K^y\times K^y}J_{K^y\times K^z}^\intercal,
\tag{B.3}
\end{equation}
where $H$ and $J$ have orthonormal columns. Then, (\ref{B.2}) implies
\begin{equation}
\phantomsection
\label{B.4}
P=\tilde{Y}HDJ^{\intercal}\tilde{Z}^\intercal.
\tag{B.4}
\end{equation}
By the orthonormality of $\tilde{Y},\tilde{Z},H$ and $J$, we have
$$\bar{U}=\tilde{Y}H, \quad \bar{V}=\tilde{Z}J, \quad \bar{\Sigma}=D .$$
Specifically,
$\bar{U}_{i\ast}=\tilde{\theta}^y_iH_{k\ast}$ for $i\in G_k^y$, and $\bar{V}_{i\ast}=\tilde{\theta}^z_iJ_{k\ast}$ for $i\in G_k^z$. Since $H$ is square matrix with orthonormal columns, ${\rm cos}(\bar{U}_{i\ast},\bar{U}_{j\ast})=0$ if $g_i^y\neq g_j^y$. Thus the argument (1) follows.

Now we proceed to calculate ${\rm cos}(\bar{V}_{i\ast},\bar{V}_{j\ast})$ for $g_i^z\neq g_j^z$. Without loss of generality, we assume $g^z_i=k,g^z_j=l$. First notice that,
\begin{equation}
\phantomsection
\label{B.5}
{\rm cos}(\bar{V}_{i\ast},\bar{V}_{j\ast})=\frac{\bar{V}_{i\ast} \bar{V}_{j\ast}^\intercal }{\|\bar{V}_{i\ast}\|_2\|\bar{V}_{j\ast}\|_2}=\frac{\tilde{\theta}^z_i\tilde{\theta}^z_jJ_{k\ast} J_{l\ast}^\intercal }{\tilde{\theta}^z_i\tilde{\theta}^z_j\|J_{k\ast}\|_2\|J_{l\ast}\|_2},
\tag{B.5}
\end{equation}
where we note that $\bar{V}_{i\ast}$ and $\bar{V}_{j\ast}$ are row vectors. We will discuss the numerator and denominator of (\ref{B.5}), respectively. By (\ref{B.3}), we have
\begin{equation}
\phantomsection
\label{B.6}
J={\Psi^z {B}^\intercal \Psi^y}HD^{-1}:={\tilde{B}^\intercal}HD^{-1},
\tag{B.6}
\end{equation}
where we define $\Psi^y {B} \Psi^z:=\tilde{B}$. Therefore, we obtain
$${\rm cos}(\bar{V}_{i\ast},\bar{V}_{j\ast})={\rm cos}((\tilde{B}_{\ast k})^\intercal H \bar{\Sigma}^{-1},(\tilde{B}_{\ast l})^\intercal H\bar{\Sigma}^{-1}),$$
where we used the fact that $H$ is orthogonal and hence $H^{\intercal}=H^{-1}$. The argument (2) holds immediately. \QEDA

\subsection{Proof of Lemma \ref{lemma4}}
First, it is worth noting that for any $1\leq k\leq K^z$, $\|{(\tilde{B}_{\ast k})^\intercal (H^{-1})^\intercal \bar{\Sigma}^{-1}}\|_2>0$, which excludes the trivial case that ${\rm cos}((\tilde{B}_{\ast k})^\intercal (H^{-1})^\intercal \bar{\Sigma}^{-1},(\tilde{B}_{\ast l})^\intercal (H^{-1})^\intercal\bar{\Sigma}^{-1})=1$ for $k\neq l$, where we have denoted $g^z_i=k,g^z_j=l(l\neq k)$. In fact, by the orthogonality of $H$, the invertibility of $\bar{\Sigma}$, and our condition that $\min_k\|\tilde{B}_{\ast k}\|_2>0$, we have
$$\|{(\tilde{B}_{\ast k})^\intercal H \bar{\Sigma}^{-1}}\|_2\geq \|\tilde{B}_{\ast k}\|_2 \sigma_{\min}(H)\sigma_{\min}(\bar{\Sigma}^{-1})>0.$$

Second, we proceed to show that under provided conditions, for any $\lambda$ and any $1\leq k<l\leq K^z$, $$\|{(\tilde{B}_{\ast k})^\intercal H \bar{\Sigma}^{-1}}-\lambda{(\tilde{B}_{\ast l})^\intercal H \bar{\Sigma}^{-1}}\|_2^2>0,$$ which implies that $${\rm cos}({(\tilde{B}_{\ast k})^\intercal H \bar{\Sigma}^{-1}},{(\tilde{B}_{\ast l})^\intercal H \bar{\Sigma}^{-1}})<1.$$ In particular, we have
\begin{align}
\phantomsection
\label{B.7}
\|{(\tilde{B}_{\ast k})^\intercal H \bar{\Sigma}^{-1}}-\lambda{(\tilde{B}_{\ast l})^\intercal H \bar{\Sigma}^{-1}}\|_2^2&=\|(\tilde{B}_{\ast k}^\intercal-\lambda \tilde{B}_{\ast l}^\intercal) \cdot H\cdot \bar{\Sigma}^{-1}\|_2^2\nonumber\\
&\geq \sigma_{\min}^2(H)\sigma^2_{\min}(\bar{\Sigma}^{-1})\|\tilde{B}_{\ast k}^\intercal-\lambda \tilde{B}_{\ast l}^\intercal\|_2^2\nonumber\\
&= \sigma_{\min}^2(H)\sigma^2_{\min}(\bar{\Sigma}^{-1})\cdot(\lambda^2\|\tilde{B}_{\ast l}\|_2^2-2\lambda \tilde{B}_{\ast k}^\intercal\tilde{B}_{\ast l}+\|\tilde{B}_{\ast k}\|_2^2)\nonumber\\
&:=\sigma_{\min}^2(H)\sigma^2_{\min}(\bar{\Sigma}^{-1})\cdot(a\lambda^2+b\lambda+c),
\tag{B.7}
\end{align}
where $a:=\|\tilde{B}_{\ast l}\|_2^2$, $b:=-2 \tilde{B}_{\ast k}^\intercal\tilde{B}_{\ast l}$, and $c:=\|\tilde{B}_{\ast k}\|_2^2$. Note that the parabola of the form $a\lambda^2+b\lambda+c$ is always larger than 0 if the discriminant $b^2-4ac:=4 (\tilde{B}_{\ast k}^\intercal\tilde{B}_{\ast l})^2-4\|\tilde{B}_{\ast l}\|_2^2\|\tilde{B}_{\ast k}\|_2^2<0$, which is equivalent to our condition that ${\rm cos}(\tilde{B}_{\ast k},\tilde{B}_{\ast l})<1$.

Finally, we provide an explicit upper bound for ${\rm cos}({(\tilde{B}_{\ast k})^\intercal H \bar{\Sigma}^{-1}},{(\tilde{B}_{\ast l})^\intercal H \bar{\Sigma}^{-1}})$. Selecting $\lambda=-\frac{2a}{b}$, we observe that
\begin{align*}
\lambda^2\|\tilde{B}_{\ast l}\|_2^2-2\lambda \tilde{B}_{\ast k}^\intercal\tilde{B}_{\ast l}+\|\tilde{B}_{\ast k}\|_2^2&\geq \frac{-b^2+4ac}{4a}:=\frac{4 (-\tilde{B}_{\ast k}^\intercal\tilde{B}_{\ast l})^2+4\|\tilde{B}_{\ast l}\|_2^2\|\tilde{B}_{\ast k}\|_2^2}{4\|\tilde{B}_{\ast l}\|_2^2}\\
&=\|\tilde{B}_{\ast k}\|_2^2(-\frac{(\tilde{B}_{\ast k}^\intercal\tilde{B}_{\ast l})^2}{\|\tilde{B}_{\ast l}\|_2^2\|\tilde{B}_{\ast k}\|_2^2}+1)\\
&\geq \underline{\iota}_n^2(1-\zeta_n^2),
\end{align*}
where the last inequality is impled by our assumptions. Combining this with (\ref{B.7}), we have for any $\lambda$ that,
\begin{align}
\phantomsection
\label{B.8}
\|{(\tilde{B}_{\ast k})^\intercal H \bar{\Sigma}^{-1}}-\lambda{(\tilde{B}_{\ast l})^\intercal H \bar{\Sigma}^{-1}}\|_2^2\geq \sigma_{\min}^2(H)\sigma^2_{\min}(\bar{\Sigma}^{-1})\underline{\iota}_n^2(1-\zeta_n^2).
\tag{B.8}
\end{align}
Denote $\mathcal B_k:=(\tilde{B}_{\ast k})^\intercal H \bar{\Sigma}^{-1}$ and $\mathcal B_l:=(\tilde{B}_{\ast l})^\intercal H \bar{\Sigma}^{-1}$ and choose $\lambda=\frac{{\mathcal B}_{k}{\mathcal B}_{l}^\intercal}{\|{\mathcal B}_{l}\|_2^2}$ in the LHS of (\ref{B.8}), we thus have
$$\frac{-({\mathcal B}_{k}{\mathcal B}_{l}^\intercal)^2+\|{\mathcal B}_{l}\|_2^2\|{\mathcal B}_{k}\|_2^2}{\|{\mathcal B}_{l}\|_2^2} \geq \sigma_{\min}^2(H)\sigma^2_{\min}(\bar{\Sigma}^{-1})\underline{\iota}_n^2(1-\zeta_n^2),$$
which indicates that
$${\rm cos}({(\tilde{B}_{\ast k})^\intercal H \bar{\Sigma}^{-1}},{(\tilde{B}_{\ast l})^\intercal H \bar{\Sigma}^{-1}}):={\rm cos}({\mathcal B}_{k},{\mathcal B}_{l})\leq \sqrt{1-\frac{\sigma_{\min}^2(H)\sigma^2_{\min}(\bar{\Sigma}^{-1})\underline{\iota}_n^2(1-\zeta_n^2)}{\|{\mathcal B}_{k}\|_2^2}}.$$
At last, by our condition, $$\|{\mathcal B}_{k}\|_2^2\leq \sigma_{n}^2(H)\sigma^2_{n}(\bar{\Sigma}^{-1})\overline{\iota}_n^2.$$
Consequently, $${\rm cos}({(\tilde{B}_{\ast k})^\intercal H \bar{\Sigma}^{-1}},{(\tilde{B}_{\ast l})^\intercal H \bar{\Sigma}^{-1}})\leq \sqrt{1-\frac{\sigma_{\min}^2(H)\sigma^2_{\min}(\bar{\Sigma})\underline{\iota}_n^2(1-\zeta_n^2)}{\sigma_{n}^2(H)\sigma^2_{n}(\bar{\Sigma})\overline{\iota}_n^2}}.$$
The proof is completed. \QEDA

\subsection{Proof of Theorem \ref{rpromisdegree}}
To fix ideas, we now recall and introduce some notation.
$\bar{U}$ and $\bar{V}$ denote the left and right $K^y$ leading eigenvectors of $P$, respectively. Accordingly, ${U}^{\rm rp}$ and ${V}^{\rm rp}$ denote the left and right $K^y$ leading eigenvectors of ${A}^{\rm rp}$. Note that the rows of $\bar{U}$ and $\bar{V}$ are all non-zero, but the rows of ${U}^{\rm rp}$ and ${V}^{\rm rp}$ might be zero. Define $\bar{U}'$ and $\bar{V}'$ be the row-normalized version of $\bar{U}$ and $\bar{V}$, respectively. Define $({U}^{\rm rp})'$ and $({V}^{\rm rp})'$ be the row-normalized version of ${U}^{\rm rp}$ and ${V}^{\rm rp}$ with their zero rows remained the same. ${\tilde{U}}^{\rm rp }$ and ${\tilde{V}}^{\rm rp }$ denote the output of the randomized spherical spectral clustering, namely, the $k$-means solution of $({U}^{\rm rp})'$ and $({V}^{\rm rp})'$, respectively. In the sequel, we discuss the performance of two types of clusters, respectively.

(1) The left side. First, by the modified Davis-Kahan-Wedin sine theorem (Theorem 19 in \citet{o2018random}), there exists a $K^y\times K^y$ orthogonal matrix $O$ such that,
\begin{equation}
\phantomsection
\label{B.9}
\|{U}^{\rm rp}- \bar{U}O\|_{\tiny \rm F}\leq \frac{2\sqrt{2K^y}}{\gamma_n}\|{A}^{\rm rp}-P\|_2.\tag{B.9}
\end{equation}
Combining (\ref{B.9}) with the results in Theorem \ref{rproappro}, we have
\begin{equation}
\phantomsection
\label{B.10}
\|{U}^{\rm rp}- \bar{U}O\|_{\tiny \rm F}\leq c\frac{2\sqrt{2K^y}}{\gamma_n}\sqrt{n\alpha_n},\tag{B.10}
\end{equation}
with probability $1-6r^{-r}-6 s^{-s}-3n^{-\epsilon}$ for any $\epsilon>0$ and some constant $c>0$. Note that the constant $c$ may be different from line to line in this proof. And without loss of generality, {we will assume the orthogonal matrix $O$ is the identity matrix $I$} in the following proof.

Then, we bound $\|{\tilde{U}^{\rm rp}}-\bar{U}'\|_{\tiny \rm F}$. We first notice that for any vectors $a$ and $b$, $$\|\frac{a}{\|a\|_2}-\frac{b}{\|b\|_2}\|_2\leq 2\frac{\|a-b\|_2}{{\rm max}(\|a\|_2,\|b\|_2)}$$
holds, and for any $a=0$, $\|0-\frac{b}{\|b\|_2}\|_2\leq 2\frac{\|0-b\|_2}{\|b\|_2}$ holds trivially. Thus we have
\begin{align}
\phantomsection
\label{B.11}
\|({U}^{\rm rp})'- \bar{U}'\|_{\tiny \rm F}^2\leq c\sum_{i=1}^n\frac{\|{(\bar{U}^{\rm rp})}_{i\ast}-\bar{U}_{i\ast}\|_2^2}{\|\bar{U}_{i\ast}\|_2^2}
\leq c\frac{\|{({U}^{\rm rp})}-\bar{U}\|_{\tiny \rm F}^2}{\min_i \|{U}_{i\ast}\|_2^2}\leq c\frac{K^yn\alpha_n\kappa^y}{\gamma_n^2},
\tag{B.11}
\end{align}
where the last inequality follows from (\ref{B.10}) and the definition of $\kappa^y$ coupled with the fact that $\|\bar{U}_{i\ast}\|_2^2=|\tilde{\theta}^y_i|^2$ (see the proof of Lemma \ref{lemma3} for details). Further, since ${\tilde{U}}^{\rm rp }$ is the $k$-means solution of $({U}^{\rm rp})'$, we can obtain
\begin{align}
\phantomsection
\label{B.12}
\|{\tilde{U}}^{\rm rp }-\bar{U}'\|_{\tiny \rm F}^2\leq \|{\tilde{U}}^{\rm rp }-({U}^{\rm rp})'\|_{\tiny \rm F}^2+\|({U}^{\rm rp})'- \bar{U}'\|_{\tiny \rm F}^2\leq 2\|({U}^{\rm rp})'- \bar{U}'\|_{\tiny \rm F}^2\leq  c\frac{K^yn\alpha_n\kappa^y}{\gamma_n^2}.\tag{B.12}
\end{align}

Next, we bound the number of misclustered nodes. Define
\begin{align}
\phantomsection
\label{B.13}
M^y=\{i:\|{\tilde{U}}^{\rm rp }_{i\ast}-\bar{U}'_{i\ast}\|_2\geq \frac{1}{\sqrt{2}}\}.
\tag{B.13}
\end{align}
By the definition of $M^y$ and (\ref{B.12}), we have
\begin{align}
\phantomsection
\label{B.14}
|M^y|\leq 2\|{\tilde{U}}^{\rm rp }-\bar{U}'\|_{\tiny \rm F}^2.
\tag{B.14}
\end{align}
Combining this with (\ref{B.12}), we have
\begin{align}
\phantomsection
\label{B.15}
\frac{|M^y|}{n}\leq c\frac{K^y\alpha_n\kappa^y}{\gamma_n^2}.
\tag{B.15}
\end{align}
By condition (\ref{C9}), we know that $|M^y|<n_k^y$ for all $k$. Hence the nodes outside those indexed by $M^y$ but within each true cluster are not empty. That is, $G_k^y\cap (\{1,...,n\}\backslash M^y)\neq\emptyset$, where $G^y_k$ denotes the set of nodes within the true cluster $k$. Now we show that these nodes are clustered correctly. On the one hand, suppose $i,j\in {\{1,...,n\}\backslash M^y}$ are in different clusters, then their estimated clusters are also different. Otherwise we have
\begin{align}
\phantomsection
\label{B.16}
&\|\bar{U}'_{i\ast}-\bar{U}'_{j\ast}\|_2\leq\|\bar{U}'_{i\ast}-{\tilde{U}}^{\rm rp }_{i\ast}\|_2+\|{\tilde{U}}^{\rm rp }_{j\ast}-\bar{U}'_{j\ast}\|_2\nonumber\\
& < \sqrt{2},
\tag{B.16}
\end{align}
where the last inequality follows from the definition of $M^y$. Since $\bar{U}'_{i\ast}$ and $\bar{U}'_{j\ast}$ are normalized vectors and by Lemma \ref{lemma3} we know that they are orthogonal with each other, the LHS of (\ref{B.16}) is $\sqrt{2}$, which contradicts with the RHS of (\ref{B.16}). On the other hand, if $i,j\in {\{1,...,n\}\backslash M^y}$ are in the same cluster, then their estimated clusters are also identical. Otherwise ${\tilde{U}}^{\rm rp}$ has more than $K^y$ distinct rows, which violates the fact that the output cluster size is $K^y$.

As a result, we have arrived the conclusion in (1).

(2) The right side. Following the same proof strategy with that in (1), we can show that
\begin{equation}
\phantomsection
\label{B.17}
\|{V}^{\rm rp}- \bar{V}\|_{\tiny \rm F}\leq c\frac{2\sqrt{2K^y}}{\gamma_n}\sqrt{n\alpha_n},\tag{B.17}
\end{equation}
and
\begin{align}
\phantomsection
\label{B.18}
\|{\tilde{V}}^{\rm rp }-\bar{V}'\|_{\tiny \rm F}^2\leq \|{\tilde{V}}^{\rm rp }-({V}^{\rm rp})'\|_{\tiny \rm F}^2+\|({V}^{\rm rp})'- \bar{V}'\|_{\tiny \rm F}^2\leq 2\|({V}^{\rm rp})'- \bar{V}'\|_{\tiny \rm F}^2\leq  c\frac{K^yn\alpha_n\kappa^z}{\gamma_n^2},\tag{B.18}
\end{align}
where the last inequality follows from the fact that $\bar{V}_{i\ast}=\tilde{\theta}_i^z(\tilde{B}_{\ast g_i})^\intercal HD^{-1} $ and the definition of $\kappa^z$.

Next, we bound the number of misclustered nodes. Define
\begin{align}
\phantomsection
\label{B.19}
M^z=\{i:\|{\tilde{V}}^{\rm rp }_{i\ast}-\bar{V}'_{i\ast}\|_2\geq {\frac{\sqrt{1-\eta(P)}}{\sqrt{2}}}\}.
\tag{B.19}
\end{align}
By the definition of $M^z$,
\begin{align}
\phantomsection
\label{B.20}
\frac{|M^z|}{n}\leq c\frac{K^y\alpha_n\kappa^z}{(1-\eta(P))\gamma_n^2}.
\tag{B.20}
\end{align}
By condition (\ref{C10}), we know that $|M^z|<n_k^z$ for all $k$. Hence the nodes outside those indexed by $M^z$ but within each true cluster are not empty. That is, $G_k^z\cap (\{1,...,n\}\backslash M^z)\neq\emptyset $, where $G^z_k$ denotes the set of nodes within the true cluster $k$. Now we show that these nodes are clustered correctly. On the one hand, if $i,j\in {\{1,...,n\}\backslash M^z}$ are in different clusters, then their estimated clusters are also different. Otherwise we have
\begin{align}
\phantomsection
\label{B.21}
&\|\bar{V}'_{i\ast}-\bar{V}'_{j\ast}\|_2\leq\|\bar{V}'_{i\ast}-({\tilde{V}}^{\rm rp })_{i\ast}\|_2+\|({\tilde{V}}^{\rm rp })_{j\ast}-\bar{V}'_{j\ast}\|_2\nonumber\\
& < \sqrt{2(1-\eta(P))},
\tag{B.21}
\end{align}
where the last inequality follows from the definition of $S^z$. Since $\bar{V}'_{i\ast}$ and $\bar{V}'_{j\ast}$ are normalized vectors and by Lemma \ref{lemma3} and the definition of $\eta(P)$, we have
\begin{align}
\phantomsection
\label{B.22}
\|\bar{V}'_{i\ast}-\bar{V}'_{j\ast}\|_2=\sqrt{1+1-2{\rm cos}(\bar{V}'_{i\ast},\bar{V}'_{j\ast})}\geq {\sqrt{2(1-\eta(P))}},
\tag{B.22}
\end{align}
which contradicts with the RHS of (\ref{B.21}). On the other hand, if $i,j\in {\{1,...,n\}\backslash M^z}$ are in the same cluster, then their estimated clusters are the also identical. Otherwise ${\tilde{V}}^{\rm rp}$ has more than $K^z$ distinct rows, which contradicts the fact that the output cluster size is $K^z$.

As a result, we have arrived the conclusion in (2). \QEDA

\section{Proofs for auxiliary theorems}
\subsection{Proofs of Theorem \ref{lemmarank}}
Recall that the SVD of $\Delta_yB\Delta_z$ is $L_{K^y\times K'}D_{K'\times K'}R^\intercal_{K'\times K^z}$, and $\bar{B}=B\Delta_z$,
we thus have
$$(\Delta_yB\Delta_z)(\Delta_yB\Delta_z)^\intercal = \Delta_yB\Delta_z^2B^\intercal\Delta_y=\Delta_y\bar{B}\bar{B}^\intercal\Delta_y=LD^2L^\intercal.$$
Without loss of generality, suppose $g_i^y=k$ and $g_j^y=l$ ($l\neq k$), we then have
\begin{align}
\phantomsection
\mu_n^2\|{U}_{i\ast}-{U}_{j\ast}\|_2^2&=\sum_{k_1=1}^{K'} \mu_n^2 (\frac{L_{kk_1}}{\sqrt{n_{k}^y}}-\frac{L_{lk_1}}{\sqrt{n_{l}^y}})^2\nonumber \\
&\geq \sum_{k_1=1}^{K'} D_{k_1k_1}^2(\frac{L_{kk_1}}{\sqrt{n_{k}^y}}-\frac{L_{lk_1}}{\sqrt{n_{l}^y}})^2\nonumber \\
&=\sum_{k_1=1}^{K'} D_{k_1k_1}^2(\frac{L_{kk_1}}{\sqrt{n_{k}^y}})^2  + \sum_{k_1=1}^{K'} D_{k_1k_1}^2(\frac{L_{lk_1}}{\sqrt{n_{l}^y}})^2 -2 \sum_{k_1=1}^{K'} D_{k_1k_1}^2\frac{L_{kk_1}L_{lk_1}}{\sqrt{n_{k}^yn_{l}^y}}\nonumber\\
&=(\bar{B}\bar{B}^\intercal)_{kk}+(\bar{B}\bar{B}^\intercal)_{ll}-2(\bar{B}\bar{B}^\intercal)_{kl}\nonumber\\
&=\|\bar{B}_{k\ast}\|_2^2+\|\bar{B}_{l\ast}\|_2^2-2\bar{B}_{l\ast}(\bar{B}_{k\ast})^\intercal \nonumber\\
&= \|\bar{B}_{k\ast}-\bar{B}_{l\ast}\|_2^2\nonumber\\
&\geq \nu_n^2,
\nonumber
\end{align}
where the first and last inequalities follow from our conditions. As a result, for $\Theta_{i\ast}\neq \Theta_{j\ast}$, we obtain $\|{U}_{i\ast}-{U}_{j\ast}\|_2\geq {\frac{\nu_n}{\mu_n}}$.\QEDA

\subsection{Proofs of Theorem \ref{limits}}
The proof follows that of Theorem 3.2 in \citet{abbe2020entrywise} closely. Before moving on, we here first provide important results which would be used to prove Theorem \ref{limits}.
\begin{theorem}[Simplification of Theorem 2.1 of \citet{abbe2020entrywise}]
\label{generalcondition}
Suppose $P$ is the population adjacency matrix of ${\rm SBM}(n,a\frac{{\rm log} n}{n},b\frac{{\rm log} n}{n}, J)$. Consider a symmetric matrix $\tilde{A}$ which satisfies $\mathbb E(\tilde{A})=P$. $u_2$ and $u_2^\ast$ are eigenvectors associated with the second largest eigenvalues of $\tilde{A}$ and $P$, respectively. Note that the two non-zero eigenvalues of $P$ are $\lambda_1^\ast=(a+b){\rm log} n/2$ and $\lambda_2^\ast=(a-b){\rm log} n/2$, respectively. Define $\Delta^\ast:=(\lambda_1^\ast-\lambda_2^\ast)\wedge \lambda_1^\ast\wedge\lambda_2^\ast=(b\wedge\frac{a-b}{2}){\rm log} n$ and $\kappa:=\lambda_1^\ast/\Delta^\ast$. Suppose the following {\rm \textbf{A1-A4}} hold,
\paragraph{A1} There exists $\gamma>0$ such that $\|P\|_{2\rightarrow\infty}\leq \gamma \Delta^\ast$.
\paragraph{A2} For any $m\in[n]$, the entries in the $m$th row and column of $\tilde{A}$ are independent with others, i.e., $\{\tilde{A}_{ij},i=m \;{\rm or}\;j=m\}$ are independent of $\{\tilde{A}_{ij}:i\neq m,j\neq m\}$.
\paragraph{A3} For some $\delta_0\in (0,1)$, $\mathbb P(\|\tilde{A}-P\|_2\leq \gamma \Delta^\ast)\geq 1-\delta_0.$
\paragraph{A4} Suppose $\phi(x)$ is continuous and non-decreasing in $\mathbb R_+$ with $\phi(0)=0$, $\phi(x)/x$ is non-decreasing in $\mathbb R_+$ and $\delta_1\in(0,1)$. For any $m\in[n]$ and $w=(w_i)\in \mathbb R^n$,
$$\mathbb P\left(\|\sum_{i=1}^nw_i(\tilde{A}-P)_{i\ast}\|_2\leq \Delta^\ast\|w\|_\infty\phi\left(\frac{\|w\|_2}{\sqrt{n}\|w\|_\infty}\right)\right)\geq 1-\frac{\delta_1}{n},$$
and $32\kappa {\rm max}\{\gamma,\phi(\gamma)\}\leq 1.$\vspace{0.3cm}\\
Then with probability at least $1-\delta_0-2\delta_1$,
\begin{equation}\label{C.1}
\min_{s\in\{\pm 1\}}\|u_2-s\tilde{A}u_2^\ast /\lambda_2^\ast\|_\infty\leq \kappa(\kappa+\phi(1))(\gamma+\phi(\gamma))\|u_2^\ast\|_\infty.
\tag{C.1}
\end{equation}\QEDA
\end{theorem}
Theorem \ref{generalcondition} is a direct simplification of Theorem 2.1 in \citet{abbe2020entrywise} provided that the population matrix $P$ is the two-block model ${\rm SBM}(n,a\frac{{\rm log} n}{n},b\frac{{\rm log} n}{n}, J)$, and thus we omit its proof. In the next corollary, we show that {\rm \textbf{A1-A4}} actually hold for $\tilde{A}={A}^{\rm rs}$ under certain parameters and thus the specified bound for (\ref{C.1}) is obtained.

\begin{corollary}
\label{specifycondition}
Suppose $P$ is the population adjacency matrix of ${\rm SBM}(n,a\frac{{\rm log} n}{n},b\frac{{\rm log} n}{n}, J)$, $A$ is one realization from ${\rm SBM}(n,a\frac{{\rm log} n}{n},b\frac{{\rm log} n}{n}, J)$ and $A^{rs}$ (see \ref{2.1}) is a sparsified adjacency matrix from $A$ with the sampling probability being $p$. If $p>1/2$, then with probability larger than $1-2n^{-\nu}-{\rm exp}\Big(-c_6np\big(1+p^{1/4}\cdot {\rm max} (1,\,\sqrt{\frac{1}{p}-1})^2\big)\Big)-4n^{-4p+1}$
\begin{equation}\label{C22}
\min_{s\in\{\pm 1\}}\|u_2-s{A^{\rm rs}}u_2^\ast /\lambda_2^\ast\|_\infty\leq \frac{C}{\sqrt{n}{\rm log log}n^{p^2}},
\tag{C.2}
\end{equation}
where $u_2$ and $u_2^\ast$ are eigenvectors associated with the second largest eigenvalues of ${A}^{\rm rs}$ and $P$, respectively, $\nu$ and $c_6$ are the same with those in Theorem \ref{rsamappro}, and $C$ is some constant depending on $a,b$ and $p$.
\end{corollary}
\paragraph{Proof of Corollary \ref{specifycondition}}
We would use Theorem \ref{limits} to prove. Take $\gamma =[(b\wedge \frac{a-b}{2})\sqrt{{\rm log} n}]^{-1} c_5/p$, where $c_5$ is the same constant as that in Theorem \ref{rsamappro}, and take $\phi(x)= \frac{2a+4}{(b\wedge \frac{a-b}{2})}(1\vee ({\rm log} (1/x)))^{-1}$. Note that $\|P\|_{2\rightarrow\infty}= \frac{{\rm log} n}{\sqrt{n}}\sqrt{\frac{a^2+b^2}{2}}$, then {\rm \textbf{A1}} is satisfied when $n$ is sufficiently large. {\rm \textbf{A2}} is trivially satisfied. By Theorem \ref{rsamappro} and the assumption that $p>1/2$, {\rm \textbf{A3}} is satisfied with $\delta_0=2n^{-\nu}-{\rm exp}\Big(-c_6np\big(1+p^{1/4}\cdot {\rm max} (1,\,\sqrt{\frac{1}{p}-1})^2\big)\Big)$. Note that $pA^{\rm rs}$ can be regarded as one realization from ${\rm SBM}(n,ap\frac{{\rm log} n}{n},bp\frac{{\rm log} n}{n}, J)$ with population matrix being $pP$. Therefore, taking $\bar{p}=pa{\rm log} n/n$, $\alpha=4/a$ and $X_i=pA_{i\ast}^{\rm rs}$ in Lemma \ref{entrybound} and after rearranging, we can obtain
$$\mathbb P\left(\left|(A-A^\ast)_{i\ast}w\right|\leq \frac{(2a+4){\rm log} n}{1\vee {\rm log}(\frac{\sqrt{n}\|w\|_\infty}{\|w\|_2}) }\|w\|_\infty \right)\geq 1-2n^{-4p}.$$ As a result, {\rm \textbf{A4}} is satisfied. And the result follows from Theorem \ref{limits}.
\QEDA

\paragraph{Proof of Theorem \ref{limits}} The proof follows that of Theorem 3.2 in \citet{abbe2020entrywise} closely.

(1) As $\sqrt{a}-\sqrt{b}>\sqrt{2/p}$, we can select $\varepsilon =\varepsilon (a,b,p)>0$ such that $(\sqrt{ap}-\sqrt{bp})^2/2-\varepsilon {\rm log}(a/b)/2>1$. Let $s\in\{\pm 1\}$ be the one that minimizes $\|u_2-sA^{\rm rs}u_2^\ast /\lambda_2^\ast\|_\infty$. By Corollary \ref{specifycondition}, we have with probability $1-o(1)$ that,
$$\sqrt{n}\min_{i\in[n]}sz_i(u_2)_i\geq \sqrt{n}\min_{i\in[n]}s^2z_i(A^{\rm rs}u_2^\ast)_{i}/\lambda_2^\ast -C({\rm log log} n^{p^2})^{-1},$$
where $C$ is defined in Corollary \ref{specifycondition}. Note that $s^2=1$ and $$\sqrt{n}z_i (pA^{\rm rs}u_2^\ast)_i/\lambda_2^\ast=\frac{2}{(a-b){\rm log} n}(\sum_{i\sim j}pA_{ij}^{\rm rs}-\sum_{i\nsim j}pA_{ij}^{\rm rs}).$$
Hence, applying Lemma \ref{diffbound}, we can obtain
$$\mathbb P\left(\sqrt{n}\min_{i\in[n]}s^2z_i(A^{\rm rs}u_2^\ast)_{i}/\lambda_2^\ast\leq \frac{2\varepsilon}{p(a-b)}\right)\leq n^{-(\sqrt{ap}-\sqrt{bp})^2/2-\varepsilon {\rm log}(a/b)/2}=o(n^{-1}).$$
By the union bound, we further obtain that with probability $1-o(1)$,
$$\sqrt{n}\min_{i\in[n]}sz_i(u_2)_i\geq \frac{2\varepsilon}{p(a-b)}-o(1)\geq \frac{\varepsilon}{p(a-b)}.$$
Choosing $\eta=\frac{\varepsilon}{p(a-b)}$, we then arrive the conclusion of part (1).

(2) Fix any $\varepsilon_0>0$ and take $\eta_0=[(ap-bp){\rm log}(a/b)/2]^{-1} \varepsilon_0$. Let $s_0\in\{\pm 1\}$ be the one that minimizes $\|u_2-sA^{\rm rs}u_2^\ast /\lambda_2^\ast\|_\infty$. Let $B_n$ be the event that \ref{C2} holds. Let $C(a,b,p)$ be the constant in Corollary \ref{specifycondition}. When $n$ is large enough, we can have $C(a,b,p)\leq \eta_0 {\rm loglog}n^{p^2}$, and thus under $B_n$, we have $\|u_2-s_0A^{\rm rs}u_2^\ast /\lambda_2^\ast\|_\infty\leq \eta_0/\sqrt{n}$. For all $i\in[n]$, we have the following observations,
\begin{align*}
\{\hat{z}_i\neq s_0z_i\}\subseteq \{s_0z_i(u_2)_i\leq 0\}\subseteq B_n^c\cup \{s_0^2 z_i(A^{\rm rs}u_2^\ast)_i/\lambda_2^\ast \leq \eta_0/\sqrt{n}\}.
\end{align*}
As mentioned before, $s^2=1$ and $$z_i (pA^{\rm rs}u_2^\ast)_i/\lambda_2^\ast=\frac{2}{(a-b){\rm log} n\sqrt{n}}(\sum_{i\sim j}pA_{ij}^{\rm rs}-\sum_{i\nsim j}pA_{ij}^{\rm rs}).$$
Applying Lemma \ref{diffbound} again and using the fact that $\eta_0=[(ap-bp){\rm log}(a/b)/2]^{-1} \varepsilon_0$, we can obtain
$$\mathbb P(z_i (A^{\rm rs}u_2^\ast)_i/\lambda_2^\ast\leq \eta_0/\sqrt{n})\leq n^{-\frac{(\sqrt{ap}-\sqrt{bp})^2}{2}+\frac{\varepsilon_0}{2}}.$$
Therefore, the expectation of misclassification rate can be bounded as follows,
\begin{align*}
\mathbb E(\hat{z},z)&=\frac{1}{n}\sum_{i=1}^n \mathbb P(B_n^c \cup \{z_i (A^{\rm rs}u_2^\ast)_i/\lambda_2^\ast\leq \eta_0/\sqrt{n}\})\\
&\leq \mathbb P(B_n^c )+ \frac{1}{n}\sum_{i=1}^n \mathbb P(\{z_i (A^{\rm rs}u_2^\ast)_i/\lambda_2^\ast\leq \eta_0/\sqrt{n}\})\\
&\leq \mathbb P(B_n^c ) +n^{-\frac{(\sqrt{ap}-\sqrt{bp})^2}{2}+\frac{\varepsilon_0}{2}}.
\end{align*}
It is not hard to observe that sufficiently large $\nu$ in \ref{C2}, $\mathbb P(B_n^c)$ is of smaller order $n^{-\frac{(\sqrt{ap}-\sqrt{bp})^2}{2}}$ as $\sqrt{ap}-\sqrt{bp}\in(0,\sqrt{2}]$. Consequently, taking small enough $\varepsilon_0$, we arrive at the conclusion of part (2). \QEDA

\section{Auxiliary lemmas}
This section includes the auxiliary lemmas that are used for proving the theorems in the paper.
\begin{lemma}[Theorem 19 in \citet{o2018random}]
\label{davis-kahan}
Consider two matrices $B$ and $C$ with the same dimensions. Suppose matrix $B$ has rank $r(B)$, and denote the $j$th largest singular value of $B$ as $\sigma_j(B)$. For integer $1\leq j\leq r(B)$, suppose matrix $V$ and $V'$ consist of the first $j$ singular vectors of $B$ and $C$, respectively. Then
$${\rm sin}(V,V')\leq2\frac{\|B-C\|_2}{\sigma_j(B)-\sigma_{j+1}(B)},$$
where ${\rm sin}(V,V'):=\|VV^\intercal-V'(V')^\intercal\|_2$.
\end{lemma}\QEDA

It can be shown that $$\|VV^\intercal-V'(V')^\intercal\|_2\geq \frac{\sqrt{2}}{2}{\rm inf}_{O\in \mathbb O_j}\|V-V'O\|_2,$$ where $\mathbb O_j$ denotes the set consisting of orthogonal square matrices with dimension $j$. Hence we further have
$${\rm inf}_{O\in \mathbb O_j}\|V-V'O\|_2\leq2\sqrt{2}\frac{\|B-C\|_2}{\sigma_j(B)-\sigma_{j+1}(B)},$$ which is actually used in this paper.

\begin{lemma}[Proposition 13 in \citet{klopp2015matrix}]\label{propsition1}
Let $X$ be an $n\times n$ random matrix with each entry $X_{ij}$ being independent and bounded such that ${\rm max}_{ij}|X_{ij}|\leq \sigma$. Define\begin{align}
\phantomsection
\sigma_1={\rm max}_i \sqrt{\mathbb E\sum_jX_{ij}^2}\quad {\rm and}\quad \sigma_2={\rm max}_j \sqrt{\mathbb E\sum_iX_{ij}^2}.\nonumber
\end{align} Then, for any $\nu>0$, there exists constant $c=c(\sigma,\nu)>0$ such that,
\begin{align}
\phantomsection
\|X\|_2\leq c\, {\rm max}(\sigma_1,\sigma_2,\sqrt{{\rm log}n}),\nonumber
\end{align}
with probability larger than $1-n^{-\nu}$.
\end{lemma}\QEDA

\begin{lemma}[Corollary 4 and Theorem 5 in \citet{gittens2009error}]\label{propsition2}
Suppose $B$ is a fixed matrix, and let $X$ be a random matrix with each entry $X_{jk}$ being independent and bounded such that ${\rm max}_{jk}|X_{jk}|\leq \frac{D}{2}$ almost surely, for which $\mathbb E(X)=B$. Then for all $\delta>0$,
\begin{align}
\phantomsection
\|X-B\|_2\leq (1+\delta)\mathbb E\|B-X\|_2,\nonumber
\end{align}
with probability larger than $1-{\rm exp}^{-\delta^2(\mathbb E\|X-B\|_2)^2/4D^2}$. Further,
\begin{align}
\mathbb E\|X&-B\|_2\leq c\, \Big(\underset{j}{\rm max}\big(\sum_k{\rm Var}(X_{jk})\big)^{1/2}\nonumber\\
&+\underset{k}{\rm max}\big(\sum_j{\rm Var}(X_{jk})\big)^{1/2}+\big(\sum_{jk}\mathbb E(X_{jk}-b_{jk})^4\big)^{1/4}\Big).\nonumber
\end{align}
\end{lemma}\QEDA

\begin{lemma}[Lemma 7 in \citet{abbe2020entrywise}]\label{entrybound}
Let $w\in \mathbb R^n$ be a fixed vector, $\{X_i\}_{i=1}^n$ be independent random variables where $X_i\sim {\rm Bernoulli}(p_i)$. Suppose $\bar{p}\geq \max_i p_i$ and $\alpha\geq 0$. Then,
$$\mathbb P\left(\left|\sum_{i=1}^nw_i(X_i-\mathbb E(X_i))\right|\geq \frac{(2+\alpha \bar{p}n)}{1\vee {\rm log}(\frac{\sqrt{n}\|w\|_\infty}{\|w\|_2}) }\|w\|_\infty    \right)\leq e^{-\alpha n \bar{p}}.$$
\end{lemma}\QEDA

\begin{lemma}[Lemma 8 in \citet{abbe2020entrywise}]\label{diffbound}
Suppose $a>b$, $\{W_i\}_{i=1}^{n/2}$ are i.i.d. ${\rm Bernoulli}(\frac{a{\rm log}n}{n})$, and $\{Z_i\}_{i=1}^{n/2}$ are i.i.d. ${\rm Bernoulli}(\frac{b{\rm log}n}{n})$, independent of $\{W_i\}_{i=1}^{n/2}$. For any $\varepsilon \in \mathbb R$, the following tail bound holds,
$$\mathbb P\left(\sum_{i=1}^{n/2}W_i-\sum_{i=1}^{n/2}Z_i\leq \varepsilon {\rm log} n\right)\leq n^{-(\sqrt{a}-\sqrt{b})^2/2+\varepsilon {\rm log}(a/b)/2}.$$
\end{lemma}\QEDA

\section{Additional experimental results}
\begin{figure}[!htbp]{}
\centering
\subfigure[]{\includegraphics[height=4.1cm,width=4.3cm,angle=0]{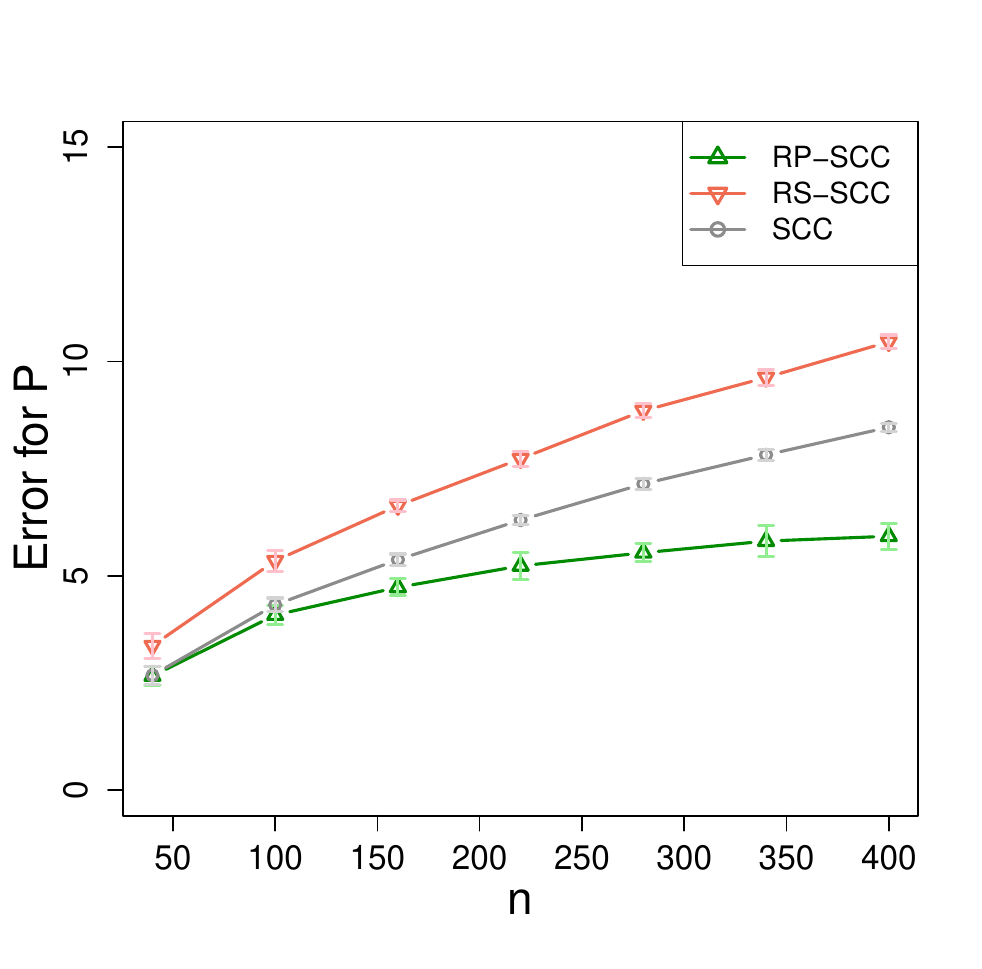}}
\subfigure[]{\includegraphics[height=4.1cm,width=4.3cm,angle=0]{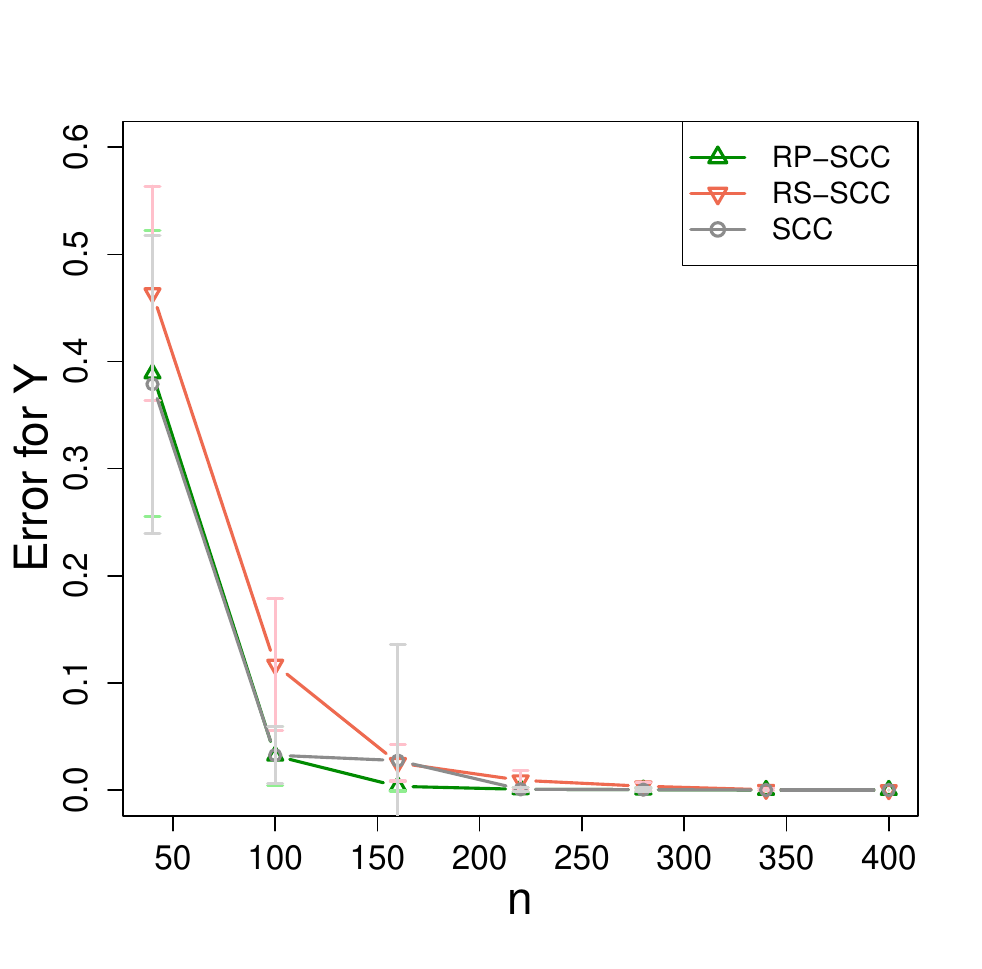}}
\subfigure[]{\includegraphics[height=4.1cm,width=4.3cm,angle=0]{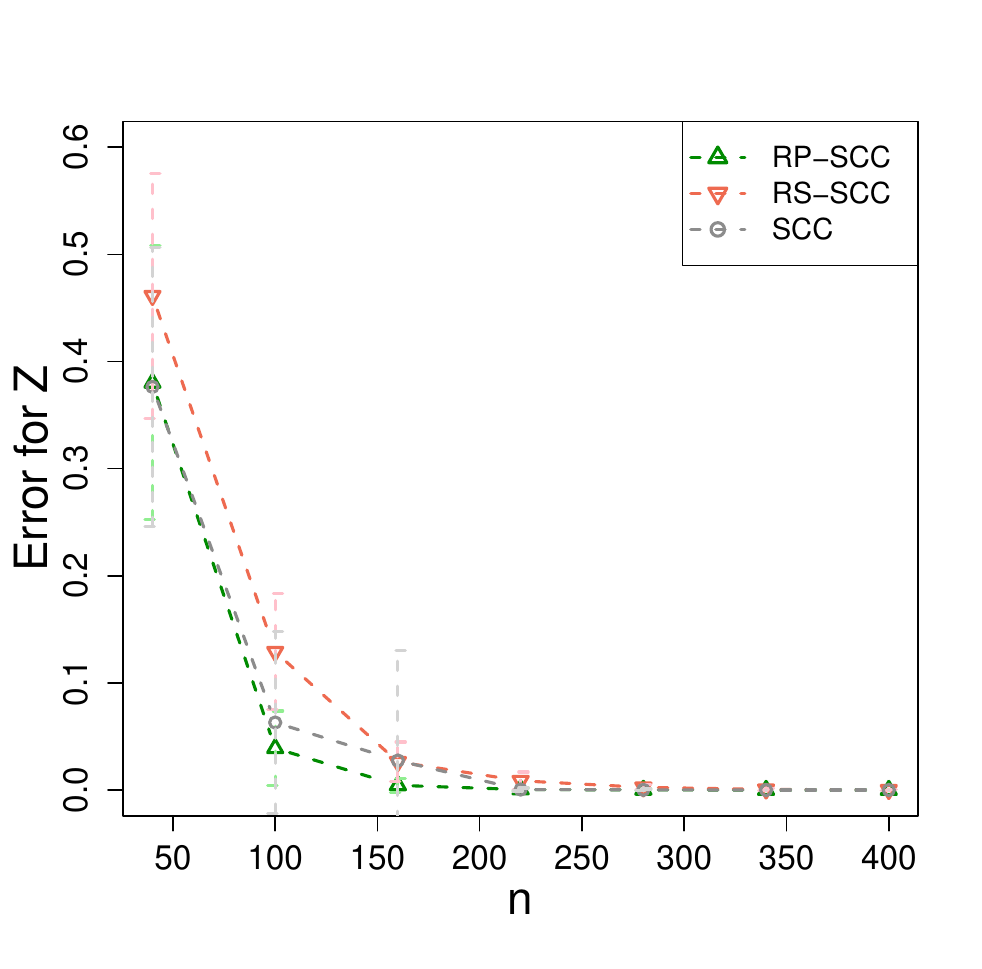}}
\caption{Simulation results of case 1 under {model set-up 2}.}\label{m2case1}
\end{figure}

\begin{figure}[!htbp]{}
\centering
\subfigure[]{\includegraphics[height=4.1cm,width=4.3cm,angle=0]{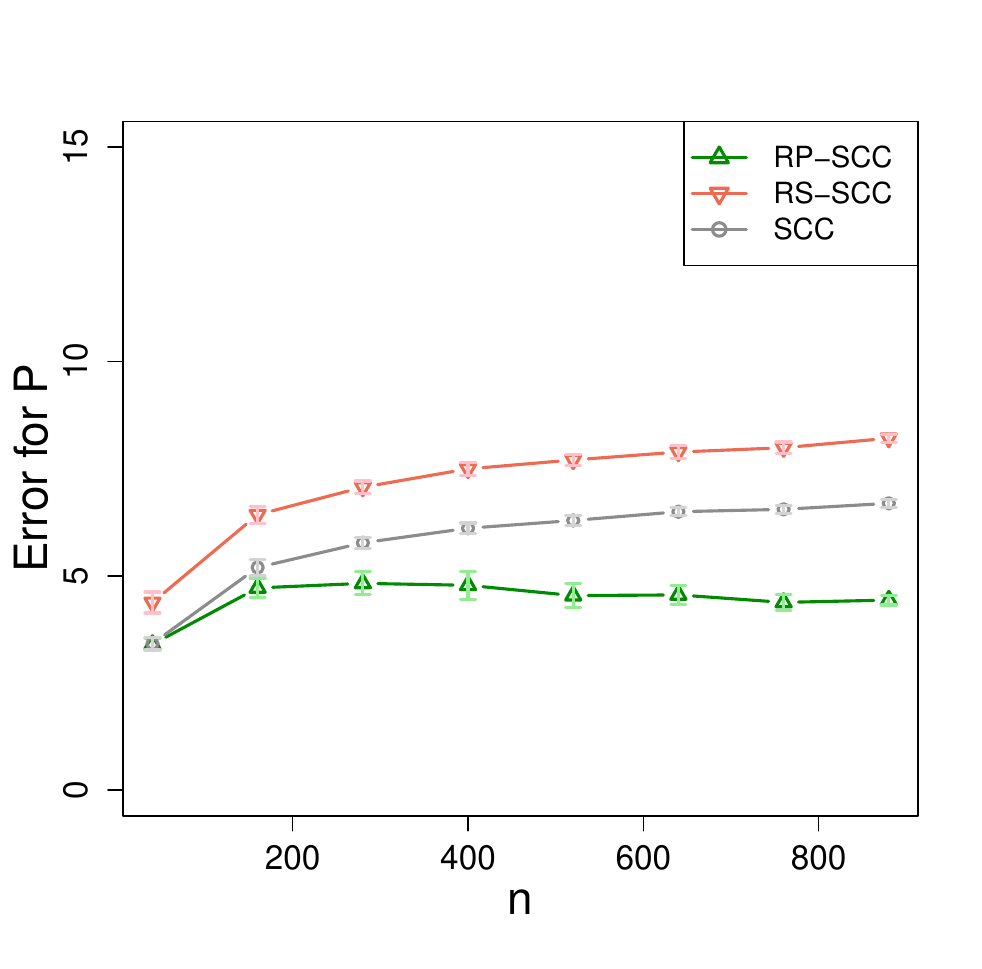}}
\subfigure[]{\includegraphics[height=4.1cm,width=4.3cm,angle=0]{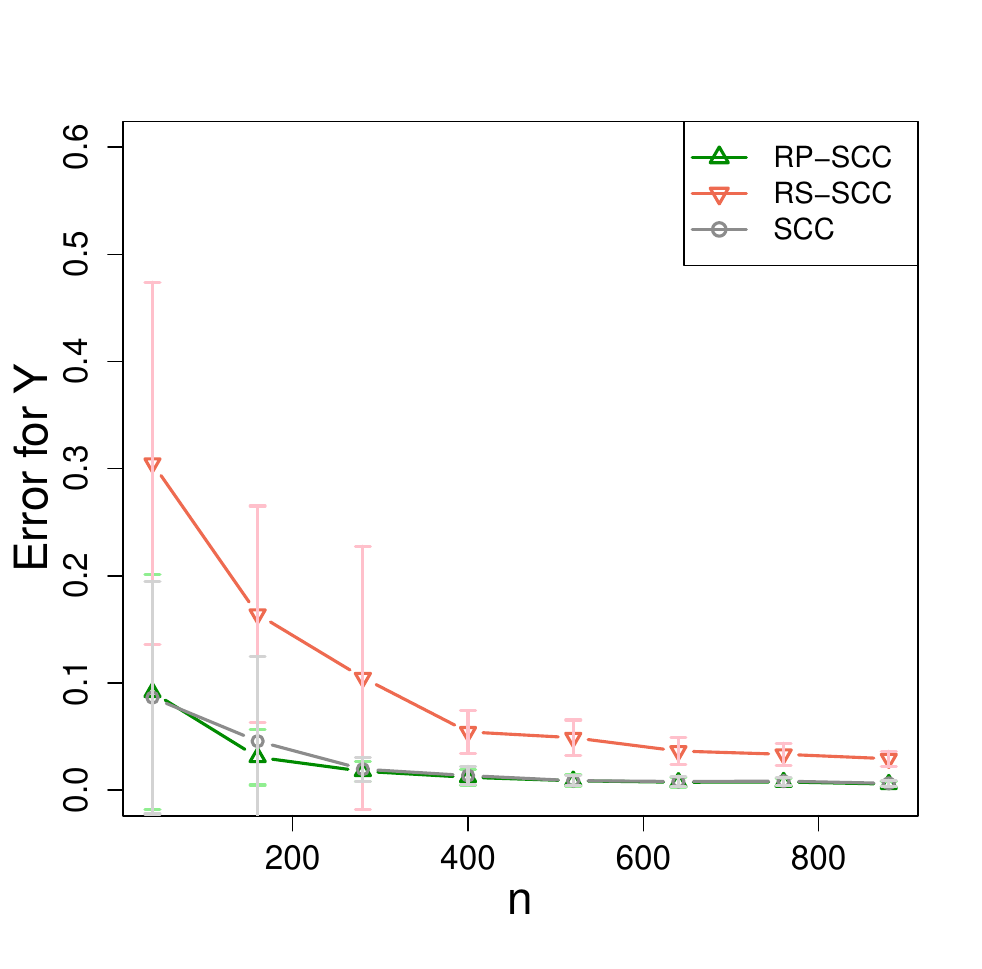}}
\subfigure[]{\includegraphics[height=4.1cm,width=4.3cm,angle=0]{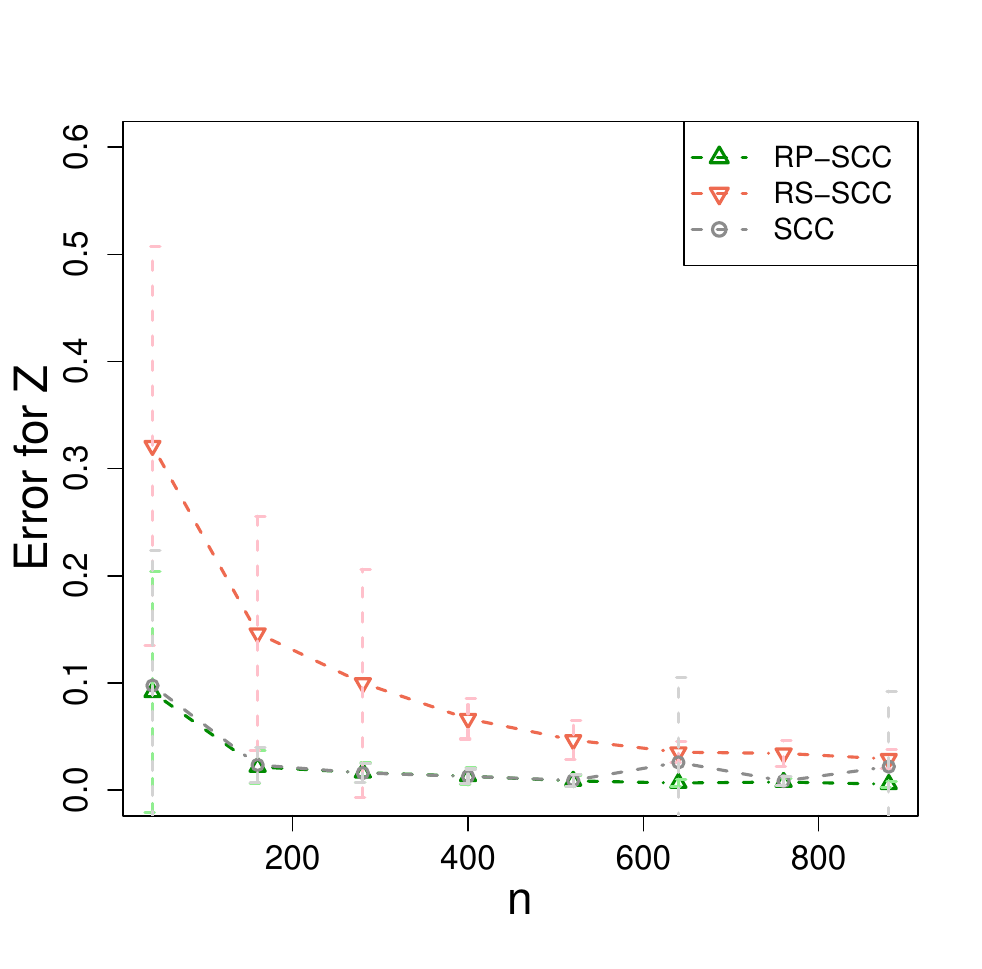}}
\caption{Simulation results of case 2 under {model set-up 2}. }\label{m2case2}
\end{figure}

\begin{figure}[!htbp]{}
\centering
\subfigure[]{\includegraphics[height=4.1cm,width=4.3cm,angle=0]{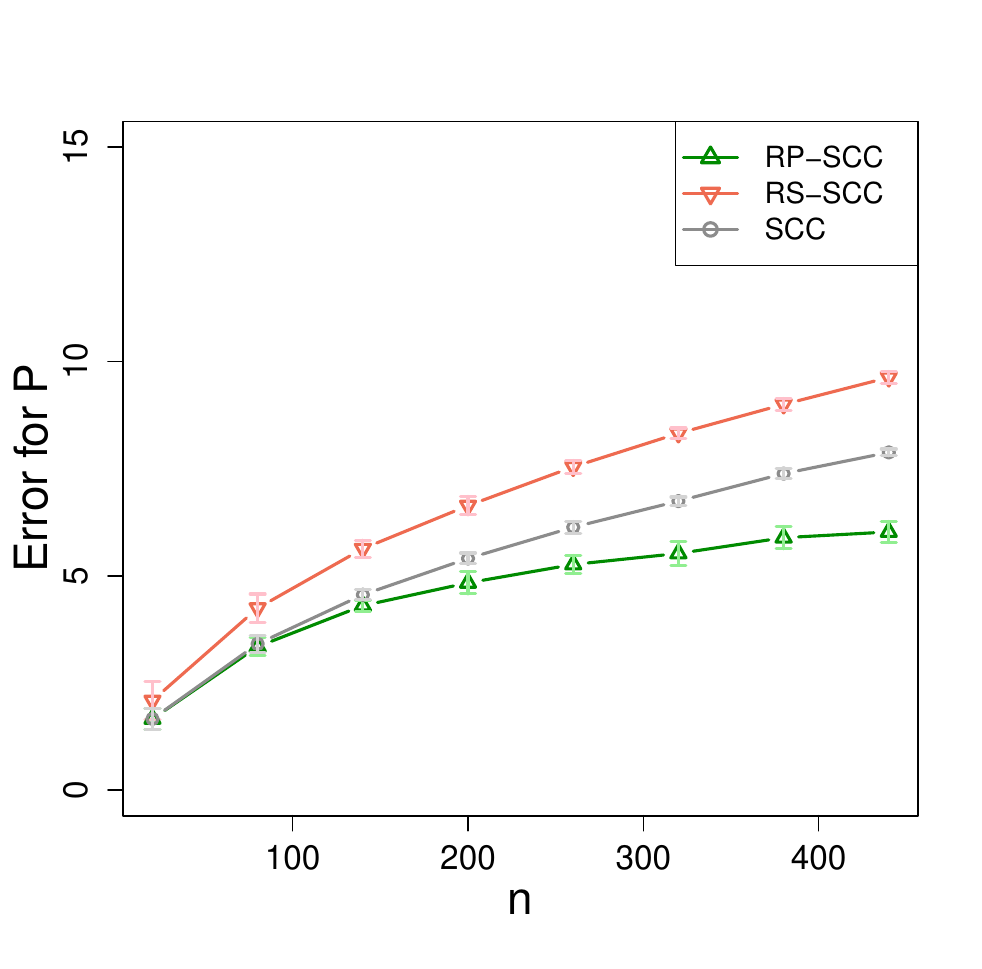}}
\subfigure[]{\includegraphics[height=4.1cm,width=4.3cm,angle=0]{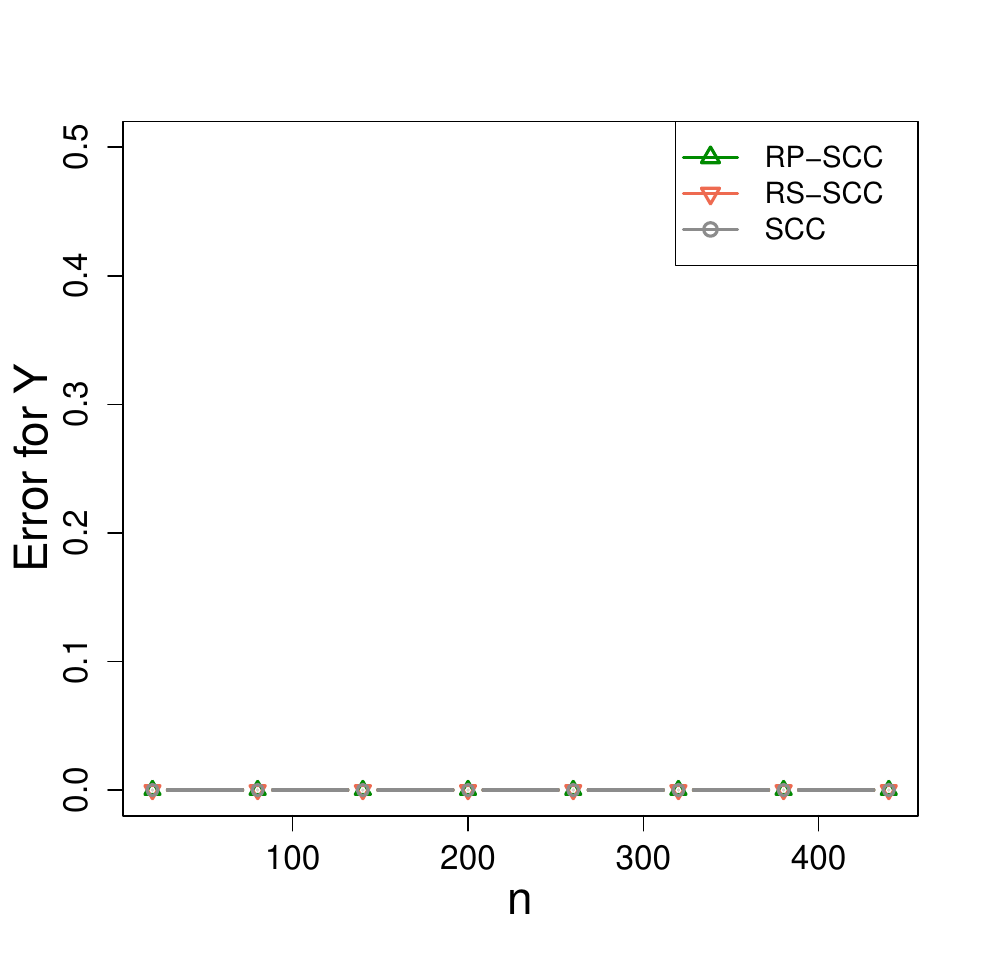}}
\subfigure[]{\includegraphics[height=4.1cm,width=4.3cm,angle=0]{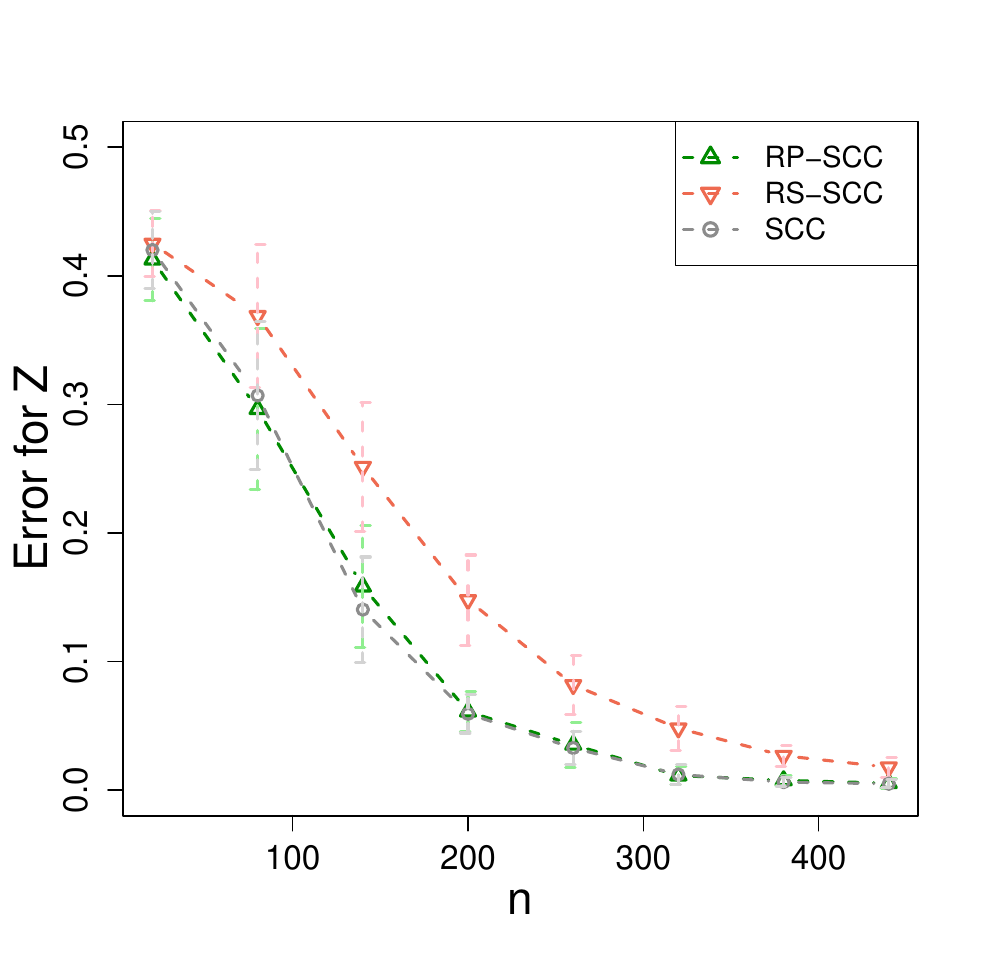}}
\caption{Simulation results of case 1 under {model set-up 3}.}\label{m3case1}
\end{figure}

\begin{figure}[!htbp]{}
\centering
\subfigure[]{\includegraphics[height=4.1cm,width=4.3cm,angle=0]{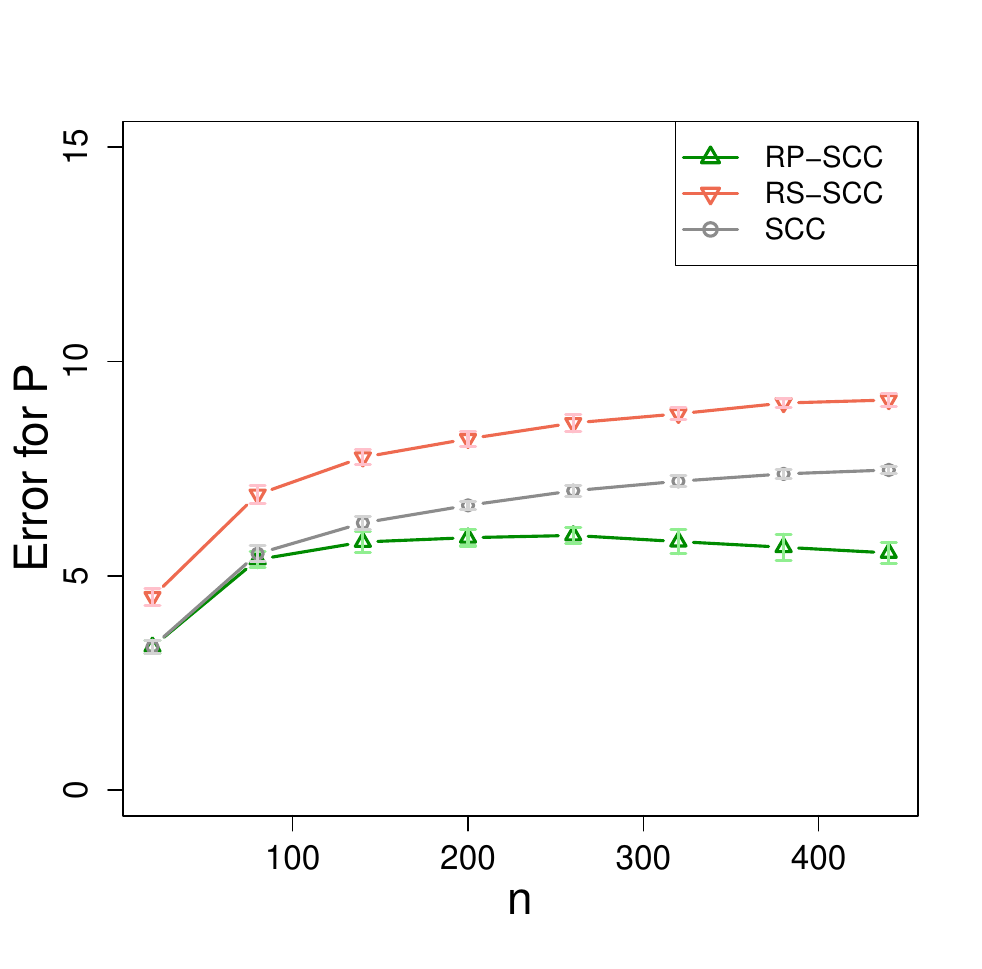}}
\subfigure[]{\includegraphics[height=4.1cm,width=4.3cm,angle=0]{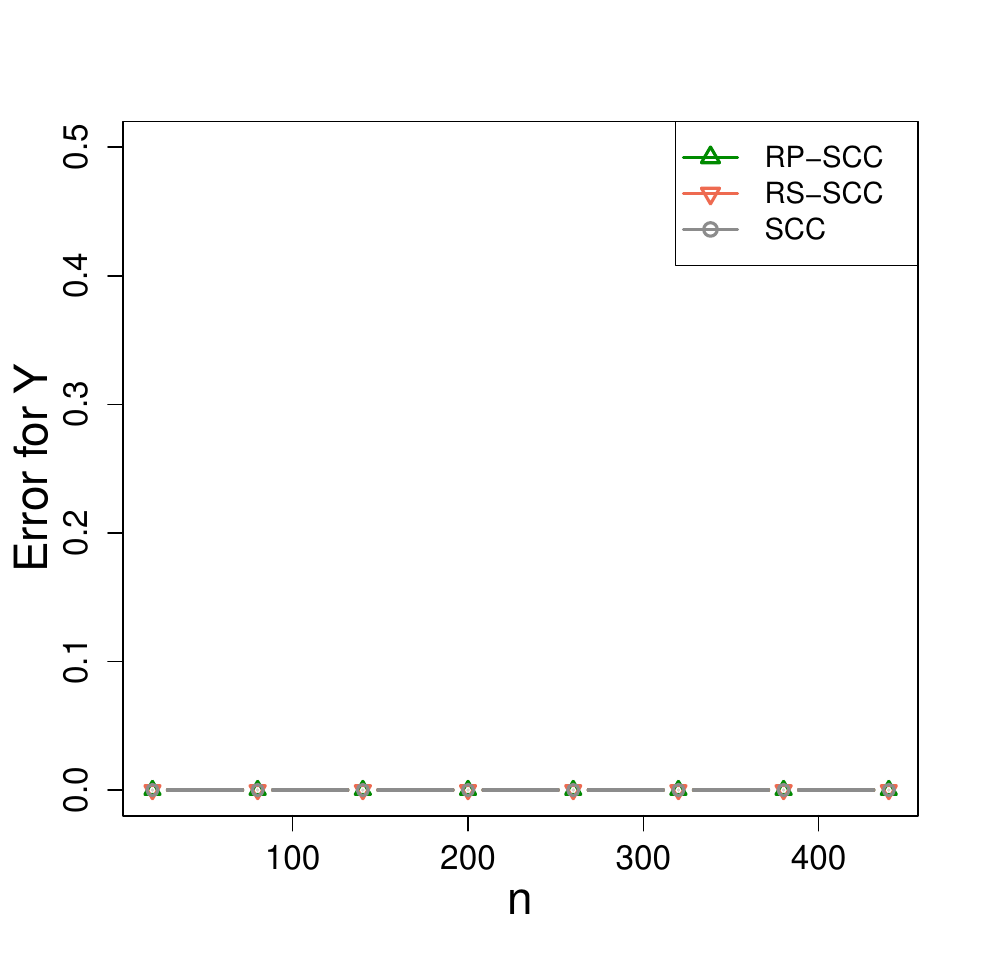}}
\subfigure[]{\includegraphics[height=4.1cm,width=4.3cm,angle=0]{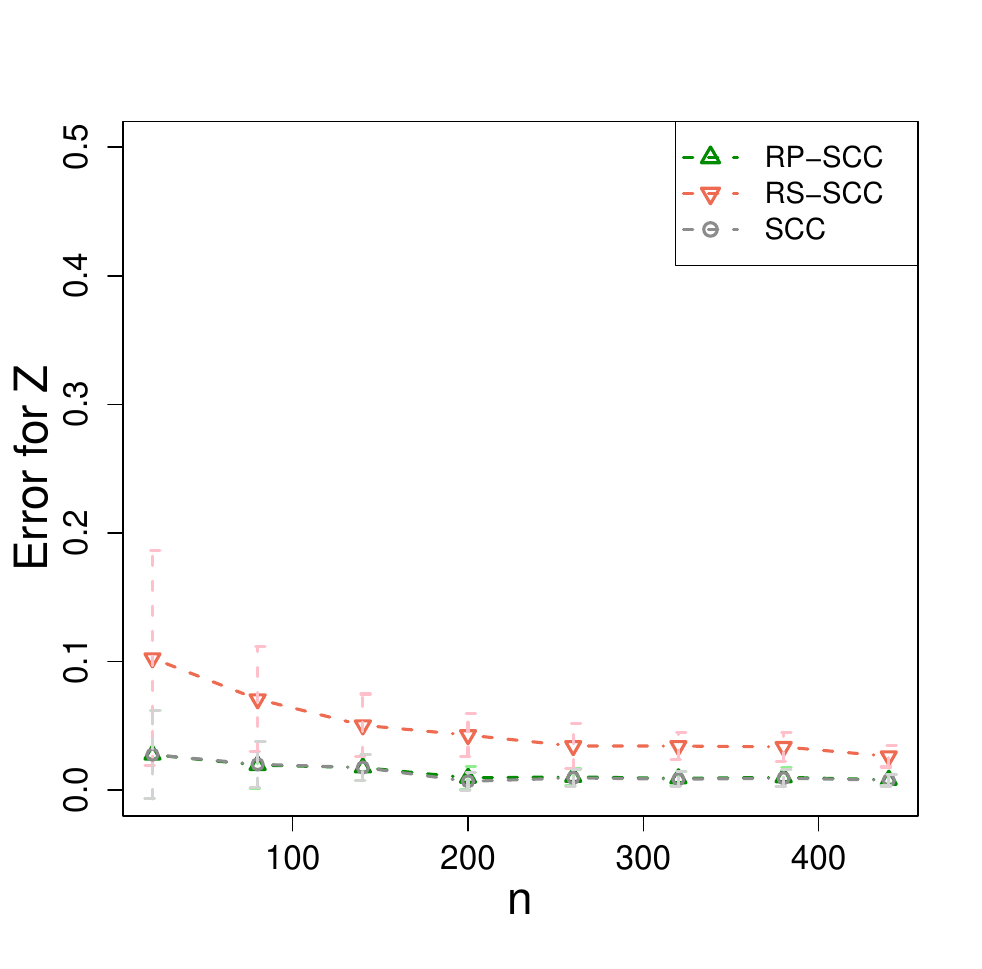}}
\caption{Simulation results of case 2 under {model set-up 3}. }\label{m3case2}
\end{figure}

\begin{figure}[!htbp]{}
\centering
\subfigure[]{\includegraphics[height=4.1cm,width=4.3cm,angle=0]{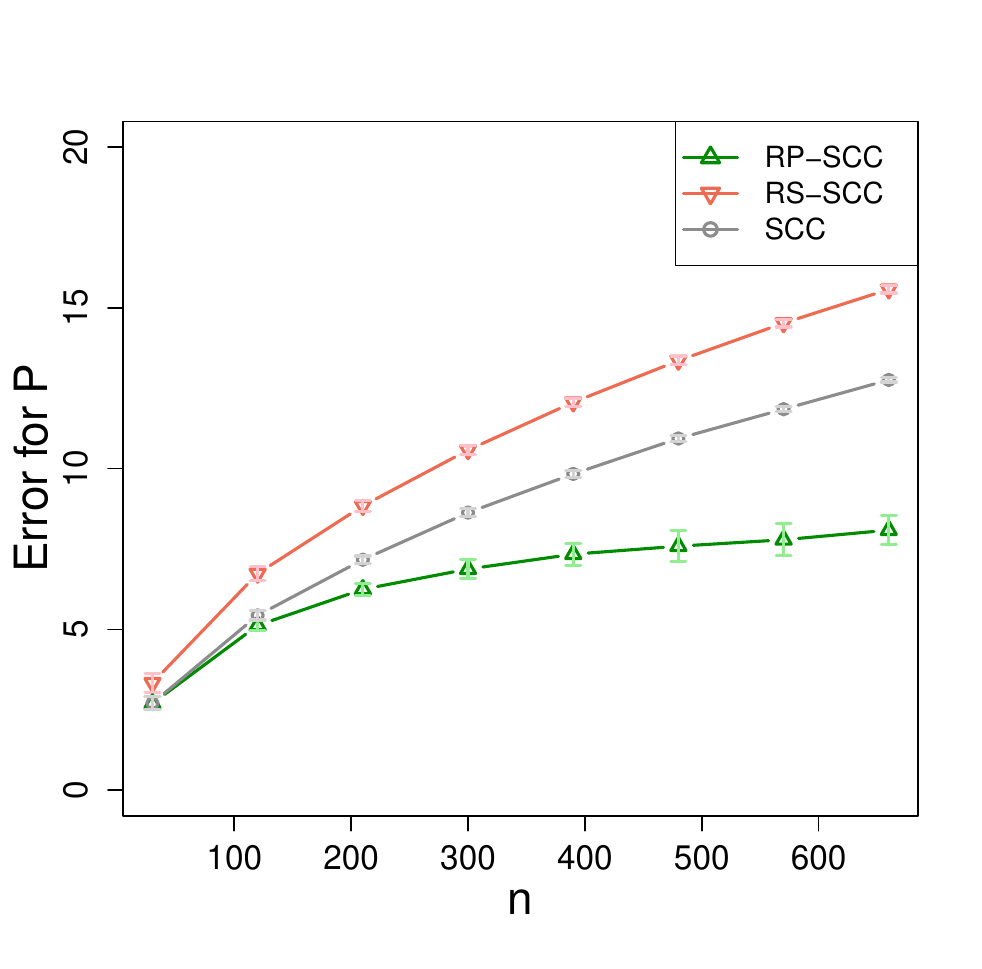}}
\subfigure[]{\includegraphics[height=4.1cm,width=4.3cm,angle=0]{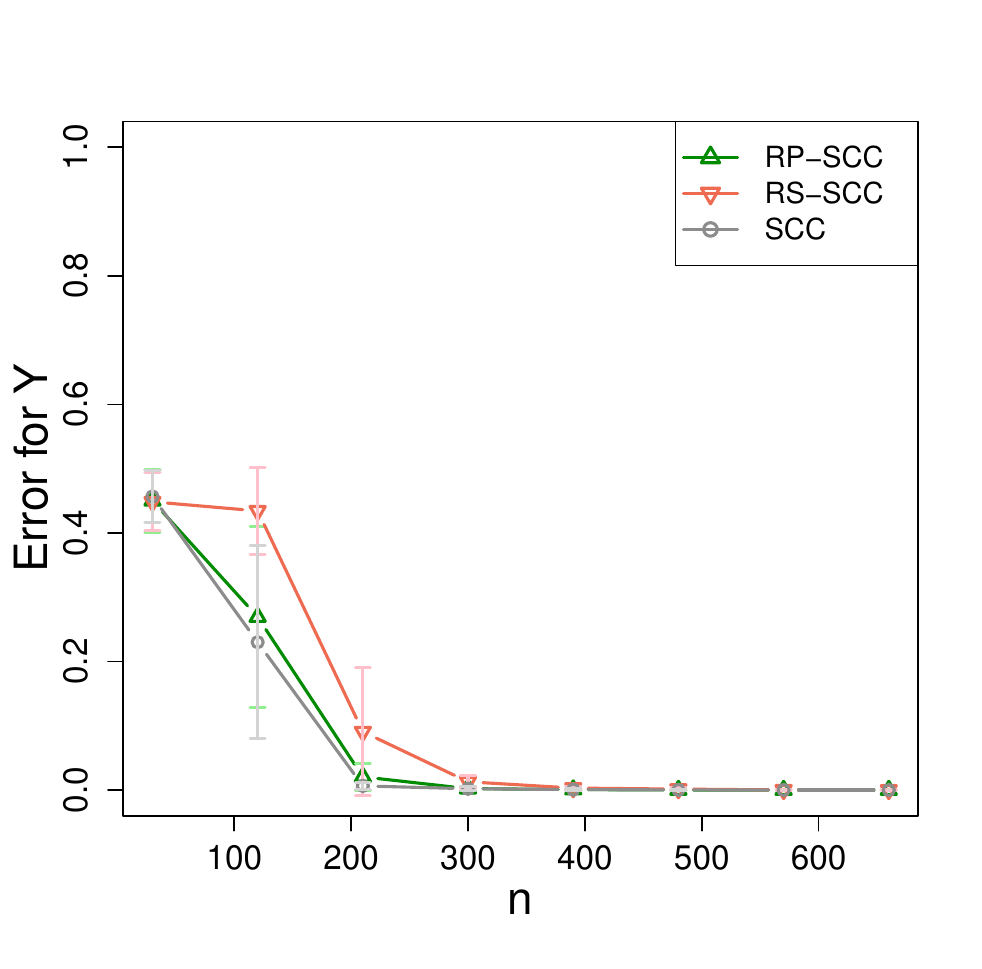}}
\subfigure[]{\includegraphics[height=4.1cm,width=4.3cm,angle=0]{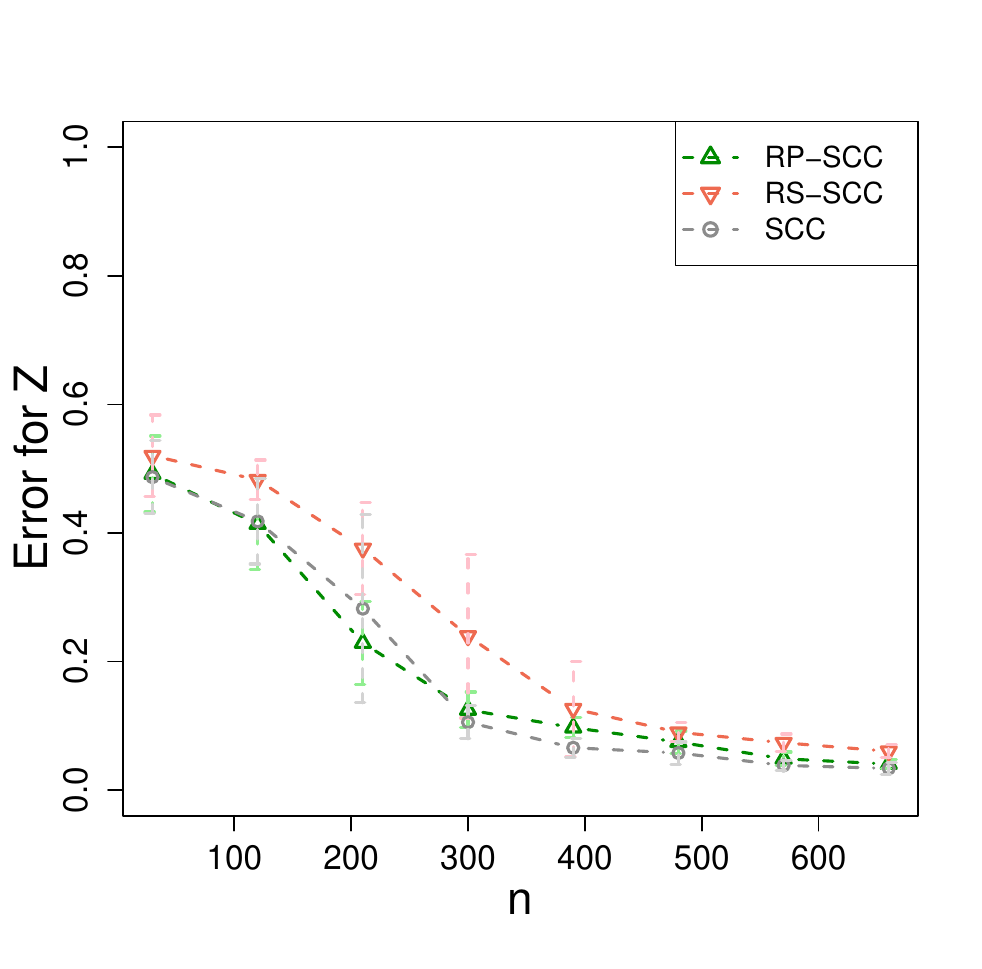}}
\caption{Simulation results of case 1 under {model set-up 4}.}\label{m4case1}
\end{figure}

\begin{figure}[!htbp]{}
\centering
\subfigure[]{\includegraphics[height=4.1cm,width=4.3cm,angle=0]{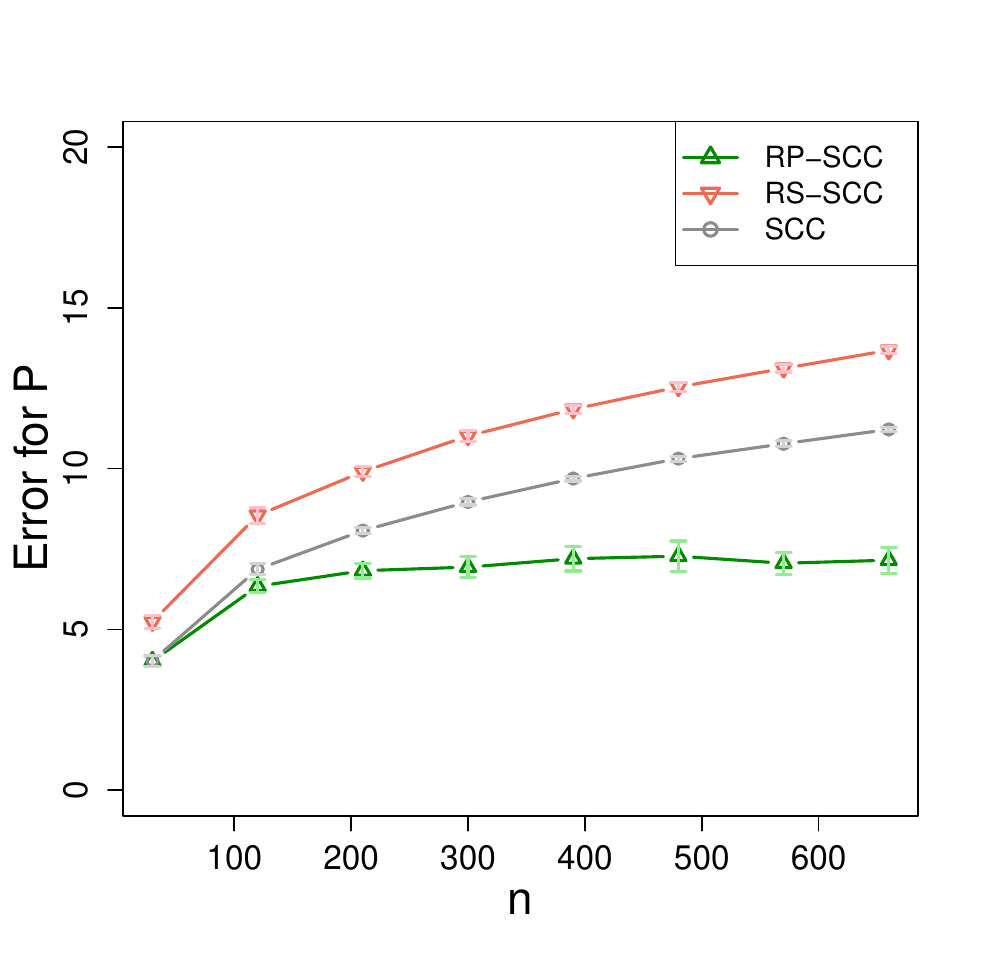}}
\subfigure[]{\includegraphics[height=4.1cm,width=4.3cm,angle=0]{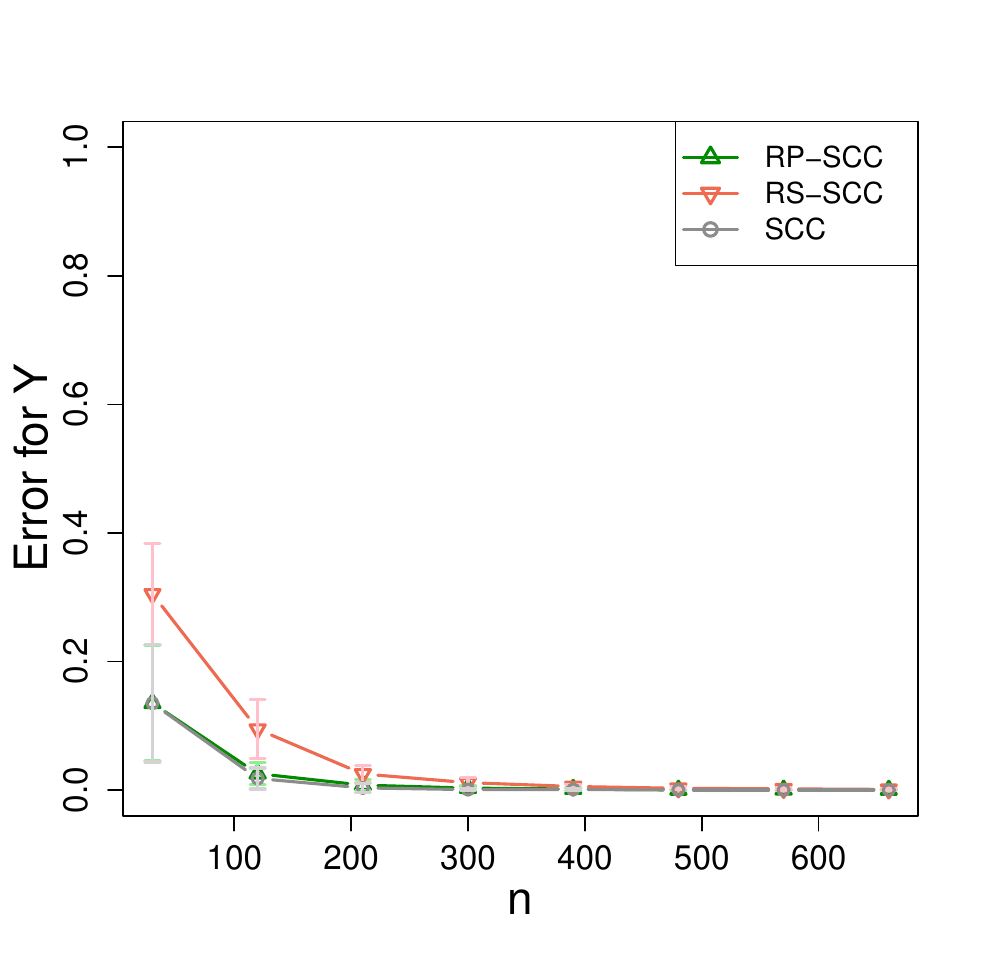}}
\subfigure[]{\includegraphics[height=4.1cm,width=4.3cm,angle=0]{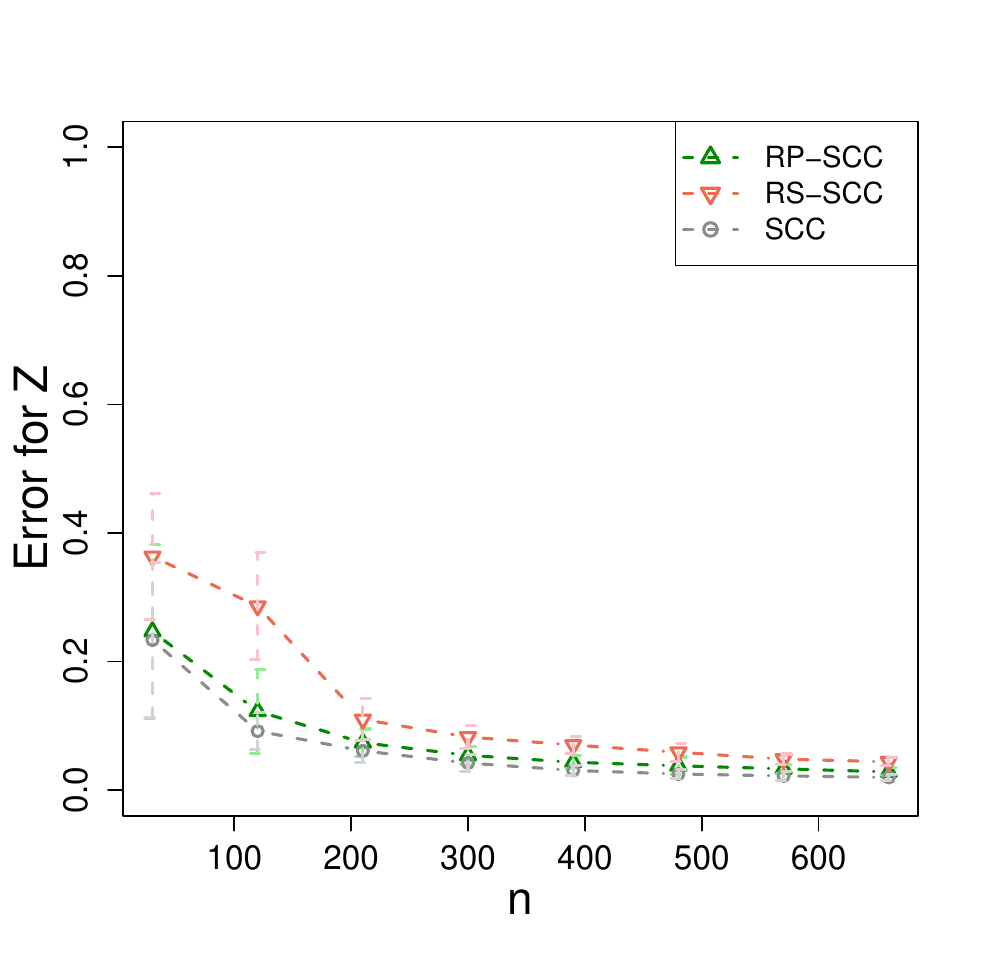}}
\caption{Simulation results of case 2 under {model set-up 4}.}\label{m4case2}
\end{figure}

\begin{figure}[!htbp]{}
\centering
\subfigure[]{\includegraphics[height=4.1cm,width=4.3cm,angle=0]{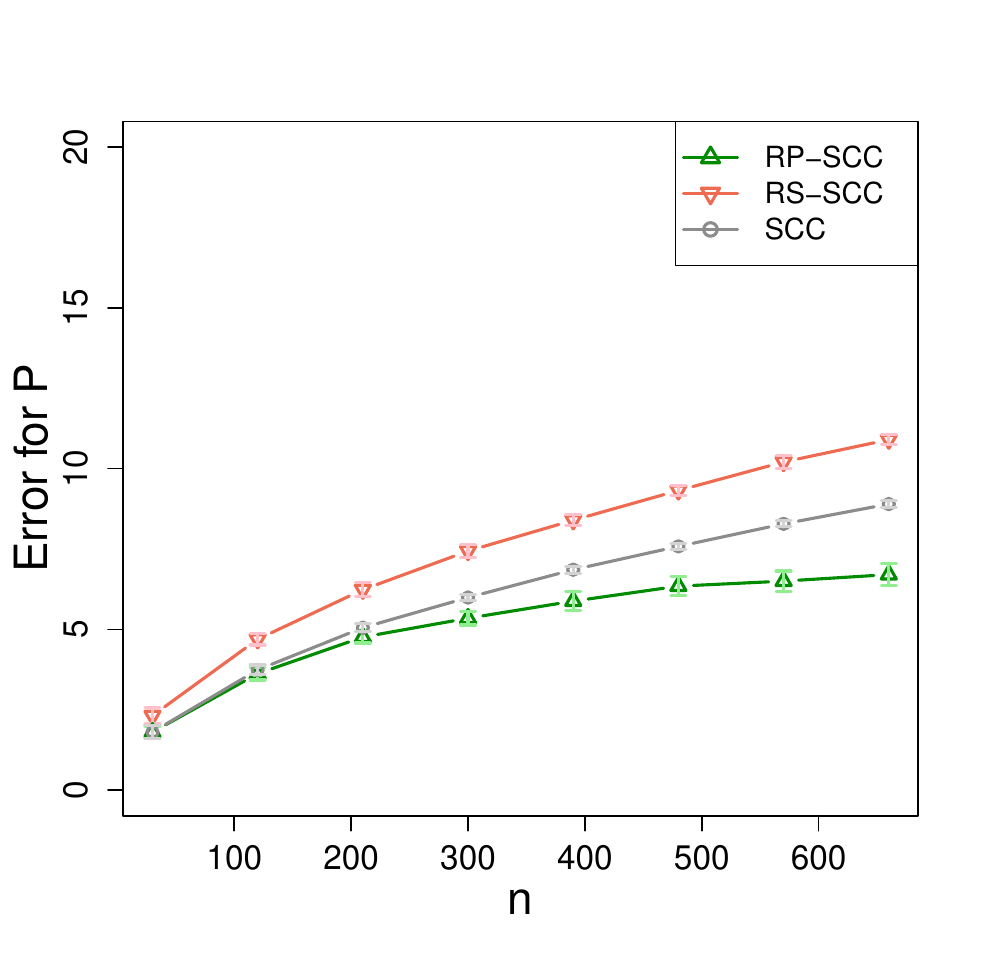}}
\subfigure[]{\includegraphics[height=4.1cm,width=4.3cm,angle=0]{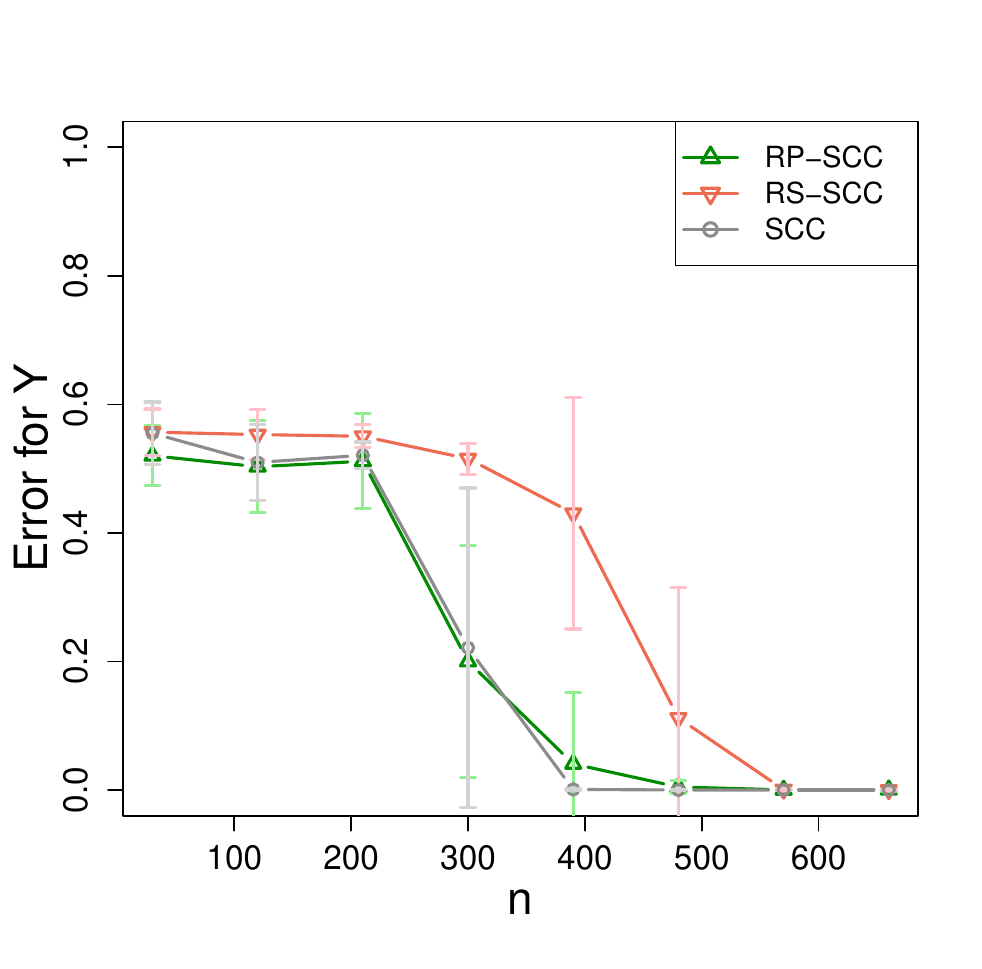}}
\subfigure[]{\includegraphics[height=4.1cm,width=4.3cm,angle=0]{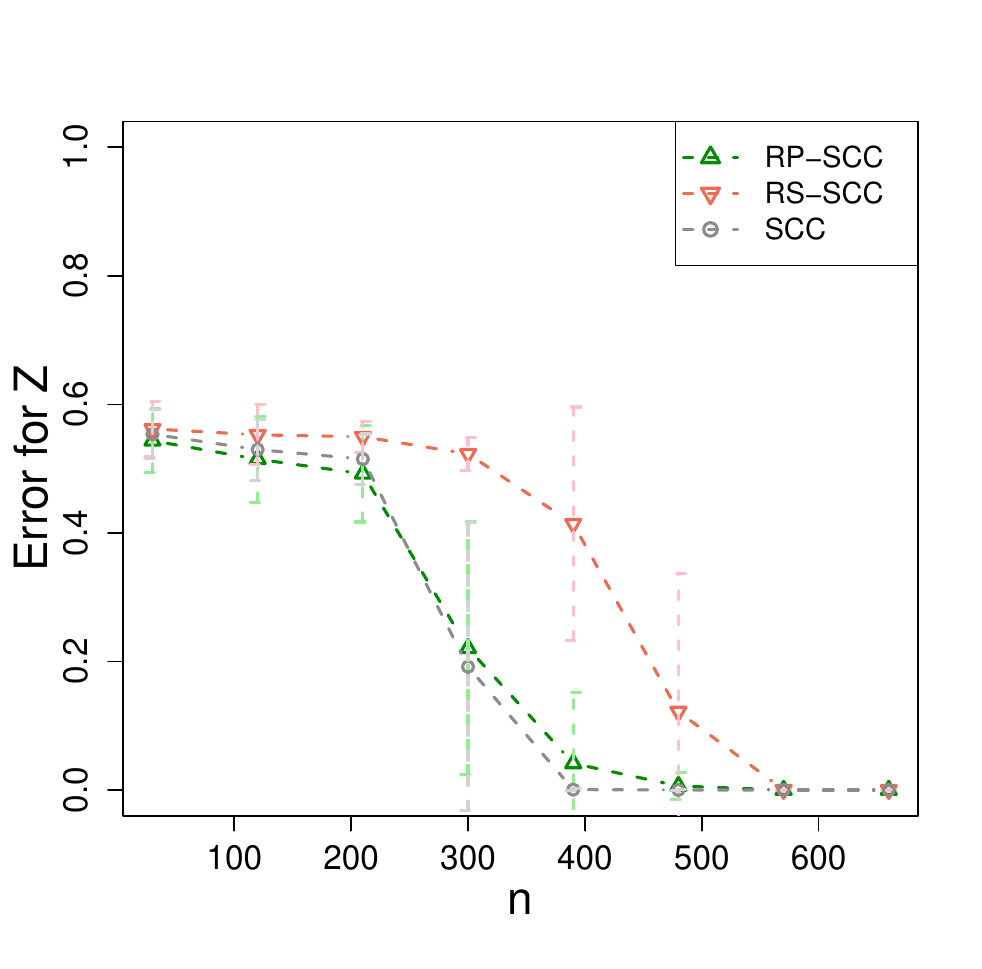}}
\caption{Simulation results of case 1 under {model set-up 5}.}\label{m5case1}
\end{figure}

\begin{figure}[!htbp]{}
\centering
\subfigure[]{\includegraphics[height=4.1cm,width=4.3cm,angle=0]{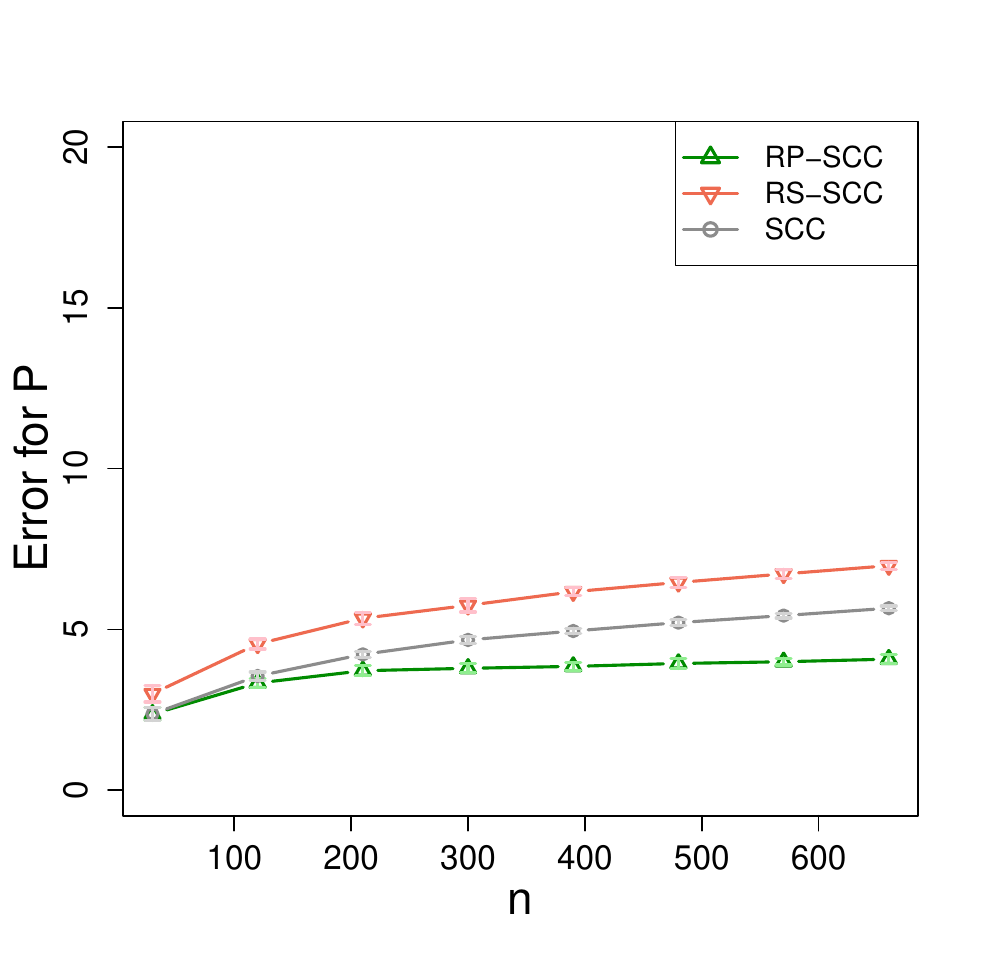}}
\subfigure[]{\includegraphics[height=4.1cm,width=4.3cm,angle=0]{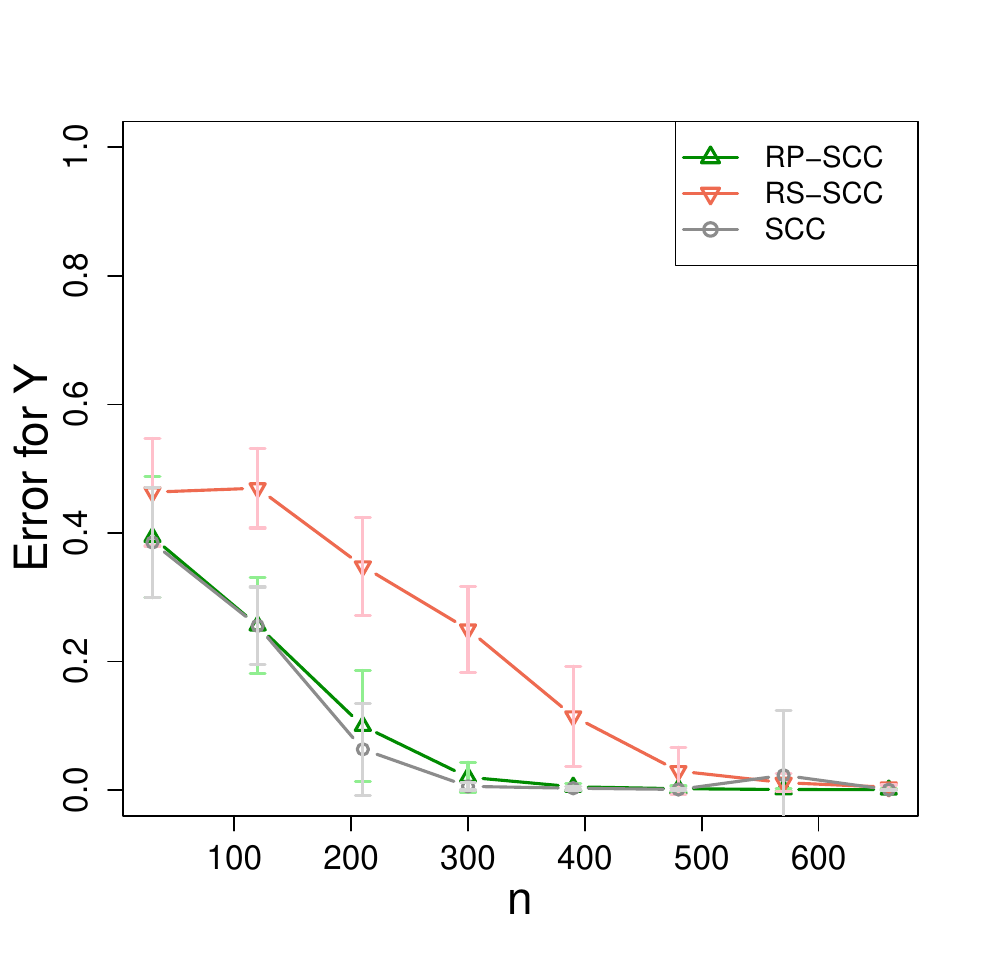}}
\subfigure[]{\includegraphics[height=4.1cm,width=4.3cm,angle=0]{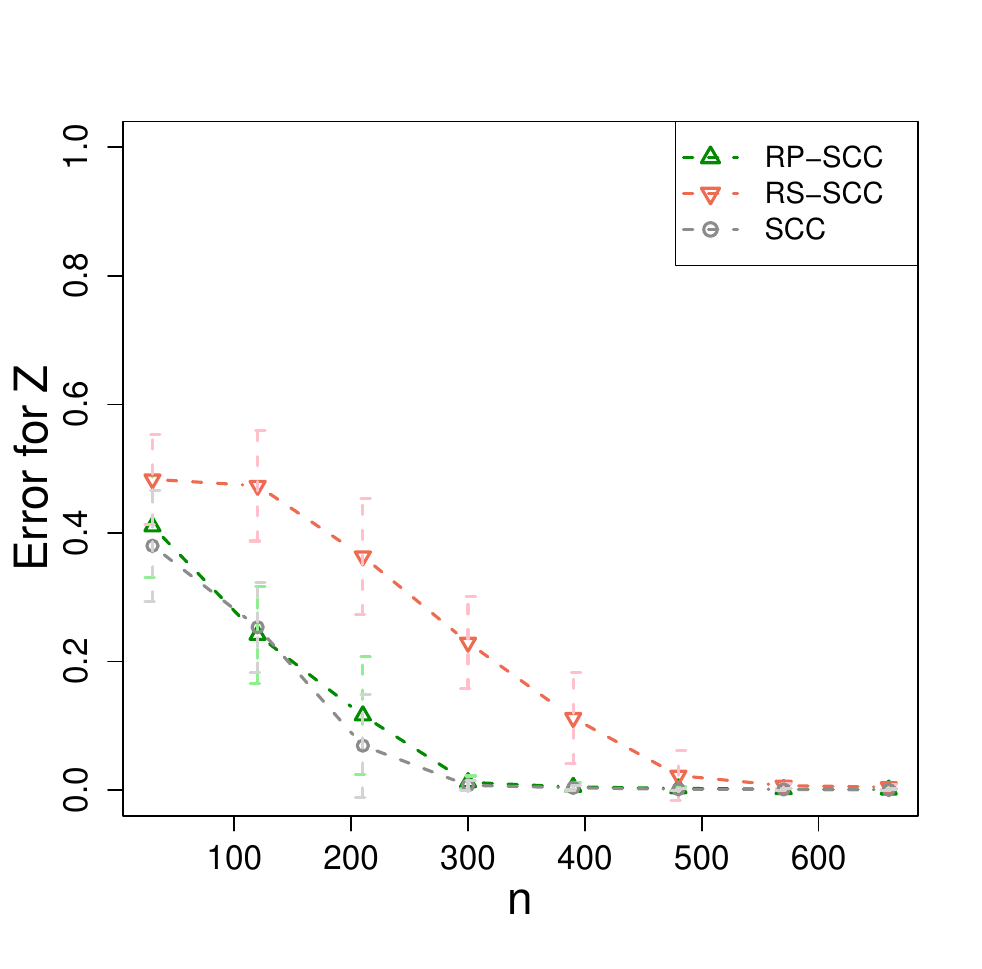}}
\caption{Simulation results of case 2 under {model set-up 5}.}\label{m5case2}
\end{figure}

\begin{figure}[!htbp]{}
\centering
\subfigure[]{\includegraphics[height=4.1cm,width=4.3cm,angle=0]{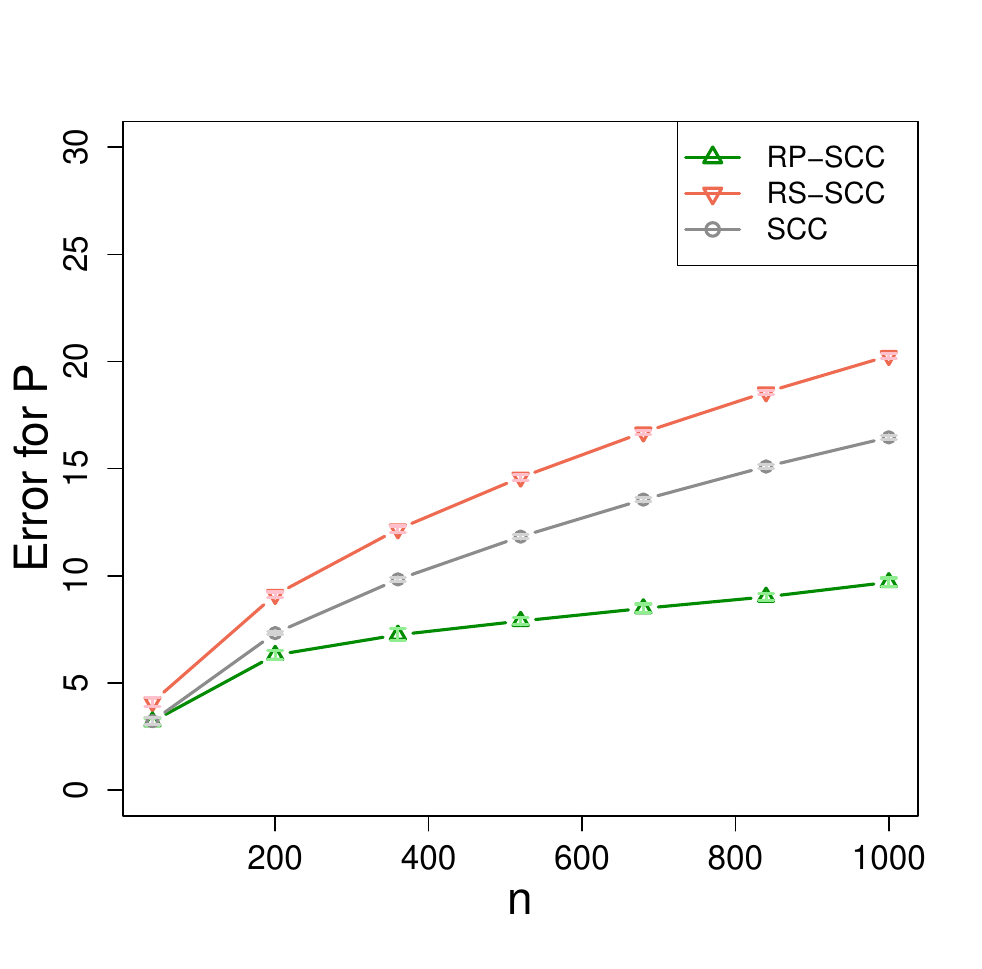}}
\subfigure[]{\includegraphics[height=4.1cm,width=4.3cm,angle=0]{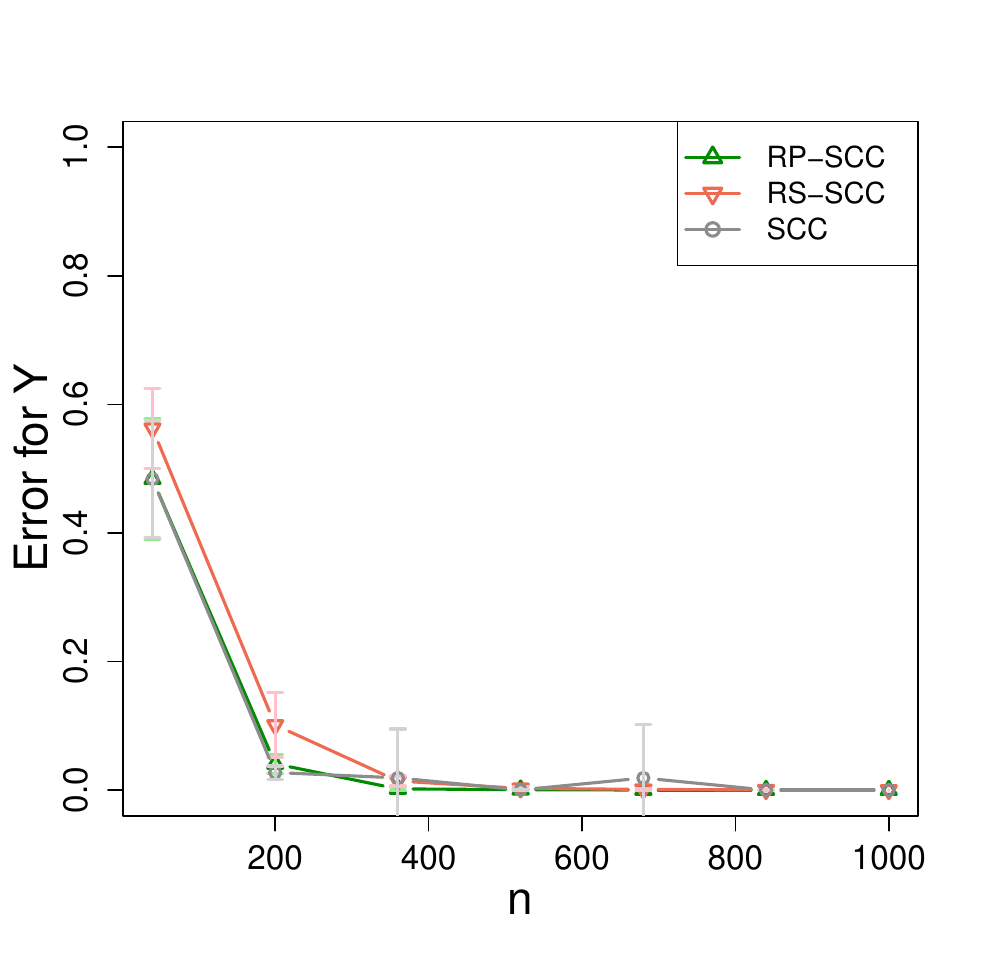}}
\subfigure[]{\includegraphics[height=4.1cm,width=4.3cm,angle=0]{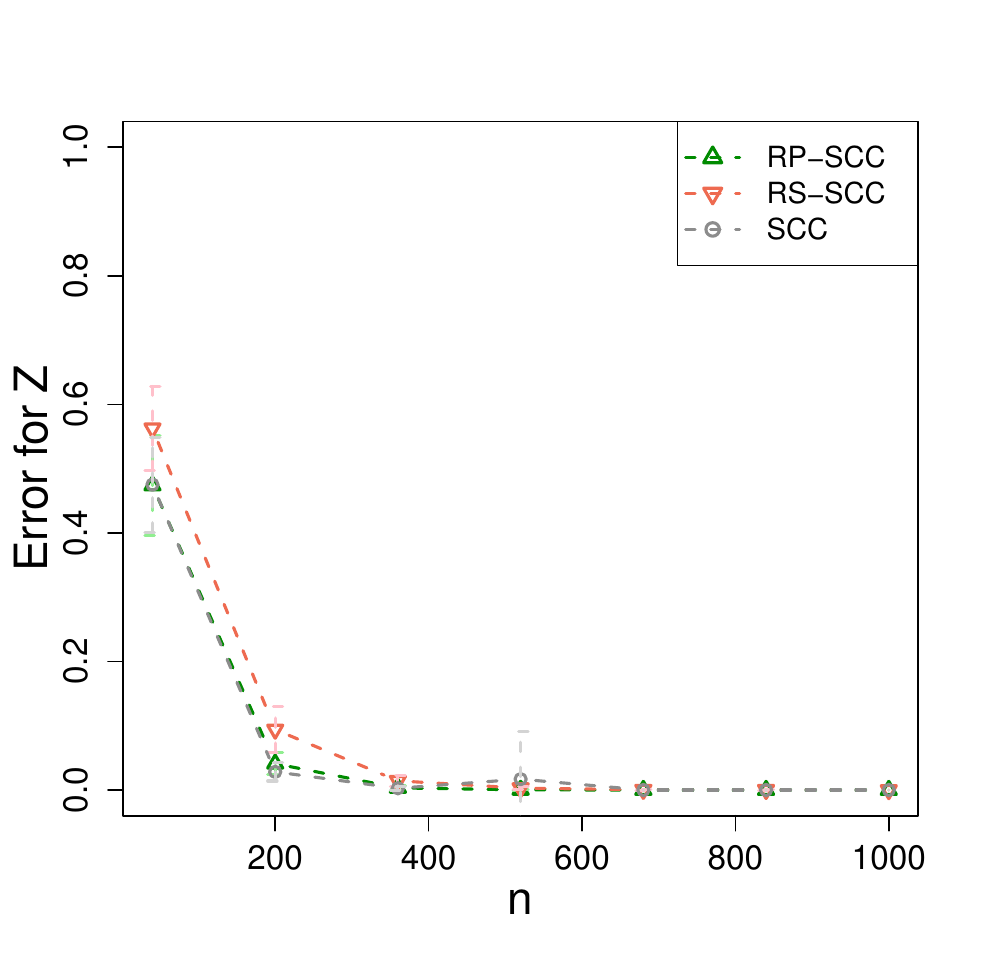}}
\caption{Simulation results of case 1 under {model set-up 6}. }\label{m6case1}
\end{figure}

\begin{figure}[!htbp]{}
\centering
\subfigure[]{\includegraphics[height=4.1cm,width=4.3cm,angle=0]{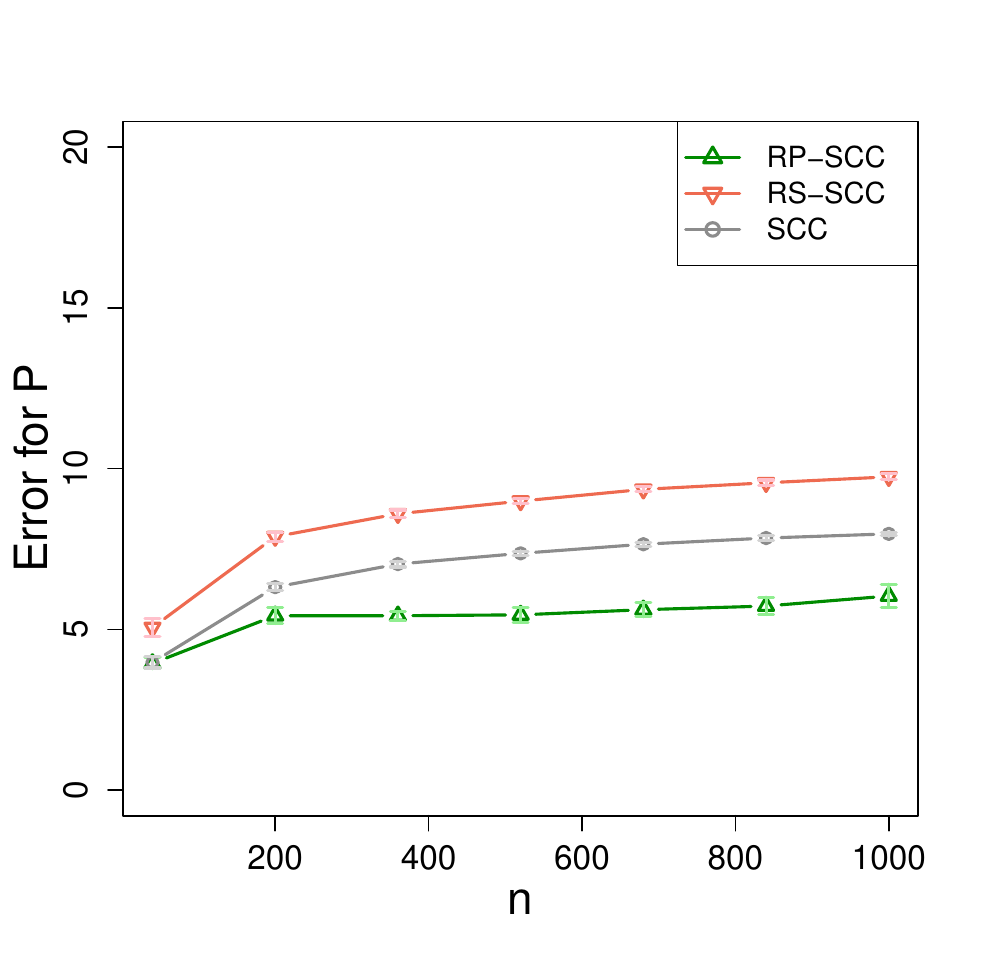}}
\subfigure[]{\includegraphics[height=4.1cm,width=4.3cm,angle=0]{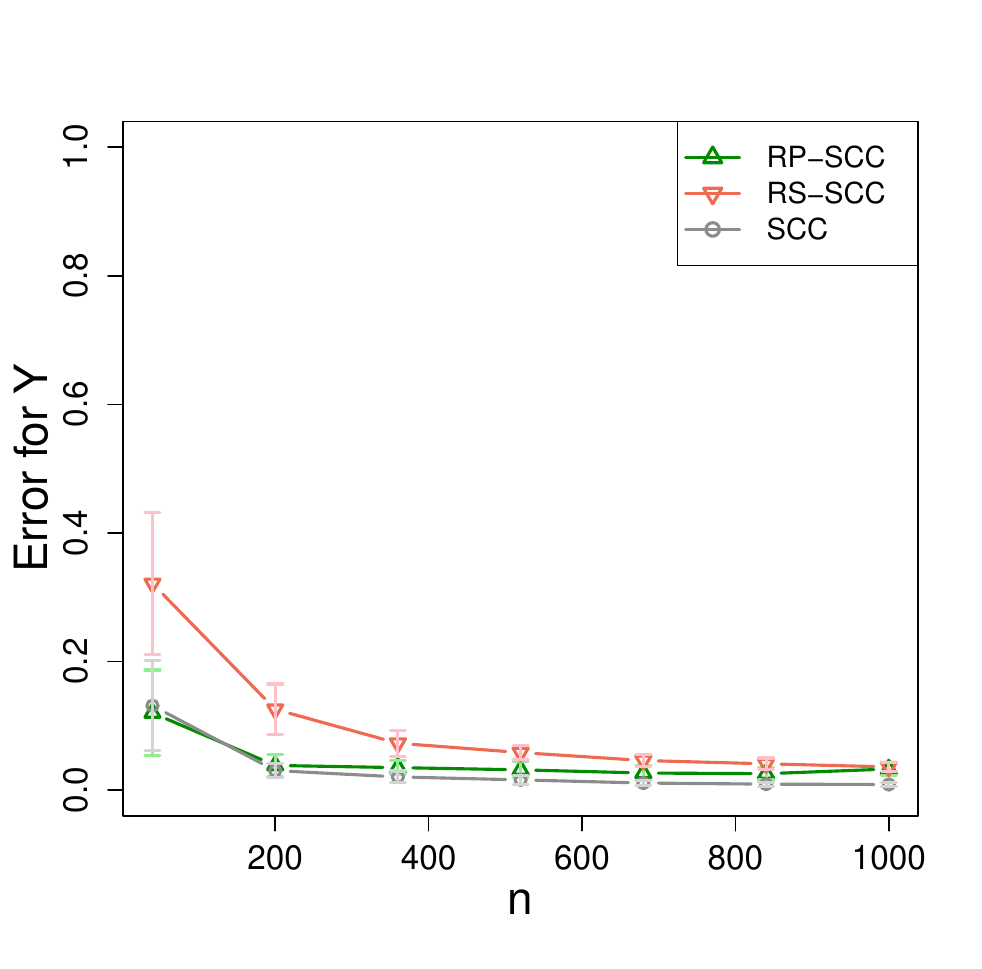}}
\subfigure[]{\includegraphics[height=4.1cm,width=4.3cm,angle=0]{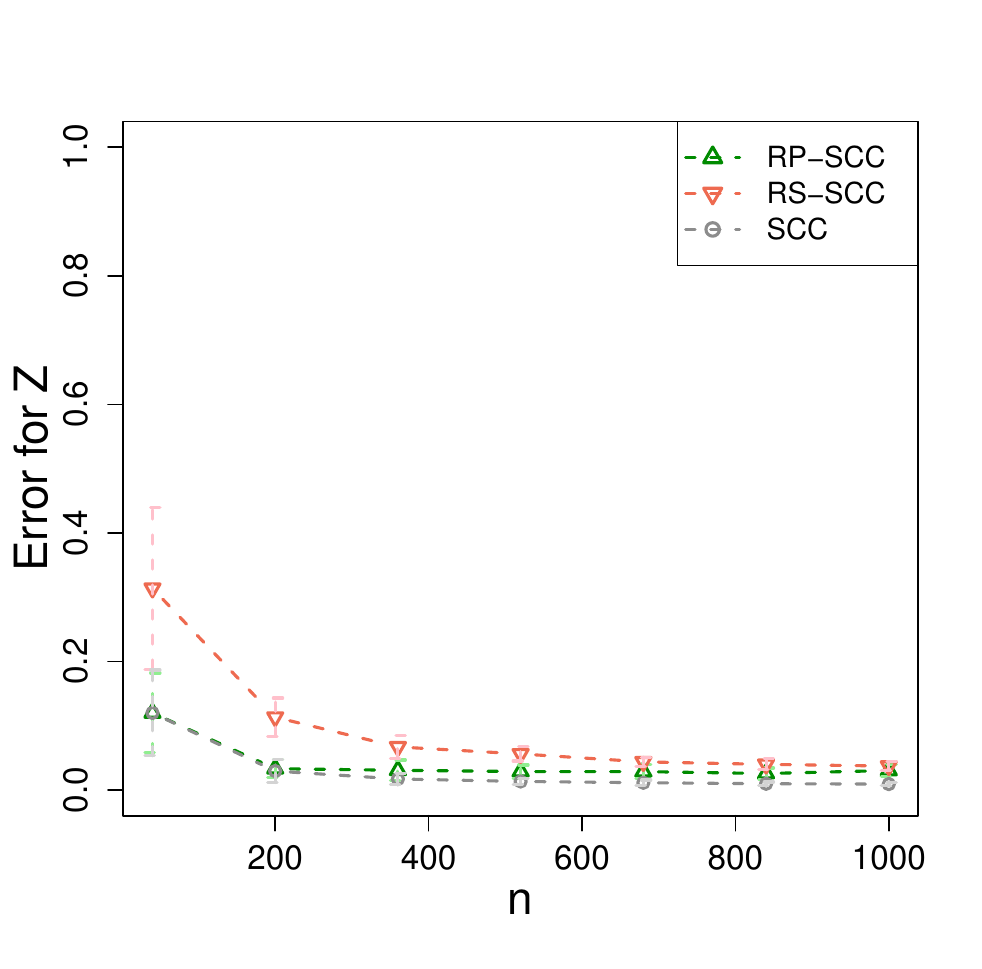}}
\caption{Simulation results of case 2 under {model set-up 6}.}\label{m6case2}
\end{figure}

\begin{figure}[!htbp]{}
\centering
\subfigure[]{\includegraphics[height=4.1cm,width=4.3cm,angle=0]{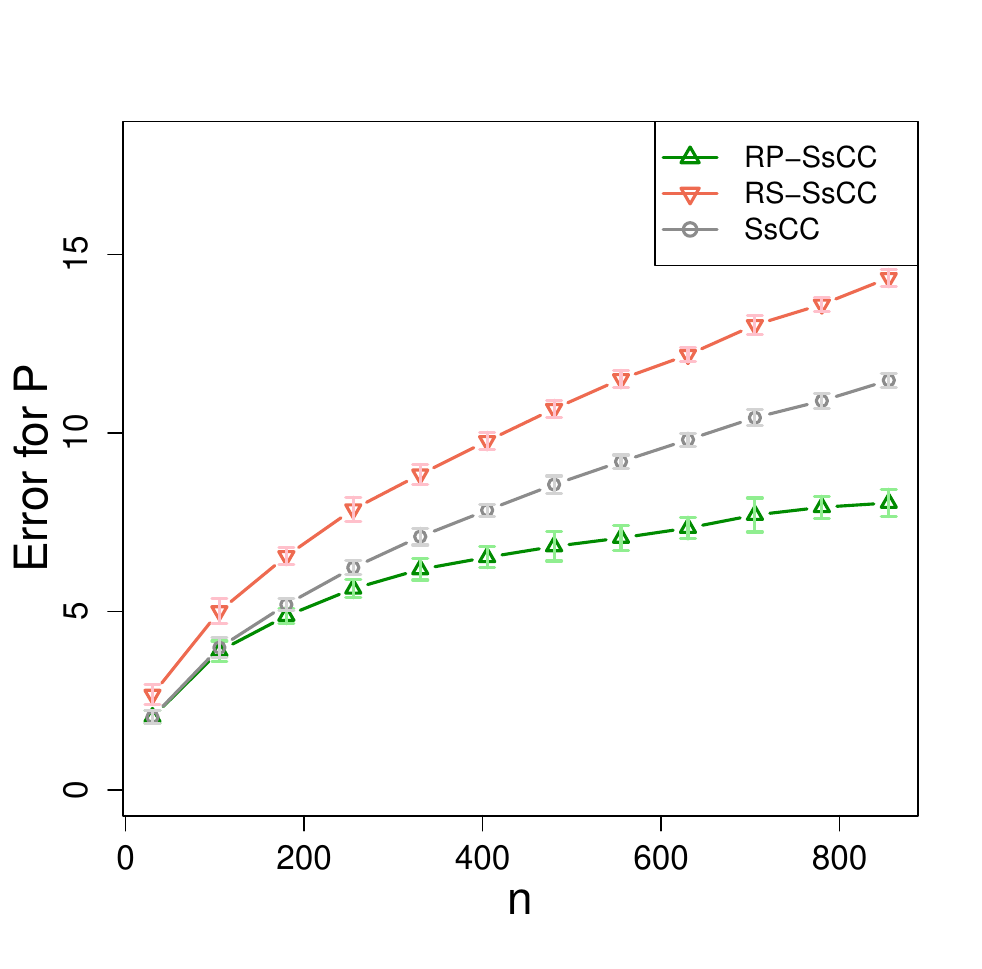}}
\subfigure[]{\includegraphics[height=4.1cm,width=4.3cm,angle=0]{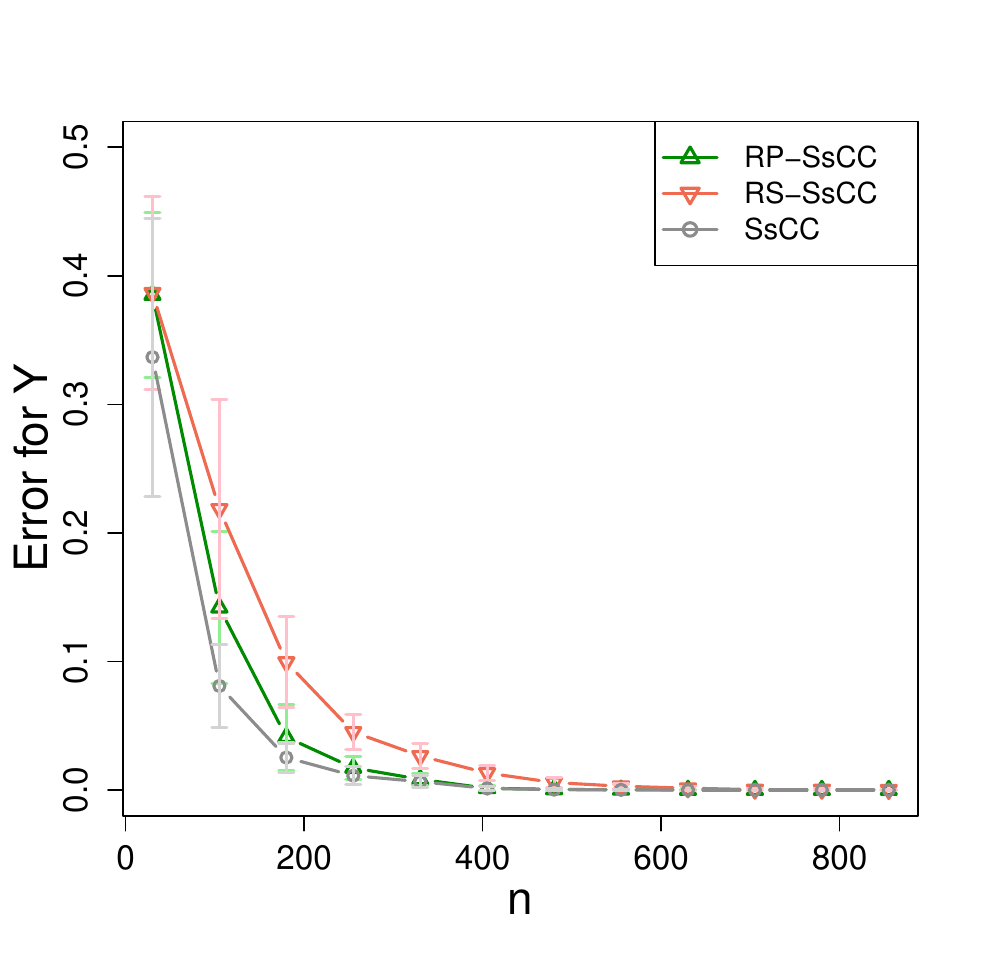}}
\subfigure[]{\includegraphics[height=4.1cm,width=4.3cm,angle=0]{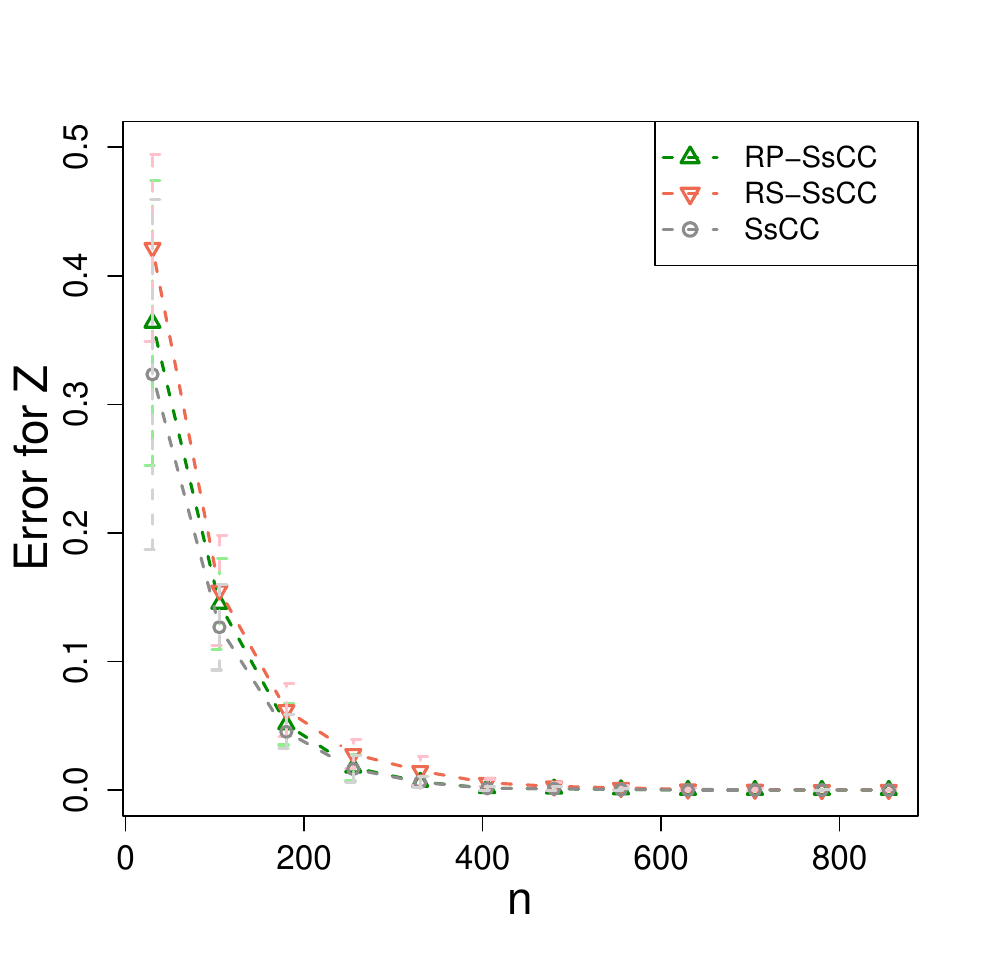}}
\caption{Simulation results of case 1 under {model set-up 7}.}\label{m7case1}
\end{figure}

\begin{figure}[!htbp]{}
\centering
\subfigure[]{\includegraphics[height=4.1cm,width=4.3cm,angle=0]{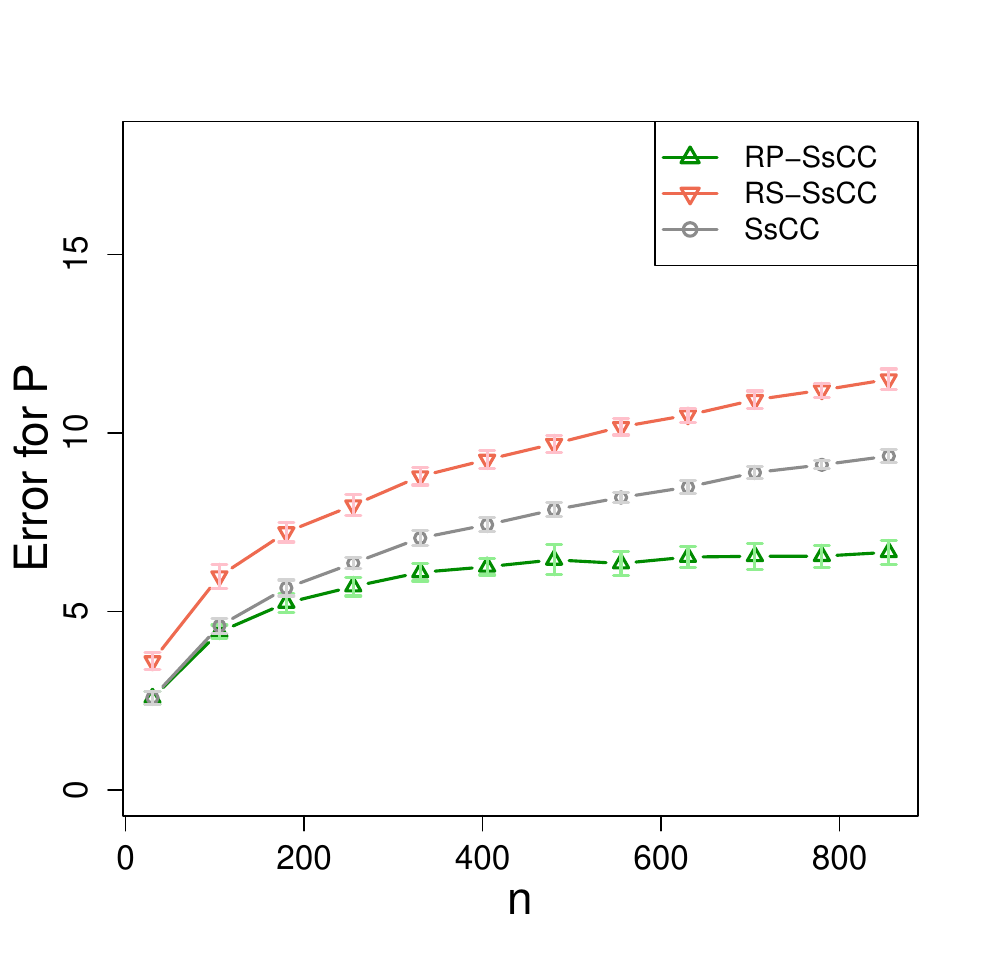}}
\subfigure[]{\includegraphics[height=4.1cm,width=4.3cm,angle=0]{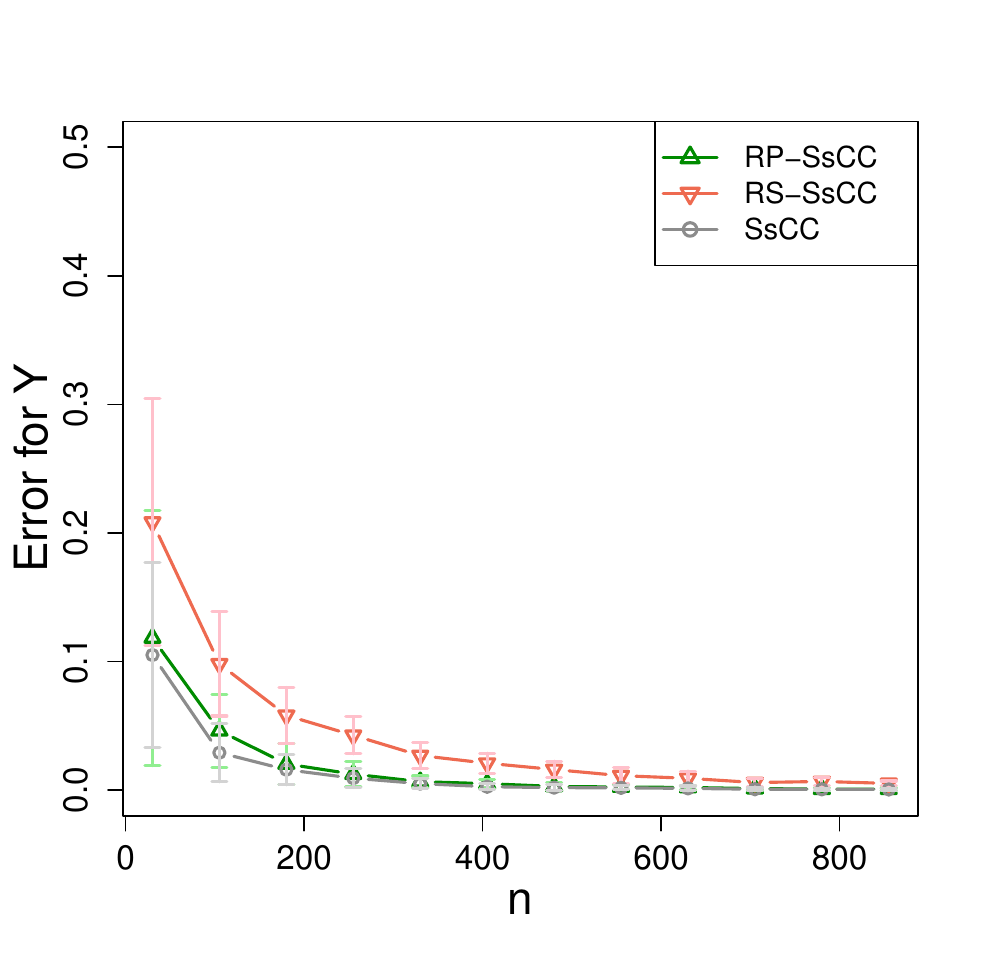}}
\subfigure[]{\includegraphics[height=4.1cm,width=4.3cm,angle=0]{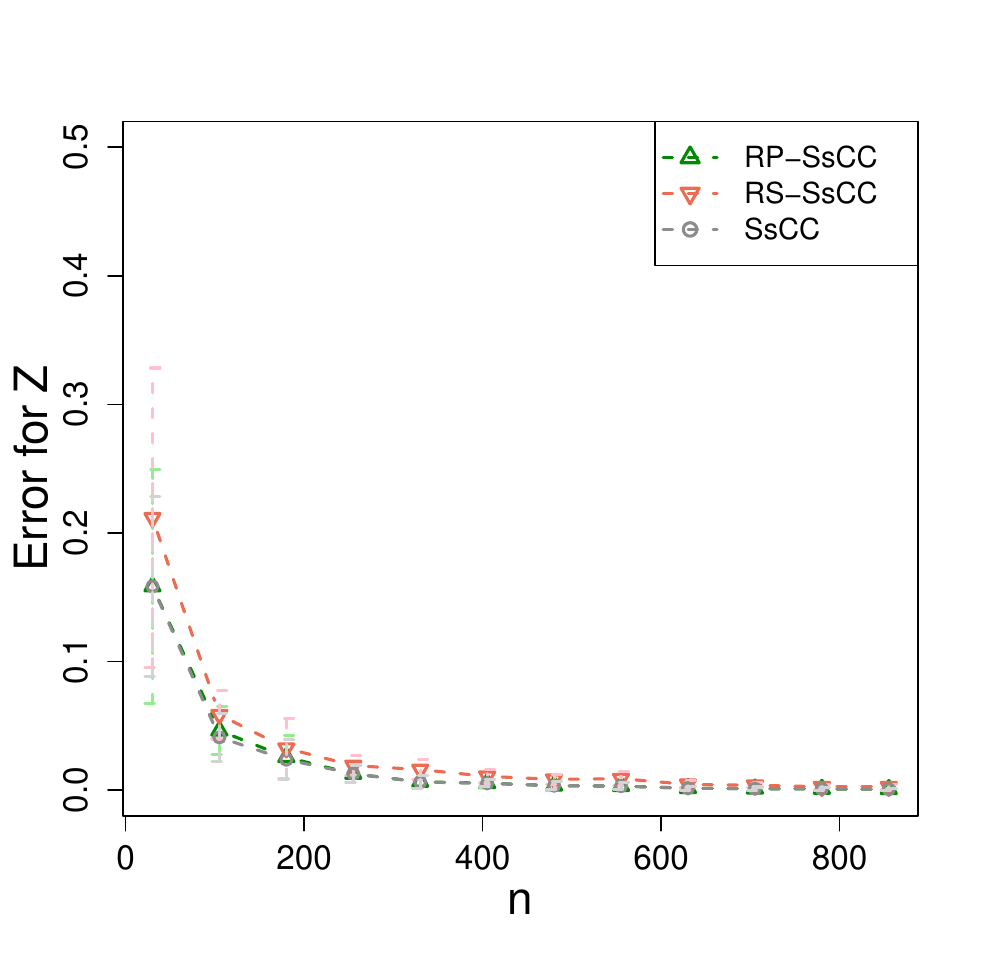}}
\caption{Simulation results of case 2 under {model set-up 7}. }\label{m7case2}
\end{figure}

\begin{figure}[!htbp]{}
\centering
\subfigure[]{\includegraphics[height=4.1cm,width=4.3cm,angle=0]{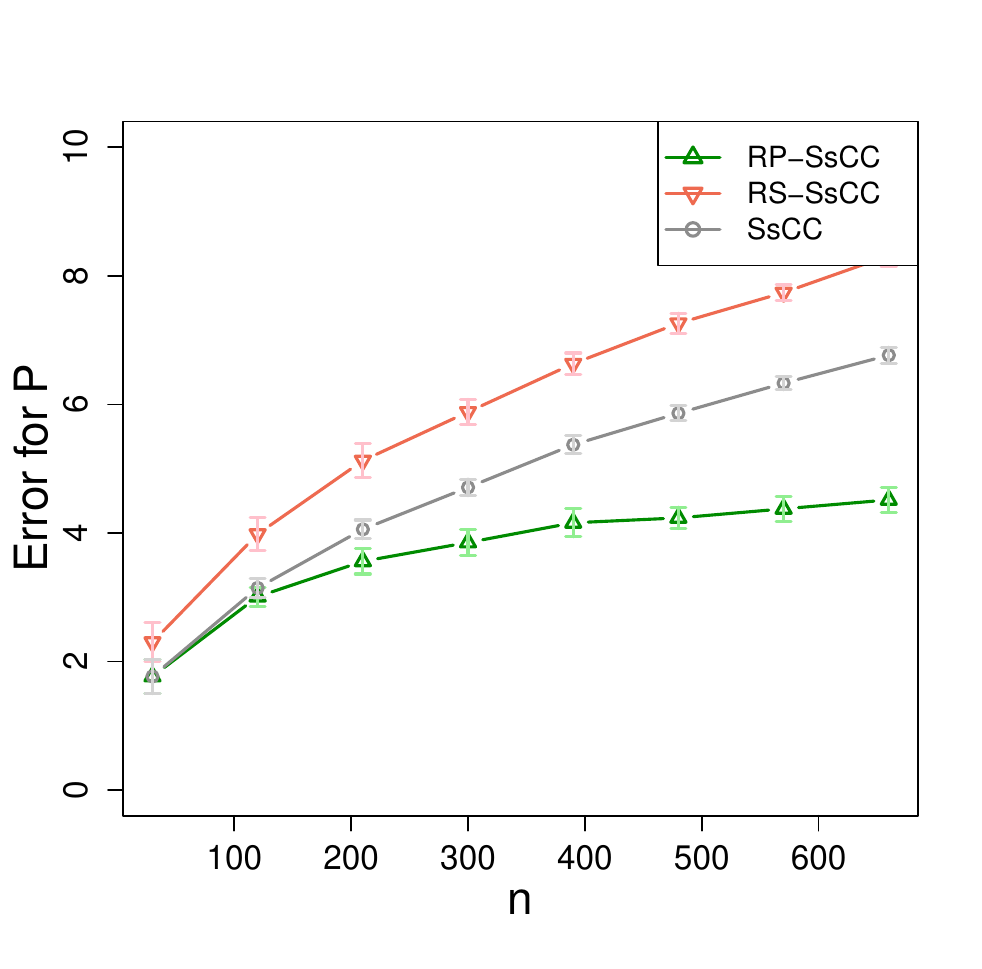}}
\subfigure[]{\includegraphics[height=4.1cm,width=4.3cm,angle=0]{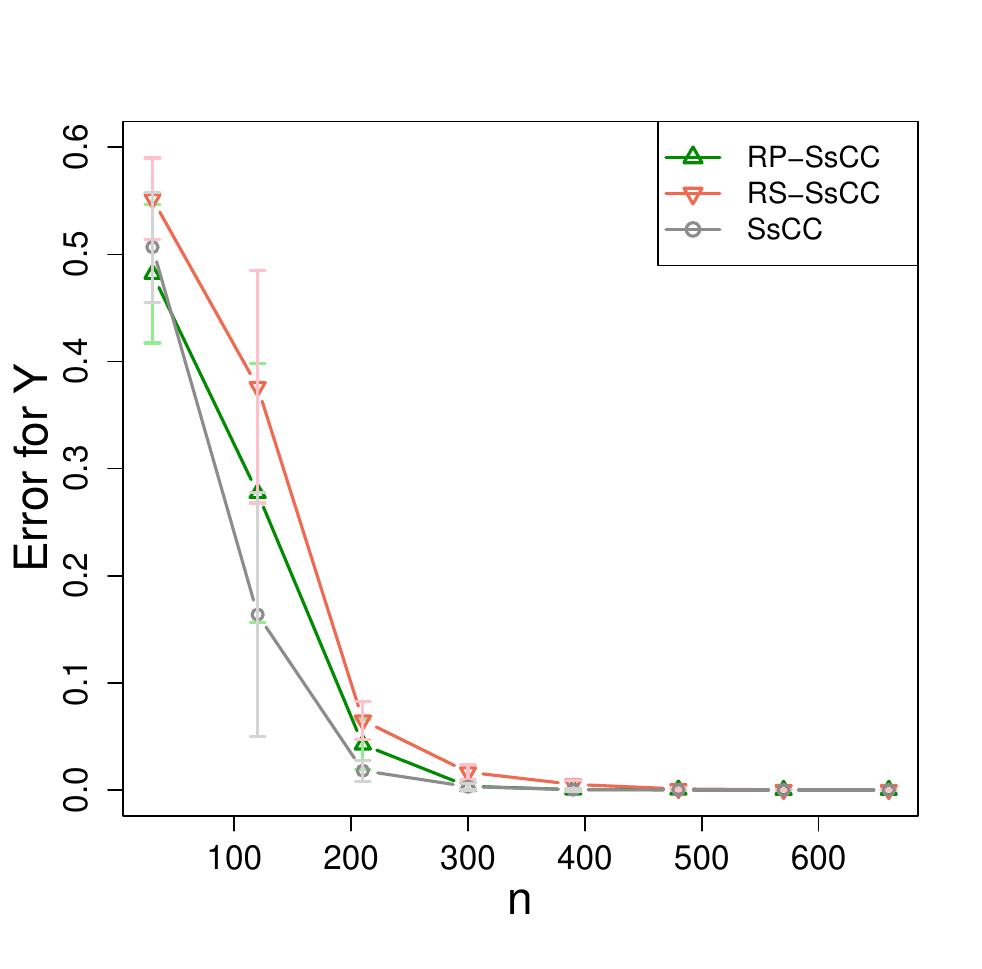}}
\subfigure[]{\includegraphics[height=4.1cm,width=4.3cm,angle=0]{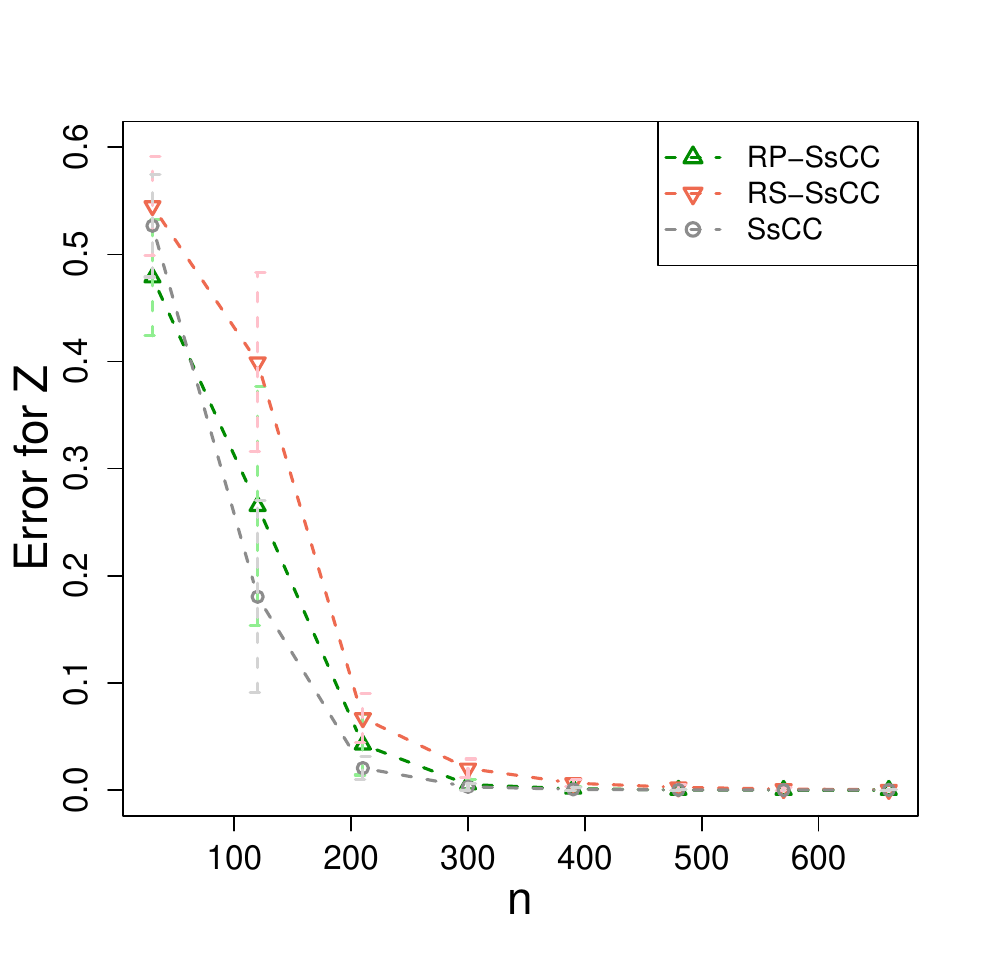}}
\caption{Simulation results of case 1 under {model set-up 8}. }\label{m8case1}
\end{figure}

\begin{figure}[!htbp]{}
\centering
\subfigure[]{\includegraphics[height=4.1cm,width=4.3cm,angle=0]{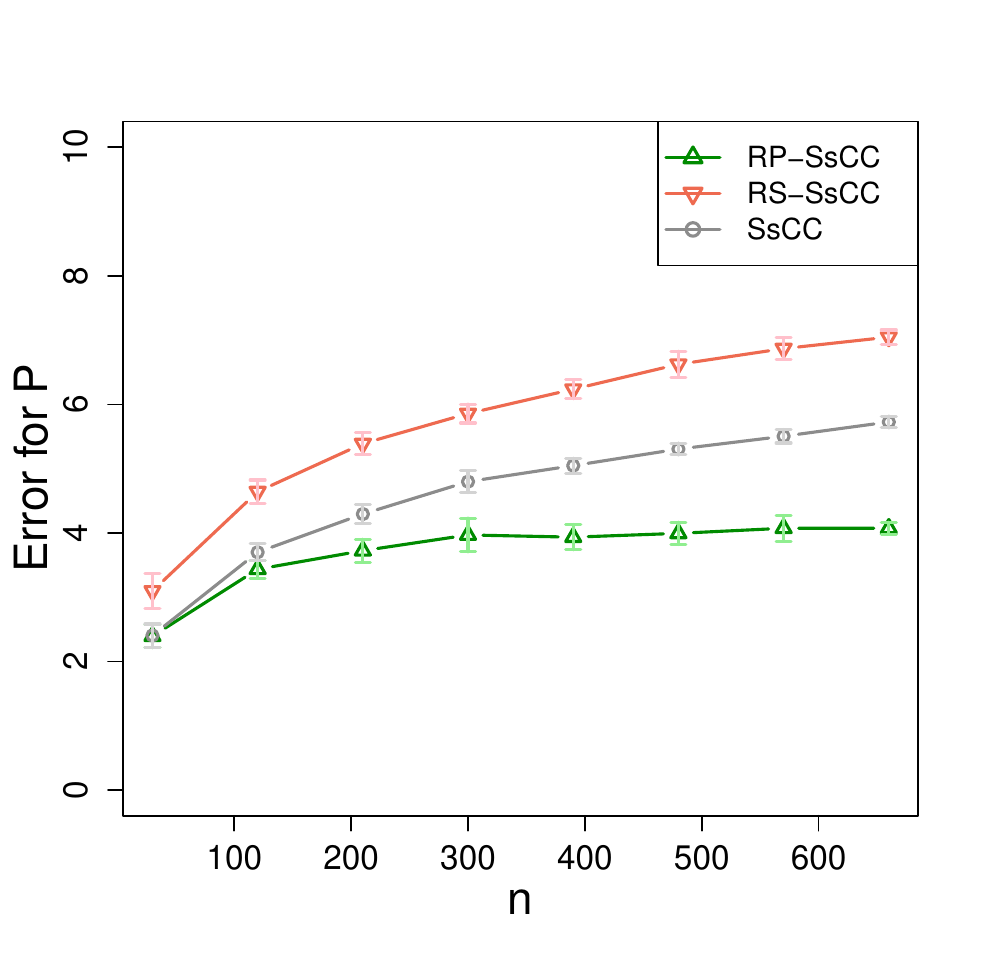}}
\subfigure[]{\includegraphics[height=4.1cm,width=4.3cm,angle=0]{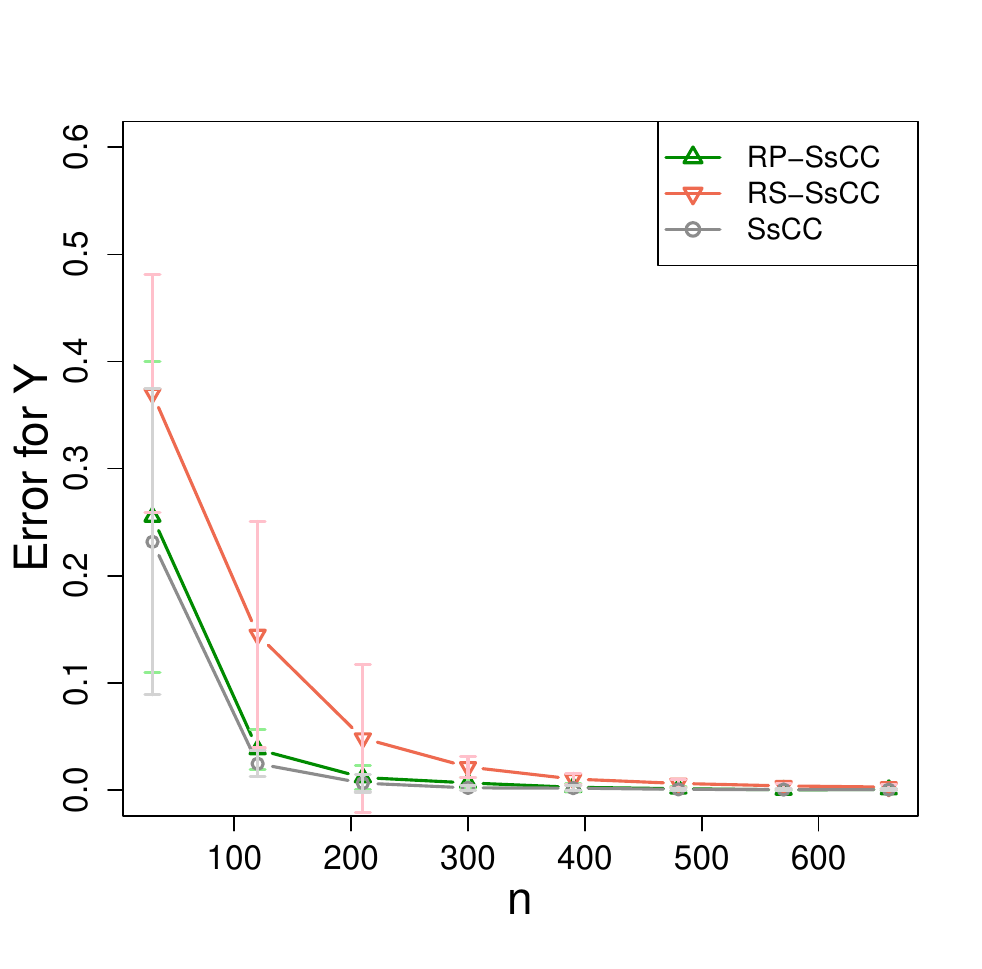}}
\subfigure[]{\includegraphics[height=4.1cm,width=4.3cm,angle=0]{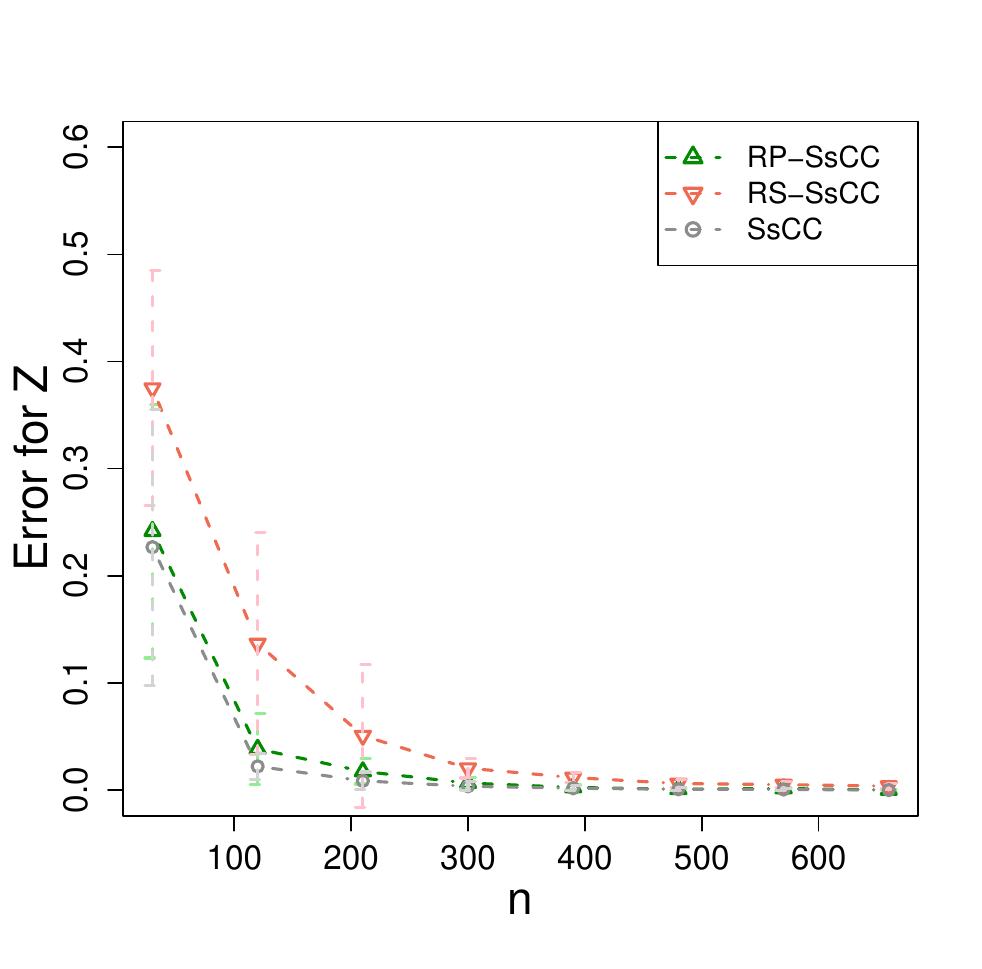}}
\caption{Simulation results of case 2 under {model set-up 8}. }\label{m8case2}
\end{figure}

\begin{figure}[!htbp]{}
	\centering
	\subfigure{\includegraphics[height=7cm,width=8.5cm,angle=0]{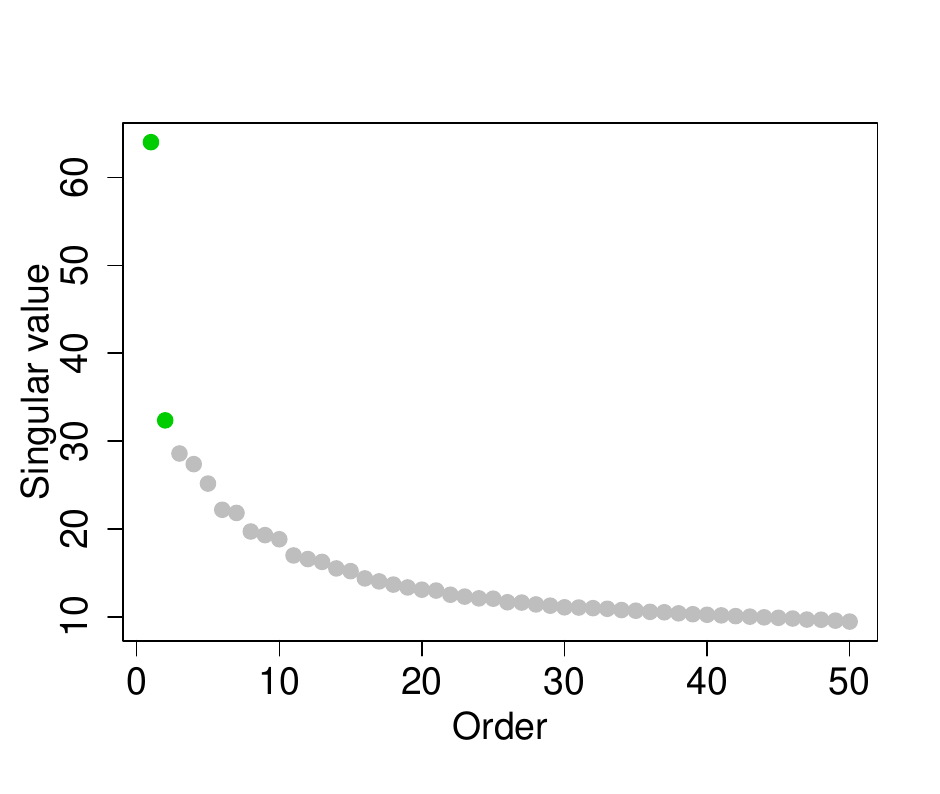}}
	\caption{The top 50 singular values of the adjacency matrix of the European email network.}\label{emailsv}
\end{figure}

\begin{figure}[!htbp]{}
	\centering
	\subfigure[SCC]{\includegraphics[height=4.8cm,width=5cm,angle=0]{citation_move_os.pdf}}
    \subfigure[RP-SCC]{\includegraphics[height=4.8cm,width=5cm,angle=0]{citation_move_rp.pdf}}
    \subfigure[RS-SCC]{\includegraphics[height=4.8cm,width=5cm,angle=0]{citation_move_rs.pdf}}
     \subfigure[\textsf{svds}]{\includegraphics[height=4.8cm,width=5cm,angle=0]{citation_move_svds.pdf}}
    \subfigure[\textsf{irlba}]{\includegraphics[height=4.8cm,width=5cm,angle=0]{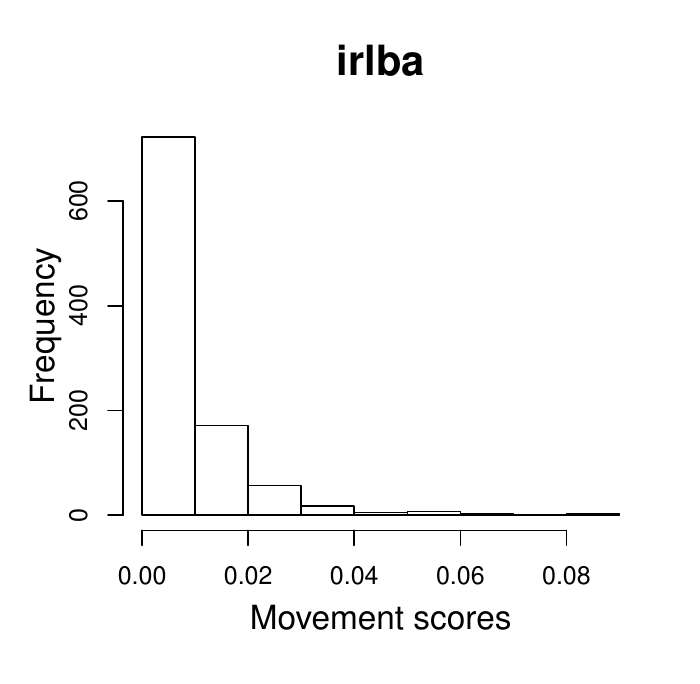}}
	\caption{Histogram of movement scores of different methods for the European email network.}\label{emailmove}
\end{figure}

\begin{figure}[!htbp]{}
	\centering
	\subfigure[Sending clusters]{\includegraphics[height=6.5cm,width=6.5cm,angle=0]{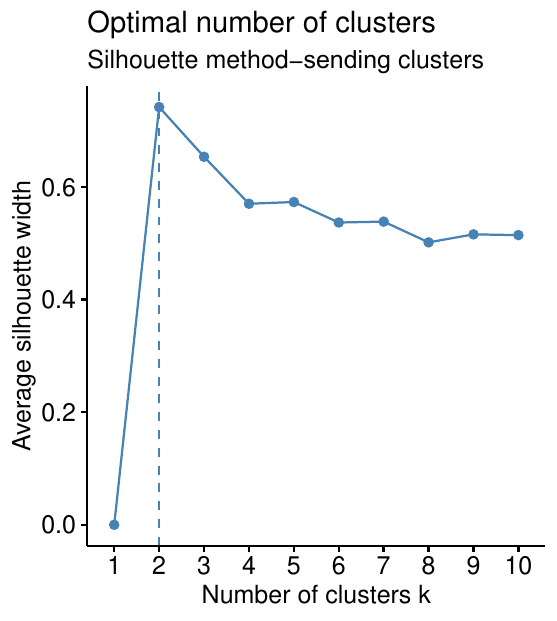}}\hspace{1.3cm}
    \subfigure[Receiving clusters]{\includegraphics[height=6.5cm,width=6.5cm,angle=0]{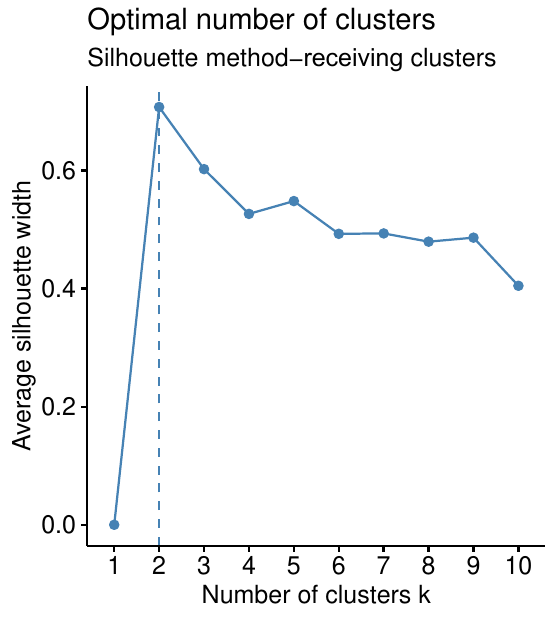}}
	\caption{Optimal number of clusters of the European email network selected by the average silhouette method based on the SCC.}\label{emailclusternumber}
\end{figure}

\begin{figure}[!htbp]{}
	\centering
	\subfigure[Sending clusters]{\includegraphics[height=5.5cm,width=5.3cm,angle=0]{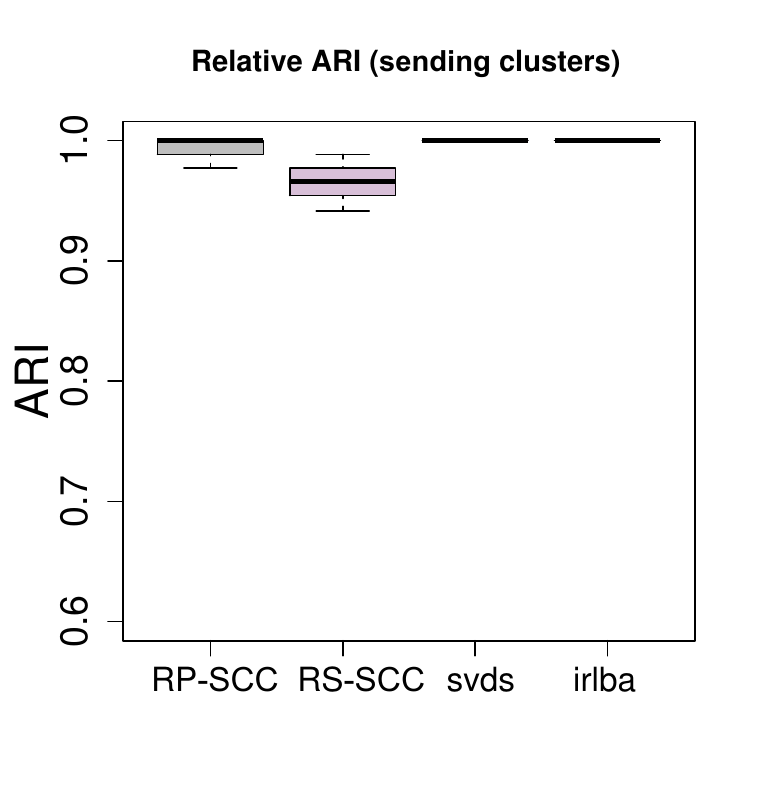}}\hspace{1.3cm}
    \subfigure[Receiving clusters]{\includegraphics[height=5.5cm,width=5.3cm,angle=0]{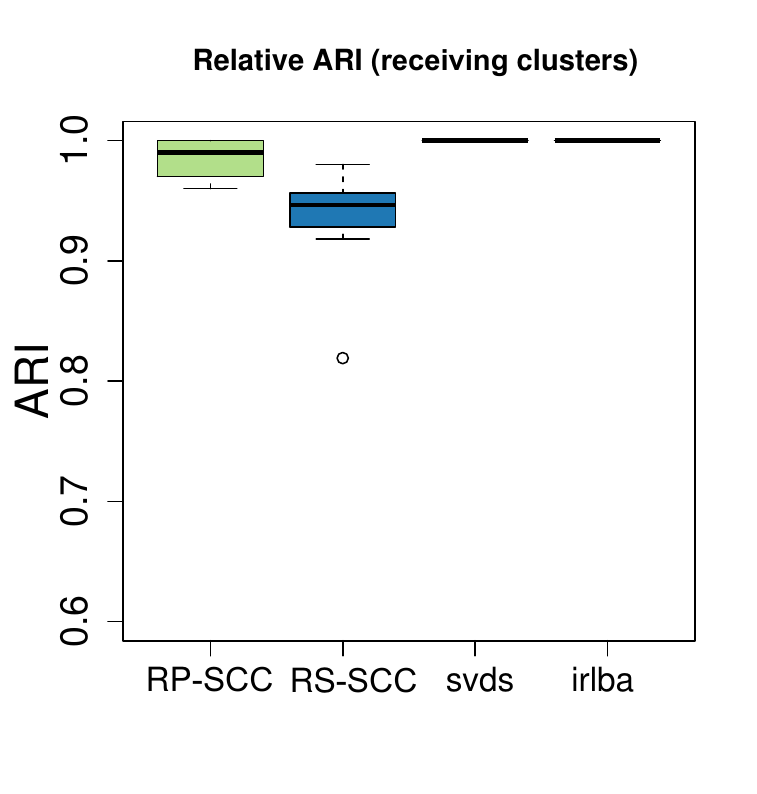}}
	\caption{Relative ARI between the SCC and SCC-based four approximate methods on the European email network.}\label{emailbox}
\end{figure}

\begin{figure}[!htbp]{}
	\centering
	\subfigure[SCC]{\includegraphics[height=4.3cm,width=4.9cm,angle=0]{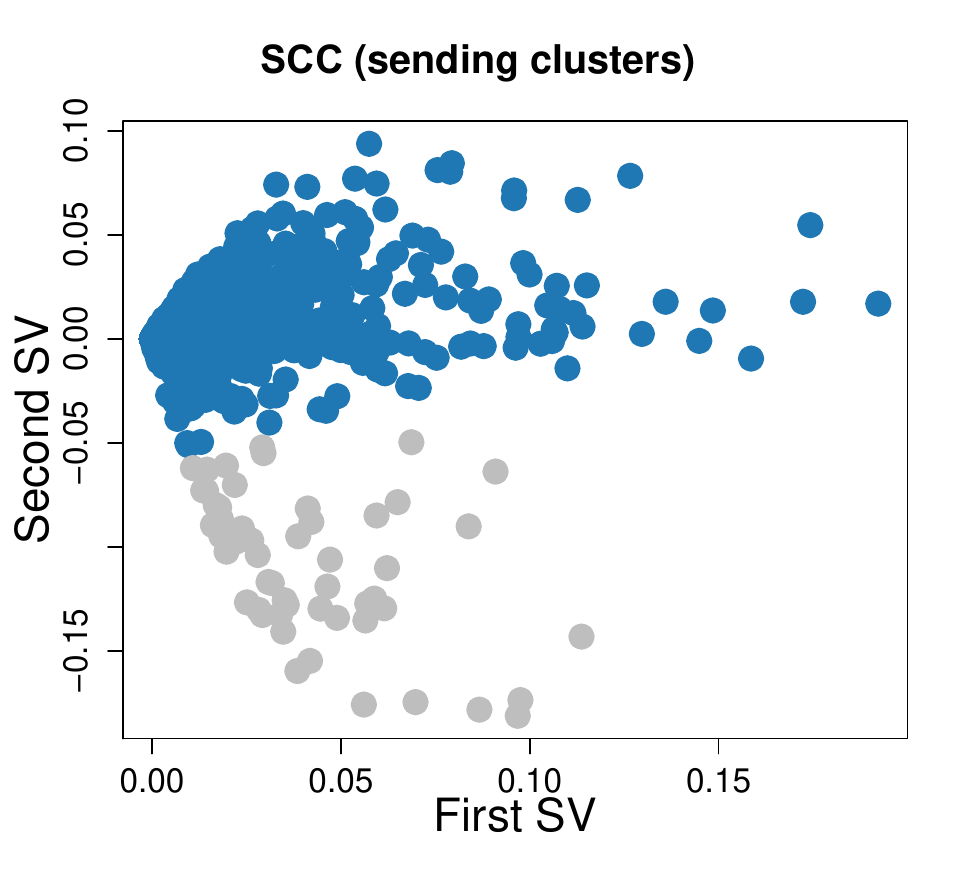}}
    \subfigure[RP-SCC]{\includegraphics[height=4.3cm,width=4.9cm,angle=0]{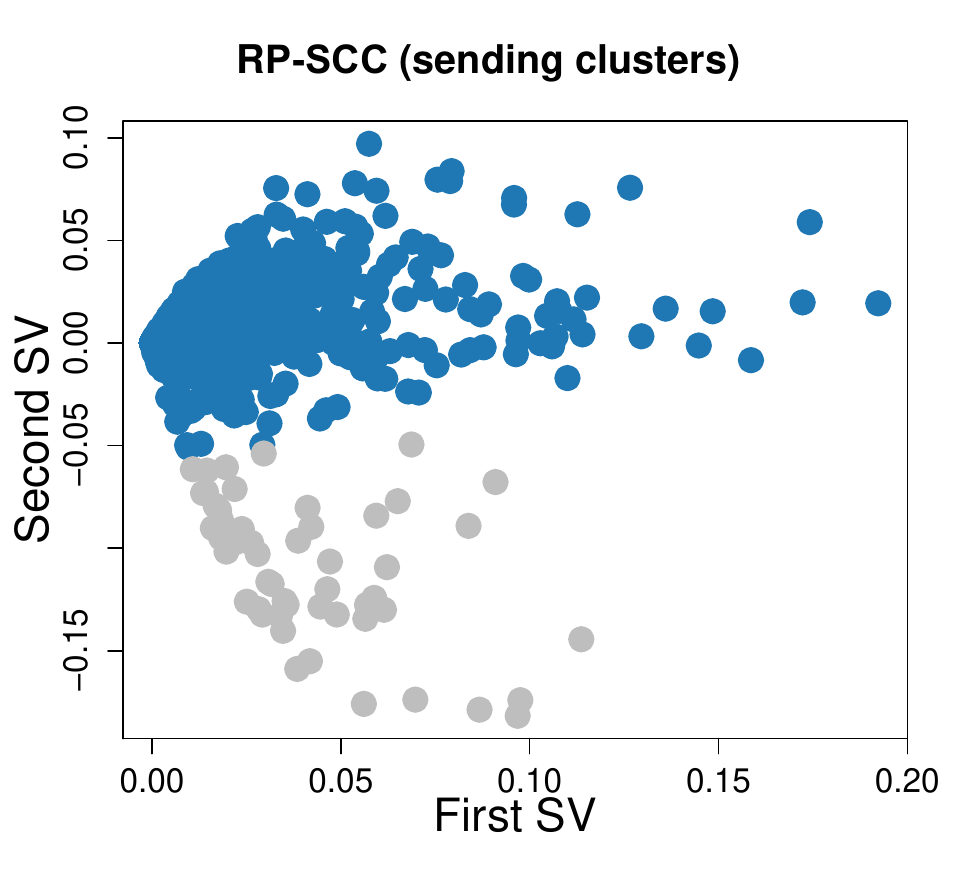}}
    \subfigure[RS-SCC]{\includegraphics[height=4.3cm,width=4.9cm,angle=0]{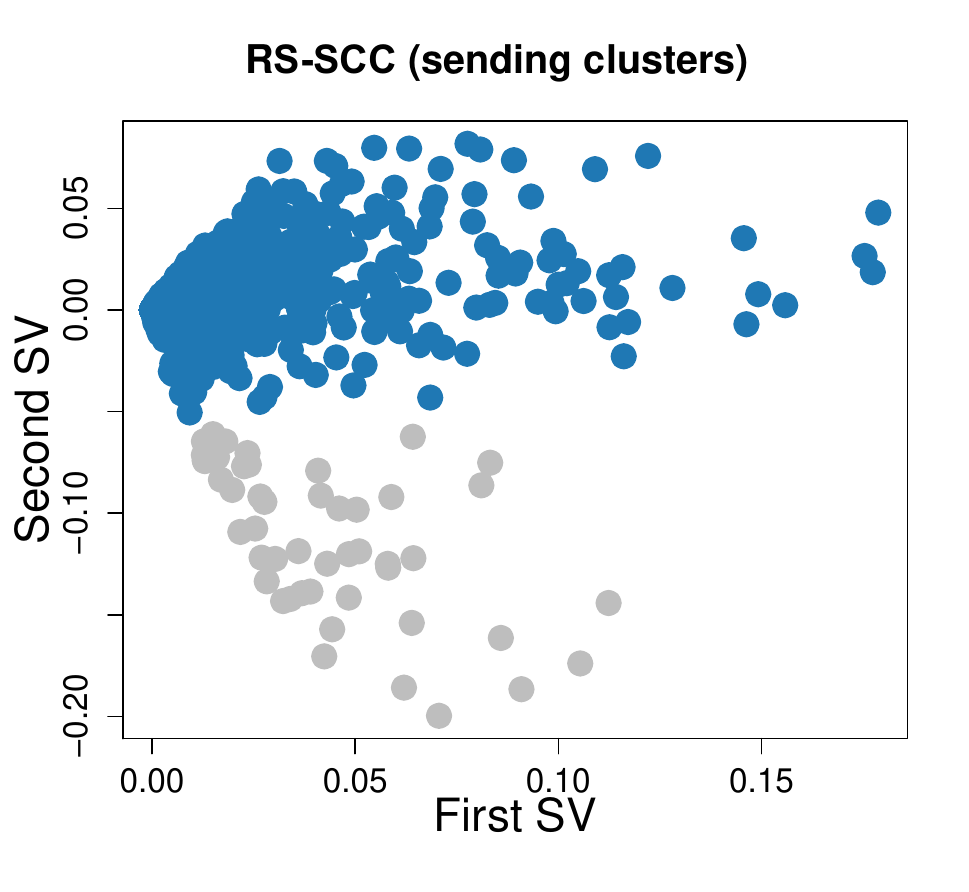}}
    \subfigure[\textsf{svds}]{\includegraphics[height=4.3cm,width=4.9cm,angle=0]{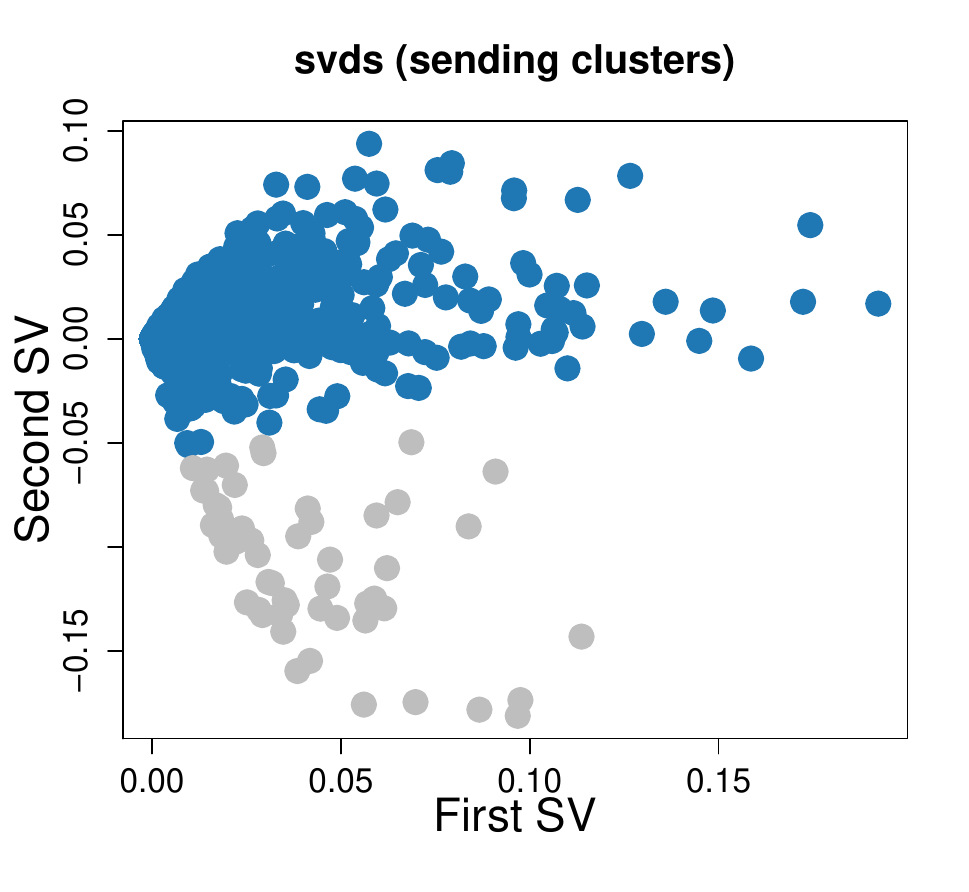}}
    \subfigure[\textsf{irlba}]{\includegraphics[height=4.3cm,width=4.9cm,angle=0]{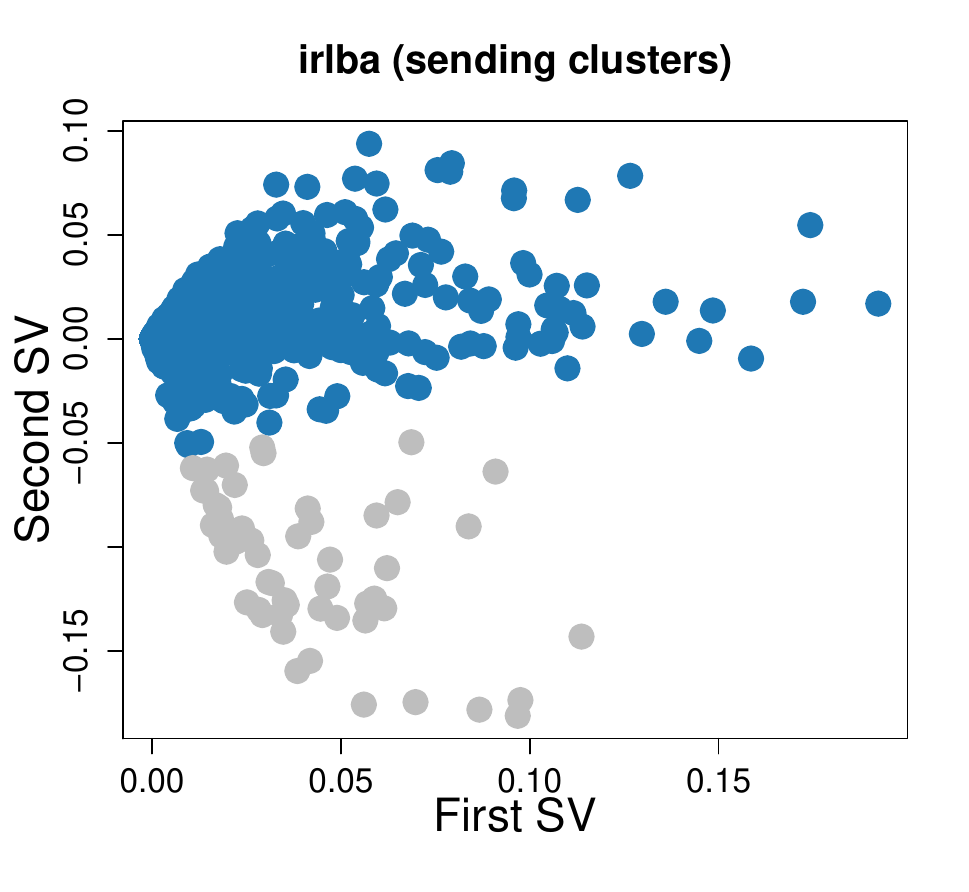}}
	\caption{Sending clusters of the European email network detected by SCC and four SCC-based approximate algorithms.}\label{emailpointsrow}
\end{figure}

\begin{figure}[!htbp]{}
	\centering
	\subfigure[SCC]{\includegraphics[height=4.3cm,width=4.9cm,angle=0]{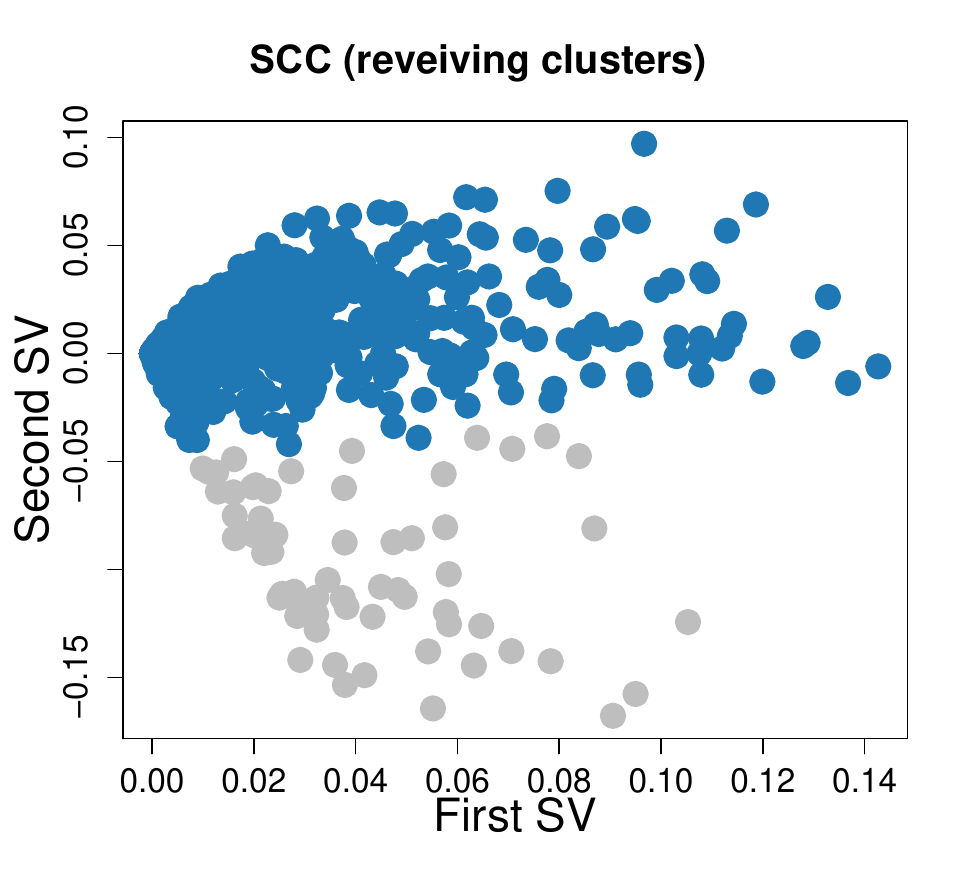}}
    \subfigure[RP-SCC]{\includegraphics[height=4.3cm,width=4.9cm,angle=0]{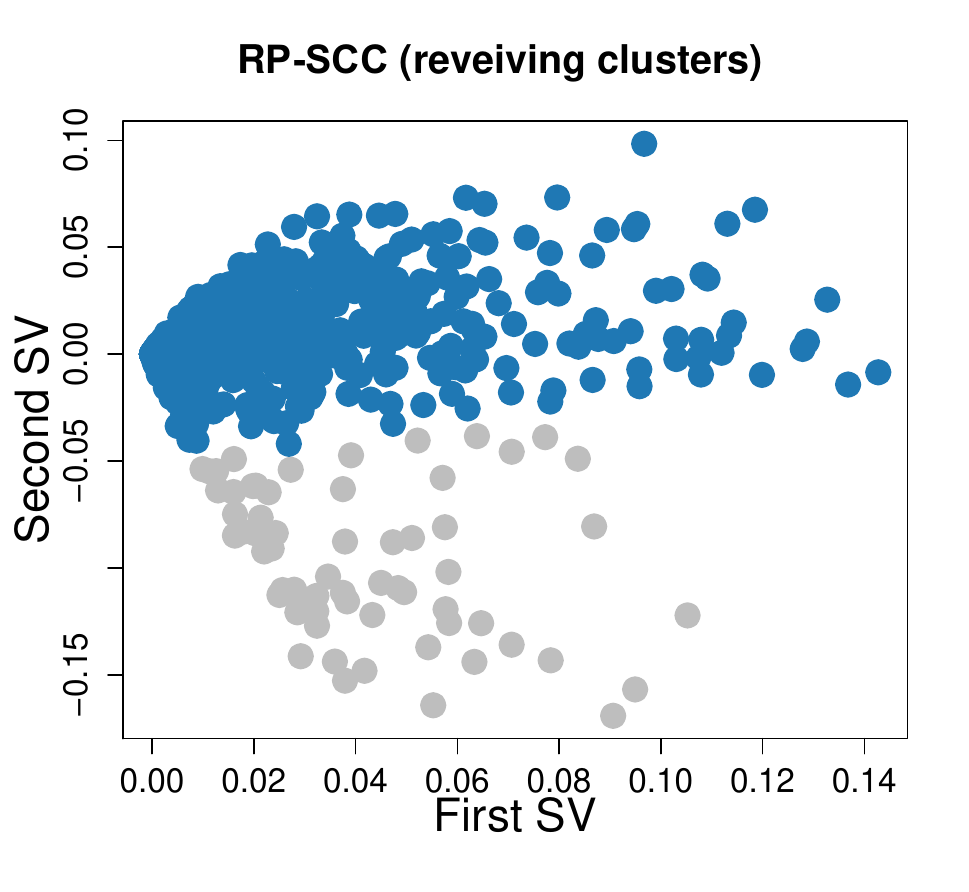}}
    \subfigure[RS-SCC]{\includegraphics[height=4.3cm,width=4.9cm,angle=0]{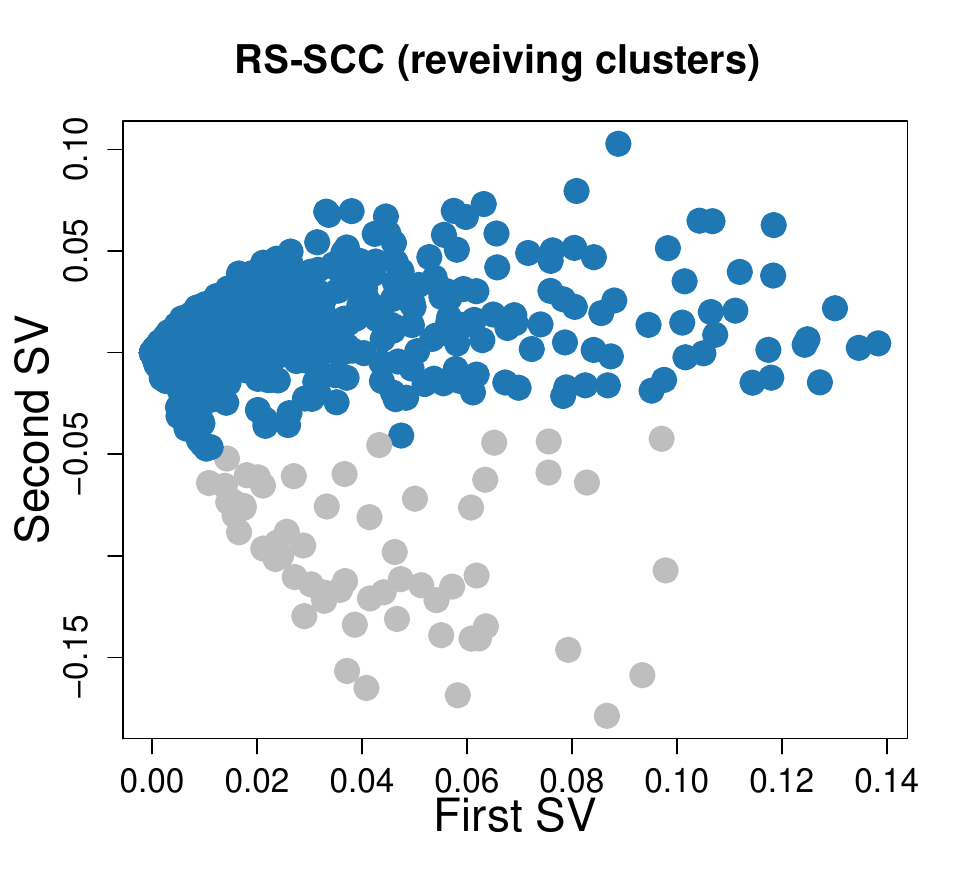}}
    \subfigure[\textsf{svds}]{\includegraphics[height=4.3cm,width=4.9cm,angle=0]{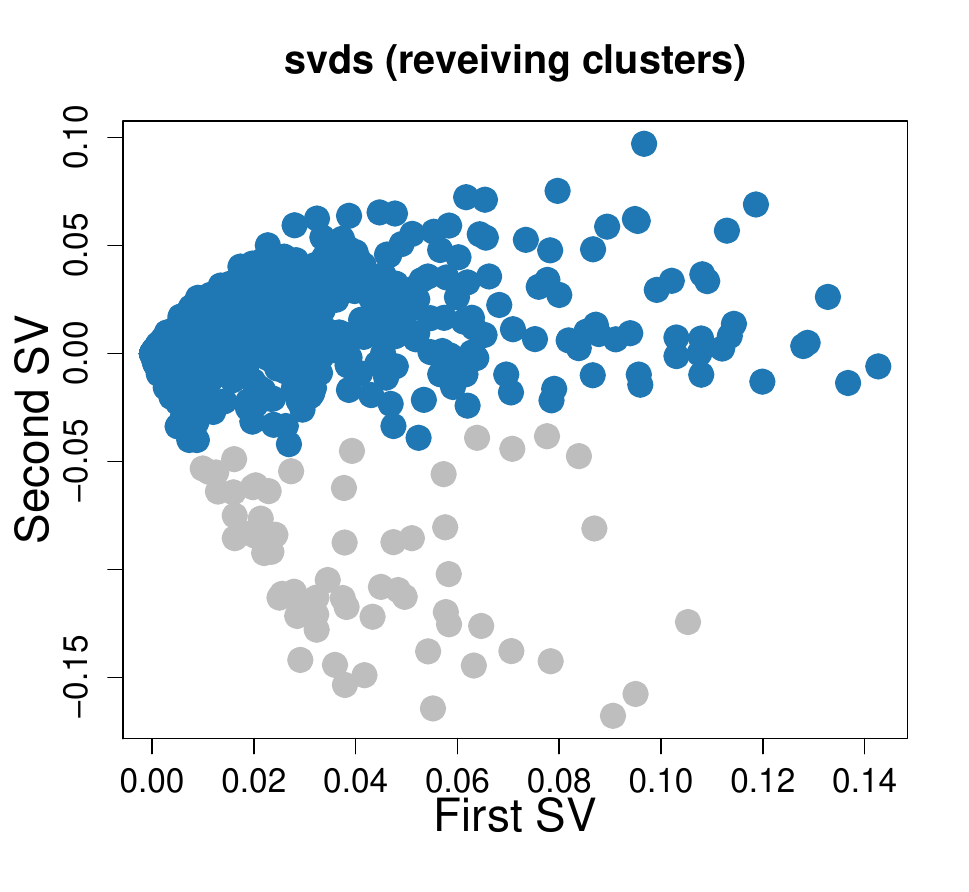}}
    \subfigure[\textsf{irlba}]{\includegraphics[height=4.3cm,width=4.9cm,angle=0]{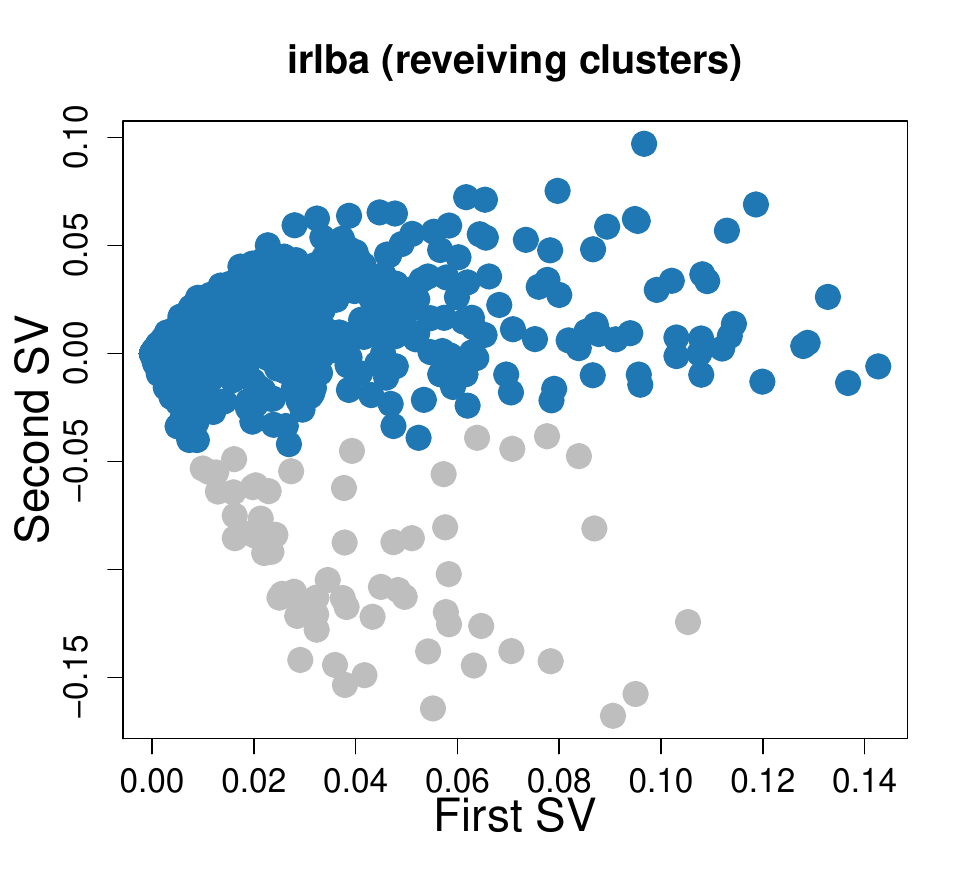}}
	\caption{Receiving clusters of the European email network detected by SCC and four SCC-based approximate algorithms.}\label{emailpointscolumn}
\end{figure}

\begin{figure}[!htbp]{}
	\centering
	\subfigure[Sending clusters]{\includegraphics[height=6.5cm,width=6.5cm,angle=0]{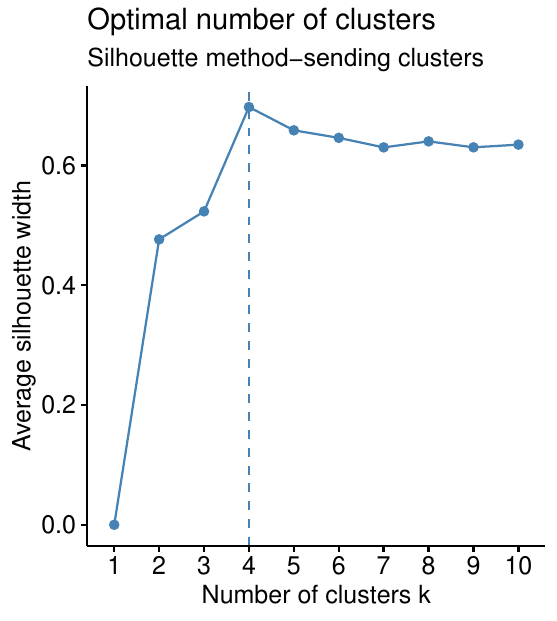}}\hspace{1.3cm}
    \subfigure[Receiving clusters]{\includegraphics[height=6.5cm,width=6.5cm,angle=0]{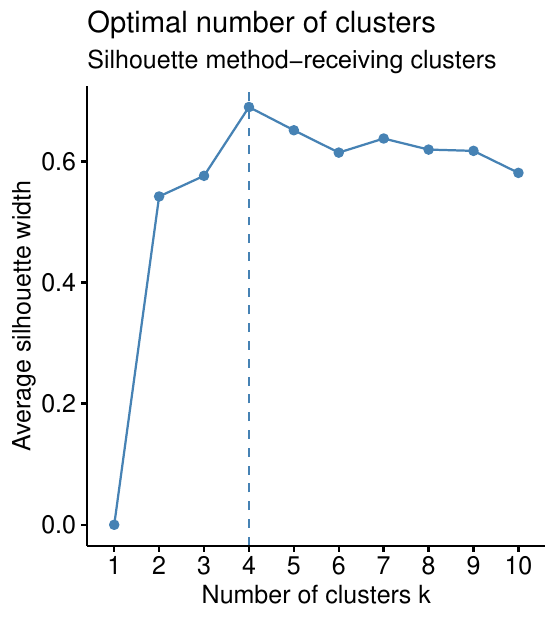}}
	\caption{Optimal number of clusters of the European email network selected by the average silhouette method based on the SsCC.}\label{emailclusternumberdc}
\end{figure}

\begin{figure}[!htbp]{}
	\centering
	\subfigure[Sending clusters]{\includegraphics[height=5.5cm,width=5.3cm,angle=0]{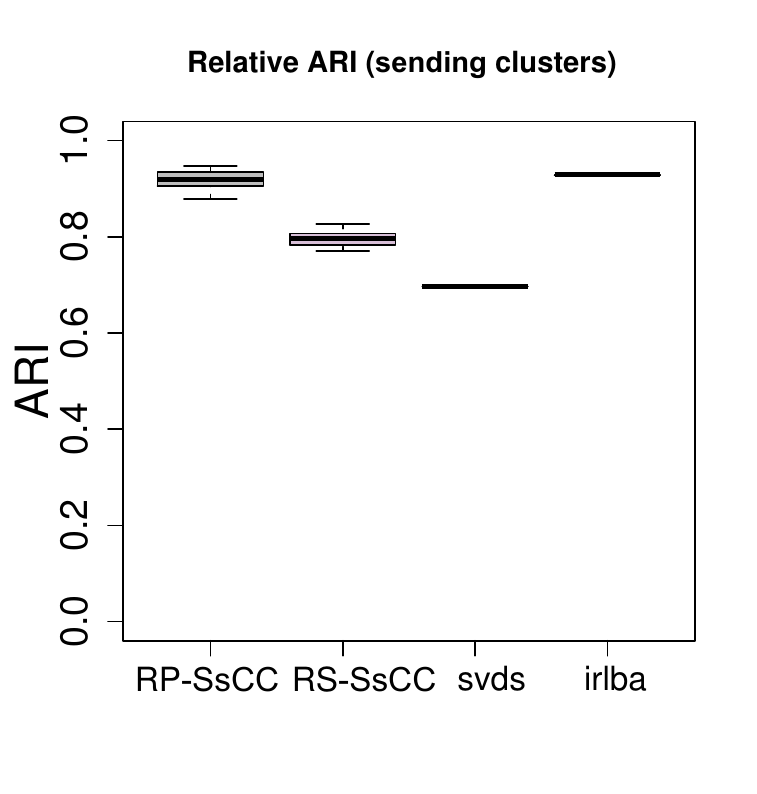}}\hspace{1.3cm}
    \subfigure[Receiving clusters]{\includegraphics[height=5.5cm,width=5.3cm,angle=0]{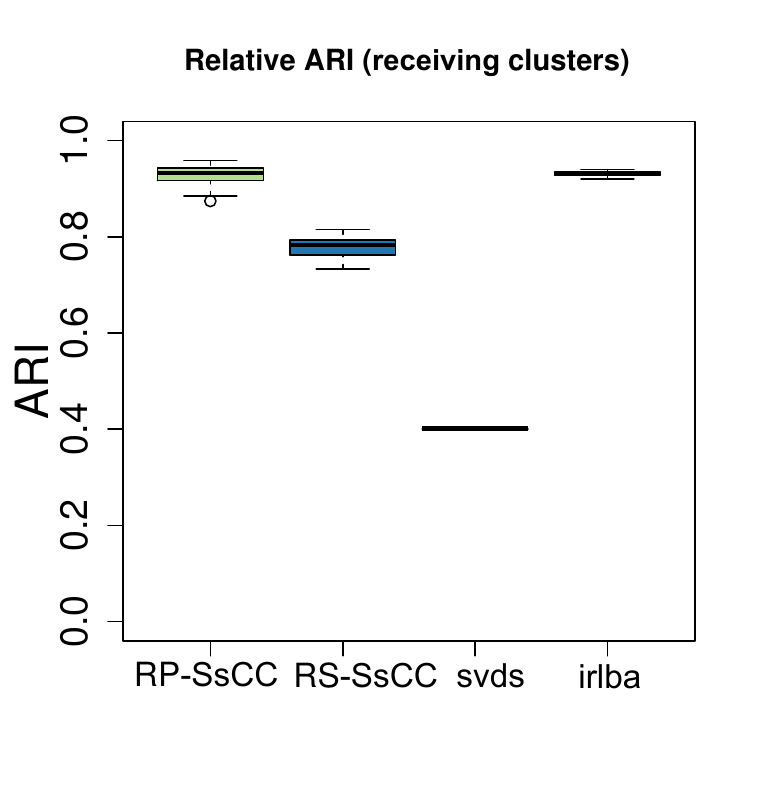}}
	\caption{Relative ARI between the SsCC and SsCC-based four approximate methods on the European email network. }\label{emailboxdc}
\end{figure}

This section provides the additional experimental results that are not shown in the main text. Figure \ref{m2case1}-\ref{m8case2} display the simulation results with respect to model set-up 2 to 8. Figure \ref{emailsv}-\ref{emailboxdc} show the results associated with the European email network.

\newpage

\newpage


\vskip 0.2in
\newpage
\bibliographystyle{plainnat}
\bibliography{randspectral}
\end{document}